\documentclass[sigconf]{acmart}

\AtBeginDocument{%
  }
    
\copyrightyear{2022}
\acmYear{2022}
\setcopyright{acmlicensed}\acmConference[CCS '22]{Proceedings of the 2022 ACM SIGSAC Conference on Computer and Communications Security}{November 7--11, 2022}{Los Angeles, CA, USA} 
\acmBooktitle{Proceedings of the 2022 ACM SIGSAC Conference on Computer and Communications Security (CCS '22), November 7--11, 2022, Los Angeles, CA, USA}
\acmPrice{15.00}
\acmDOI{10.1145/3548606.3560662} \acmISBN{978-1-4503-9450-5/22/11}

\usepackage{mdframed,lipsum}
\usepackage{amsmath,amsfonts}
\usepackage{graphicx}
\usepackage{xcolor}
\usepackage{textcomp}
\usepackage{bm}
\usepackage{balance}  
\usepackage{float}
\usepackage{xcolor}
\usepackage{color, colortbl}
\usepackage{diagbox}
\usepackage{amsmath}
\usepackage{url}
\usepackage{multirow}
\usepackage[boxruled,linesnumbered]{algorithm2e}
\usepackage{subcaption}
\usepackage{mathtools}
\usepackage{amsfonts}
\usepackage{threeparttable}
\usepackage{booktabs}
\usepackage{enumitem}
\usepackage{subfiles}

\def\g{{\bf g}}

\def\0{{\bf 0}}
\def\1{{\bf 1}}

\newcommand{\nop}[1]{}
\newcommand{\Wendy}[1]{{{\textcolor{black}{\textbf{Wendy:}}}{\textcolor{red}{\textbf{#1}}}}}
\newcommand{\xiuling}[1]{{{\textcolor{black}{\textbf{xiuling:}}}{\textcolor{brown}{\textbf{#1}}}}}

\newcommand{\TargetM}{T}
\newcommand{\TargetG}{G}
\newcommand{\ShadowG}{\TargetG^{S}}
\newcommand{\ShadowM}{\TargetM^{S}}
\newcommand{\PartialG}{{\TargetG}^A}
\newcommand{\Ttrain}{\TargetM^{\text{train}}}
\newcommand{\Ttest}{\TargetM^{\text{test}}}
\newcommand{\PIA}{P}
\newcommand{\ExternalK}{\mathbb{K}}
\newcommand{\Ptrain}{\PIA^{\text{train}}}
\newcommand{\Ptest}{\PIA^{\text{test}}}
\newcommand{\Property}{\mathbb{P}}

\def\gat{\textsc{GAT}\xspace}
\def\gcn{\textsc{GCN}\xspace}

\usepackage{colortbl}
\definecolor{asparagus}{rgb}{0.53, 0.66, 0.42}
\definecolor{carrotorange}{rgb}{0.93, 0.57, 0.13}
\definecolor{lightthulianpink}{rgb}{0.9, 0.56, 0.67}

\newenvironment{ditemize}{%
\begin{list}{{\bf $\bullet$}}{%
\setlength{\itemsep}{0pt}\setlength{\rightmargin}{0pt}%
\setlength{\leftmargin}{1.2em}\setlength{\parsep}{0pt}}}{
\end{list}}

\begin{document}

\title{Group Property Inference Attacks Against \\Graph Neural Networks}

\author{Xiuling Wang, Wendy Hui Wang}
\email{{xwang193, hwang4}@stevens.edu}
\affiliation{%
  \institution{Stevens Institute of Technology}
  \city{Hoboken}
  \state{NJ}
  \country{USA}
  \postcode{07030}
}

\begin{abstract}
With the fast adoption of machine learning (ML) techniques, sharing of ML models is becoming popular.  However, ML models are vulnerable to privacy attacks that leak information about the training data. In this work, we focus on a particular type of privacy attacks named {\em property inference attack} (PIA) which infers the sensitive properties of the training data through the access to the target ML model. 
In particular, we consider Graph Neural Networks (GNNs) as the target model, and distribution of particular groups of nodes and links in the training graph as the target property. While the existing work has investigated PIAs that target at graph-level properties, no prior works have studied the inference of node and link  properties at group level yet. 

In this work, we perform the first systematic study of {\em group property inference attacks} (GPIA) against GNNs. First, we consider a taxonomy of threat models under both black-box and white-box settings with various types of adversary knowledge, and design six different attacks for these settings. We evaluate the effectiveness of these attacks through extensive experiments on three representative GNN models and three real-world graphs. 
Our results demonstrate the effectiveness of these attacks whose accuracy outperforms  the baseline approaches. Second, we analyze the underlying factors that contribute to GPIA's success, and show that the target model trained on the graphs with or without the target property represents some dissimilarity in model parameters and/or model outputs, which enables the adversary to infer the existence of the property. Further, we design a set of defense mechanisms against the GPIA attacks, and demonstrate that these mechanisms can reduce attack accuracy effectively with small loss on GNN model accuracy.   

\end{abstract}



\maketitle

\vspace{-0.1in}
\section{Introduction}

Advances in machine learning (ML) in recent years have enabled a large array of applications such as data analytics, autonomous systems, and security diagnostics. Building good ML models, however, require computational resources and possibly significant financial investment.  
Small companies as well as developers and researchers with limited resources may not be able to afford such impractical cost. 
This motivates the creation of online ML model  marketplaces where ML models are shared and traded \cite{caffezoo,aws,modzy}. However, since these models may be trained on the data that contains sensitive information, it raises an important question: {\em how much does the model reveal about the training data? }

Recent studies have identified a number of attacks to infer the sensitive information in the training data. For example, membership inference attacks \cite{shokri2017membership,nasr2018machine} infer whether a particular data sample was used in the training of the ML models. Model inversion attacks  reconstruct the training examples given the access to the target model  \cite{fredrikson2014privacy,fredrikson2015model,wu2016methodology}. These attacks focus on the privacy of {\em individual records} in the dataset. On the other hand, property inference attacks \cite{ateniese2013hacking,ganju2018property,melis2019exploiting} infer the aggregate information (property) of the dataset.  

In this paper, we consider Graph Neural Networks (GNNs) as the target model. We consider the {\em group properties} that are defined over distribution of nodes and links as the attack target. An example of the node-level group property is that a professional network graph contains more male users than female ones, while an example of the link-level group properties is that a  social network graph has more links between White users than those between  African American users.  
Obtaining these properties by the adversary may directly violate the intellectual property (IP) of the model owner \cite{zhou2021property}. 

In general, property inference exploits the idea that ML models trained on similar datasets will represent some similarity in the model parameters and/or the model outputs \cite{ateniese2013hacking,ganju2018property}. 
Following this idea, various PIA models have been designed to attack the  classification models \cite{ateniese2013hacking}, deep neural networks \cite{ganju2018property,melis2019exploiting}, and generative adversarial networks \cite{zhou2022property}.  
All these works mainly focus on ML models trained on tabular and image data. 
Few work \cite{zhang2021inference,suri2021formalizing,zhang2021leakage} has studied PIA against GNN models over graph data. None of these works have investigated the leakage of nodes and links at group level.  More detailed comparison between these works and ours can be found in Table \ref{tab:comp-literature} and Section \ref{sc:relatedwork}.

Intuitively, the attacker  can infer the data properties by inferring the values of the property features (e.g., Gender) from the model outputs through  the attribute inference attacks (AIAs) \cite{song2020information,duddu2020quantifying,wen2021meta}. However, the effectiveness of AIAs highly relies on several assumptions such as strong correlation between the property features and the label as well as the adversary knowledge of the property features of a subset of nodes in the original graph \cite{song2020information,duddu2020quantifying}. These assumptions may not hold for PIAs. Indeed, both the prior work \cite{zhang2021leakage} and our analysis show that PIA can incur leakage even when the property features are weakly correlated with the task label. 

\begin{table*}[t]
\small
     \centering
     {
     \begin{tabular}{c|c|c|c|c|c}\hline
     \multirow{2}{*}{}&\multirow{2}{*}{{\bf GNN model}} & \multicolumn{2}{c|}{{\bf Adversary knowledge}}& \multicolumn{2}{c}{{\bf Target properties}}  \\\cline{3-6}
     &&{\bf White-box} & {\bf Black-box} &{\bf Graph-level} & {\bf Group-level} \\\hline
     \cite{suri2021formalizing}&GCN & Model loss & $\times$ & Avg. node degree& $\times$ \\\hline
     \cite{zhang2021leakage}&GCN & $\times$ & Model prediction & $\times$ & Node group distribution \\\hline
     \cite{zhang2021inference}&GraghSAGE & Graph embedding& $\times$ & \# of nodes \& edges, graph density & $\times$ \\\hline
     Ours&GCN, GraghSAGE, GAT & Node embedding & Model prediction & $\times$ & Node \& link group distribution \\\hline
     \end{tabular}
     }
     \caption{Comparison between the existing works on property inference attacks against GNNs and ours.}
    \label{tab:comp-literature}
     \vspace{-0.15in}
\end{table*} 

In this paper, we perform the first systematic investigation of vulnerability of GNNs against {\em group property inference attacks} (GPIA).  
We consider both {\em black-box} and {\em white-box} settings. Under the former setting, the adversary only can access the output of the target model (e.g., posterior probability), while under the latter setting, he/she can access the architecture and parameters (e.g., node embeddings) of the target model. 
For both settings, we consider a comprehensive taxonomy of the threat model with various types of adversary knowledge, and design six attacks for these settings. 
All these attacks are designed as the classification models which are trained to distinguish the positive graphs (i.e., the graphs with the properties) from the negative ones (i.e., the graphs without the properties)   by the behaviors and outputs of either the target model or a {\em shadow model} that mimics the target model on these graphs. 

We evaluate the effectiveness of our proposed  attacks on three representative GNN models, namely, GCN \cite{kipf2017semisupervised}, GraphSAGE \cite{hamilton2018inductive}, and GAT \cite{velickovic2018graph}, and three real-world graph datasets. 
The results demonstrate that our attacks are effective under various settings. For example, a black-box GPIA attack with only 20\% of the training graph available in the adversary knowledge can achieve the accuracy in the range of [0.9, 1] and [0.72, 0.92]  for node and link properties respectively. The attack accuracy remains to be effective when the adversary transfers the knowledge learned from a shadow graph to infer the properties in the target graph. For example, the attack accuracy can be as high as 0.66 when the adversary uses the Facebook social network graph as the shadow graph to infer whether the Pokec social network graph has more male users than female ones. Furthermore, our attacks greatly outperform the  baseline methods that utilize the attribute inference attacks (AIAs) \cite{song2020information,duddu2020quantifying,wen2021meta} and the meta-classifier \cite{ateniese2013hacking,ganju2018property,zhang2021leakage}. 

Next, we analyze the main factors that contribute to GPIA's success.
We found that, due to the (indirect) correlation between the property feature and the label as well as the non-negligible disparity in the {\em influence} of different node/link groups on the target model, the model parameters (node embeddings) and model outputs by the target model trained on the data with the target property $\Property$ is distinctly dissimilar to those obtained from the data without $\Property$, which enables the adversary to infer the existence of $\Property$.

Further, we design three defense mechanisms to mitigate the vulnerabilities of GPIA under both black-box and white-box settings. For the former setting, we add Laplace noise to perturb the posterior output. For the latter setting, we design two defense mechanisms, namely adding Laplace noise on the node embeddings and compressing node embeddings. We evaluate the performance of these defense mechanisms, and show that these defense mechanisms can reduce the attack accuracy significantly with either a small amounts of noise or a small compression ratio. Furthermore, both defense mechanisms address the trade-off between privacy and model accuracy; the model accuracy is still acceptable when the defense is sufficiently strong. 

In summary, we make the following contributions in this paper:
\begin{ditemize}
    \item We design the first set of attacks against GNNs that can infer the  properties of groups of nodes and links in the training graph. 
    \item We perform extensive empirical studies and demonstrate the effectiveness of our proposed attacks. 
    \item We analyze the main factors that contribute to the success of the attacks. 
    \item We propose three defense mechanisms and demonstrate their effectiveness against the proposed attacks.  
\end{ditemize}

\vspace{-0.1in}
\section{Related work}
\label{sc:relatedwork}

\noindent{\bf Privacy attacks against GNNs. }
Many studies have explored the privacy vulnerability of GNNs. Based on  which assets in GNN models are considered as sensitive and the adversary tries to obtain, these privacy attacks can be grouped into two categories: (1) {\em privacy attacks on GNN models} that aim to extract information about the model’s structure and parameters; and (2) {\em privacy attacks on training data} that aim to infer the sensitive information in the training graph. There has been few studies on privacy attacks on GNN models: Wu et al. \cite{wu2021model} designed the {\em model extraction attack} that aims to reconstruct a duplicated GNN model. 
In terms of privacy attacks on training data of GNNs, He et al. \cite{he2020stealing} designed the link stealing attacks to infer if some specific links exist in the training graph. 
Duddu et al. \cite{duddu2020quantifying} designed three privacy attacks against GNNs - a {\em membership inference attack} that infers whether a graph node was in the training data, a {\em graph reconstruction attack} that reconstructs the target graph, and an {\em attribute inference attack} that infers the sensitive attributes.  
He et al. \cite{he2021nodelevel} proposed the node-level membership inference attacks against GNNs. Wu et al. \cite{wu2022linkteller} considered the data partition setting where each data holder has either node features or edge information, and proposed a link-level membership inference attack to infer the existence of links. Zhang et al. \cite{zhang2021inference} designed three inference attacks against GNNs: (1) a {\em property inference attack} that infers the  graph-level information such as graph density and number of nodes/edges of the training graph; (2) a {\em subgraph inference attack} that  infers  whether a given subgraph is contained in the training graph; and (3) a  {\em graph reconstruction attack} that reconstructs the structure of the training graph. 

\noindent{\bf Property inference attacks against ML models.}
Ateniese et al. \cite{ateniese2013hacking} first proposed the concept of the  property inference attack against ML models. They design a white-box PIA model and demonstrate its effectiveness against SVM and HMM models. 
In the following years, the design of PIA has been extended to fully connected neural networks \cite{ganju2018property}, Convolutional Neural Networks (CNNs) \cite{parisot2021property}, collaborative learning  \cite{melis2019exploiting}, federated learning \cite{wang2019eavesdrop,wang2019beyond}, and GANs \cite{zhou2022property}. Mahloujifar et al. \cite{chase2021property} designed a property inference poisoning attack by which the adversary can learn a particular property in the training data by injecting specially crafted poison data in the data. Unlike their work, we assume the adversary has no update access to the training data. 

\noindent{\bf Property inference attacks against GNNs.} 
Very few works \cite{suri2021formalizing,zhang2021leakage,zhang2021inference} have studied property inference attacks against GNNs. Suri et al. \cite{suri2021formalizing} proposed a generic definition of PIA which defines the attack goal as distinguishing between two possible training distributions. They assume that the adversary has the access to (transformed) data distributions, while we assume the adversary only has access to either node embeddings or posterior probabilities. Zhang et al. \cite{zhang2021leakage} studied the leakage of properties of node group distribution in the centralized multi-party setting. They show that PIA can incur leakage even when the property attribute is not correlated with the label. 
Probably the property inference attack in  \cite{zhang2021inference} is the most relevant to our work. However, it differs from our attack fundamentally from the following perspectives. First,  \cite{zhang2021inference} considers the properties at graph level (e.g., number of nodes/edges and graph density), while we consider the  properties of nodes and links at group level. Second,  \cite{zhang2021inference} considers the graph embedding (i.e., the vector representation of the whole graph) in the adversary knowledge, while we consider node embeddings. These major differences in the type of properties and adversary knowledge lead to fundamentally different design of PIA models.
Furthermore, besides the empirical results to demonstrate the effectiveness of PIA and its defenses, we provide in-depth investigation of which factor(s) contribute to GPIA's success.

\vspace{-0.1in}
\section{Graph Neural Network}
 

In general, GNNs take an input graph $\TargetG(V, E)$, along with a set of node features, to generate a representation vector $z_i$ (node embedding) for each node $v_i\in V$. 
One of the defining features of GNN models is that is uses a form of {\em neural message passing} by which vector messages are exchanged between nodes in the graph and updated using neural networks.  

In particular, during each message-passing iteration\footnote{The different iterations of message passing are also sometimes known as the different “layers” of the GNN.} in a GNN, the embedding  $z_i^{(\ell)}$ corresponding to each node $v_i\in V$ at layer $\ell$ is updated according to $v_i$'s graph neighborhood $\mathcal{N}(v_i)$ (typically 1-hop neighborhood). This update process can be expressed as:
\begin{equation}
    \label{eqn:aggregate}
    \begin{split}
    z_i^{\ell+1} & = \textsf{UPDATE}^{\ell}(z_i^{\ell}, \textsf{AGGREGATE}^\ell(\{z_j^{\ell},\forall v_j\in \mathcal{N}(v_i)\})),  
    \end{split}  
    \vspace{-0.2in}
\end{equation}
where $\textsf{UPDATE}$ and $\textsf{AGGREGATE}$ are arbitrary differentiable functions (e.g., neural networks). 
The initial embeddings at $\ell = 0$ are set to the input features for all the nodes, i.e., $z^0_i = x_i, \forall v_i\in V$. 

After $k$ iterations of message passing, a $\textsf{Readout}$ function pools the node embeddings at the last layer and produces the prediction results. The $\textsf{Readout}$ function varies by the learning tasks. In this paper, we consider node classification as the learning task. For this task, often the $\textsf{Readout}$ function is a softmax function. The prediction output for each node $v$ is a vector of probabilities, each corresponding to the predicted probability (posterior) that $v$  is assigned to a class.

In this paper, we consider three representative GNN models, namely {\bf Graph Convolutional Network (GCN)} \cite{kipf2017semisupervised}, {\bf GraphSAGE} \cite{hamilton2018inductive}, and {\bf Graph Attention network (GAT)} \cite{velickovic2018graph}. These three models mainly differ on either \textsf{AGGREGATE} and \textsf{UPDATE} functions. 
More details of the two functions for the three GNN models can be found in Appendix \ref{appendix:gnn}. 

\vspace{-0.1in}
\section{Problem Formulation}
\label{sc:setting}
\begin{table}[t!]
\centering
\setlength\tabcolsep{5pt}
\begin{tabular}{c|c}
\toprule
Symbol & Meaning \\\hline
$v/e(v_i, v_j)$ & node/link between two nodes $v_i, v_j$\\\hline
$A$/$X$& Property/non-property feature\\\hline
$\Property$ & Target property \\\hline
$\TargetG$/$\ShadowG$ & Target/shadow graph\\\hline
$\TargetM$/$\ShadowM$ & Target/shadow model\\\hline 
$\PIA$ & GPIA attack classifier\\\hline 
$\Ttrain$, $\Ttest$ & Training and testing datasets of target model $\TargetM$\\\hline 
$\Ptrain$, $\Ptest$ & Training and testing datasets of GPIA model $\PIA$ \\\hline 
$Z^i$ & Node embedding generated at the $i$th-layer of $\TargetM$\\\hline
\bottomrule
\end{tabular}
\caption{\label{tab:notation} Notations}
\vspace{-0.25in}
\end{table}
Given a graph 
$\TargetG(V, E)$ 
and a GNN model $\TargetM$ trained on $\TargetG$, the goal of GPIA is to infer whether $\TargetG$ has a group property $\Property$ from the access to $\TargetM$. 
Table \ref{tab:notation} lists the common notations used in the paper. 

\subsection{Group Properties} 

In this paper, we consider two types of properties that the adversary aims to infer: {\em node  group properties} (node properties)  that specify the aggregate information of particular node groups;
and 
{\em link  group properties} (link properties) that specify the aggregate information of particular link groups. 
The property can be either binary or non-binary. 
An example of the binary property is whether the graph contains more female nodes than male ones. 
An example of the non-binary property is whether the graph has 75\%, or 50\%, or 25\% female nodes. In this paper, we only consider binary properties. 
If a graph has the property $\Property$, we say it is a {\em positive} graph. Otherwise, it is a {\em negative} graph.  

{\bf Node/link groups.} We assume the nodes are associated with a set of  features $P$ (called as {\em property features}) on which the grouping of nodes and links will be defined. For simplicity, we only consider one property feature in this paper. The rest of the node features are called as {\em non-property features}. 
 Typical examples of the property features include the demographic features such as gender and race. The grouping of nodes and links is specified by adding value-based constraints (VBCs) on the property features. For example,  $\texttt{gender=“Male”}$ defines the male group. 
 \nop{
 The grouping of nodes and links will be specified by adding value-based constraints (VBCs) on the property features, where the VBCs can be either point-based (i.e., gender = “Male”) or range-based (i.e., age $\in[20, 25])$. 
We use $V_{C}$ and $E_{C}$ to denote the set of nodes and links that satisfy VBC $C$ respectively. For example, $V_{\text{gender=“Male”}}$ defines the male group. }

{\bf Node properties.} The node properties are specified on the property features with aggregate functions and arithmetic comparison operators. 
In this paper, we consider COUNT() as the  aggregate function, and five arithmetic comparison operators including $<$, $\leq$, $>$, $\geq$, $=$, and $\neq$. 
An example of the node property $\Property$ is “\texttt{COUNT(Male) > COUNT(Female)}”. 
    
 {\bf Link properties.} The link properties are specified on property features of both end nodes in the links, with aggregate functions and arithmetic comparison operators.
 An example of the link property is
“\texttt{COUNT(Male-Male) $>$ COUNT(Female-Female)}”, i.e.,  there are more links between male users than between female users.

\nop{
\subsection{Target Model} 

In this paper, we mainly consider Graph Neural Networks (GNNs) as the target model and node classification as the learning task. Specifically, given a number of pre-defined class labels, the goal of node classification is to predict the class(es) that each node belongs to. The output of the target GNN model $\TargetG$ takes the format of a set of  probabilities (posteriors), where each node is assigned with a vector of posterior probability, each indicating the likelihood that the node belongs to a specific class.
}

\subsection{Adversary Knowledge} 
The adversary may  have additional background knowledge $\ExternalK$ which can be categorized along three dimensions: 
\begin{ditemize}
\item {\em Partial graph} $\PartialG$: the adversary has a subgraph $\PartialG\subset\TargetG$. 
\item {\em Shadow graph} $\ShadowG$: the adversary has a shadow graph (or multiple graphs) $\ShadowG$ which contains its own structure and node attributes. $\ShadowG$ may have different domain and data distribution from  $\TargetG$; 
\item  {\em Target model} $\TargetM$: We consider two types of adversary knowledge of $\TargetM$: the {\em white-box} access to $\TargetM$, which reveals the model architecture, parameters, and the loss function, and the {\em black-box} access which allows the adversary to obtain the target model output (i.e., posteriors) only. We also assume that the adversary has the knowledge of the number of classes for the target model.  
\end{ditemize} 

The assumption of the white-box setting is reasonable and quite common nowadays \cite{ateniese2013hacking,ganju2018property}. For example, some online platforms \cite{caffezoo,modzy} share their models openly, including their parameters, thereby providing white-box access. On the other hand, ML-as-a-service services (e.g. \cite{aws,bigml,googleclound}) that provide an API for users to query for predictions but keep their models inaccessible to users are typical examples of black-box settings. 


\nop{
\subsection{Problem Definition} 
Our goal of GPIA is to design a binary classifier $\PIA$ that predicts whether the training graph $\Ttrain$ has a pre-defined property $\Property$.  
Then, the attack's goal can be formulated as: 
\[\PIA: \ExternalK, \Property  \rightarrow L\]
where $\mathcal{K}$ denotes the adversary knowledge, and $L$ is the set of class labels for property prediction.  In this paper, we only consider binary property (i.e., $L=\{0, 1\}$). We will discuss how to extend to non-binary properties in Section \ref{sc:conclusion}.
}

\section{Methodology}
 \label{sc:method}

\begin{table}
    \centering
    \begin{tabular}{c|c|c|c}
    \hline
         \multirow{2}{*}{\textbf{Attack}}& \multicolumn{3}{c}{\textbf{Adversary knowledge}} \\\cline{2-4}
          & {$\ShadowG$} & Access to {$\TargetM$} & {$\PartialG$}\\\hline
          $A_1$ & $\times$ & White-box & $\checkmark$ \\\hline
          $A_2$ & $\times$ & Black-box & $\checkmark$ \\\hline
         $A_3$ & $\checkmark$ & White-box & $\times$ \\\hline
         $A_4$ & $\checkmark$ & Black-box & $\times$ \\\hline
          $A_5$ & $\checkmark$ & White-box & $\checkmark$ \\\hline
         $A_6$ & $\checkmark$ & Black-box & $\checkmark$ \\\hline
    \end{tabular}
    \caption{Attack taxonomy ($\ShadowG$: shadow graph; $\TargetM$: target model; $\PartialG$: partial graph).}
    \label{tab:attacks}
    \vspace{-0.3in}
\end{table}

Given a target graph $\TargetG$ and a GNN model $\TargetM$ trained on $\TargetG$, the adversary aims to infer if $\TargetG$ has the property $\Property$ by either the white-box access to node embeddings  or the black-box access to posterior probabilities output by $\TargetM$. An example for the former case is that the data owner uploads node embeddings to a third-party service provider such as Google's Embedding Projector service\footnote{https://projector.tensorflow.org/} to perform downstream analysis tasks, while an example for the latter case is that the data owner uploads the posterior probability (e.g., by a GNN-based recommender system \cite{fu2020fairness}) to a third-party online optimization solver such as Gurobi\footnote{https://www.gurobi.com} for optimization. Another possible attack scenario is  the collaborative setting under which the attacker and other parties train a model jointly by sharing either the model predictions or node embeddings  \cite{zhang2021leakage}. The attacker is curious to infer the properties of other parties' data from their shared embeddings/predictions.

Formally, the attack's goal is to design a binary classifier $\PIA$ that can be formulated as: $\PIA: \ExternalK, \Property  \rightarrow L$, 
where $\ExternalK$ denotes the adversary knowledge, and $L$ is the set of class labels for property prediction.  In this paper, we only consider binary property (i.e., $L=\{0, 1\}$). We will discuss how to extend to non-binary properties in Section \ref{sc:conclusion}.

Whether the adversary has each of $\PartialG$, $\TargetM$, and $\ShadowG$ in $\ExternalK$ is a binary choice. However, we assume at least one of  $\PartialG$ and $\ShadowG$ is available for training of GPIA model, as the adversary always can obtain some public graphs from external resources as the shadow graphs if the partial graph is not available. Therefore, we have a comprehensive taxonomy with six different threat models based on different combinations of $\PartialG$, $\TargetM$, and $\ShadowG$ in $\ExternalK$. We design six GPIA attack classifiers for these threat models, and summarize the taxonomy of our  attacks in Table \ref{tab:attacks}. Next, we describe the details of the black-box attacks ($A_2, A_4, A_6$) first, followed by the details of the white-box attacks ($A_1, A_3, A_5$).  
Enlightened by the existing PIA works  \cite{zhang2021leakage,zhang2021inference}, our attacks also use shadow models. However, due to the assumption of different adversary knowledge (see Table \ref{tab:comp-literature}), the design of our shadow models is fundamentally different from these works in the design of attack features.
\nop{
\begin{table}[t!]
    \centering
    \begin{tabular}{c|c|c|c}
    \hline
       \multirow{2}{*}{Attack} & \multirow{2}{*}{Attack type} & \multicolumn{2}{c}{\bf Feature dimension size}\\\cline{3-4}
       & & Before alignment & After alignment \\\hline
        $A_1$ & White-box & & \\\hline
        $A_2$ & Black-box & &N/A \\\hline
        $A_3$ & White-box & & \\\hline
        $A_4$ & Black-box & & N/A\\\hline
        $A_5$ & White-box & & \\\hline
        $A_6$ & Black-box & & N/A\\\hline
    \end{tabular}
    \caption{Comparison of six attacks}
    \label{tab:attack-feature}
    \vspace{-0.2in}
\end{table}
}
\nop{
We design two types of attacks in terms of different adversary knowledge $G_{ack}$: 
\begin{itemize}
    \item {\bf Black-box attack} for the adversary who can access the target model output only. As we only consider the node classification task in this paper, we consider the probability posteriors generated by the target GNN model as the input to the black-box attack.   
    \item {\bf White-box attack} when the adversary has access to the target model, including the model's architecture and parameter values. 
\end{itemize}
}

\vspace{-0.1in}
\subsection{Black-box Attacks}
\label{sc:black-box-ttack}
 \begin{figure*}[t!]
\begin{center}

\begin{subfigure}[b]{.9\textwidth}
      \centering
    \includegraphics[width=\textwidth]{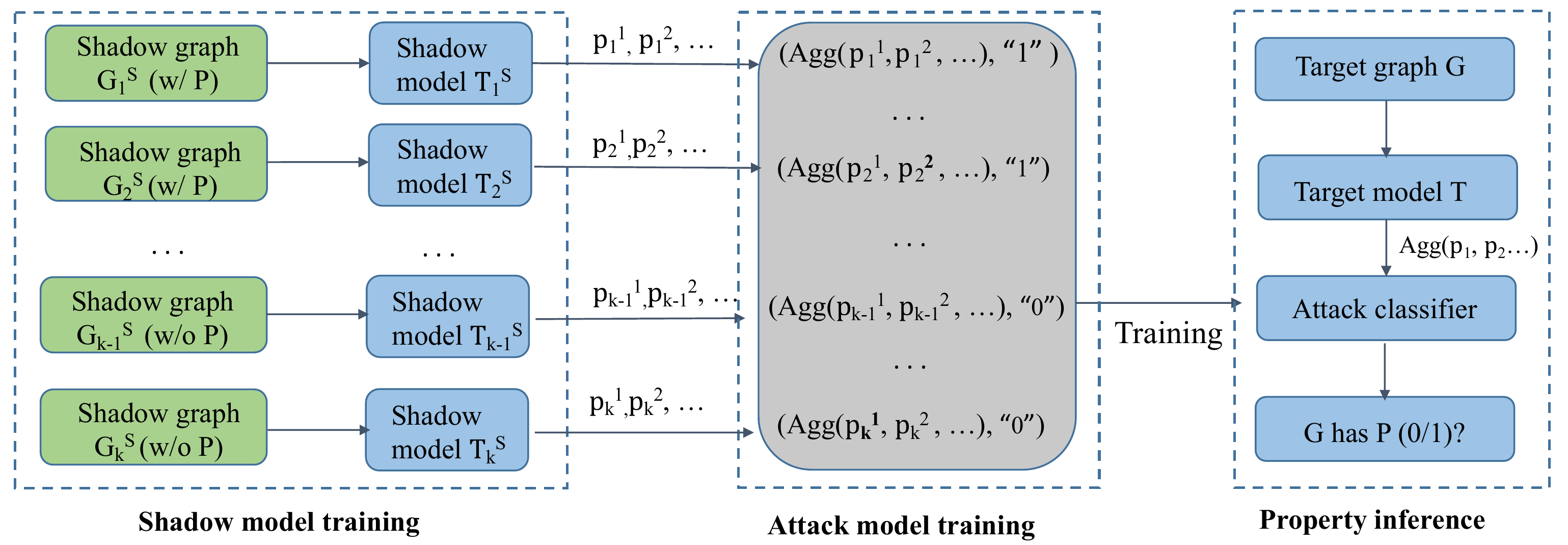}
\vspace{-0.1in}
\caption{\label{figure:pia-attack} Overview of the black-box GPIA. \textsf{Agg}() is the aggregation function that generates GPIA features from posteriors.}
\end{subfigure}
\end{center}
\vspace{-0.1in}
\end{figure*}

The black-box attack includes three phases: {\em shadow model training}, {\em attack model training}, and {\em property attack inference} (Figure \ref{figure:pia-attack}). Next, we explain the details of these three phases. 

{\bf Shadow model training phase.} To collect the data $\Ptrain$ to train the GPIA classifier $\PIA$, first, the adversary trains $k\geq 1$ shadow models ${\ShadowM_{1}}$, $\dots, {\ShadowM_k}$. 
The training data for each shadow model ${\ShadowM_{i}}$ is a subgraph ${\ShadowG_{i}}$ that is randomly sampled from  the partial graph $\PartialG$ (Attack $A_2$), the shadow graph $\ShadowG$ (Attack $A_4)$, or both (Attack $A_6$). 
Each shadow training graph ${\ShadowG_{i}}$ may or may not have the property $\Property$. 
In this paper, we assume all shadow graphs have the same size. Let $n_s$ be the number of nodes in the shadow graphs. Intuitively, to ensure the shadow models mimic the behaviors of the target model, they should be trained in the way that the output of each shadow model  ${\ShadowM_{i}}$ on the shadow training dataset ${\ShadowG_{i}}$ is close to the output of the target model $\TargetM$ on ${\ShadowG_{i}}$. In this paper, we follow the prior works \cite{ateniese2013hacking,ganju2018property} and consider the strongest attack scenario that the shadow models are identical to the target model, i.e.,  they have the same architecture and parameters. 

{\bf Attack model training phase.} 
Before training the GPIA classifier, the adversary constructs the attack training data $\Ptrain$ by the following procedure.  
 For each trained shadow model $\ShadowM_{i}$ and its training data $\ShadowG_i$, the adversary aggregates the set of posterior probability values generated by  $\ShadowM_{i}$ on $\ShadowG_i$ into a vector $\vec{v_i}$.  The vector $\vec{v_i}$ is inserted into the GPIA training dataset $\Ptrain$ as the features, which is associated with a GPIA label “1” if $\ShadowG_i$ is positive, and “0” otherwise. 
 
How to aggregate multiple posterior probability values into one vector as the GPIA features? We consider two different approaches:
\vspace{-0.2in}
\begin{ditemize}
    \item  {\em Concatenation:} Given $n_s$ nodes in the shadow graph, each associated with $\ell$ posterior probabilities, there are $N=n_s\times\ell$ posterior probability $p_1, \dots, p_N$ in total. These $N$ probability values are concatenated 
    into a vector $\vec{v}= <p_1, \dots, p_N>$ as GPIA features.  
    \item  {\em Element-wise difference} (EWD): 
For each node $v$, which is associated with $\ell$ posterior probability values $p_1, \dots, p_\ell$, we calculate the {\em average element-wise difference} $p^{\text{diff}}$ of $v$ as follows:
    \[p^{\text{diff}}=\frac{1}{\ell(\ell-1)}\sum_{1\leq i, j\leq \ell, i\neq j}(|p_i - p_j|).\] 
    
    Intuitively, the more (less, resp.) skewed posterior distribution, the higher (lower, resp.) $p^{\text{diff}}$ will be. In a uniform distribution of posteriors, $p^{\text{diff}}=0$.
    After the element-wise difference of each node is calculated, all values are concatenated into a vector $\vec{v}= <p_1^{\text{diff}}, \dots, p_{n_s}^{\text{diff}}>$ as GPIA features. 
     
    \end{ditemize}
    
    The intuition behind the EWD method is that the distribution of posterior output of positive and negative graphs is significantly different, where such difference can be captured by the element-wise difference value.


After $\Ptrain$ is generated, the adversary trains the GPIA classifier $\PIA$ on $\Ptrain$. In this paper, we consider three types of  classifiers, namely Multi-layer Perceptron (MLP), Random Forest (RF), and Linear Regression (LR). 

{\bf Property attack inference phase.} At inference time, the adversary computes the aggregated posterior probability output by the target model $\TargetM$ on the target graph $\TargetG$, using the same posterior aggregation function in the training phase. Then the adversary feeds the aggregated posterior as the input feature of the testing sample to $\PIA$, and obtains the predicted GPIA label. 
\subsection{White-box Attacks}
\label{sc:white-attack}

Unlike the black-box attacks that need shadow models, the white-box attacks do not need any shadow model due to its white-box access to the target model. Therefore, the white-box attacks only include two phases, namely, {\em attack model training} and {\em property attack inference}. Next, we discuss the details of these two phases. 

{\bf Attack model training phase.}  since the adversary has the white-box access to the target model $\TargetM$, he will construct the attack training data $\Ptrain$ by using the model parameters of $\TargetM$ as the features in $\Ptrain$. The motivation behind this is that the parameters of the models trained on the positive graphs will be more similar  than those trained on the negative graphs.
Following this, we design the following method to construct $\Ptrain$. 
For each shadow graph $\ShadowG_{i}$, the adversary uses it to train $\TargetM$ and obtains all the parameters of $\TargetM$, where the parameters are the node embeddings of $\ShadowG_{i}$. The shadow graph $\ShadowG_{i}$ can be randomly sampled from  the partial graph $\PartialG$ (Attack $A_1$), the shadow graph $\ShadowG$ (Attack $A_3)$, or both (Attack $A_5$). Then the adversary  aggregates these node embeddings into a vector $\vec{v_i}$, and inserts $\vec{v_i}$ into $\Ptrain$ as the features. 
He further associates $\vec{v_i}$ with a GPIA label “1” if $\ShadowG_i$ is positive, and “0” otherwise. 

In general, given a GNN model of $k$ layers, the adversary  can choose any $k'\leq k$ layers, and collect the node embeddings of these $k'$ layers to generate GPIA features. We use $A_i^{j_1, \dots, j_t}$ to indicate that the attack $A_i$ uses the model parameters (i.e., node embeddings) at the $j_1$-th, $\dots$, $j_t$-th layers of GNN. For example, $A_1^{2}$ indicates the attack $A_1$ that utilizes the embedding at the 2nd layer of the target model $\TargetM$, and $A_1^{1, 2}$ indicates the attack $A_1$ that utilizes the embeddings at both the 1st and 2nd  layers of $\TargetM$. 
 There are $2^{k}-1$ possible choices of choosing these $k'$ layers in total. 
 Besides these embeddings, the posterior probabilities also can be included to generate features in the same way as in the black-box setting (Section \ref{sc:black-box-ttack}). We will investigate the impact of choosing different amounts of node embeddings on GPIA performance later (Section \ref{sc:exp-pia}). 

Next, we discuss how to aggregate a set of node embeddings into one vector as the GPIA feature. 
We consider the following three aggregation methods in this paper. We adapt two pooling methods that have been widely used for Convolutional Neural Networks  \cite{boureau2011ask,weng1992cresceptron,lecun1998gradient}, namely {\em max-pooling} and {\em mean-pooling}. Both pooling methods take a set of network parameters in the format of vectors as the input, and summarize these vectors as a single vector of fixed length. Max-pooling preserves the most prominent features, while mean-pooling has a smoothing effect.
\begin{ditemize}
\vspace{-0.2in}
   \item {\em Concatenation}: Given $n_s$ node embeddings $z_1, \dots, z_{n_s}$ of the shadow graph, they are concatenated into a vector $\vec{v}=<z_1, \dots, z_{n_s}>$ of dimension $n_s$ as GPIA features.  
    \item {\em Max-pooling}: Given $n_s$ node embeddings $z_1, \dots, z_{n_s}$ from the shadow graph, we generate a vector $\vec{v}=<z^{max}_1, \dots, z^{max}_{n_s}>$ as the GPIA features, where $z^{max}_i$ is the maximum of all value in the embedding $z_i$. 
    \item {\em Mean-pooling}: Given $n_s$ node embeddings $z_1, \dots, z_{n_s}$, we generate a vector $\vec{v}=<z^{mean}_1, \dots, z^{mean}_{n_s}>$, where $z^{mean}_i$ is the mean of all values in $z_i$. 
\end{ditemize}


Different aggregation methods generate different GPIA features, and thus lead to different attack performance. We will investigate the impact of different embedding aggregation methods on GPIA performance in Section \ref{sc:exp-pia}. 

{\bf Property attack inference phase.} At inference time, the adversary collects the model parameters of the target model $\TargetM$ trained on the target graph $\TargetG$, and aggregates the parameters into a vector as the input feature of the testing sample,  by using the same embedding aggregation function in the training phase. Therefore, the feature $\Ptest$ is a vector whose size is the same as the number of nodes in $\TargetG$, whereas the feature of $\Ptrain$ is a vector whose size is the same as the number of nodes in $\ShadowG$. Since $\TargetG$ and $\ShadowG$ may have different number of nodes, the feature of $\TargetG$ and $\ShadowG$ can be of different sizes. This raises the challenge of how to predict on $\Ptest$ if its feature is not of the same size as that of $\Ptrain$.  

To address this challenge, we consider four different methods to align the features of $\Ptrain$ and $\Ptest$ to be of same dimensions: (1) \underline{Sampling}: the most straightforward approach is to ensure that $\TargetG$ and $\ShadowG$ have the same number of nodes. This can be achieved as the adversary  can obtain the knowledge of the number of nodes in $\TargetG$ by counting the number of node embeddings via its white-box access to the target model. Then the adversary samples the same number of nodes from $\ShadowG$. This method is applicable when the number of nodes in $\ShadowG$ is no less than that of $\TargetG$; (2) \underline{TSNE projection} \cite{van2008visualizing}: it projects high-dimensional data to either two or three-dimensional data. We apply TSNE on the features of $\Ptrain$ and $\Ptest$ to project them into the same two-dimensional space, regardless of their original dimensions; (3) \underline{PCA dimension reduction} \cite{minka2000automatic}: We apply PCA, a widely-used dimension reduction method in the literature, on the feature vector of 
    both $\Ptrain$ and $\Ptest$ and project them into a space of the same dimension; (4) \underline{Autoencoder dimension compression}: Autoencoder \cite{hinton2006reducing} compresses the dimensions in the way that the data in the high-dimensional space can be reconstructed from the representation of lower dimension with small error.  We apply Autoencoder on the features of 
    $\Ptrain$ and $\Ptest$ to compress both into the same space of a lower dimension. 
    To reduce the amounts of information loss by compression, we only compress the feature vector of the larger dimension into the space of the feature vector of smaller one. 

Different alignments methods incur different amounts of information loss on the resulting embeddings, and thus lead to different GPIA performance. We will investigate the impact of different alignment methods on GPIA performance in Section \ref{sc:exp-pia}.

\nop{
{\bf Meta-classifiers \cite{ateniese2013hacking,ganju2018property} versus ours.} The training data $\Ptrain$ of six GPIA classifiers have different feature dimension sizes. We summarize the  number of samples and features of $\Ptrain$  for these attacks in Table \ref{tab:attack-dimension}. If we follow the meta-classifier method \cite{ateniese2013hacking,ganju2018property}, all the parameters of the shadow model can be considered as the features of $\Ptrain$. If we only consider node embeddings as the parameters of GNN models, the feature dimension of $\Ptrain$ for the meta-classifier is the same as the number of nodes in the shadow graph, which can be thousands or  more. 
Training of such meta-classifier is expected to suffer from {\em curse of dimensionality} \cite{}. 
On the contrary, our attacks only have one feature, which contains the information aggregated from all node embeddings in the shadow graph. We will show that the aggregated information  still enables effective PIA (Section \ref{sc:exp-pia}). We note that, although our GPIA features may be of large size, the alignment methods  will reduce the feature length. 
}

\vspace{-0.1in}
\section{Evaluation}
\label{sc:exp-pia}

In this section, we aim to demonstrate the effectiveness of GPIA through answering the following three research questions:
\begin{ditemize}
    \item \textbf{RQ1} - How effective is GPIA on representative GNN models and real-world graph datasets?
\item{\bf RQ2} - Why GPIA work? 
\item{\bf RQ3} - How various factors (e.g., attack classifier models, embedding/posterior aggregation methods, and complexity of GNN models) affect GPIA effectiveness?
\end{ditemize}

\begin{table}[t!]
    \centering
    \begin{tabular}{c|c|c|c|c}
    \hline
         \textbf{Dataset} & \textbf{\# nodes}  & \textbf{\# edges} & \textbf{\# features}  & \textbf{\#   classes}\\\hline
        \textbf{Pokec} &45,036 & 170,964 & 5 & 2\\\hline
        \textbf{Facebook} &4,309 & 88,234 &1,284 & 2 \\ \hline 
        \textbf{Pubmed}  &19,717 & 44,338& 500 & 3\\\hline
    \end{tabular}
    \caption{Description of datasets}
    \label{tab:data}
    \vspace{-0.3in}
\end{table}
\nop{
\begin{table*}[t!]
    \centering
    \begin{tabular}{|c|c|c|c|c|c|}\hline
        Graph & \# of   nodes & \# of   Edges &\# of   Node features & Classification task / \# of   Classes& Properties to be attacked  / \# of   Classes\\\hline
         Pokec &45036 &170964 & 6&public / 2 &portion of male / 2\\\hline
         Pokec &45036 &170964 & 6&public / 2 &portion of (age > 20) / 2\\\hline
         Pokec &45036 &170964 & 6&gender / 2 &portion of (weight > 60KG) / 2\\\hline
         Facebook &4039 &88234 & 1284 &gender / 2 &portion of (educatiion type $\ge$ college) / 2\\\hline
         Pubmed &19717 &44338 & 500&paper type / 3 &portion of (appearance of word 'insulin') / 2\\\hline
    \end{tabular}
    \caption{Datasets and their properties to be attacked}
    \label{tab:data}
\end{table*}
}
\begin{table*}[t!]
    \centering
    \begin{tabular}{c|c|c|c|c}\hline
        \multirow{2}{*}{\bf Property} & \multirow{2}{*}{\bf Type} &\multirow{2}{*}{\bf Graph} & {\bf Property}  & \multirow{2}{*}{\bf Description} \\
         & & & {\bf feature} & \\\hline
        $P_1$ & \multirow{3}{*}{Node} & Pokec &  Gender &  COUNT(Male)$>$ COUNT(Female)\\\cline{1-1}\cline{3-5}
         $P_2 $ & & Facebook & Gender& COUNT(Male)$>$ COUNT(Female) \\\cline{1-1}\cline{3-5}
         $P_3 $ & & Pubmed & Keyword &  COUNT(publications with “IS”) $>$ COUNT(publications without “IS”)\\\hline\hline
         $P_4 $ &\multirow{3}{*}{Link} & Pokec & Gender & COUNT(same-gender links) $>$ COUNT(diff-gender links) \\
         \cline{1-1}\cline{3-5}
         $P_5 $ & & Facebook & Gender & COUNT(same-gender links) $>$ COUNT(diff-gender links)\\\cline{1-1}\cline{3-5}
        $P_6$ & & Pubmed & Keyword  & COUNT(links btw. papers with “IS”) $>$ COUNT(links btw. papers with “ST”) \\\hline
    \end{tabular}
    \caption{Properties to be attacked by GPIA. same-gender (diff-gender, resp.) links indicate those links between users of the same (different, resp.) gender. “IS” = “Insulin”; “ST” = “Streptozotocin”. }
    \label{tab:property}
    \vspace{-0.25in}
\end{table*}

\subsection{Experimental Setup}
\label{sc:exp}

All the experiments are executed on Google COlab with Tesla P100 (16G) and 200GB memory. All the algorithms are implemented in Python with PyTorch. Our code and datasets are available online\footnote{https://anonymous.4open.science/r/PIA-CE14/}.

{\bf Datasets.}
We consider three real-world datasets, namely {\em Pokec}, {\em Facebook}, and {\em Pubmed} datasets, that are popularly used for graph learning in the literature: (1) {\bf Pokec} social network graph\footnote{https://snap.stanford.edu/data/soc-pokec.html} is collected from the most popular on-line social network in Slovakia;  
(2) {\bf Facebook} social network graph\footnote{https://snap.stanford.edu/data/ego-Facebook.html}  consists of Facebook users as nodes and their friendship relationship as edges;  
and (3) {\bf Pubmed} Diabetes dataset\footnote{https://linqs-data.soe.ucsc.edu/public/Pubmed-Diabetes}  consists of scientific publications from Pubmed database that are classified into three classes. Each publication node is associated with 500 unique keywords as the features. The links between publications indicate the citation relationship. 
Table \ref{tab:data} summarizes the information of the three datasets. More details of the three datasets can be found in Appendix \ref{appendix:dataset}. 
The purpose of pick two graphs in one domain (social network graphs) and one graph from a different domain is for the validation of the effectiveness of transfer attacks ($A_3$ and $A_4$). 

{\bf Target GNN models.} 
We consider three state-of-the-art GNN models, namely GCN \cite{kipf2017semisupervised}\footnote{We use implementation of GCN at https://github.com/tkipf/pygcn}, GraphSAGE \cite{hamilton2018inductive} and GAT \cite{velickovic2018graph}\footnote{We use the implementation of both GraphSAGE and GAT from DGL package available at https://github.com/dmlc/dgl}, that are widely used by the ML  community. For each hidden layer, the number of neurons is 64, which is the same as the dimension of node embedding. We set the number of epoches for training as 1,500, and use early stop with the tolerance as 50. We set the dimension of node embeddings to 64 for all the three datasets. 

{\bf Properties and property groups.}  For each dataset, we design one node property and one link property to be attacked. The properties  are summarized in Table \ref{tab:property}. 
We pick the keywords “Insulin” (IS) and “streptozotocin” (ST) for Pubmed dataset as they are the  keywords of the highest and lowest TF-IDF weight respectively. The successful attacks on these properties can reveal the gender distribution  in Facebook and Pokec social network graphs, and the frequency distribution of particular keywords (which can be sensitive) in Pubmed graph. We also measure the size of the property groups, and show the results in Appendix \ref{appendix:group-ratio}. 

{\bf Implementation of attack classifier.} We use three types of attack classifiers for both attacks, namely Multi-layer Perceptron (MLP), Random Forest (RF), and Linear Regression (LR). 
We use the implementation of the three classifiers provided by sklearn package.\footnote{https://scikit-learn.org/}
We set up the MLP classifier of three hidden layers, with the number of neurons for each layer as 64, 32, 16 respectively. We use ReLU as the activation function for the hidden layers and Sigmoid for the output layer. We train 1,000 epochs with a learning rate of 0.001.  We use cross-entropy loss as the loss function and Adam optimizer.
For RF classifier, we set the maximum depth as 150 and the minimum number of data points allowed in a leaf node as 1. 
For LR classifier, we use the L2 norm as the penalty term, and liblinear\footnote{Liblinear libary: https://www.csie.ntu.edu.tw/~cjlin/liblinear/} as the optimization solver. We set the maximum number of iterations as 100 and the early-stop tolerance as 1e-4. 

{\bf Partial graphs.} We randomly sample 1,000 subgraphs from each dataset as the partial graph. The size of each partial graph is 20\%, 25\%, and 30\% of Pokec, Facebook, and Pubmed datasets respectively. 

{\bf GPIA training and testing data.} For $A_1 \& A_2$, we randomly sample 1,000 subgraphs from the same dataset to generate the training and testing data for GPIA. Each subgraph is of the same size as the partial graph. The training/testing split is 0.7/0.3, with the same number of positive and negative subgraphs in both training and testing data. There is no overlap of either links or subgraphs between training and testing data. However, it is challenging to enforce no node overlapping between training and testing data, especially for the datasets with a small number of nodes (e.g., Facebook dataset), as a large portion of sampled subgraphs in the training data will have highly similar structure. Therefore, we allow  a small amounts of node overlap between training and testing data (3\%, 5\%, and 4\% for Pokec, Facebook, and Pubmed dataset respectively).  We will show the impact of node non-overlapping between training and testing data on attack accuracy in Section \ref{sc:att-settup}.  For attacks $A_3 \& A_4$, we sample 700 subgraphs from the shadow graph as the GPIA training data, and 300 subgraphs from the target graph as the  testing data. There is no node/link overlap between training and testing data for $A_3$ and $A_4$. For attacks $A_5 \& A_6$, we sample some subgraphs from the partial graph plus some subgraphs from the shadow graph (700 in total) as the training data, and 300 subgraphs from the target graph as the testing data. We consider various size ratios (1:10, 1:4, 1:2, 1:1, 2:1, 4:1, and 10:1) between partial and shadow graphs in the training data. Similar to $A_1$ \& $A_2$, there is no link overlap but a small node overlap between training and testing data for $A_5$ and $A_6$, where the node overlap ratio does not exceed 5\%.

{\bf Metrics.}  
We measure classification accuracy as the GNN model performance. We measure {\em attack accuracy} $AC$ as the effectiveness of the proposed attacks.
In particular, $AC = \frac{N_c}{N}$, 
where $N_c$ is the number of graphs that are correctly predicted by GPIA (either as positive or negative), and $N$ is the total number of graphs in the testing data. Higher AC indicates that GPIA is more effective.

{\bf Baselines.} 
We consider three approaches as baselines for comparison with our GPIA model: (1) {\bf Attribute inference attack (AIA)  (Baseline-1)}: we follow \cite{song2020information} and design an AIA that predicts the values of property features by the access to the embeddings/posteriors. Then we evaluate PIA accuracy based on the predicted values of property features. To ensure fair comparison between AIA and GPIA, we consider the same partial graphs in the adversary knowledge of GPIA for AIA.  (2) {\bf K-means clustering (Baseline-2)}: we apply k-means clustering ($k=2$) on node embeddings and posteriors. Then we measure the average distance between the centroid of each cluster to the embedding/posteriors, and pick the cluster of the smaller distance; (3)  {\bf Meta-classifier (Baseline-3)}: We follow \cite{ateniese2013hacking,ganju2018property,zhang2021leakage}  and use a meta-classifier as the GPIA  classifier. We also have two additional threshold-based baseline methods. More details of this method and its comparison with ours can be found in Appendix \ref{appendix:more-baseline}.
\vspace{-0.12in}
\begin{figure*}[t!]
    \centering
\begin{subfigure}[b]{.8\textwidth}
      \centering
    \includegraphics[width=\textwidth]{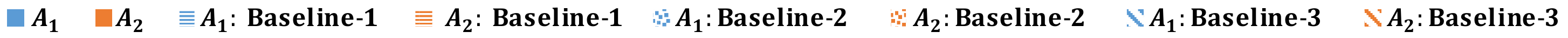}
     \vspace{-0.2in}
    \end{subfigure}\\
\begin{tabular}{ccc}
    \begin{subfigure}[b]{.32\textwidth}
      \centering
    \includegraphics[width=\textwidth]{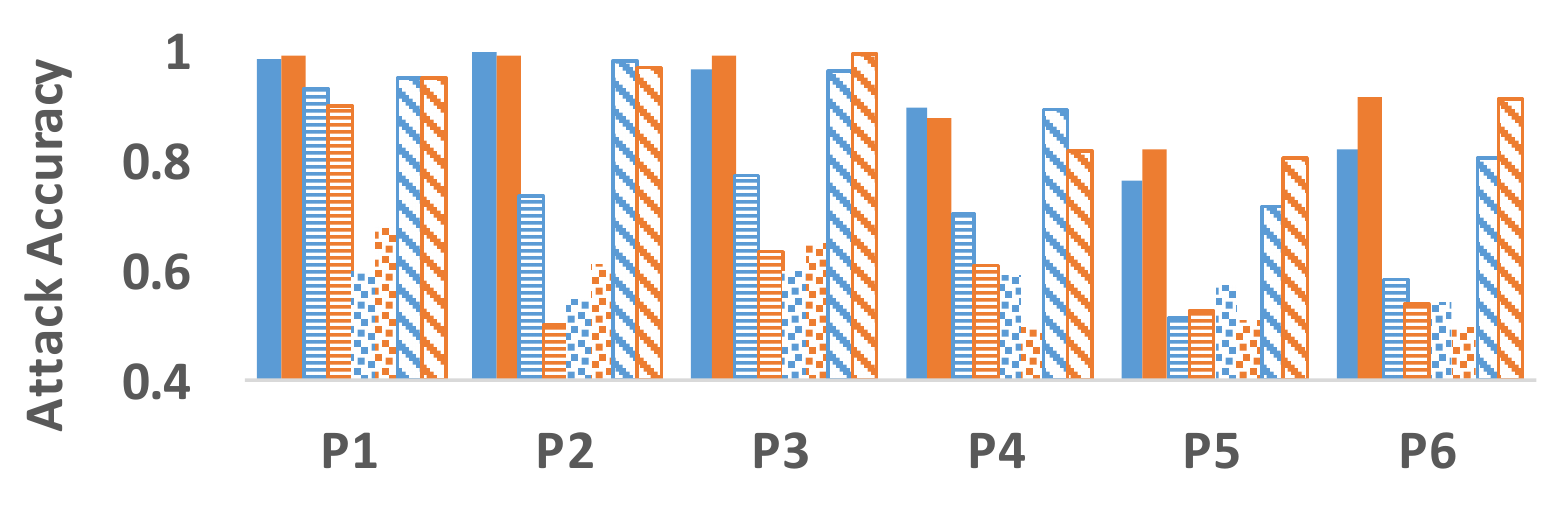}
     \vspace{-0.2in}
    \caption{GCN}
    \end{subfigure}
    &
    \begin{subfigure}[b]{.32\textwidth}
    \centering
    \includegraphics[width=\textwidth]{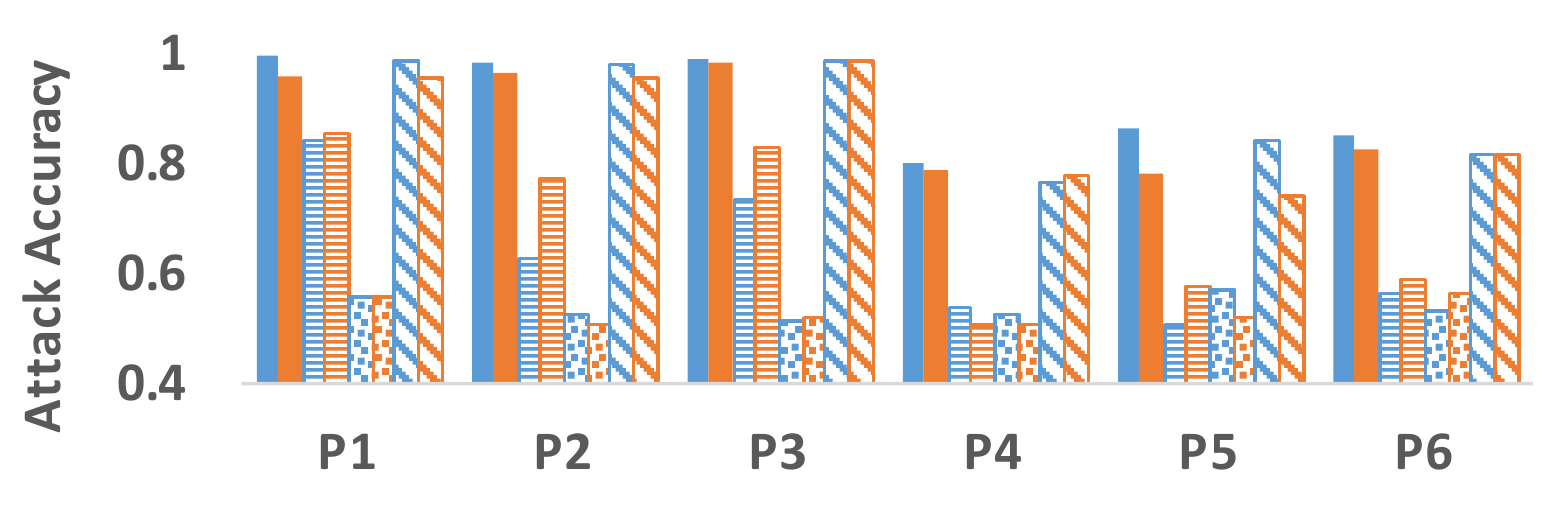}
    \vspace{-0.2in}
    \caption{GraphSAGE}
    \end{subfigure}
    &
    \begin{subfigure}[b]{.32\textwidth} 
    \centering
    \includegraphics[width=\textwidth]{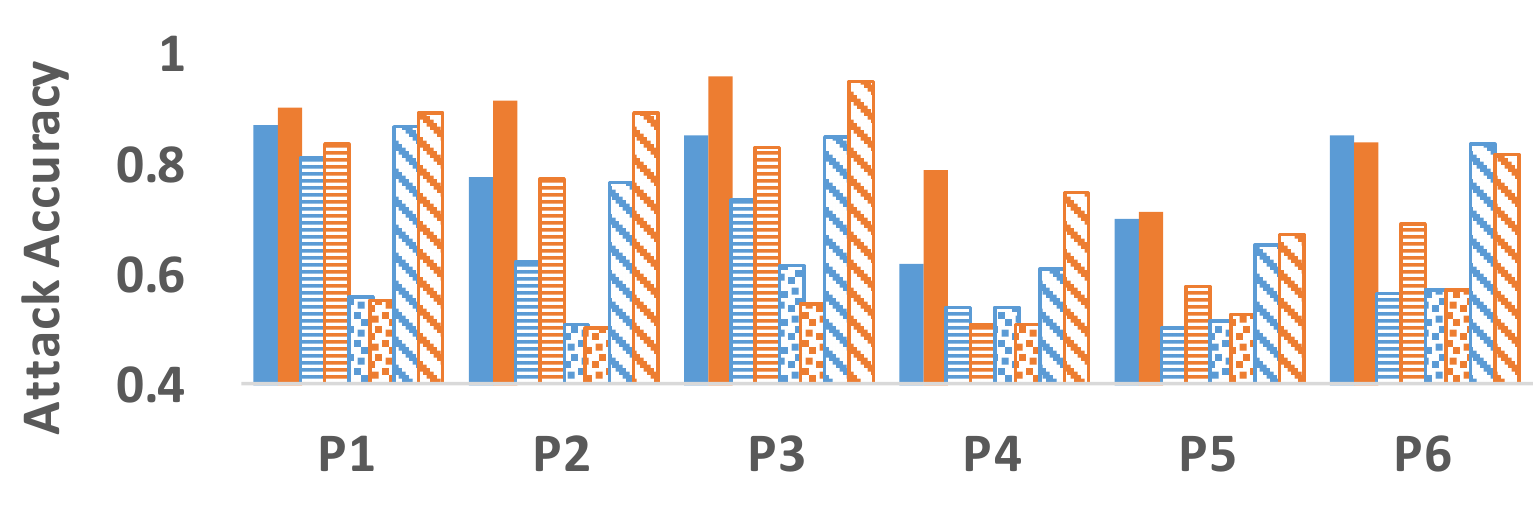}
    \vspace{-0.2in}
    \caption{GAT}
    \end{subfigure}
\end{tabular}    
    \vspace{-0.2in}
\caption{\label{fig:pia_acc} Attack accuracy of $A_1$ and $A_2$. $A_1$ and $A_2$ are indicated in different colors respectively, while our approaches, Baseline-1, Baseline-2, and Baseline-3 are indicated in solid fill, horizontal stripe fill, sphere fill, and diagonal shape fill respectively. }
\vspace{-0.05in}
\end{figure*}

\subsection{GPIA Performance (RQ1)}
We launch the attacks $A_1$ - $A_6$ to attack GCN, GAT, and GraphSAGE, and measure their accuracy. To have a fair comparison of attack accuracy across different settings, we ensure that GPIA training data is of the same size for all the attacks.  We use RF and MLP as the white-box and black-box attack classifiers, max-pooling as the embedding aggregation method, concatenation as the posterior aggregation method, and TSNE projection as the embedding alignment method, as these setups produce the best attack performance. More results of the attack performance under different setups can be found in Section \ref{sc:att-settup}. As the possible hidden layers that the attacker collects node embeddings from is exponential to the number of hidden layers, we only consider GNNs of two hidden layers in this part of experiments to ease explanation. 

{\bf Performance of target GNN models.} Before we evaluate GPIA effectiveness, we evaluate the performance of the three GNN models, aiming to justify why they are worthy to be attacked. The performance of the three models can be found in Appendix \ref{appendix:gnn-performance}. First, all the three models perform well, with their classification accuracy significantly higher than the random guess. Therefore, these models are ready for the launch of GPIA. Second, the three models have good generalizability as the training-testing gap is in a small range of [0.02, 0.08]. Thus the three models do not have overfitting. 

{\bf Attacks $A_1$ and $A_2$.} 
For $A_1$, we consider three variants whose GPIA features are generated from  the embeddings at the first layer ($A_1^1$),  the second layer ($A_1^2$), and both layers ($A_1^{1,2}$). 
We measured the performance of these variants and reported the best GPIA performance. More details of how different amounts of embeddings collected from different layers affect GPIA performance will be discussed in Section \ref{sc:att-settup}.

Figure \ref{fig:pia_acc} shows the attack accuracy of our proposed attacks and the baselines. 
First, we observe that the attack accuracy of our $A_1$ and $A_2$ attacks ranges in [0.62, 1], which is significantly higher than 0.5 (random guess). In some settings (e.g., Figure \ref{fig:pia_acc} (a)), the attack accuracy can be as high as close to 1, even under the black-box setting. This demonstrates the effectiveness of GPIA against these target models. 
Furthermore, both $A_1$ and $A_2$ outperform  Baseline-1 in all the settings. Although the superiority of $A_1$ and $A_2$ to Baseline-1 is marginal on $P_1$ for GCN and GAT (Figure \ref{fig:pia_acc} (a) \& (c)), it is significant for the rest of settings. We believe AIA is much less effective than GPIA for property inference is because there is no strong correlation between the property feature and the label in all the three graphs. More details of the correlation between the property feature and the label can be found in Appendix \ref{appendix:label-correlation}. 
Similarly, we observe that the accuracy of $A_1$ and $A_2$ is also much higher than Baseline-2. We also observe that the performance of $A_1$ and $A_2$ is similar to Baseline-3. Thus using either one classifier or stacking multiple classifiers into a meta-classifier does not impact the GPIA performance. 

Second, although the attack performance varies across different types of properties, the attack accuracy of GPIA against the three node properties ($P_1$ - $P_3$) is noticeably higher than the link properties ($P_4$ - $P_6$) in general. The only exception is for the property $P_6$ and GAT as the target model (Figure \ref{fig:pia_acc} (c)), where the attack accuracy of $P_6$ is close to that of $P_3$, and higher than $P_1$ and $P_2$. One possible reason that GPIA is more successful against the node properties than the link properties is that it only needs to infer the node feature distribution over node features, but it has to infer both node feature distribution and graph structure for the link properties.  

Third, we observe that the accuracy of the white-box and black-box attacks is very close. The difference between them is negligible in most of the cases. 
Interestingly, the white-box attack does not always outperform the black-box attack,  even though its features may include those features used by the black-box attack, possibly due to overfitting of the GPIA classifier by including more features. 
This demonstrates the power of the property inference - the black-box access to the target model is sufficient to launch the attack. 


\begin{figure*}[t!]
    \centering
\begin{tabular}{cc}
  {\bf Node group properties}
  &
  {\bf Link group properties}
  \\
\begin{tabular}{ccc}
    \begin{subfigure}[b]{.135\textwidth}
      \centering
    \includegraphics[width=\textwidth]{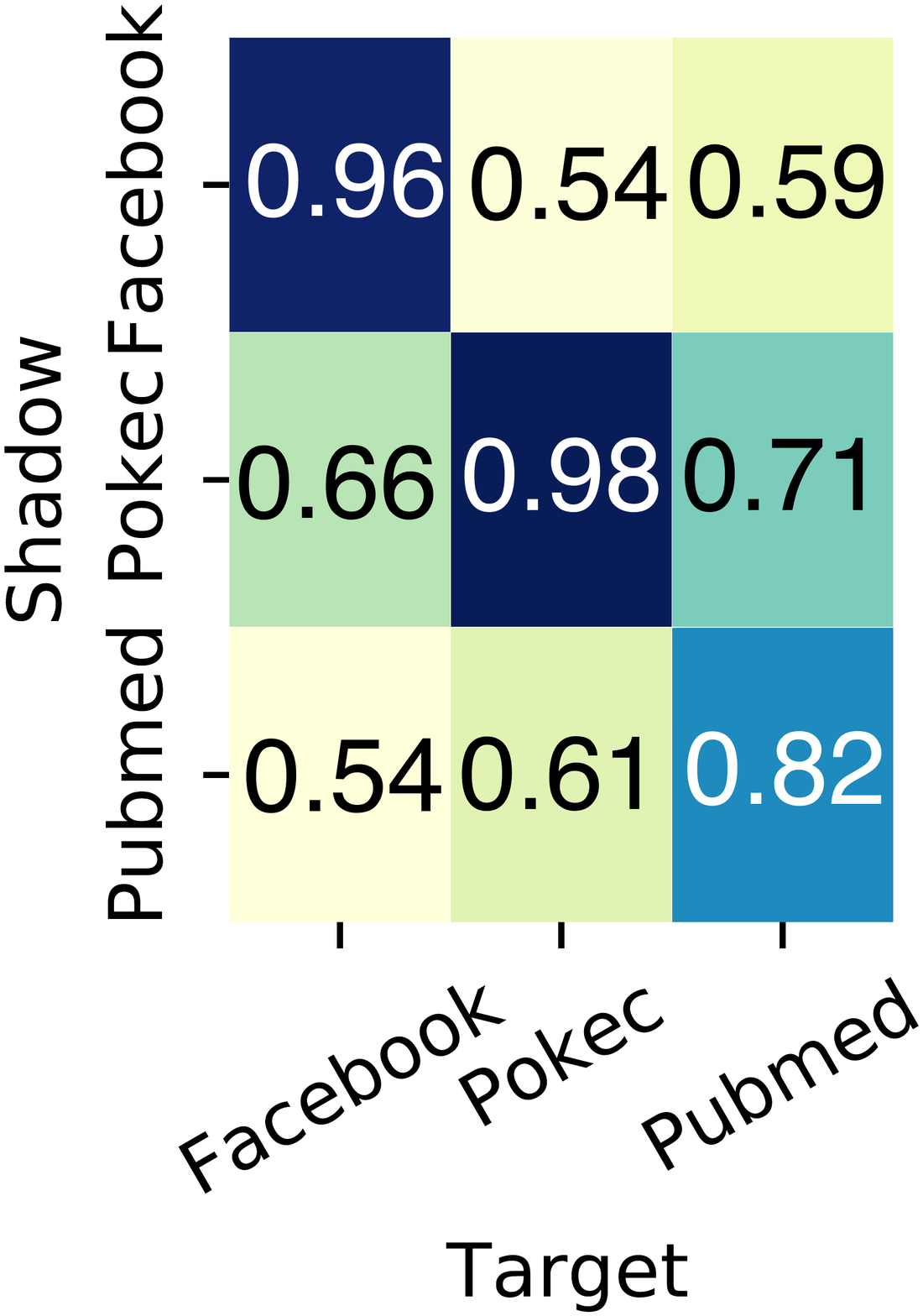}
     \vspace{-0.2in}
    \caption{$A_3^1$ attack}
    \end{subfigure}
    &
    \begin{subfigure}[b]{.135\textwidth}
    \centering
    \includegraphics[width=\textwidth]{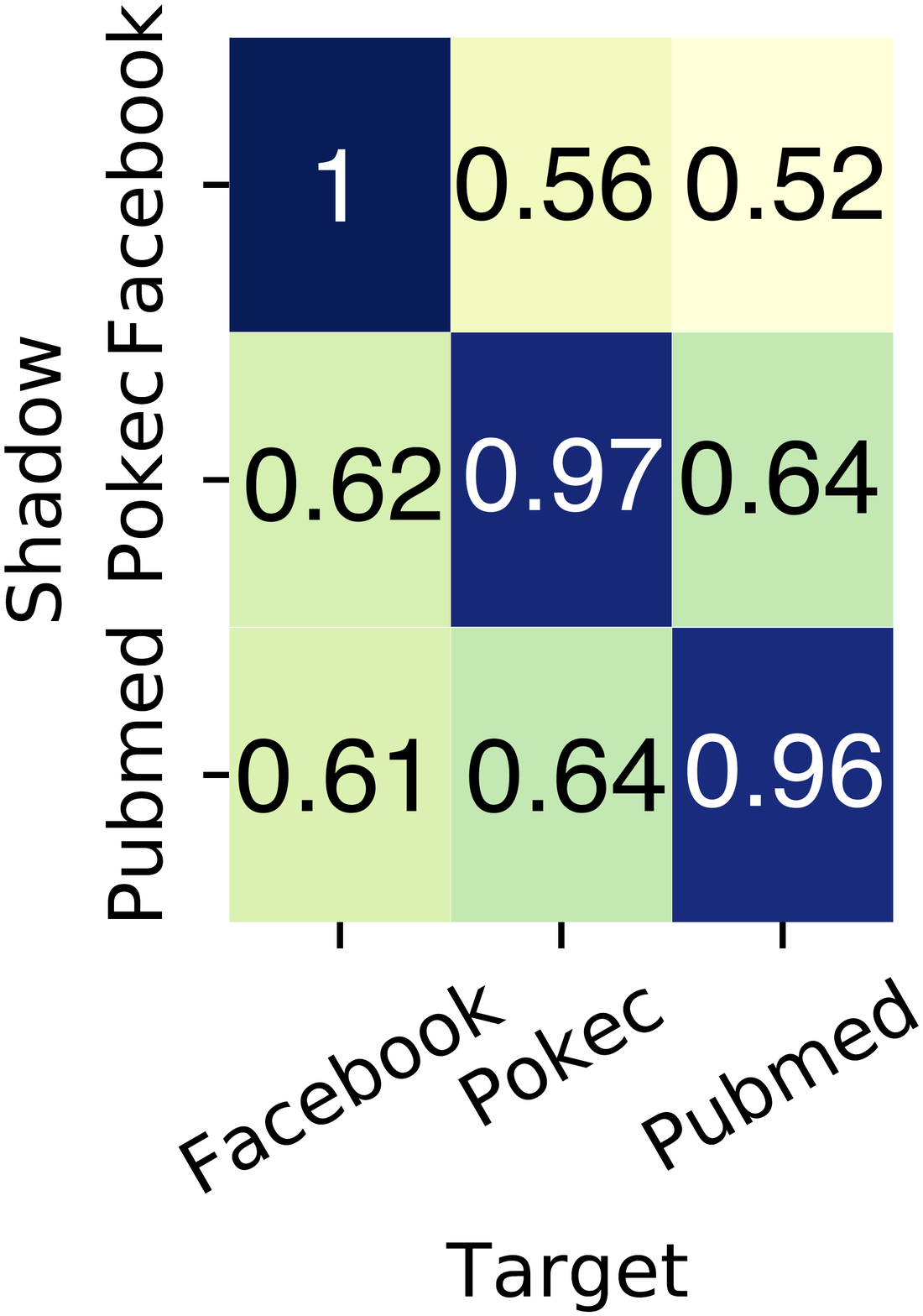}
    \vspace{-0.2in}
    \caption{$A_3^2$ attack}
    \end{subfigure}
    &
    \begin{subfigure}[b]{.135\textwidth} 
    \centering
    \includegraphics[width=\textwidth]{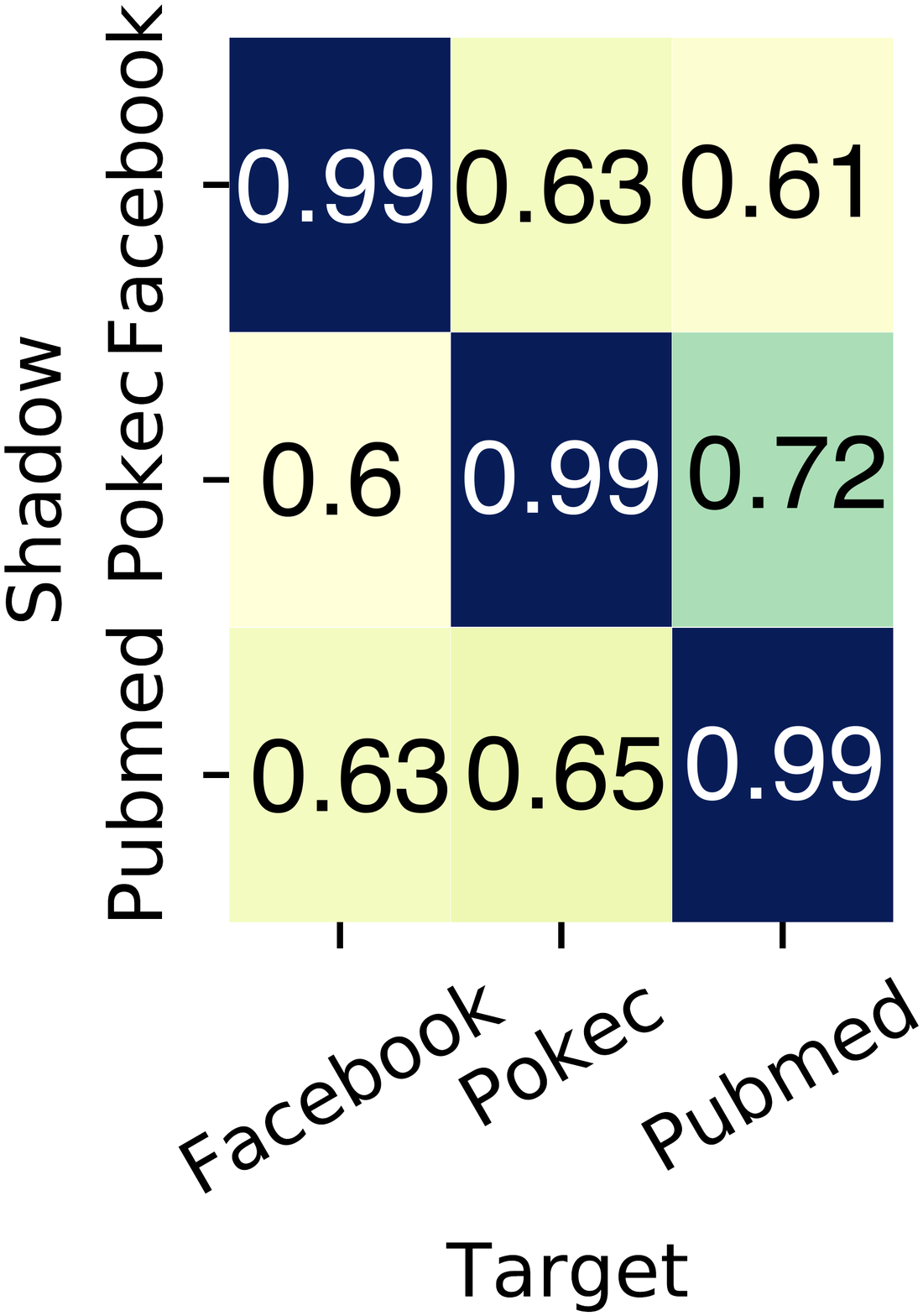}
    \vspace{-0.2in}
    \caption{$A_4$ attack}
    \end{subfigure}
\end{tabular}
&
\begin{tabular}{ccc}
        \begin{subfigure}[b]{.135\textwidth}
      \centering
    \includegraphics[width=\textwidth]{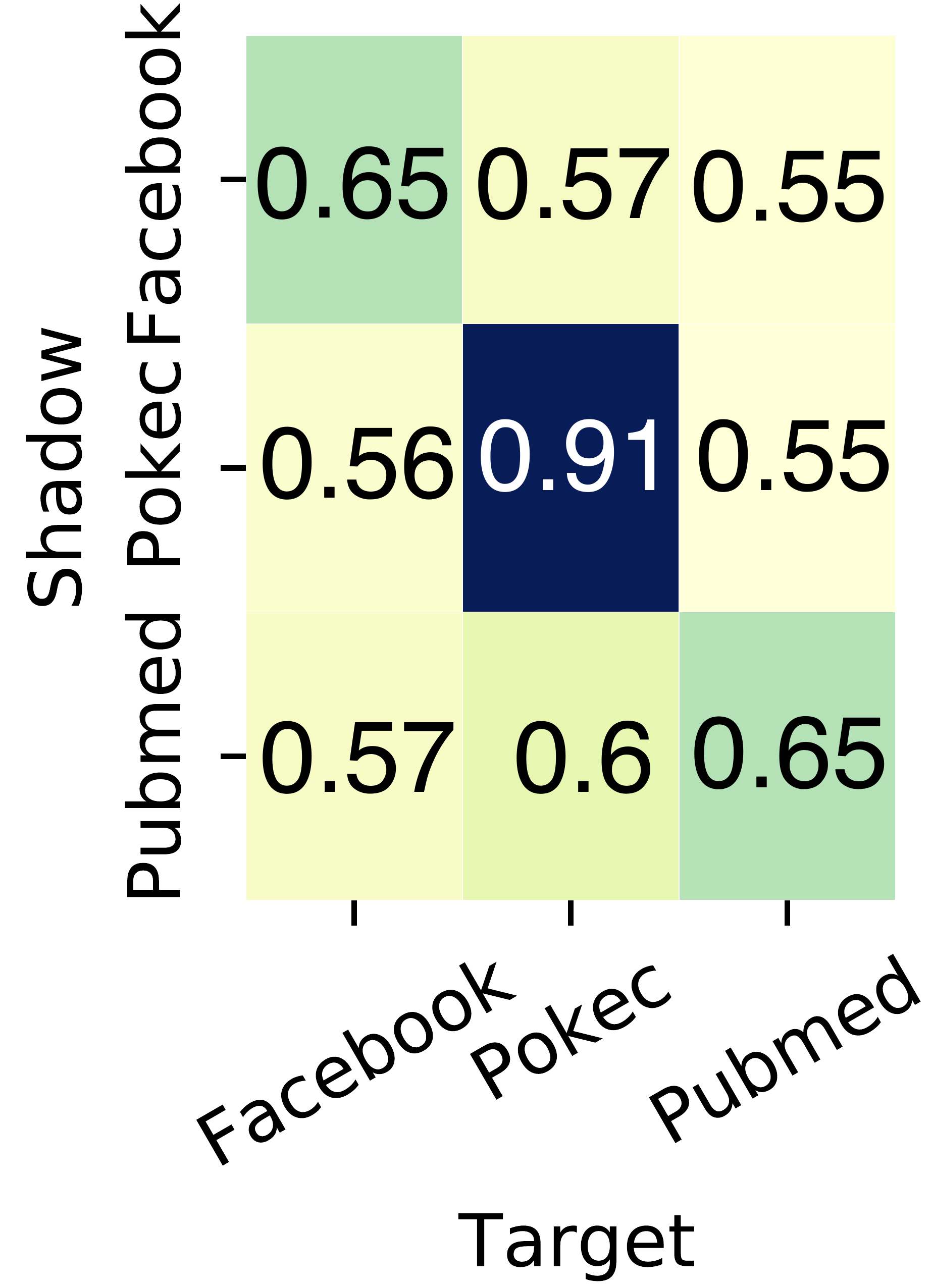}
     \vspace{-0.2in}
    \caption{$A_3^1$ attack}
    \end{subfigure}
    &
    \begin{subfigure}[b]{.135\textwidth}
    \centering
    \includegraphics[width=\textwidth]{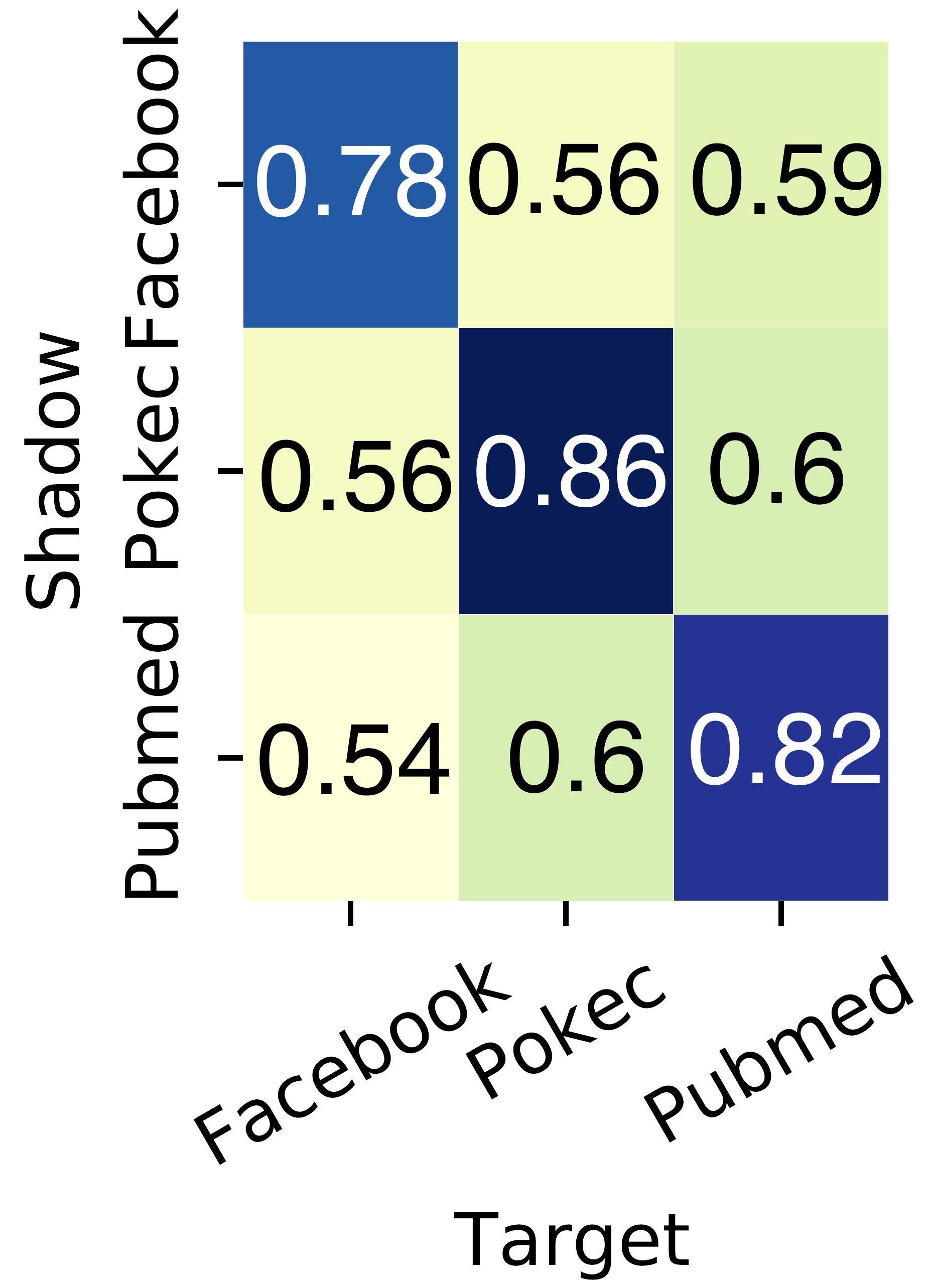}
    \vspace{-0.2in}
    \caption{$A_3^2$ attack}
    \end{subfigure}
    &
    \begin{subfigure}[b]{.135\textwidth} 
    \centering
    \includegraphics[width=\textwidth]{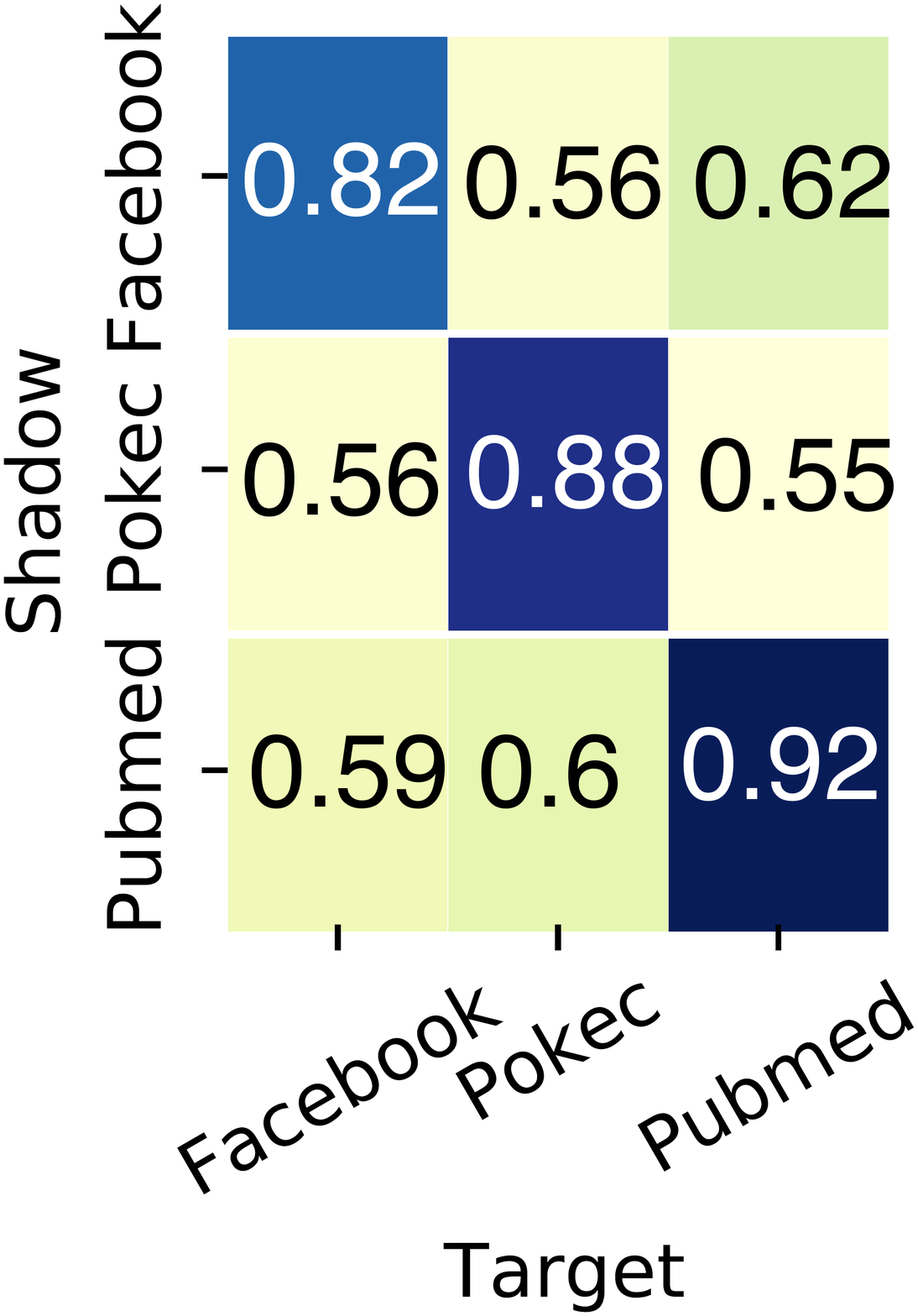}
    \caption{$A_4$ attack}
    \end{subfigure}
  \end{tabular}
\end{tabular}    
    \vspace{-0.15in}
\caption{\label{fig:attack34-acc} Attack accuracy of $A_3$ and $A_4$ when the GCN model is the target model. $A_3^1$ and $A_3^2$ indicate the $A_3$ attack that uses the model parameters at Layer 1 and Layer 2 of GCN models respectively.} 
\end{figure*}
\begin{figure*}[t!]
    \centering
\begin{subfigure}[b]{.55\textwidth}
      \centering
    \includegraphics[width=\textwidth]{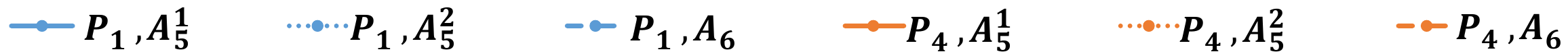}\\
    \end{subfigure}
\begin{tabular}{ccc}
    \begin{subfigure}[b]{.28\textwidth}
      \centering
    \includegraphics[width=\textwidth]{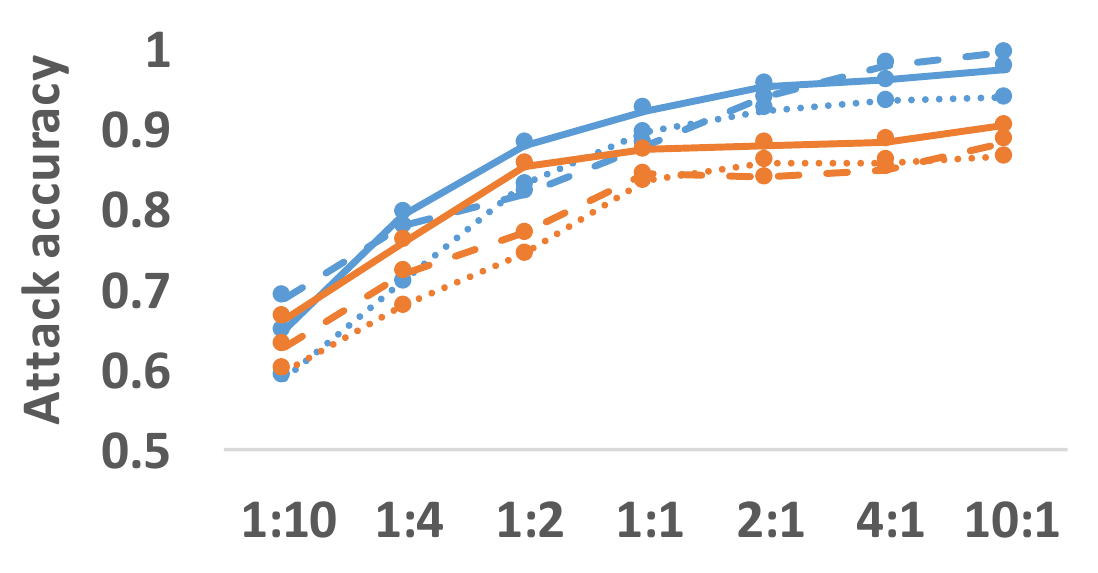}
     \vspace{-0.2in}
    \caption{GCN}
    \end{subfigure}
    &
    \begin{subfigure}[b]{.28\textwidth}
    \centering
    \includegraphics[width=\textwidth]{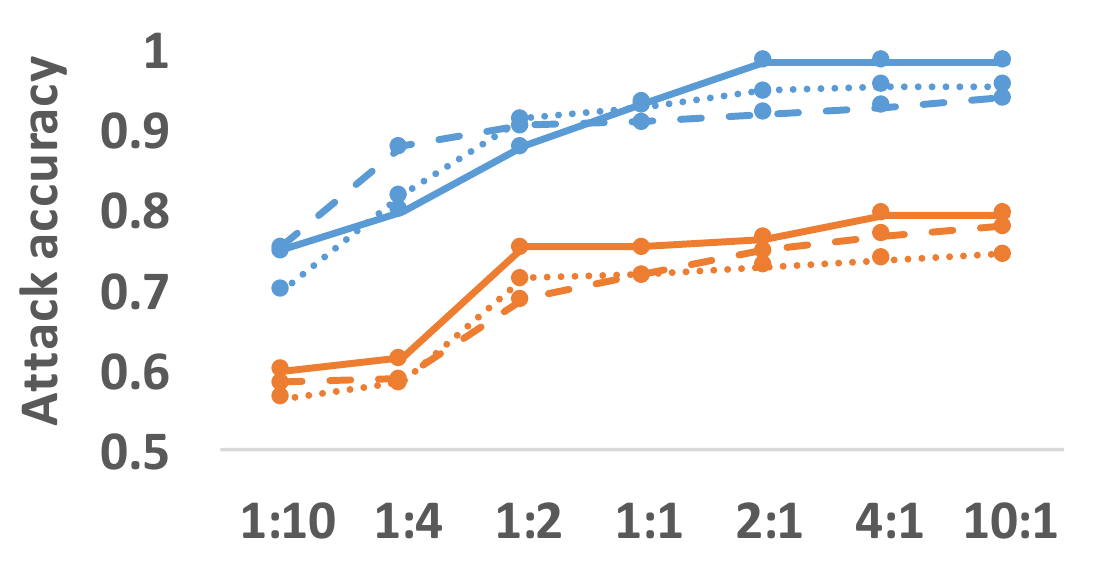}
    \vspace{-0.2in}    \caption{GraphSAGE}
    \end{subfigure}
    &
    \begin{subfigure}[b]{.28\textwidth} 
    \centering
    \includegraphics[width=\textwidth]{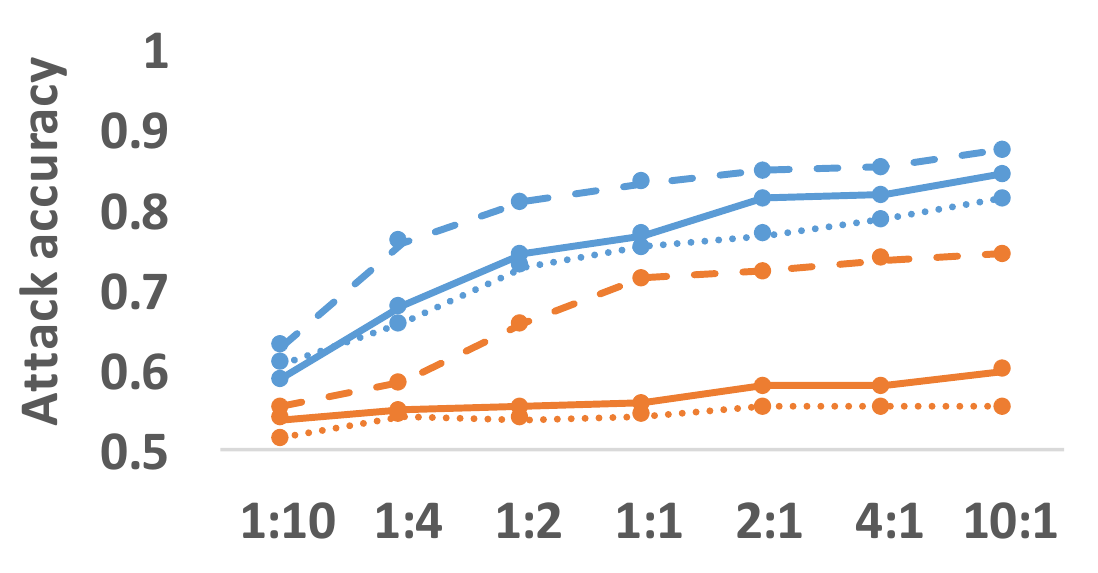}
    \vspace{-0.2in}
    \caption{GAT}
    \end{subfigure}
\end{tabular}    
    \vspace{-0.15in}
\caption{\label{fig:attack56-acca-pokec-fb} Attack accuracy of $A_5$ and $A_6$ against the properties $P_1$ and $P_4$. The partial and shadow graphs are sampled from Pokec and Facebook datasets respectively. X-axis shows the size ratio between partial and shadow graphs.} 
\end{figure*}

{\bf Attacks $A_3$ and $A_4$.} Figure \ref{fig:attack34-acc} presents the results of $A_3$ and $A_4$. The results of GAT and GraphSAGE models are included in Appendix \ref{appendix:attack34-gs-gat}. In all the settings,  $A_3$ and $A_4$ are effective as their accuracy is higher than 0.5 (random guess). The attack accuracy can be as high as 0.66 when the adversary uses the Facebook social network graph as the shadow graph to infer whether the Pokec social network graph has disproportionate distribution between male and female users. In other words, GPIA can transfer the knowledge learned from a graph to infer the properties of another graph. 
However, the accuracy of both $A_3$ and $A_4$ is worse than $A_1$ and $A_2$. The reason behind this that some amounts of information of the properties embedded in node embeddings/posteriors is  lost due to the feature alignment methods used for $A_3$/$A_4$. 
We also observe that GPIA sometimes performs better under the settings that shadow and target datasets belong to different domains than the settings where they belong to the same domain.  For example, GPIA accuracy can be as high as 0.72 when Pubmed and Pokec datasets are the target and shadow datasets respectively (Figure \ref{fig:attack34-acc} (c)), but it is only 0.6 when the target dataset is changed to Facebook dataset while keeping Pokec dataset as the shadow dataset, although both Pokec and Facebook datasets are social network graphs. 
We analyze the reason behind this observation, and found that the distribution of the GPIA attack features for positive and negative graphs in the shadow graph can be similar to that of the target graph even though they are from different domains and/or of different structure. For example, the distribution of the attack features over positive and negative graphs of Pokec dataset is more similar to   Pubmed dataset   than Facebook dataset (Appendix  \ref{appendix:distribution}). Thus the attacker can transfer such knowledge learned from the shadow graph for property inference on the target graph successfully.

{\bf Attacks $A_5$ and $A_6$.} We vary the portions of $\ShadowG$ and $\PartialG$ in the adversary knowledge when we evaluate the effectiveness of $A_5$ and $A_6$. 
Figure \ref{fig:attack56-acca-pokec-fb} present the attack accuracy result of both properties $P_1$ (node property) and $P_4$ (link property) with the partial graph  sampled from Pokec dataset and the shadow graph sampled from Facebook dataset. Note that both Pokec and Facebook are social network graphs. 
The performance of other settings are shown in Appendix \ref{appendix:attack56-2}. 
We have the following observations. First,  the accuracy of $A_5$ and $A_6$ in all the settings is higher than 0.5, i.e., both attacks are effective. Furthermore, the attack accuracy increases as the size of the partial graph grows. In particular, when the partial graph size dominates the shadow graph size (e.g., when the ratio exceeds 4:1),  the attack accuracy against the property $P_1$ can be close to 1 for both $A_5$ and $A_6$ for GCN model, and no less than 0.7 for GAT and GraphSAGE. 
Second, the attack accuracy of $A_5$ and $A_6$ on $P_1$ is lower than that of the attacks $A_1$ and $A_2$ (i.e., only the partial graph is available), similarly for $P_4$. This is because the features collected from the shadow graphs may not have consistent distribution with those collected from the partial graph, and thus becomes “noise” and degrades GPIA performance.  
On the other hand, the attack accuracy of $A_5$ and $A_6$ is higher than that of $A_3$ and $A_4$. This is unsurprising as, compared with $A_3/A_4$, $A_5$ and $A_6$ utilizes the additional knowledge learned from the partial graph to improve its accuracy.



\vspace{-0.2in}
\subsection{Why Does GPIA Work? (RQ2)} 
\label{sc:RQ2}
As the experimental results have demonstrated the effectiveness of GPIA, next, we analyze why GPIAs can infer the existence of property in the training graph successfully. Conducting the theoretical analysis is very challenging due to the complexity in both training data and GNN models. Thus we discuss why GPIAs work based on practical evaluations. 

{\bf Correlation between property feature and label.} Intuitively, the attacker can infer the properties from the model output possibly because the property feature is strongly correlated with the task (i.e., the class label). Following this intuition, we measure the Pearson correlation between the property feature and class label of the three graph datasets, and show the results in Appendix \ref{appendix:label-correlation}. Essentially, the correlation between the property feature and the task label is weak for all the three datasets. Then why the properties can be leaked even when there is weak correlation between the property features and the label? This is possible due to the strong correlation between the property and non-property features in the data. For example, there is a strong Pearson correlation (0.81) between \texttt{gender} (property feature) and \texttt{height} (non-property feature) in Pokec dataset, between  "Insulin" (property feature) and "dietaries" (non-property feature) in Pubmed dataset (Pearson correlation 0.41), and between \texttt{gender} (property feature) and \texttt{education year} (non-property feature) in Facebook dataset (Pearson correlation 0.92). As these non-property features are correlated with the task label, the information of the properties still can be leaked regardless of whether training data contained the property feature or not.  

{\bf Non-negligible disparate influence across different groups.} 
As observed by the recent studies, ML models are “biased” in the sense that they behave differently across different groups in the training data  \cite{chouldechova2018frontiers,pessach2022review}.
Following this, we measure the disparity in GNN model accuracy across different node/link groups in Facebook and Pokec datasets. The results are included in Appendix \ref{appendix:disparity}. We observe the existence of accuracy disparity to some extent for all the three GNN models. In particular, the disparity is significant on Facebook dataset, where the difference in node classification accuracy across male and female groups can be as large as 0.13. 
The disparity demonstrates that the GNN models behave differently for different node/link groups. 

To have a deeper understanding of GNN models' behaviors towards different node/link groups, we measure the {\em influence score} of individual node/link to quantify the impact of a node/link on the GNN model performance. An intuitive idea measuring the influence of a given training node/link on a GNN model is to ask the counterfactual~\cite{feldman2020does,koh2017understanding}: {\em what would happen to the model behaviors if the model did not see the node/link}? Answering this counterfactual enables to connect the model's behaviors with the training data.

To quantify the effect of the counterfactual, we measure the difference in the model behaviors when it is trained with and without a particular node/link. We use gradients to capture the model behaviors. Formally, assuming 
$\g_{v}\stackrel{\Delta}{=}\nabla\mathcal{L}(\TargetG\backslash\{v\})$, 
that is, $\g_{v}$ is the set of gradients induced from the training of  the target model $\TargetM$ given the graph $\TargetG$ excluding the node $v$, where $\mathcal{L}$ is the loss function of $\TargetM$. 
Then the {\em influence score $I(v)$ of a node $v$} on $\TargetM$ is measured as 
\[I(v)\stackrel{\Delta}{=}distance(\g_{v}, g),\] where $g$ is the set of gradients induced by $\TargetM$ on $\TargetG$. Intuitively, higher influence score indicates the node $v$ impacts more on $\TargetM$. Similarly, 
assuming $g_{e}\stackrel{\Delta}{=}\nabla\mathcal{L}(\TargetG\backslash\{e\})$ for a given edge $e\in G$, the {\em influence score $I(e)$ of an edge $e$}  on $\TargetM$ is measured as follows:
\vspace{-0.05in}
\begin{equation}
\label{eqn:if}
I(e)\stackrel{\Delta}{=}distance(g_{e}, g).
\end{equation}
Various distance functions (e.g., Euclidean distance and cosine similarity) can be used. We use cosine similarity as the distance function. After we compute the influence score of individual nodes and links, we compute the average influence score of nodes/links in a particular group as the influence score of the group.


\begin{table*}[t!]
    \centering
    \begin{tabular}{c|c|c|c|c|c|c|c|c|c|c}
    \hline
     \multirow{3}{*}{{\bf Model}}&\multicolumn{5}{c|}{{\bf Node group}}&\multicolumn{5}{c}{{\bf Link group}}  \\ \cline{2-11}
     &\multicolumn{2}{c|}{{\bf Influence score}}&\multicolumn{2}{c|}{{\bf Model loss}}&{\bf Loss}&\multicolumn{2}{c|}{{\bf Influence score}}&\multicolumn{2}{c|}{{\bf Model loss}}&{\bf Loss}\\ \cline{2-5}\cline{7-10}
     &Male&Female$^*$&Male&Female$^*$&{\bf gap}&Same-gender$^*$&Diff-gender&Same-gender$^*$&Diff-gender&{\bf gap}\\ \cline{1-11}
     GCN&\cellcolor{asparagus}{0.038}&0.017&\cellcolor{carrotorange}{0.647}&0.298&0.01 &\cellcolor{asparagus}{0.022}&0.012&0.346&\cellcolor{carrotorange}{0.604}&0.004\\ \cline{1-11}
     GraphSAGE&\cellcolor{asparagus}{0.022}&0.015&\cellcolor{carrotorange}{0.558}&0.255&0.003 &0.0039&\cellcolor{asparagus}{0.011}&0.353&\cellcolor{carrotorange}{0.586}&0.0007\\ \cline{1-11}
     GAT&0.057&\cellcolor{asparagus}{0.087}&0.153&\cellcolor{carrotorange}{0.234}&0.004 &\cellcolor{asparagus}{0.034}&0.03&0.21&\cellcolor{carrotorange}{0.313}&0.003\\ \hline
    \end{tabular}
  \caption{Influence scores and GNN model loss per group, and loss gap between positive and negative graphs (Facebook dataset). The group of the larger size is marked with $^*$. Between two groups, the group of higher influence score and higher loss is marked \textcolor{asparagus}{green} and \textcolor{carrotorange}{orange} respectively. }
    \label{tab:inf-score}
    \vspace{-0.2in}
\end{table*}

\nop{
 \begin{table}[t!]
    \centering
    \begin{tabular}{c|c|c}
    \hline
     \multirow{2}{*}{{\bf GNN model}}&\multicolumn{2}{c}{{\bf Node group}}  \\ \cline{2-3}
     &Male&Female$^*$\\ \cline{1-3}
     GCN&\cellcolor{asparagus}{0.65}&0.30\\ \cline{1-3}
     GraphSAGE&\cellcolor{asparagus}{0.56}&0.26\\ \cline{1-3}
     GAT&0.15&\cellcolor{asparagus}{0.23}\\ \hline
    \end{tabular}
    \caption{GNN loss of different node groups (Facebook dataset). The group of the larger size is marked with $^*$. Between two subgroups, the group of higher loss is marked \textcolor{asparagus}{green}. }
  \end{table}
}
\nop{
\begin{table}[t!]
    \centering
    \begin{tabular}{c|c|c|c|c}
    \hline
     \multirow{2}{*}{{\bf GNN model}}&\multicolumn{2}{c|}{{\bf Node group}}&\multicolumn{2}{c}{{\bf Link group}}  \\ \cline{2-5}
     &Male&Female$^*$&Same-gender$^*$&Diff-gender\\ \cline{1-5}
     GCN&\cellcolor{asparagus}{0.038}&0.017&\cellcolor{asparagus}{0.022}&0.012\\ \cline{1-5}
     GraphSAGE&\cellcolor{asparagus}{0.022}&0.015&0.0039&\cellcolor{asparagus}{0.011}\\ \cline{1-5}
     GAT&0.057&\cellcolor{asparagus}{0.087}&\cellcolor{asparagus}{0.034}&0.03\\ \hline
    \end{tabular}
  \caption{Influence scores of different node/link groups (Facebook dataset). The group of the larger size is marked with $^*$. Between two subgroups, the group of higher influence score is marked \textcolor{asparagus}{green}. \xiuling{In the graph I used her, the group size of female is larger than that of male. }}
    \label{tab:inf-score}
    \vspace{-0.2in}
\end{table}
}
Since influence measurement is time consuming as it needs model retraining, we take a set of samples of Facebook dataset, with each sample  containing $\sim$ 700 nodes, and compute the average influence score for male and female groups (for property $P_2$), as well as the same-gender and diff-gender links (for property $P_4$) in these samples. From the results reported in Table \ref{tab:inf-score}, we observe that {\em all GNN models have noticeable disparity in the influence scores across different groups}. {\em Moreover, which group has higher influence is not solely determined by its group size. It is also dependent on the target model.} For example, as shown in Table \ref{tab:inf-score}, the Male group in Facebook dataset  has higher influence score on GCN and GraphSAGE but lower score on GAT than the Female group. 

We also measure the impact of disparate group influence on their model performance, and report the average model loss of different groups in Table \ref{tab:inf-score}. The results show the non-negligible disparity in model  loss across different groups. We also observe that the node groups of higher influence score also have higher loss. However, this does not hold for the link groups, as their influence is measured at link level, while their loss is calculated at node level (for node classification task). 
 

{\bf Negligible loss gap between positive and negative graphs.} Can the disparate influence and model loss of different node/link groups lead to different target model performance over positive and negative graphs, and thus enables GPIA? To answer this question, we generate a set of positive and negative graphs from the Facebook samples we used in Table \ref{tab:inf-score}, and measure the gap between the average loss of the target model trained on positive graphs and that on negative graphs. Our results (“Loss gap” column in  Table \ref{tab:inf-score}) show that the loss gap between positive and negative graphs is indeed negligible - the loss gap does not exceed 0.01 for all target models (loss values in the range of [0.12, 0.44]). This is not surprising, as making positive graphs to negative graphs (and vice versa) essentially changes the size of a group (e.g., from a minority group to a majority group). However, changing group size does not necessarily lead to higher or lower model loss averaged over the whole graph, given that there is no relationship between group size and its loss.   
Furthermore, our empirical analysis (Appendix  \ref{appendix:overfit-pia}) shows that there is no linear relationship between GPIA accuracy and the loss gap between positive and negative graphs. In particular, GPIA accuracy neither increases or decreases consistently with the growth of the loss gap. Therefore, the loss gap between positive and negative graphs does not contribute to GPIA's success.

{\bf Dissimilar distribution of embeddings/posteriors of positive and negative graphs.}
As different node/link groups have different influence on the target model, how these groups are distributed in the training graph affects the model parameters (embeddings) and posterior outputs obtained from positive and negative graphs. To justify this, we visualize the distribution of GPIA features aggregated from embeddings/posteriors output by the target model on positive and negative graphs (Appendix \ref{appendix:distribution}). We observe that the distribution of GPIA features aggregated from embeddings and posteriors of positive and negative graphs are distinctly dissimilar and well distinguishable. Such dissimilarity is thus utilized by the GPIA classifier to infer the existence of the property in the training data.


\nop{
\subsection{Why PIAs Work (RQ2)?}
As the experimental results have demonstrated the effectivenss of GPIA, next, we analyze why PIA can infer the existence of property in the training graph successfully. Conducting the theoretical analysis of why PIA can work is very challenging due to the complexity in both training data and GNN models. Thus we discuss why PIAs work based on practical evaluations.  



\begin{table*}[t!]
\small
    \centering
    \scalebox{1}{\begin{tabular}{c|c|c|c|c|c|c|c|c|c|c|c|c|c|c|c|c|c|c}
    \hline
    Property&\multicolumn{3}{c|}{$P_1$}&\multicolumn{3}{c|}{$P_2$}&\multicolumn{3}{c|}{$P_3$}&\multicolumn{3}{c|}{$P_4$}&\multicolumn{3}{c|}{$P_5$}&\multicolumn{3}{c}{$P_6$}\\\cmidrule{1-19}
    \multirow{2}{*}{Output}& \multicolumn{2}{c|}{Embedding} & Prob. & \multicolumn{2}{c|}{Embedding} & Prob.& \multicolumn{2}{c|}{Embedding} & Prob. & \multicolumn{2}{c|}{Embedding} & Prob. & \multicolumn{2}{c|}{Embedding} & Prob.& \multicolumn{2}{c|}{Embedding} & Prob.\\\cline{2-19}
    & $Z^1$ & $Z^2$ &$O$& $Z^1$ & $Z^2$ &$O$& $Z^1$ & $Z^2$ &$O$& $Z^1$ & $Z^2$ &$O$& $Z^1$ & $Z^2$ &$O$& $Z^1$ & $Z^2$ &$O$ \\ \hline
    GCN&0.33&0.45&0.19&0.17&0.32&0.4&0.16&0.33&0.24&0.31&0.37&0.21&0.16&0.30&0.38&0.14&0.23&0.16\\\cmidrule{1-19}
    GraphSAGE&0.2&0.31&0.05&0.17&0.29&0.3&0.56&0.40&0.21&0.17&0.16&0.06&0.38&0.32&0.26&0.35&0.13&0.15\\\cmidrule{1-19}
    GAT&0.29&0.35&0.11&0.31&0.29&0.42&0.27&0.32&0.25&0.37&0.14&0.17&0.36&0.26&0.33&0.13&0.25&0.13\\\hline
    \end{tabular}}
    \caption{\label{tab:compare1} Importance of property features on target model behaviors, measured as the JSD distance between embedding/posteriors of positive and negative graphs.}
\end{table*}

\nop{\begin{table*}[t!]
    \centering
    \begin{tabular}{|c|cccccc|cccccc|}\hline
   \multirow{2}{*}{\backslashbox{Attack}{GNN}}&\multicolumn{12}{c|}{GCN}\\\cline{2-13}        
   &$P_1 $& $P_1^{-P}$ &$P_1^{-P,X_1}$&$P_1^{-P, X_1,X_2}$& $P_1^{I}$&$P_1^{'}$&$P_4 $& $P_4^{-P}$ &$P_4^{-P,X_1}$&$P_4^{-P, X_1,X_2}$& $P_4^{I}$&$P_4^{'}$\\\hline
        $A_1^{1}$ &0.98&0.98&0.97&0.97&0.54&0.97&0.91&0.90&0.91&0.91&0.51&0.89\\\hline
        $A_1^{2}$ &0.97&0.98&1&1&0.53&0.92&0.86&0.88&0.88&0.83&0.5&0.87\\\hline
        $A_2$ &0.99&0.99&1&1&0.67&0.91&0.88&0.88&0.86&0.85&0.5&0.8\\\hline
    \end{tabular}
    \caption{\label{tab:compare1}Impact of property features and correlated features on PIA. $T^{-P}$ denotes PIA against the target model that does not use the property feature $P$.
    $T^{-P, X_1}$ ($T^{-P, X_1, X_2}$, resp.) denotes PIA against the target model that does not use the property feature $P$ and the feature $X_1$ ($X_2$ resp.) that has the top-1 (top-2, resp.) strongest correlation with $P$. 
    \Wendy{To-do: show the distance between embedding/posteriors for data with or without property (e.g. flip gender value).}}
\end{table*}    

}


\begin{figure*}[t!]
\centering
    \centering
    \large{\bf Pokec dataset}
    \vspace{0.1in}
    \\
    \begin{tabular}{cc}
    \begin{subfigure}[b]{.15\textwidth}
     \includegraphics[width=\textwidth]{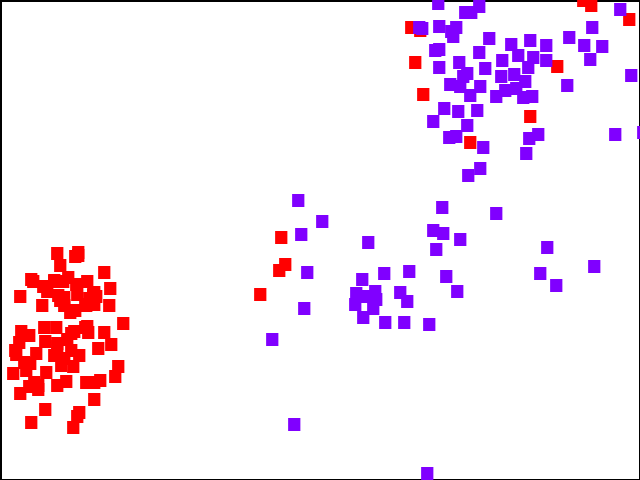}
     \vspace{-0.2in}
    \caption{Embedding $Z^1$}
    \end{subfigure}
     \begin{subfigure}[b]{.15\textwidth}
      \centering
     \includegraphics[width=\textwidth]{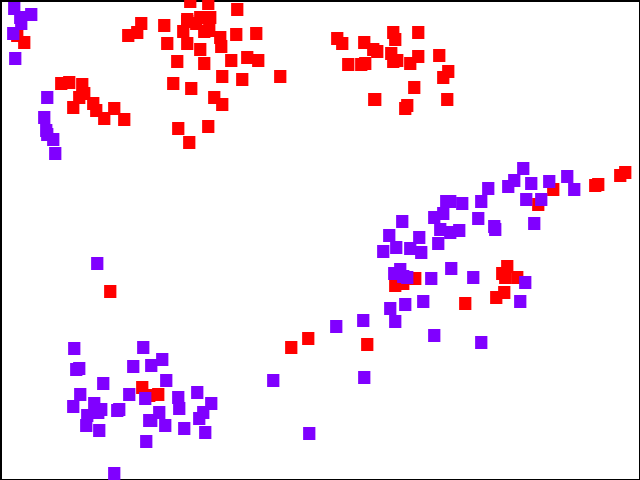}
     \vspace{-0.2in}
    \caption{Embedding $Z^2$}
    \end{subfigure}
    \begin{subfigure}[b]{.15\textwidth}
      \centering
     \includegraphics[width=\textwidth]{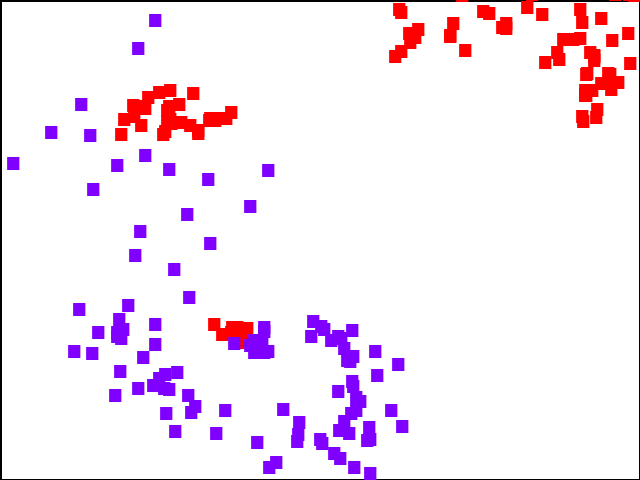}
     \vspace{-0.2in}
    \caption{Posteriors}
    \end{subfigure}
    &
    \begin{subfigure}[b]{.15\textwidth}
      \centering
     \includegraphics[width=\textwidth]{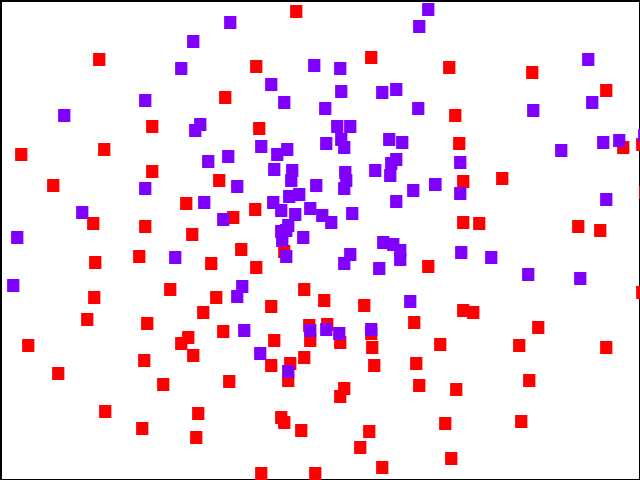}
     \vspace{-0.2in}
    \caption{Embedding $Z^1$}
    \end{subfigure}
     \begin{subfigure}[b]{.15\textwidth}
      \centering
     \includegraphics[width=\textwidth]{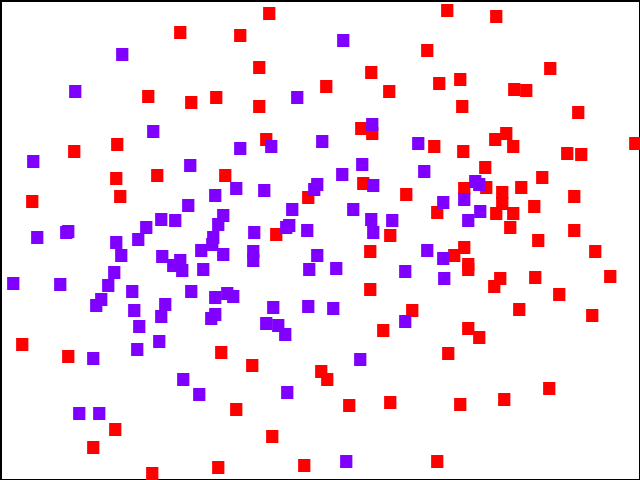}
     \vspace{-0.2in}
    \caption{Embedding $Z^2$}
    \end{subfigure}
    \begin{subfigure}[b]{.15\textwidth}
      \centering
     \includegraphics[width=\textwidth]{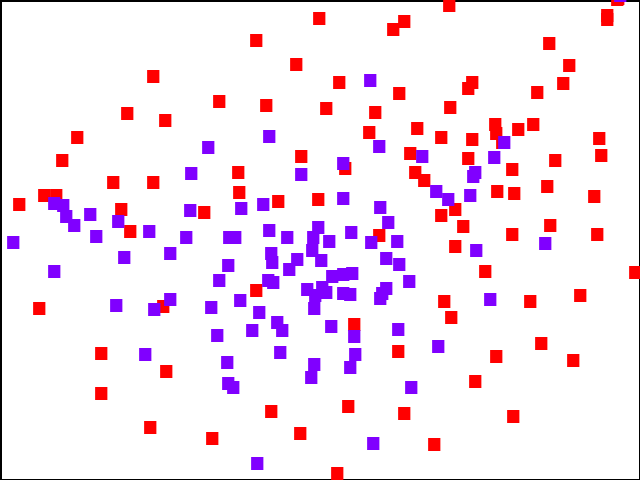}
     \vspace{-0.2in}
    \caption{Posteriors}
    \end{subfigure}
    \\
    {\bf Node-based property $P_1$}
    &
    {\bf Link-based property $P_4$}
    \\
     \end{tabular}
       \vspace{0.1in}
      \large{\bf Facebook dataset}
      \\
      \begin{tabular}{cc}
     \begin{subfigure}[b]{.15\textwidth}
     \includegraphics[width=\textwidth]{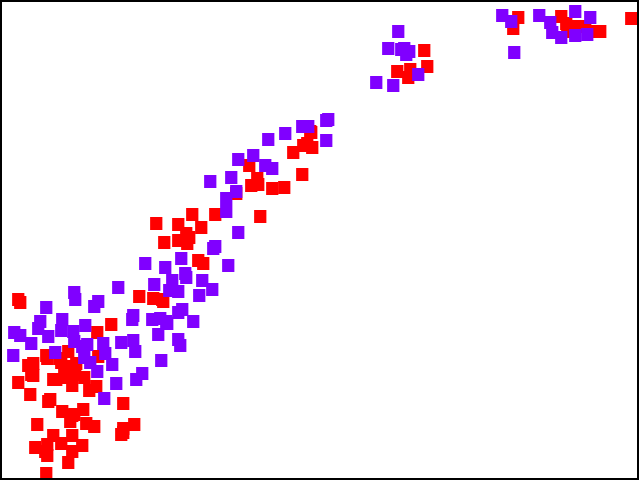}
     \vspace{-0.2in}
    \caption{Embedding $Z^1$}
    \end{subfigure}
     \begin{subfigure}[b]{.15\textwidth}
      \centering
     \includegraphics[width=\textwidth]{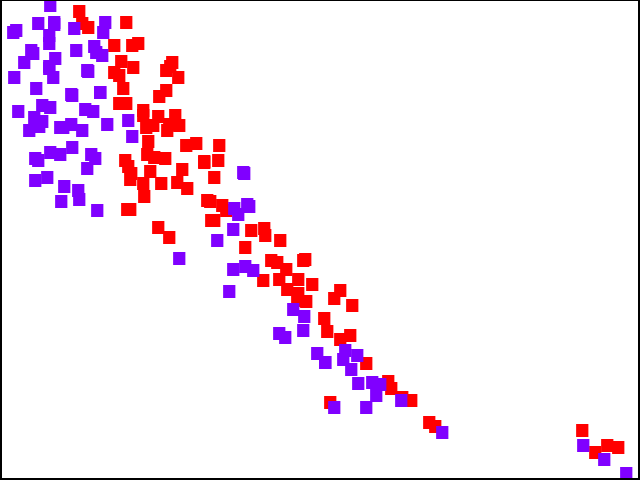}
     \vspace{-0.2in}
    \caption{Embedding $Z^2$}
    \end{subfigure}
    \begin{subfigure}[b]{.15\textwidth}
      \centering
     \includegraphics[width=\textwidth]{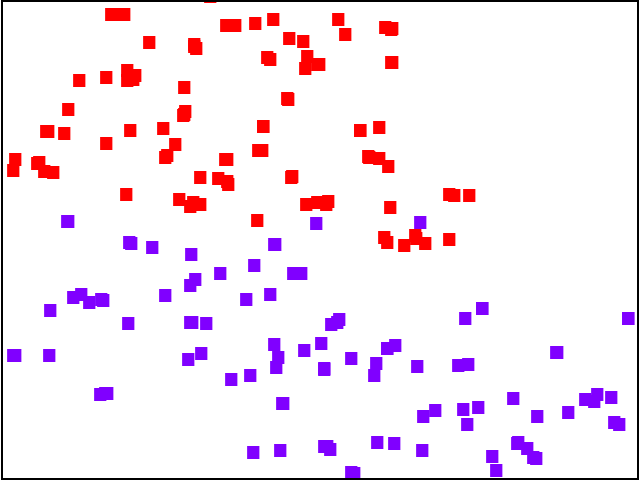}
     \vspace{-0.2in}
    \caption{Posteriors}
    \end{subfigure}
    &
    \begin{subfigure}[b]{.15\textwidth}
      \centering
     \includegraphics[width=\textwidth]{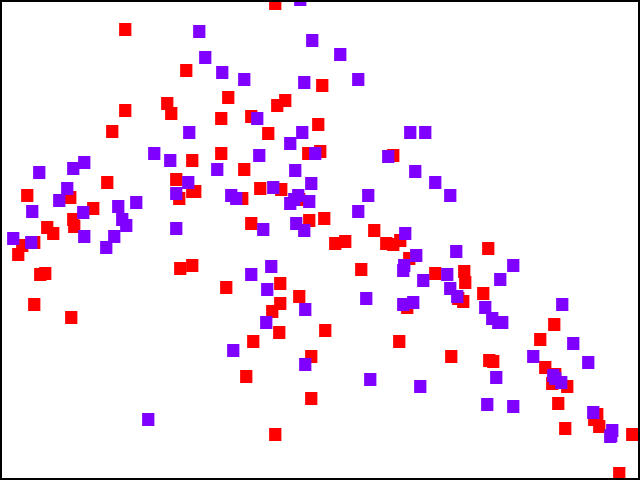}
     \vspace{-0.2in}
    \caption{Embedding $Z^1$}
    \end{subfigure}
     \begin{subfigure}[b]{.15\textwidth}
      \centering
     \includegraphics[width=\textwidth]{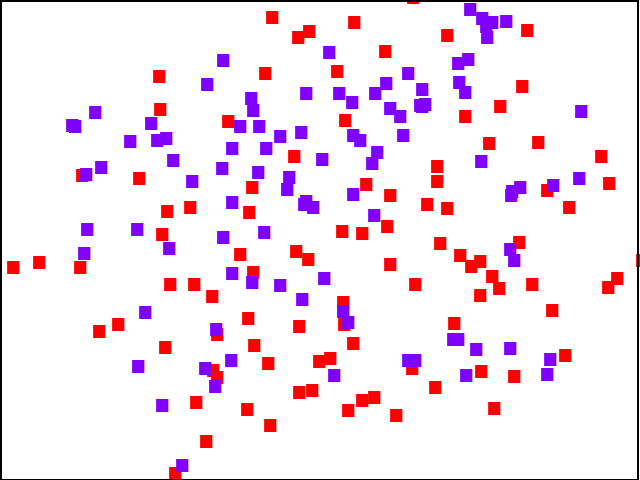}
     \vspace{-0.2in}
    \caption{Embedding $Z^2$}
    \end{subfigure}
    \begin{subfigure}[b]{.15\textwidth}
      \centering
     \includegraphics[width=\textwidth]{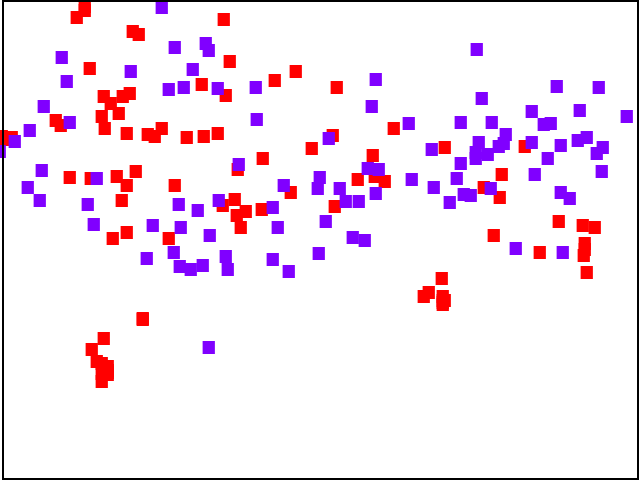}
     \vspace{-0.2in}
    \caption{Posteriors}
    \end{subfigure}
    \\
    {\bf Node-based property $P_2$}
    &
    {\bf Link-based property $P_5$}
     \end{tabular}
    \\
       \vspace{0.1in}
      \large{\bf Pubmed dataset}
      \vspace{0.1in}
      \\
      \begin{tabular}{cc}
        \begin{subfigure}[b]{.15\textwidth}
     \includegraphics[width=\textwidth]{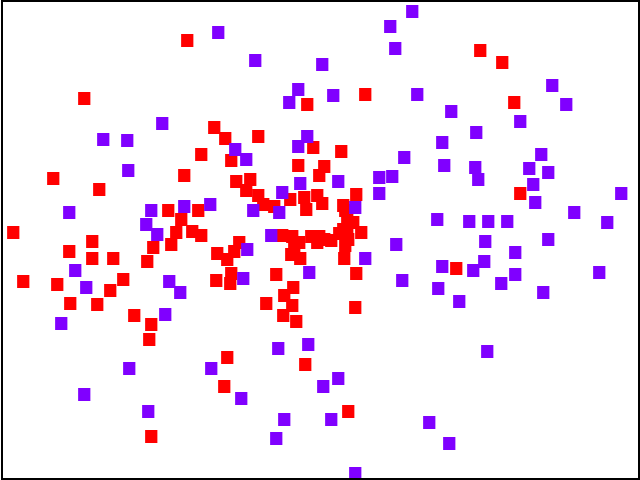}
     \vspace{-0.2in}
    \caption{Embedding $Z^1$}
    \end{subfigure}
     \begin{subfigure}[b]{.15\textwidth}
      \centering
     \includegraphics[width=\textwidth]{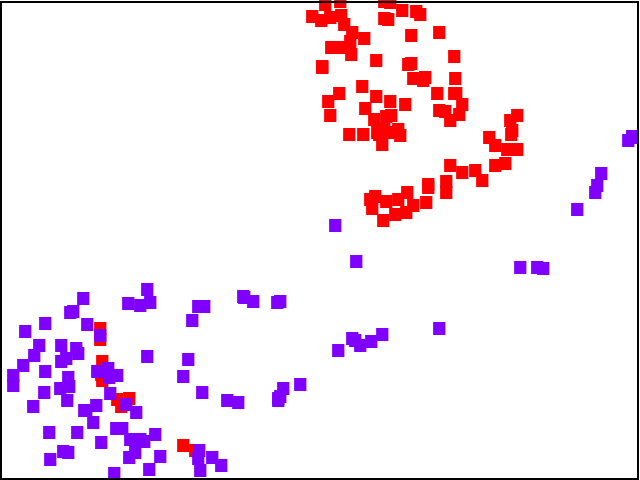}
     \vspace{-0.2in}
    \caption{Embedding $Z^2$}
    \end{subfigure}
    \begin{subfigure}[b]{.15\textwidth}
      \centering
     \includegraphics[width=\textwidth]{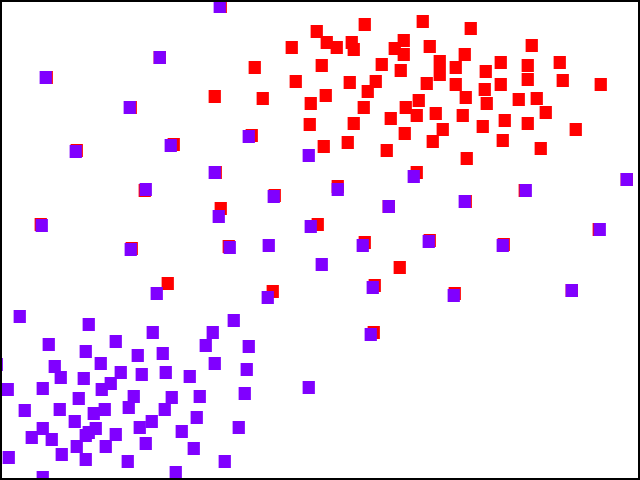}
     \vspace{-0.2in}
    \caption{Posteriors}
    \end{subfigure}
    &
    \begin{subfigure}[b]{.15\textwidth}
      \centering
     \includegraphics[width=\textwidth]{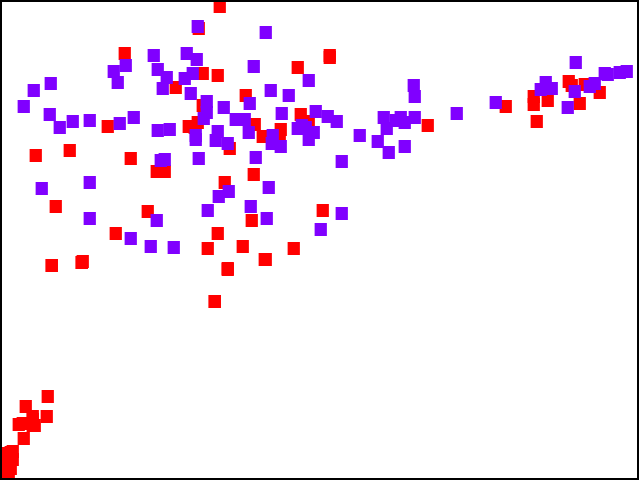}
     \vspace{-0.2in}
    \caption{Embedding $Z^1$}
    \end{subfigure}
     \begin{subfigure}[b]{.15\textwidth}
      \centering
     \includegraphics[width=\textwidth]{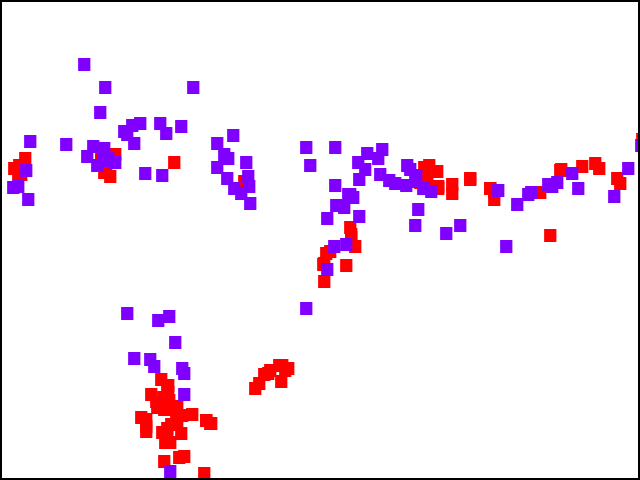}
     \vspace{-0.2in}
    \caption{Embedding $Z^2$}
    \end{subfigure}
    \begin{subfigure}[b]{.15\textwidth}
      \centering
     \includegraphics[width=\textwidth]{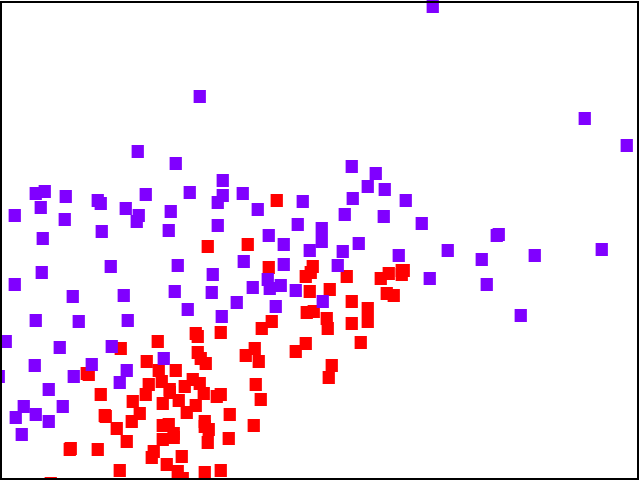}
     \vspace{-0.2in}
    \caption{Posteriors}
    \end{subfigure}
    \\
    {\bf Node-based property $P_3$}
    &
    {\bf Link-based property $P_6$}
    \end{tabular}
\caption{\label{fig:dis_pokec} TSNE visualization of the distribution of GPIA features (aggregation of node embeddings and posteriors by GCN model). Blue and red dots denote the node embeddings/posteriors generated from positive and negative graphs respectively.} 
\end{figure*}

\nop{
\begin{figure*}[t!]
\centering
    \centering
    \begin{tabular}{cc}
    \begin{subfigure}[b]{.15\textwidth}
     \includegraphics[width=\textwidth]{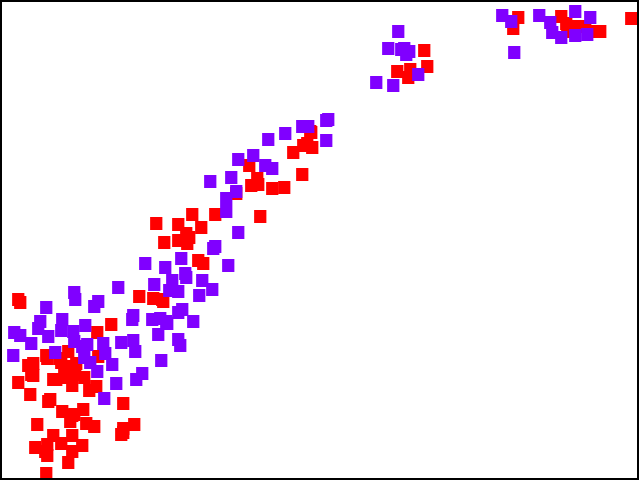}
     \vspace{-0.2in}
    \caption{Embedding $Z^1$}
    \end{subfigure}
     \begin{subfigure}[b]{.15\textwidth}
      \centering
     \includegraphics[width=\textwidth]{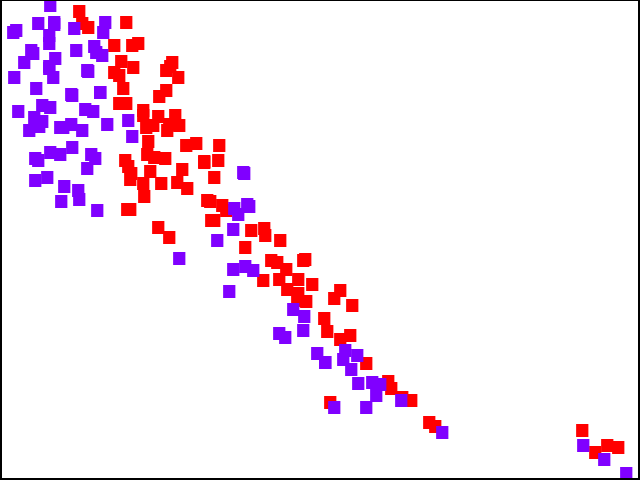}
     \vspace{-0.2in}
    \caption{Embedding $Z^2$}
    \end{subfigure}
    \begin{subfigure}[b]{.15\textwidth}
      \centering
     \includegraphics[width=\textwidth]{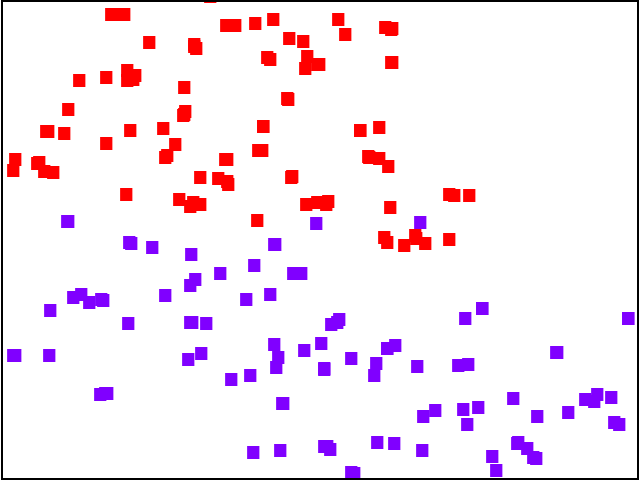}
     \vspace{-0.2in}
    \caption{Posteriors}
    \end{subfigure}
    &
    \begin{subfigure}[b]{.15\textwidth}
      \centering
     \includegraphics[width=\textwidth]{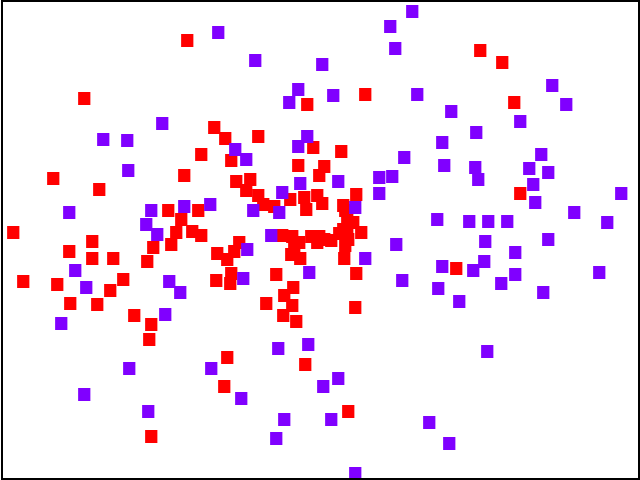}
     \vspace{-0.2in}
    \caption{Embedding $Z^1$}
    \end{subfigure}
     \begin{subfigure}[b]{.15\textwidth}
      \centering
     \includegraphics[width=\textwidth]{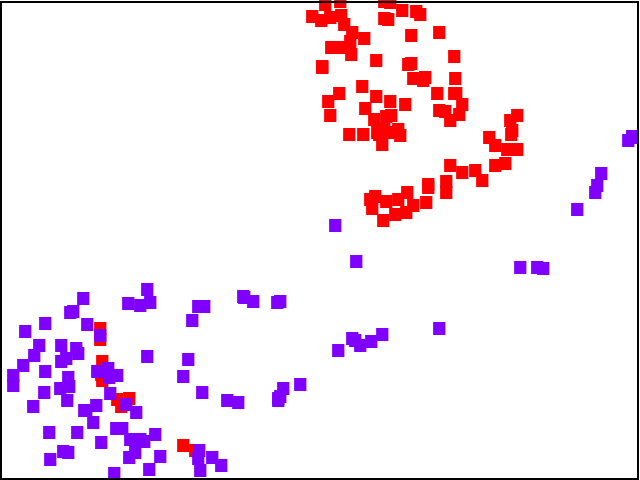}
     \vspace{-0.2in}
    \caption{Embedding $Z^2$}
    \end{subfigure}
    \begin{subfigure}[b]{.15\textwidth}
      \centering
     \includegraphics[width=\textwidth]{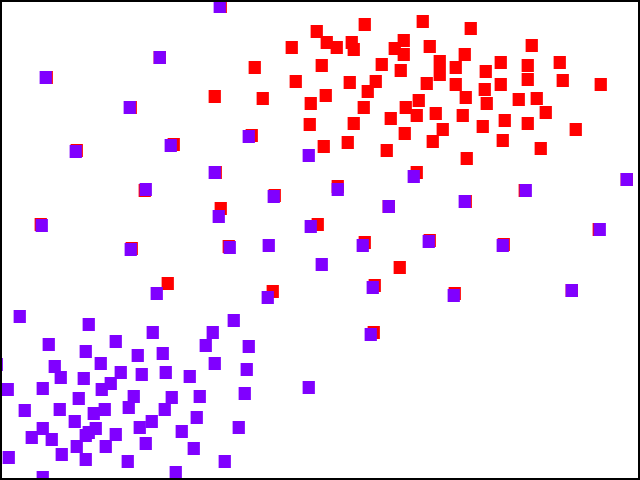}
     \vspace{-0.2in}
    \caption{Posteriors}
    \end{subfigure}
    \\
    {\bf Node-based property $P_2$}
    &
    {\bf Node-based property $P_3$}
    \end{tabular}
\caption{\label{fig:dis_fb-pubmed} TSNE visualization of the distribution of the input feature of GPIA (aggregation of node embeddings or target model outputs by GCN model) on Facebook and Pubmed dataset. Blue and red dots denote the node embeddings/posteriors generated from positive and negative graphs respectively. \xiuling{Should I also add the link level distributions here? Or just show Figure \ref{fig:dis_fb-pubmed} (c) and (f)?}}
\end{figure*}
}




{\bf GPIA Independence from correlation between property feature and class label.} One possible factor that contributes to GPIA's success is that there exists a strong correlation between the property feature and the class label. Such correlation can make the target model rely heavily on the property feature and thus its output varies significantly for positive and negative graphs. To validate this, we measure the Pearson correlation between the property feature and class label of the three graph datasets. However, it turns out that all the three graphs have weak Pearson correlations between the property feature and the label; the Pearson correlation value is 0.248, -0.01, and 0.107 for Pokec, Facebook, Pubmed datasets respectively. This shows that the correlation between property feature and class label doe not play an important role in the success of GPIAs. 

{\bf Data distribution of property features on target model.} Although there does not exist a strong correlation between the property feature and class label, the property feature may still impact the target model in the way that different data distributions on the property features change the target model output. To formalize such impact, we ask the counterfactual: {\em what would the target model behave if the values of the property feature were changed}? Note that changing the value of the property feature can change the graph from positive (negative, resp.) to negative (positive, resp.) consequently.  
To answer this question, we generate a number of pairs of positive and negative graphs that have the same graph structure but only differ at the property feature. 
In particular, given a graph $\TargetG$, we generate its {\em reverse} version $\bar{\TargetG}$ so that $\TargetG$ and $\bar\TargetG$ have the same graph structure but opposite property states, i.e., $\bar\TargetG$ is negative (positive, resp.) if $\TargetG$ is positive (negative, resp.). 
For both node properties $P_1$ and $P_2$ (on Pokec and Facebook graphs), we generate $\bar\TargetG$ by flipping the values on their property feature (i.e., change male/female to female/male). While for both link properties $P_4$ and $P_5$ on these two graphs, we generate $\bar\TargetG$ by converting the same-gender edges into diff-gender ones and the diff-gender edges into same-gender ones via changing the gender value. 
For both properties $P_3$ and $P_6$ on Pubmed graph, we generate $\bar\TargetG$ by flipping the  keywords “Insulin” with “Streptozotocin”. 
Next, we run the same GNN model on both $\TargetG$ and $\hat{\TargetG}$, and measure the distance between node embeddings and posterior outputs generated from them. 
To measure the distance of node embeddings, we follow \cite{hinton2002stochastic} to transfer the node embeddings into a probability distribution based on the scaled squared Euclidean pairwise distance between node embeddings. Then we use Jensen-Shannon divergence (JSD) \cite{majtey2005jensen} to measure the distance between the two probability distributions from $\TargetG$ and $\bar\TargetG$. 
The distance between posterior outputs is measured as JSD between the two posterior output distributions. The JSD distance is within the range [0, 1], where smaller value indicates the closer distributions. Finally, for each dataset, we generate 1,000 pairs of positive and negative graph pairs for node properties ($P_1$ - $P_3$) and 100 pairs for link properties ($P_4$ - $P_6$), and compute the average JSD distance of these pairs. 

We report the results of average JSD distance between embeddings/posteriors across positive and negative graphs in Table \ref{tab:compare1}. We observe although the distance varies for different target models and properties, there exists noticeably large JSD distance in many settings. The distance can be as large as 0.56 ($P_3$ with $Z^1$ and GraphSAGE). There are only 7 out of 54 settings whose JSD distance is less than 0.15. This demonstrates that the data distribution on the property feature have substantial influence on the behaviors of the target model, and thus explains why PIA can work. 


{\bf Distinguishable distribution of embeddings/posteriors from positive and negative graphs.} Intuitively, GPIA can predict the presence of the property because the output (either node embeddings or posteriors) of the positive a and negative graphs are distinguishable. To validate this, first, we compare the distribution of graph embeddings and posteriors generated from positive and negative graphs. Figure \ref{fig:dis_pokec} illustrates the distribution of both graph embedding and posteriors generated by GCN model on a pair of positive and negative graphs sampled from Pokec graph. We ensure both graphs only differ on their property feature. 
First, for the node property (Figure \ref{fig:dis_pokec} (a) - (c)), both node embeddings and posteriors from positive and negative graphs are well separated. Furthermore, node embeddings from the negative graph (red dots) are closer to each other than those (blue dots) from the positive graph. This also applies to the posteriors. This enables PIA to predict the property $P_1$ by utilizing the distance between node embeddings and posteriors. Second, for the link property (Figure \ref{fig:dis_pokec} (d) - (e)),  embeddings and posteriors do not show the same well-separateness pattern as the link property. However, node embeddings from the positive graph (blue dots) are closer to each other than those (red dots) from the positive graph. Thus GPIA is able to predict the presence of $P_4$ in Pokec dataset, but with a lower GPIA accuracy than the property $P_1$ (as shown in Figure \ref{fig:pia_acc} (a)). 
We have similar observations for both GAT and GraphSAGE models; more details can be found in Appendix \ref{appendix:dis_pokec2}. 
}

\subsection{Impact Factors of GPIA Performance (RQ3)}
\label{sc:att-settup}

In this section, we investigate how various factors impact GPIA performance. We consider the following factors: type of attack classifier model, type of embedding /posterior aggregation methods, type of dimension alignment methods, complexity of GNN models, group size ratio, and node overlapping between GPIA training and testing 
data. We also consider the impact of different amounts of embeddings on GPIA, and observe that GPIA performance stays stable for all the settings. We thus omit this part of discussions and put the results in Appendix \ref{appendix:multi-ft2}. 

{\bf Type of attack classifier models.} 
We measure GPIA accuracy when MLP, RF, and LR are used as the attack classifiers, and include the results in Appendix \ref{appendix:pia-classifier-graphsage-gat} due to the limited space. 
The main observation is that, while the three attack classifiers deliver similar performance in most of the settings, LR never outperforms MLP and RF. Furthermore, RF outperforms MLP slightly in  most of the white-box attacks,  while  MLP has slightly better performance than RF for the  black-box attacks.  Therefore, we recommend RF and MLP as the white-box and black-box attack classifier respectively. 
\begin{figure}[t!]
    \centering
\begin{subfigure}[b]{.25\textwidth}
      \centering
    \includegraphics[width=\textwidth]{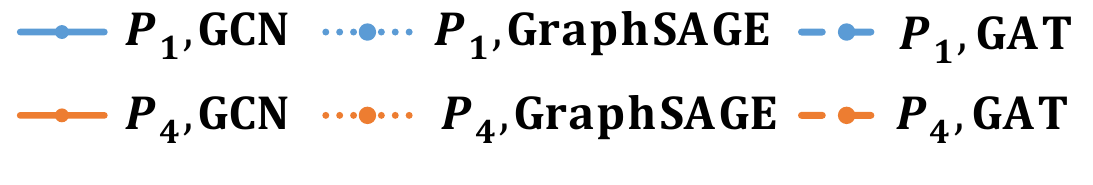}\\
    \end{subfigure}
\begin{tabular}{ccc}
    \begin{subfigure}[b]{.2\textwidth}
      \centering
    \includegraphics[width=\textwidth]{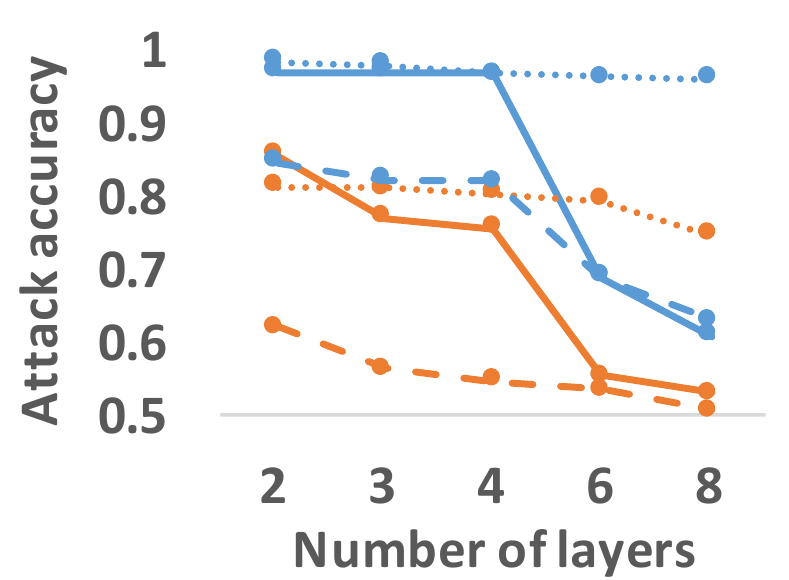}
     \vspace{-0.2in}
    \caption{Attack $A_1$}
    \end{subfigure}
    &
    \begin{subfigure}[b]{.2\textwidth}
    \centering
    \includegraphics[width=\textwidth]{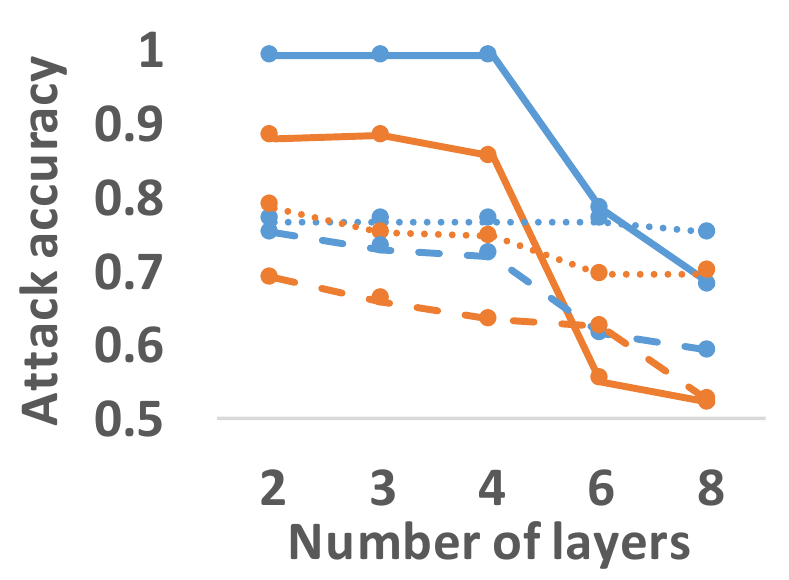}
    \vspace{-0.2in}
    \caption{Attack $A_2$}
    \end{subfigure}
\end{tabular}    
   \vspace{-0.1in}
\caption{\label{fig:more-layers} Impact of GNN model complexity on GPIA. }
\vspace{-0.2in}
\end{figure} 

\nop{
\begin{figure}[t!]
    \centering
      \centering
    \includegraphics[width=0.3\textwidth]{text/figure/agg-post-legend(1).pdf}\\
    \includegraphics[width=0.25\textwidth]{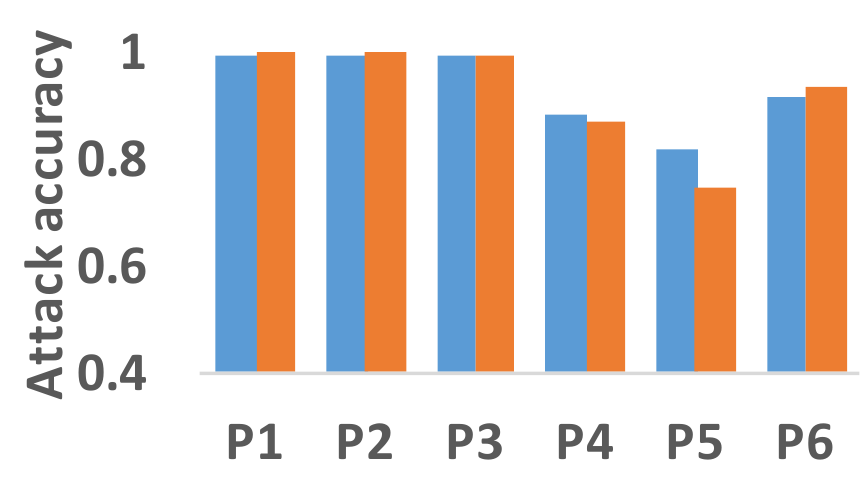}
     \vspace{-0.1in}
\caption{\label{fig:dif-ft-posterior-aggregation} Impacts of posterior aggregation method (element-wise difference) on GPIA performance (GCN as target model). } 
 \vspace{-0.1in}
\end{figure}
}

{\bf Embedding/posterior aggregation methods.} We measure the impacts of the three embedding aggregation methods (i.e., concatenation, max-pooling and mean-pooling) on GPIA performance, and include the results in Appendix \ref{appendix:agg-embedding-graphsage-gat}. 
The main observation is that max-pooling outperforms the other two methods in most of the settings. Therefore, we recommend  max-pooling as the embedding aggregation method.

We also measured the impact of the posterior aggregation methods (concatenation and element-wise difference) on GPIA performance. Our observation is that using concatenation method has either similar or slightly better GPIA performance than that by element-wise difference. 
The results can be found in Appendix \ref{appendix:agg-posterior-graphsage-gat}.

{\bf Dimension alignment methods.}
We measure GPIA performance for the four alignment methods, namely, sampling, TSNE projection, PCA dimension reduction, and Autoencoder compression, on the three target models. 
We put the results in Appendix \ref{sc:pca-encoder-acc} due to the limited space. 
The results suggest using TSNE as the dimension alignment method, as it delivers the best attack accuracy. On the other hand, the Autoencoder method always delivers the worst GPIA accuracy among all four alignment methods. 

{\bf Complexity of GNN models.} 
We define the network complexity by both the number of hidden layers and the total number of neurons in the network, and measure the GPIA performance against the target model of various complexity. We vary the number of hidden layers from 2 to 8, with 64 neurons at each layer, and use the embedding at the final hidden layer to launch the attack $A_1$. We only consider $A_1$ and $A_2$ in this set of experiments. 

Figure \ref{fig:more-layers} shows the results on Pokec dataset. We observe that both $A_1$ and $A_2$ are less effective on complex GNNs than the simple ones. For example, when the number of hidden layers increases to 8, the  accuracy of both $A_1$ and $A_2$ against GAT becomes close to 0.5.  
Although this is against our initial hypothesis that more complex models would intrinsically learn more information from the training dataset and hence be more sensitive to GPIA, our observation is indeed consistent with the prior PIA studies when CNNs are the target model \cite{parisot2021property} that it is not necessary that more complex models are more vulnerable to PIA. 

{\bf Group size ratio.}
So far our studies show that GPIA is successful for groups of disparate sizes. Next, we study if the size ratio between different property groups impacts GPIA performance. We consider property $P_2$ on Facebook dataset, vary the size ratio between Male and Female groups, and measure GPIA accuracy for these settings. The results can be found in Appendix \ref{appendix:group-size}. We observe that the attack accuracy is low ($\leq$ 0.6) when the group size ratio is 1:1, and the attack accuracy grows with the increase of the group size ratio. The accuracy can be as high as 1 when the group size ratio increases to 1:3. This demonstrates that GPIA performance is affected by group prevalence - it may fail if the property has a near 50\% prevalence. 


{\bf Node non-overlapping between GPIA training and testing data.} As our results of attacks $A_1 \& A_2$ were evaluated over GPIA training and testing data that have small amounts of node overlap, we generate GPIA training and testing data with no node overlapping, and evaluate the accuracy of $A_1 \& A_2$. The results can be found in Appendix \ref{appendix:non-overlap_pokec}. The main observation is that the attack accuracy for the non-overlapping setting is very close to that for the node-overlapping setting (Figure \ref{fig:pia_acc}). Thus a small amount of node overlapping between GPIA training and testing data does not affect GPIA accuracy significantly. 

\vspace{-0.1in}
\section{Defense Mechanisms}
\label{sc:defense}

In this section, we present our defense mechanisms against GPIA.

\vspace{-0.1in}
\subsection{Details of Defense Mechanisms}

{\bf Defense against black-box attacks.} As GPIA features are generated from the posteriors of the target model, we perturb these posteriors to defend against GPIA. In particular, for each node $v\in G$ and its associated posterior probabilities, 
 we add noise on each probability \cite{zhang2021inference} where the noise follows the Laplace distribution whose density function is given by $\frac{1}{2b} e^{ -\frac{x-\mu}{b}}$ ($b$: noise scale, $\mu$: the location parameter of the Laplace distribution). 
    
        

\begin{figure*}[t!]
\centering
    \centering
    \begin{subfigure}[b]{.65\textwidth}
      \centering
     \includegraphics[width=\textwidth]{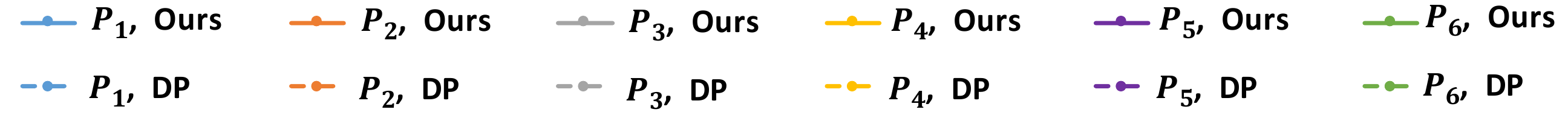}
     \vspace{-0.1in}
    \end{subfigure}
    \begin{tabular}{cc}
    \begin{subfigure}[b]{.3\textwidth}
      \centering
     \includegraphics[width=\textwidth]{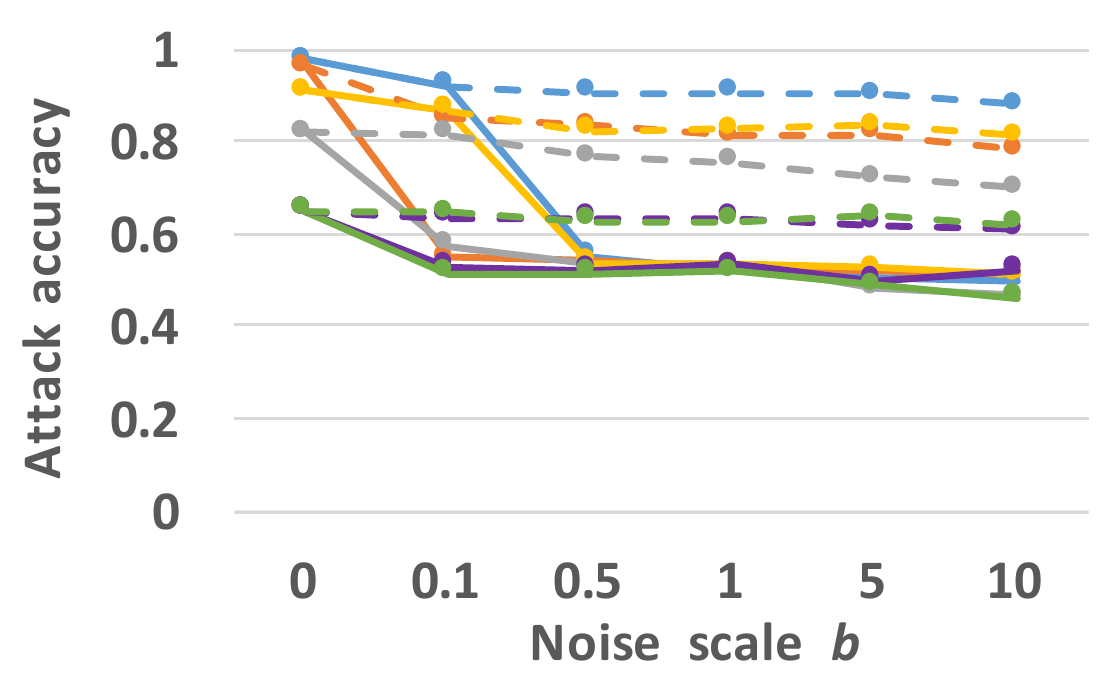}
     \vspace{-0.2in}
    \caption{\label{fig:gcn-lap-embed1} Attack $A^1_1$}
    \end{subfigure}
     \begin{subfigure}[b]{.3\textwidth}
      \centering
     \includegraphics[width=\textwidth]{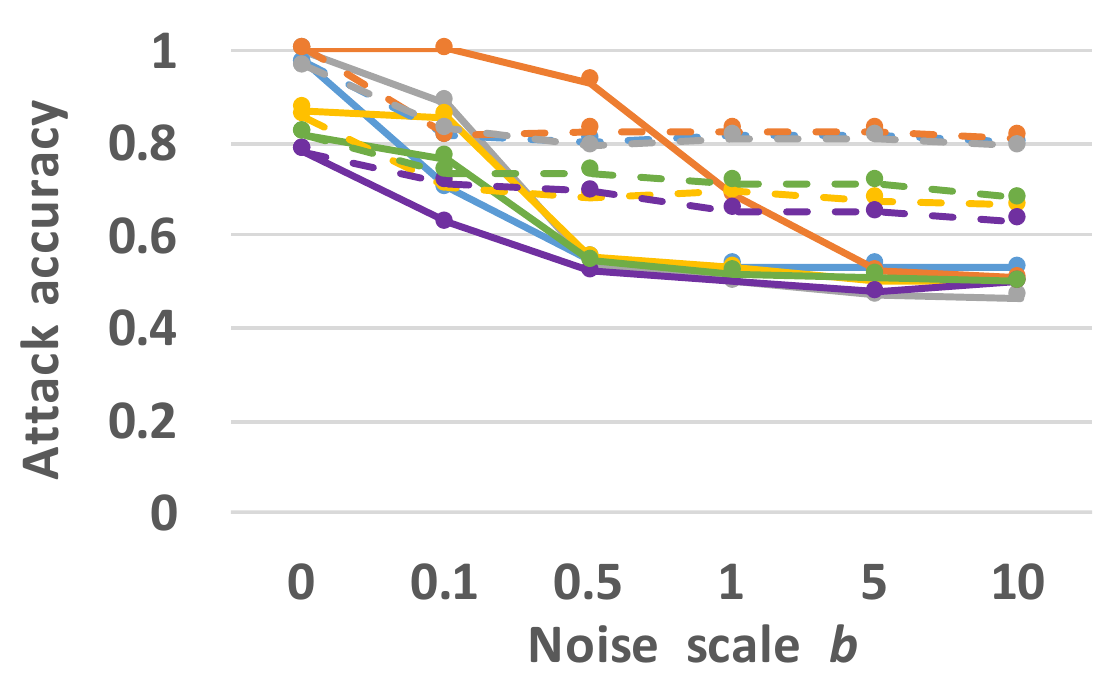}
     \vspace{-0.2in}
    \caption{\label{fig:gcn-lap-embed2} Attack $A^2_1$}
    \end{subfigure}
    \begin{subfigure}[b]{.3\textwidth}
      \centering
     \includegraphics[width=\textwidth]{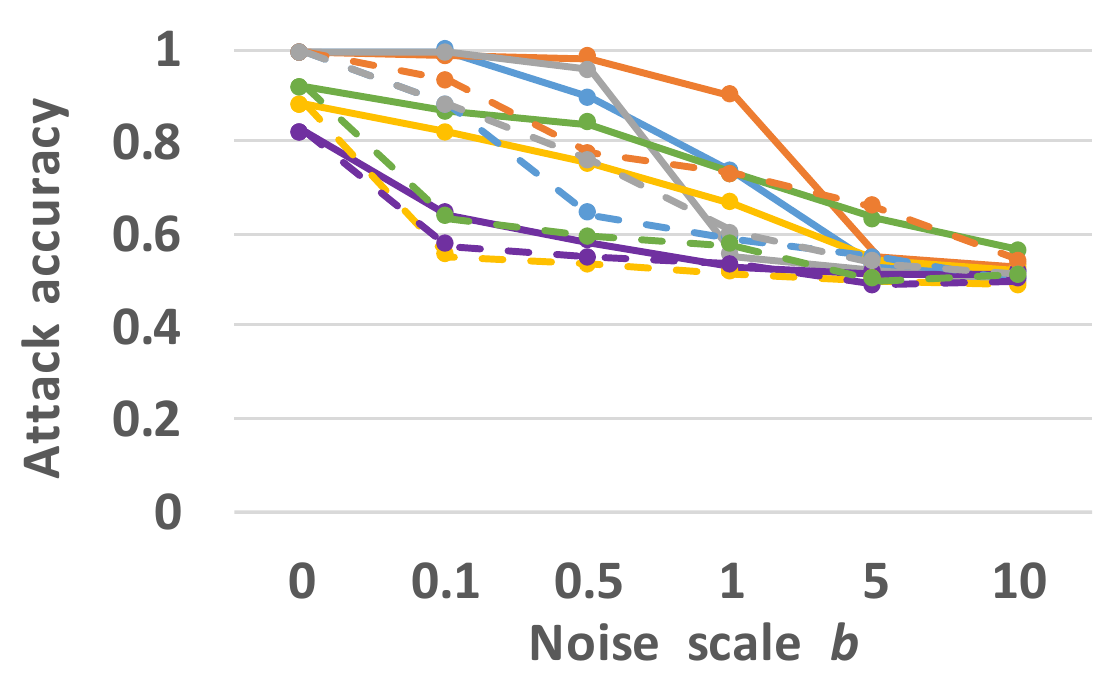}
     \vspace{-0.2in}
    \caption{\label{fig:gcn-lap-post} Attack $A_2$}
    \end{subfigure}
    \end{tabular}
    \vspace{-0.15in}
\caption{\label{fig:gcn-lap} Defense effectiveness of the noisy posterior/embedding defense method (GCN as the target model). The noisy embedding defense is used against $A^1_1$ and $A^2_1$, while the noisy posterior defense is used against $A_2$. }
\end{figure*}

 \begin{figure*}[t!]
\centering
\begin{tabular}{cc}
    \begin{subfigure}[b]{.45\textwidth}
      \centering
     \includegraphics[width=\textwidth]{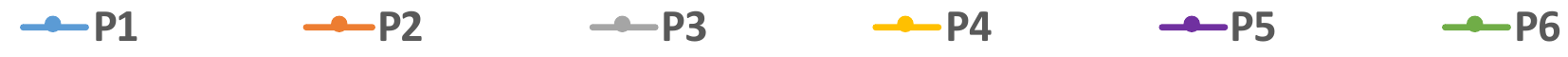}
     \vspace{-0.15in}
    \end{subfigure}
    &
     \begin{subfigure}[b]{.3\textwidth}
      \centering
     \includegraphics[width=\textwidth]{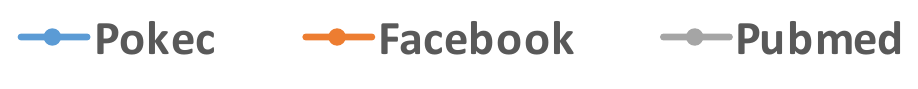}
     \vspace{-0.2in}
    \end{subfigure}
    \\
    \begin{tabular}{cc}
    \begin{subfigure}[b]{.23\textwidth}
      \centering
     \includegraphics[width=\textwidth]{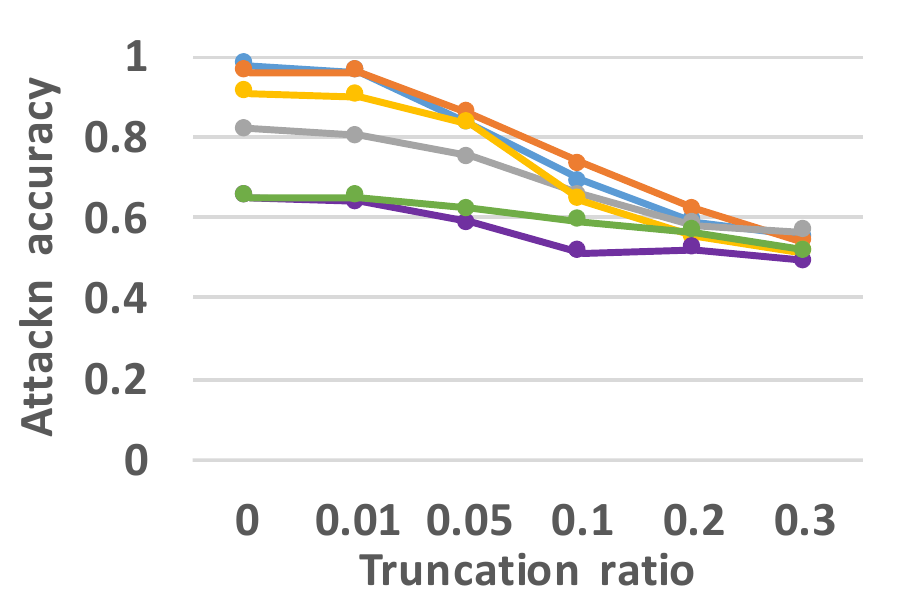}
     \vspace{-0.2in}
    \caption{\label{fig:gcn-trunc-embed1} Attack $A^1_1$}
    \end{subfigure}
     \begin{subfigure}[b]{.23\textwidth}
      \centering
     \includegraphics[width=\textwidth]{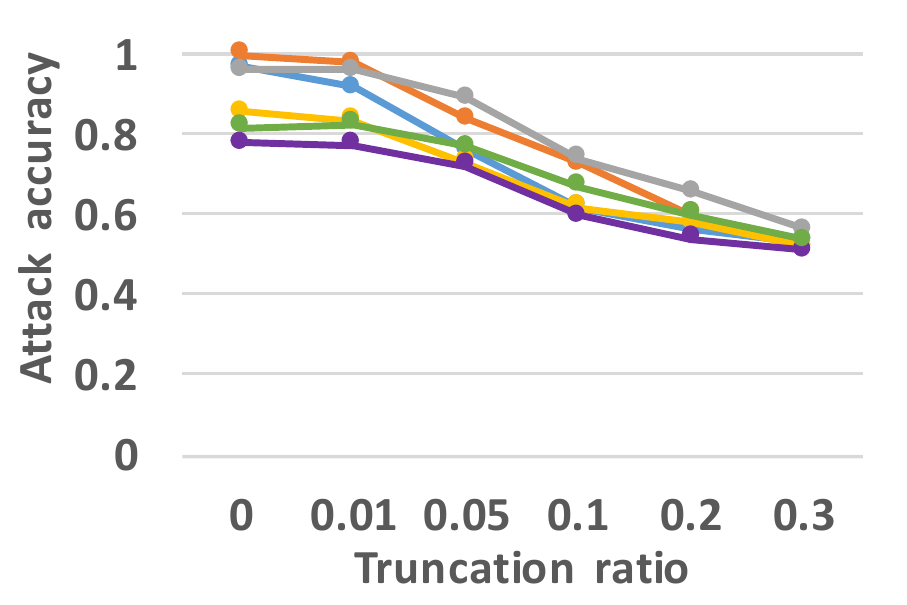}
     \vspace{-0.2in}
    \caption{\label{fig:gcn-trunc-embed2} Attack $A^2_1$}
    \end{subfigure}
    \end{tabular}
    &
    \begin{tabular}{cc}
    \begin{subfigure}[b]{.24\textwidth}
      \centering
     \includegraphics[width=\textwidth]{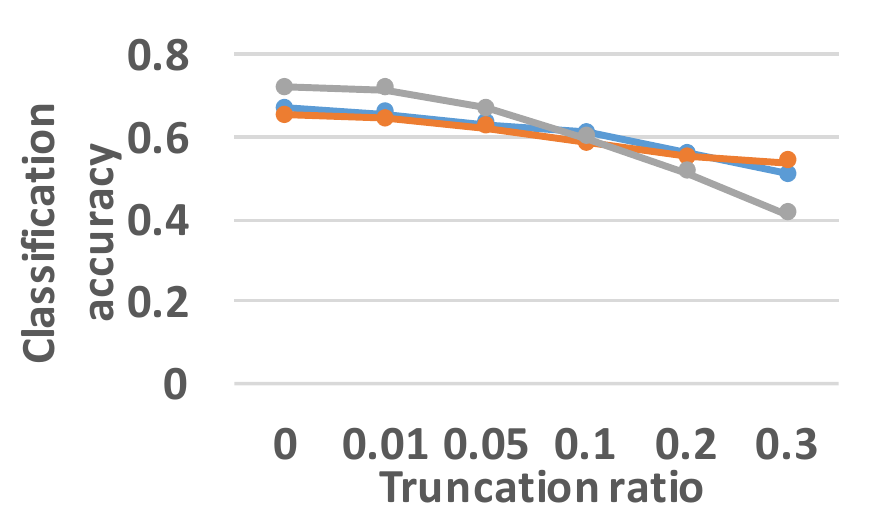}
     \vspace{-0.2in}
    \caption{\label{fig:gcn-trunc-quality1}Truncate embedding $Z^1$}
    \end{subfigure}
    \begin{subfigure}[b]{.24\textwidth}
      \centering
     \includegraphics[width=\textwidth]{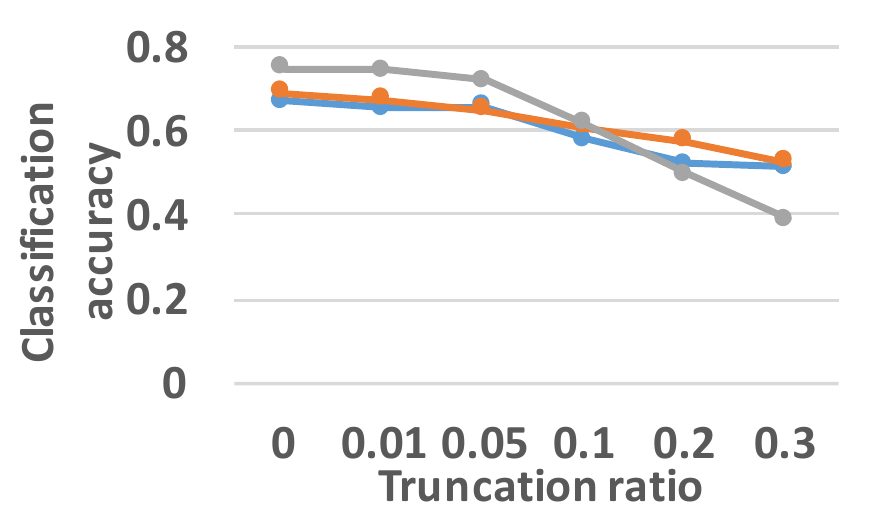}
      \vspace{-0.2in}
    \caption{\label{fig:gcn-trunc-quality2} Truncate embedding $Z^2$}
    \end{subfigure}
    \end{tabular}
    \\
    {\bf Attack accuracy}
         &
    {\bf Target model accuracy}
    \\
\end{tabular}
    \vspace{-0.1in}
\caption{\label{fig:gcn-trunc} Performance of the embedding truncation defense  (GCN as the target model).}
\vspace{-0.1in}
\end{figure*}

\begin{figure*}[t!]
\centering
    \centering
    \begin{subfigure}[b]{.8\textwidth}
      \centering
     \includegraphics[width=\textwidth]{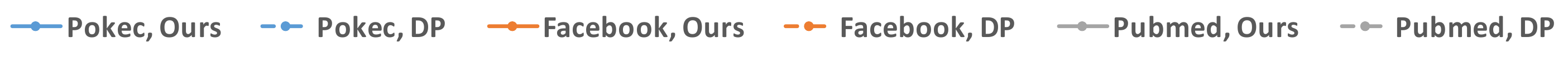}
    \end{subfigure}
    \begin{tabular}{cc}
    \begin{subfigure}[b]{.3\textwidth}
      \centering
     \includegraphics[width=\textwidth]{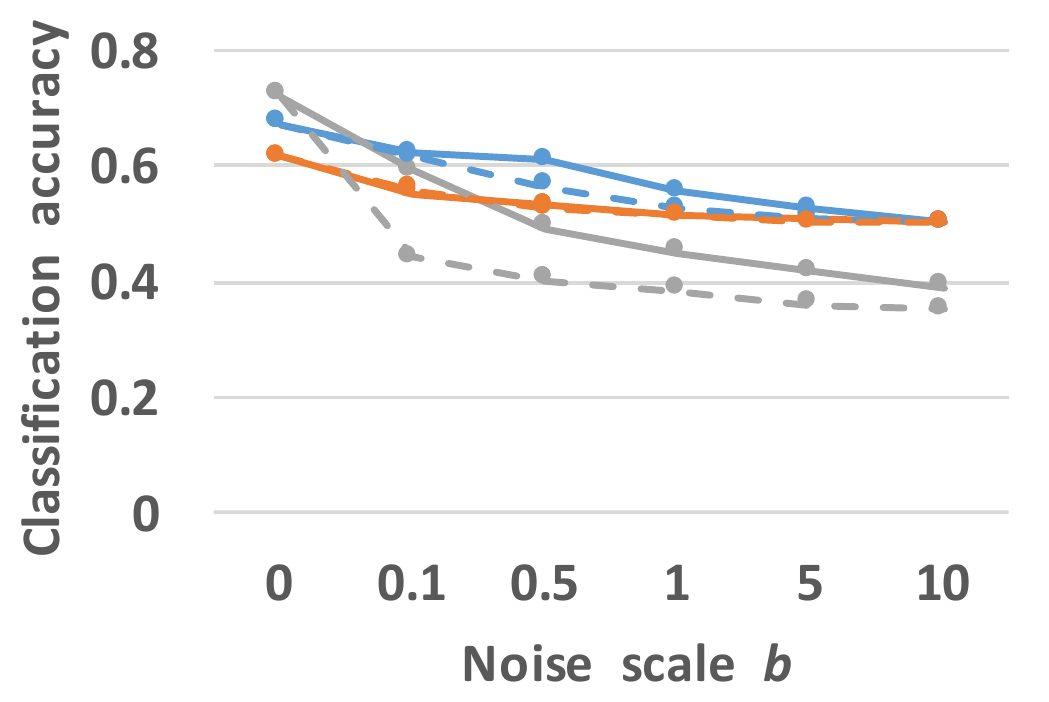}
     \vspace{-0.2in}
    \caption{\label{fig:gcn-lap-embed1-acc} Noise added to $Z^1$}
    \end{subfigure}
     \begin{subfigure}[b]{.3\textwidth}
      \centering
     \includegraphics[width=\textwidth]{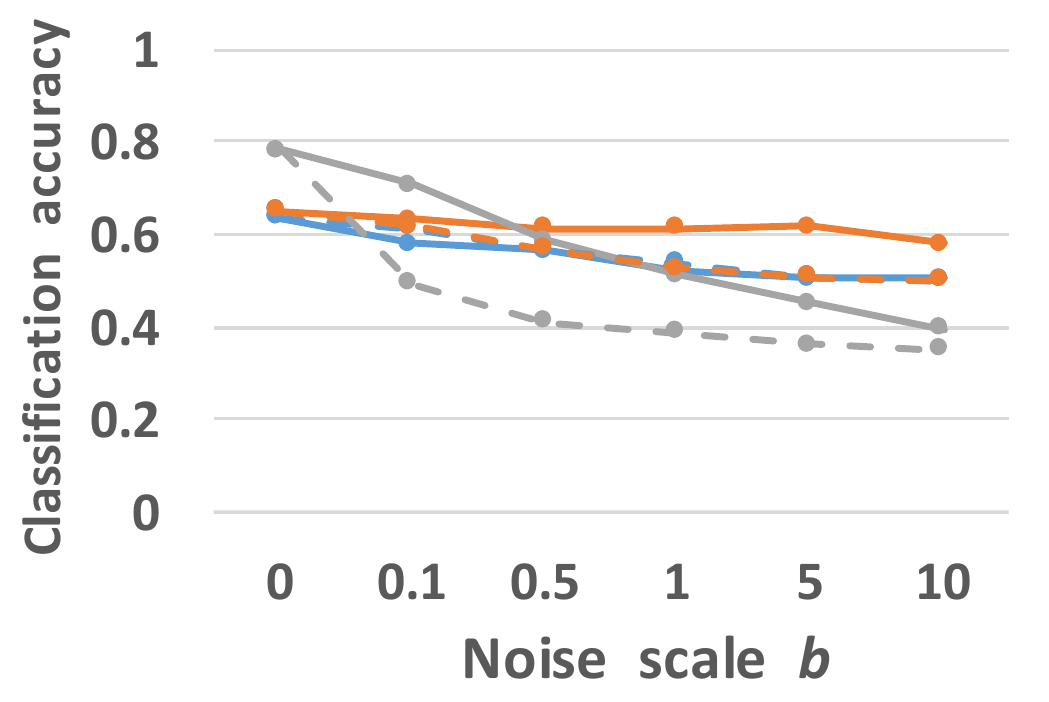}
     \vspace{-0.2in}
    \caption{\label{fig:gcn-lap-embed2-acc} Noise added to $Z^2$}
    \end{subfigure}
    \begin{subfigure}[b]{.3\textwidth}
      \centering
     \includegraphics[width=\textwidth]{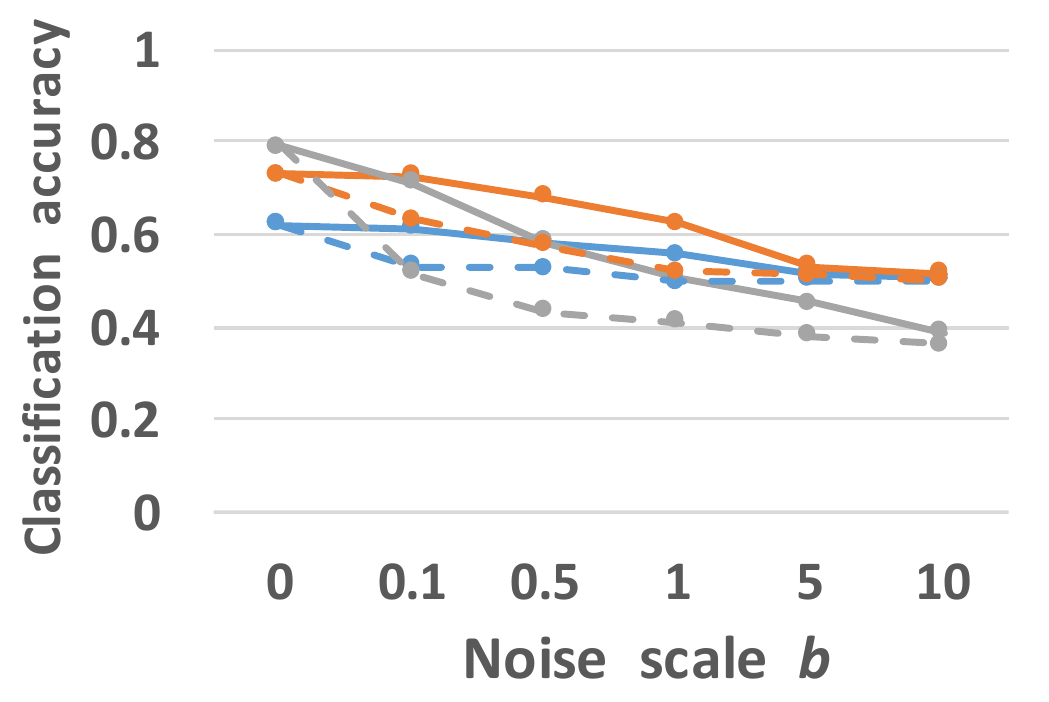}
     \vspace{-0.2in}
    \caption{Noise added to posteriors\label{fig:gcn-lap-post-acc}}
    \end{subfigure}
    \end{tabular}
    \vspace{-0.15in}
\caption{\label{fig:gcn-lap-acc} Target model accuracy under the noisy posterior/embedding defense (GCN as the target model). }
\end{figure*}

Besides the noisy posterior mechanism, we  evaluated two alternative methods: (1) {\em top-k posterior output} method: For each node $v\in G$ and its associated posterior probability values, we keep the top-k largest posteriors as the output. GPIA will be launched on the top-k posterior output; (2) {\em label-only output} method that the target model outputs the classification label instead of the posteriors. Our results show that both methods fail to either decrease GPIA accuracy significantly or provide acceptable target model accuracy. Thus we will not discuss these two defense mechanisms.

{\bf Defense against white-box attacks.} We design two types of defense mechanisms that mitigate GPIA effectiveness by modifying the node embeddings: (1)  {\em  noisy embedding}: For each node $v\in G$, let $z$ be its node embedding. We add Laplace noise on $z$, where the noise follows the Laplace distribution whose density function is the same as noisy posterior method; and (2) {\em embedding truncation}: An embedding of dimension $d$ is converted to another embedding of lower dimension $d'= d\times (1-r)$, where $r\in(0, 1)$ is the {\em truncation ratio}. Higher $r$ indicates more dimensions to be truncated and less information is kept in the embedding. We randomly pick $d'<d$ dimensions from the original embedding.  Different node embeddings may have different dimensions to be truncated even under the same truncation ratio. 
  
For the embedding truncation method, we implemented and evaluated three embedding dimension reduction methods including PCA \cite{minka2000automatic}, TSNE projection \cite{van2008visualizing}, and Autoencoder \cite{hinton2006reducing}. However, all of them fail to provide strong defense against GPIA, as they still preserve large amounts of information in the embedding which can be utilized by GPIA. Thus we will not present the details of these alternative truncation methods. 

\vspace{-0.15in}
\subsection{Evaluation of Defense Mechanisms}
We evaluate both effectiveness of the proposed defense methods and their impact on target model accuracy. We only consider attacks $A_1$ and $A_2$ as the defense effectiveness against these two attacks are expected to be applied to $A_3 - A_6$ due to their similarities. 

{\bf  Setup of defense mechanisms.} 
For both noisy embedding and noisy posterior defense mechanisms, we set the noise scale  $b=\{0.1,0.5,1,5,10\}$. For the embedding embedding truncation defense, we consider the compression ratio $r=\{0.01, 0.05, 0.1, 0.2, 0.3\}$. 
The setup of the target model is the same as in Section \ref{sc:exp-pia}. 

{\bf  Metrics.} We measure {\em defense effectiveness} as the accuracy of GPIA   against the GNN with defense. 
We measure {\em target model accuracy} as the accuracy of node classification by the target model. 


{\bf  Baseline.} Differential privacy (DP) \cite{dwork2014algorithmic} has been shown as effective against inference attacks on ML models \cite{shokri2017membership,jayaraman2019evaluating}.  Therefore, we use differentially private deep learning method \cite{abadi2016deep} that adds Laplace noise to the gradients as the baseline. We set the noise scale $\epsilon=\frac{1}{b}$ (i.e., $\epsilon=$ \{10,5,1,0.5,0.1\}), where $b$ is the noise scale value for the noisy embedding/posterior scheme. Lower $\epsilon$ indicates stronger noise scale and thus strong privacy protection.  
 
{\bf  Effectiveness of defense.} We add the noise to the embeddings $Z^1$ and  $Z^2$ to defend against $A^1_1$ and $A^2_1$. We do not consider adding noise to both layers as more noise will lead to higher loss of target model accuracy. Figure \ref{fig:gcn-lap} shows the attack accuracy results under the noisy embedding/posterior defense when GCN is the target model. The defense performance on GraphSAGE and GAT are shown in Appendix \ref{appendix:gs-gat-defense1}. First, we observe that the noisy embedding/posterior defense can reduce the attack accuracy effectively by achieving the accuracy around 0.5 in all the settings. Second, different attacks require different amounts of noise to achieve the same degree of protection. For example, the defense against $A^1_1$ and $A^2_1$ requires the noise $b= 0.5$ and $b=5$ to reduce the attack accuracy  to be close to 0.5 respectively. Different properties also require different amounts of noise to be added to achieve the same degree of protection. For example, the defense against $A^2_1$ on $P_2$ and $P_5$ requires noise of scale $b=5$ and $b=0.5$ to reach the attack accuracy around 0.5. 

We also observe that the defense power of the DP baseline is weaker than our method in the defense against the white-box attack $A_1$ (Figure \ref{fig:gcn-lap} (a) \& (b)), even with the  noise scale as large as 10 (i.e., DP noise scale $\epsilon=0.1$). 
Indeed, similar observations have been made that DP is ineffective against GPIA for other target models such as HMMs and SVMs \cite{ateniese2013hacking}. 
Indeed, although DP provides theoretical guarantee of privacy protection against {\em individual data points}, it is unclear if it can provide sufficient protection over GPIA inference of aggregate information of a group of samples.  
Further, the data independence assumption of DP \cite{liu2016dependence} is indeed violated in the context of GNNs, as the edges in graph are dependent and correlated. However, we also observe that DP is effective against the black-box attack $A_2$ (Figure  \ref{fig:gcn-lap} (c)), and outperforms our method when the noise scale $b<5$ in most of the settings.  

Figure \ref{fig:gcn-trunc} (a) and (b) demonstrate the effectiveness of the embedding truncation defense method with GCN as the target model.  The defense performance on GraphSAGE and GAT are shown in Appendix \ref{appendix:gs-gat-defense2}.  The embedding truncation defense can reduce the attack accuracy of $A^1_1$ to be  close to 0.5 when the truncation ratio is as small as 0.1 (i.e., remove 10\% of embeddings).  However, it requires more noise (truncation ratio as large as 0.3) to reduce the attack accuracy of $A^2_1$ to be close to 0.5. We believe this is because $A^2_1$ is stronger than $A^1_1$ as it encodes more information in the embedding that can be utilized by the attack. 

{\bf Target model accuracy under defense.} For the noisy posterior/embedding defense mechanism, we show the result of GCN accuracy in  Figure \ref{fig:gcn-lap-acc}. The results of GraphSAGE and GAT are included in  Appendix \ref{appendix:gs-gat-defense1}. We observe that GCN accuracy downgrades when more noise is added to the embeddings/posteriors. The accuracy loss varies for different datasets. For example, the accuracy loss never exceeds 10\% for Facebook when the noise is added to $Z^2$, but becomes as large as 50.7\% for Pubmed dataset (Figure \ref{fig:gcn-lap-acc} (b)). Indeed, Pubmed dataset is the most sensitive to the noise among the three datasets, as it witnesses the largest amounts of accuracy loss. Nevertheless, the target model accuracy is acceptable when the defense is sufficient (i.e., the attack accuracy is close to 0.5). For example, consider Pokec dataset and noise scale $b=1$, the accuracy of attack $A_1^1$ is  mitigated to around 0.5 (Figure \ref{fig:gcn-lap} (a)), while the target model accuracy is still 0.6 (Figure \ref{fig:gcn-lap-acc} (a)), which is higher than random guess for a binary classification task. 
We also observe that the target model accuracy by our defense always outperforms that of DP baseline. This demonstrates that adding noise on embeddings and posteriors better address the trade-off between defense and target model accuracy than adding noise on gradients. 

For the embedding truncation defense, we show the result of GCN in  Figure \ref{fig:gcn-trunc} (c) \& (d). The results of GraphSAGE and GAT are shown in  Appendix \ref{appendix:gs-gat-defense2}. We observe that the target model accuracy downgrades when the truncation ratio increases, and the accuracy loss varies for different datasets and  different embeddings that are truncated. For example, the target model accuracy loss is 13.9\% and 36.1\% for Facebook and Pubmed datasets respectively when $Z^1$ is truncated. Pubmed dataset witnesses the highest accuracy loss among the three datasets. Second, in terms of the trade-off between privacy and accuracy, the embedding truncation method loses to the noisy embedding method, as its target model accuracy is lower than that by the noisy embedding method under similar attack accuracy. For example, by the embedding truncation method, the target model accuracy is 0.58 for Facebook dataset (Figure \ref{fig:gcn-trunc} (d)) when the attack accuracy of $A^2_1$ against all properties becomes around 0.5 (Figure \ref{fig:gcn-trunc} (b)). This is slightly lower than that by the noisy embedding method, where the target model accuracy is 0.61 for Facebook dataset (noise scale $b=5$ in Figure \ref{fig:gcn-lap-acc} (b)) when the attack accuracy of $A^2_1$ against all properties becomes around 0.5 (noise scale $b=5$ in Figure \ref{fig:gcn-lap} (b)). 

{\bf Why are defense mechanisms effective?} 
Intuitively, our defense mechanisms add perturbations on individual embeddings and posterior  probabilities. Then why can they defend against the property inference at group level? To answer this question, we recall that one of root causes of GPIA is the disparate model loss across different groups (Section \ref{sc:exp-pia}). We measure the loss of all groups after the defense mechanisms are applied, and observe that the loss disparity across different groups is mitigated to some extent by the perturbation added to embeddings/posteriors. There is more mitigation of loss  disparity when more perturbation is added (i.e., stronger defense).  More details of how loss disparity across different groups is mitigated by defense can be found in Appendix \ref{appendix:defense-mitigation}.

\vspace{-0.1in}
\section{Conclusion}
\label{sc:conclusion}

In this paper, we propose the first systematic study of GPIA against GNNs. We design six GPIA attacks for both white-box and black-box settings, and demonstrate the attack effectiveness through extensive experiments. 
We  analyze the main factors that contribute to the success of GPIA. We also present various defense mechanisms against the proposed attacks, and demonstrate the effectiveness of these mechanisms. 

{\bf Limitations and future work.}  Next, we discuss  the limitations of our work and several research directions for the future work. 

\underline{Properties at subgraph level.} So far, we only consider the group properties at node and link levels. In general, the properties at subgraph level, e.g., imbalanced data distribution across different communities, are sensitive. Thus an interesting direction for the future research is to extend our GPIA model to deal with subgraph-level properties. The design strategy of the subgraph-based PIAs can be similar to GPIA: we generate the training data from positive and negative shadow/partial graphs, and train the classifier on the generated data. 

\underline{Non-binary properties.} Our attacks only deal with binary properties. A more powerful attack can be predicting from multiple classes, for example, inferring the population ratio of particular racial groups in the given graph. 
One straightforward solution to non-binary properties is simply replacing the binary GPIA classifiers with multi-class ones using the same GPIA features. 
An alternative solution is to use meta-classifiers \cite{ateniese2013hacking, ganju2018property} for inference. 

\underline{Sparse graphs.} The success of GPIA relies on its training data that consists of sufficient number of positive graphs sampled from shadow/partial graphs. This may not be achievable on shadow/target graphs that are sparse, as  their samples are likely to contain few links and thus fail to meet the properties, especially the link-level ones. In this case, the adversary may need to apply link prediction algorithms \cite{lu2011link,al2006link} to add links to the shadow/partial graphs in the GPIA training data.

\underline{Fairness-enhancing methods as defense.} As the accuracy disparity of GNN models is one of the potential factors that contribute to the success of GPIA, a straightforward approach is to apply the existing fairness-enhancing methods for graph embeddings (e.g., \cite{rahman2019fairwalk,bose2019compositional}) to ensure the target model to achieve group fairness, i.e., the graph embeddings are independent from the property features and thus the accuracy disparity of downstream tasks across different groups is minimized. We will evaluate the effectiveness of GPIA against those “fair” graph embeddings, and further investigate the relationship between fairness and GPIA. 


\begin{acks}
We thank the anonymous reviewers for their feedback. This project was supported by the National Science Foundation (\#CNS-2029038; \#CNS-2135988). Any opinions, findings, and conclusions or recommendations expressed in this paper are those of the authors and do not necessarily reflect the views of the funding agency.
\end{acks}

\newpage
\bibliographystyle{plain}
\balance
\bibliography{bib}

\appendix

\section{Details of Three GNN models}
\label{appendix:gnn}

\begin{table}[t]
\small
     \centering
     \begin{tabular}{c|c|c}\hline
    {\bf Model} & {\bf \textsf{Aggregate} function} & {\bf \textsf{Update} function} \\\hline
     GCN & All neighbor nodes & ReLU (embeddings) \\\hline
     GAT & All neighbor nodes & ReLU (embeddings + weights) \\\hline
     GraghSAGE & A subset of neighbor nodes& ReLU (embedding) \\\hline
     \end{tabular}
     \caption{Comparison of three GNN models.}
    \label{tab:gnnproperty}
     \vspace{-0.15in}
\end{table}

In this paper, we consider three representative GNN models, namely {\bf Graph Convolutional Network (GCN)} \cite{kipf2017semisupervised}, {\bf GraphSAGE} \cite{hamilton2018inductive}, and {\bf Graph Attention network (GAT)} \cite{velickovic2018graph}. These three models mainly differ on either \textsf{AGGREGATE} and \textsf{UPDATE} functions. 
Table \ref{tab:gnnproperty} summarizes the major difference in the two functions of the three GNN models. 

Next, we briefly describe the \textsf{AGGREGATE} and \textsf{UPDATE} functions of these models.

{\bf Graph Convolutional Networks (GCN) \cite{kipf2017semisupervised}}. The \textsf{AGGREGATE} function of GCN is defined as following: 
 \begin{equation}
\label{eqn:aggregategcn}
    Z^{\ell+1}=\textsf{UPDATE}(\bar{D}^{-\frac{1}{2}}\bar{A}\bar{D}^{-\frac{1}{2}}Z^{\ell}\theta^{\ell}),
\end{equation}
where $\bar{A}=A+I_N$ is the adjacency matrix of the graph $\TargetG$ with added self-connections, $I_N$ is the identity matrix, $\bar{D}_{ii}=\sum_j\bar{A}_{i,j}$ for all nodes $i,j\in\TargetG$, $\theta^{\ell}$ are layer-specific trainable parameters.
GCN uses ReLU as the \textsf{UPDATE} function. 
  
  {\bf GraphSAGE} \cite{hamilton2018inductive}  differs from GCN in the  \textsf{AGGREGATION} function. Unlike GCN that use the complete 1-hop neighborhood at each iteration of message passing, GraphSAGE samples a certain number of neighbour nodes randomly at each layer for each node. The message-passing update of  GraphSAGE is formulated as:
  \begin{equation}
  \begin{aligned}
\label{eqn:aggregategraphsage}
    z_{i}^{\ell+1}&=\textsf{CONCAT}(z_{i}^{\ell}, \textsf{AGGREGATE}^\ell(\{z_j^{\ell},\forall v_j\in \tilde{\mathcal{N}}(v_i)\})), 
    \end{aligned}
\end{equation}
  where $\tilde{\mathcal{N}}(v_i)$ is the sampled neighbours of node $v_i$. There are multiple choices of \textsf{AGGREGATE} functions such as mean, LSTM, and pooling aggregators. The \textsf{UPDATE} remains the same as ReLU.

  {\bf Graph Attention Networks (GAT)} \cite{velickovic2018graph} adds attention weights to the {\textsf AGGREGATE} function. In particular, the aggregation function at the $(\ell+1)$-th layer by the $t$-th attention operation is formulated as:
    \begin{equation}
  \begin{aligned}
\label{eqn:aggregategraphgat}
    z_{i}^{\ell+1,t}&=\textsf{UPDATE}(\sum_{\forall v_j \in \mathcal{N}(v_i)\cup{v_i}} \alpha_{ij}^{t}z_{i}^{\ell}),\\
    \end{aligned}
\end{equation}
where $\alpha_{ij}^t$ is the attention coefficient computed by the $t$-th attention mechanism to measure the connection strength between the node $v_i$ and its neighbor $v_j$. 
The \textsf{UPDATE} function concatenates all node embeddings corresponding to $T$ attention mechanisms
   \begin{equation}
  \begin{aligned}
\label{eqn:aggregategraphgat-update}
    z_{i}^{\ell+1} = ||_{t=1}^{T}ReLU(z_{i}^{\ell,t}\mathbf{W}^{t\ell}),
    \end{aligned}
\end{equation}
where $||$ denotes the concatenation operator and $\mathbf{W}^{t\ell}$ denote the corresponding weight matrix at layer $\ell$.

\section{Datasets and Their Characteristics}
\label{appendix:data}

\subsection{Datasets}
\label{appendix:dataset}
We use three datasets, namely Pokec, Facebook, and PubMed datasets, in our paper. Below are the details of these datasets. 

 {\bf Pokec}\footnote{https://snap.stanford.edu/data/soc-pokec.html} dataset is an online social collected in Slovak. It contains 632,803 nodes and 30,622,564 edges. Each node in the graph has the anonymized features such as gender, age, hobbies, interest, and education. We sampled the nodes with complete features as the original graph. The graph contains 45,036 nodes and 170,964 edges. We take the public, gender, age, heights, weight, region as node features. 

{\bf Facebook}\footnote{https://snap.stanford.edu/data/ego-Facebook.html} dataset consists of 4,039 nodes and 88,234 edges. Each node in the graph has the following features: birthday, education, work, name, location, gender, hometown, and language. All the values of the features were anonymized for privacy protection. Specifically, the gender values were anonymized as values 77 and 78. 
    We de-anonymize these values by their frequency. Since Gender 77 and 78 take 38.7\% and 61.3\% respectively, we de-anonymized value 77 to {\em male}, and value 78 to {\em female}, according to Facebook statistics\footnote{Facebook User Statistics: \url{https://tinyurl.com/y87bfs3o}} The education type were anonymized as values 53, 54, 55. Following the external knowledge of education types  \cite{jure2012advances}, we de-anonymize the education types as  college, graduate school, and high school. 

{\bf Pubmed Diabetes dataset}\footnote{https://linqs-data.soe.ucsc.edu/public/Pubmed-Diabetes} dataset consists of 19,717 scientific publications from PubMed database pertaining to diabetes classified into one of three classes. The citation network consists of 44,338 links. Each publication in the dataset is described by a TF/IDF weighted word vector from a dictionary which consists of 500 unique words, e.g., male, female, children, cholesterol, and insulin.  

\subsection{Size Ratio of Property Groups}
\label{appendix:group-ratio}
We measured the size ratio between the property groups specified in $P_1 - P_6$, and show the results in Table \ref{tab:group-size}. The main observation is that the group distribution is not uniform for all the three datasets. 

\begin{table}[t!]
\small
    \centering
    {
    \begin{tabular}{c|c|c}
    \hline
         \textbf{Dataset} & \textbf{Property}  & \textbf{Group size ratio} \\\hline
        \multirow{2}{*}{Pokec}& $P_1$ & Male: Female  = 0.76\\\cline{2-3}
        &$P_4$ & Same-gender links : diff-gender links = 0.61\\\hline
        \multirow{2}{*}{Facebook}& $P_2$ & Male: Female= 1.58\\\cline{2-3}
        &$P_5$ & Same-gender links: diff-gender links= 1.29\\\hline
        \multirow{2}{*}{Pubmed}& $P_3$ & With "IS": w/o "IS" = 1.54\\\cline{2-3}
        &$P_6$ &links between "IS" : links between "ST" = 1.41\\\hline
    \end{tabular}}
    \caption{Group size ratio for each property.}
    \label{tab:group-size}
    \vspace{-0.2in}
\end{table}

\subsection{Correlations between Property Features and Labels}
\label{appendix:label-correlation}
Table \ref{tab:label-correlation} shows the Pearson correlations between the property features and label. We observed that all the three graphs have weak Pearson correlations (no more than 0.3) between the property feature and the label.

\begin{table}[t!]
     \centering
     \small
     {
     \begin{tabular}{c|c|c|c}\hline
         Dataset &Property feature &Label &Pearson correlation\\\hline
         Pokec& \texttt{Gender}& \texttt{Public/private} &0.248\\\hline
         Facebook& \texttt{Gender}& \texttt{Education} &-0.01\\\hline
         Pubmed&\texttt{Keyword}& \texttt{Publication type} &0.107\\\hline
     \end{tabular}
     }
     \caption{\label{tab:label-correlation} Pearson correlation between property feature and label.} 
     \vspace{-0.2in}
 \end{table}

\section{Performance of Target Model}
\label{appendix:target}

\nop{
 \begin{table*}[t!]
     \centering
     \begin{tabular}{c|c|c|c|c|c|c}\hline
         \multirow{2}{*}{\backslashbox{GNN}{Dataset}}&\multicolumn{2}{c|}{Pokec} &\multicolumn{2}{c|}{Facebook} &\multicolumn{2}{c}{Pubmed}\\\cline{2-7}
         & Train acc. & Test acc. & Train acc. & Test acc. &Train acc. & Test acc. \\\hline
          GCN &0.69 & 0.66&0.78 & 0.72 & 0.87& 0.80\\\hline
          GraphSAGE &0.67 & 0.65& 0.7& 0.65 & 0.88& 0.80\\\hline
         GAT & 0.71& 0.67& 0.67& 0.64&0.84 & 0.79\\\hline
          
     \end{tabular}
     \caption{\label{tab:target-accuracy} Performance of the three GNN models. The performance is evaluated as the accuracy of node classification. \Wendy{Merge this table with Table \ref{tab:disp-target-model}}. Let me know if you want to discuss the structure of the merged table.} 
     \vspace{-0.2in}
 \end{table*}
}

\begin{table*}[t!]
    \centering
    {\begin{tabular}{c|c|c|c|c|c|c|c|c|c|c|c|c}\hline
  \multirow{3}{*}{\bf GNN models}&\multicolumn{6}{c|}{\bf Facebook dataset}&\multicolumn{6}{c}{\bf Pokec dataset}\\\cline{2-13}
   &\multicolumn{2}{c|}{Overall}&\multicolumn{2}{c|}{Male}&\multicolumn{2}{c|}{Female}&\multicolumn{2}{c|}{Overall}&\multicolumn{2}{c|}{Male}&\multicolumn{2}{c}{Female}\\\cline{2-13}      
   &Train &Test &Train  &Test &Train  &Test &Train  &Test &Train  &Test &Train &Test \\\hline
   GCN&0.78&0.72&0.83&0.79&0.73&0.69&0.69&0.66&0.68&0.64&0.7&0.68\\\hline
   GraphSAGE&0.7&0.65&0.76&0.69&0.65&0.6&0.67&0.65&0.66&0.68&0.69&0.68\\\hline
   GAT&0.67&0.64&0.71&0.67&0.61&0.59&0.74&0.67&0.71&0.67&0.72&0.69\\\hline
   \end{tabular}}
   \caption{\label{tab:disp-target-model}Disparity in target model accuracy across different groups in Facebook and Pokec datasets.}
   \end{table*}

\subsection{Model Accuracy over Whole Population} 
\label{appendix:gnn-performance}
Table \ref{tab:disp-target-model} (“Overall” column) shows the GNN classification performance results. All three GNN models have good node classification performance - the accuracy is much higher than that of random guess. Furthermore, all the three GNN models have good generalizability with small train-test accuracy gap.

\subsection{Disparity in Model Accuracy across Groups}
\label{appendix:disparity}
Table \ref{tab:disp-target-model} (“Male” and "Female" columns)  shows the disparity in target model accuracy across different groups on Facebook and Pokec datasets. We measure target model accuracy as the accuracy of node classification. Both datasets present accuracy disparity to some extent for both accuracy evaluation measurement. In particular, Facebook dataset shows significant disparity as high as 0.13 between male and female groups.

\section{Additional Results of Attack Performance}

\subsection{Accuracy of $A_3$ and $A_4$ against GraphSAGE and GAT}
\label{appendix:attack34-gs-gat}

\begin{figure*}[t!]
    \centering
\begin{tabular}{cccccc}
   
    \begin{subfigure}[b]{.15\textwidth}
      \centering
    \includegraphics[width=\textwidth]{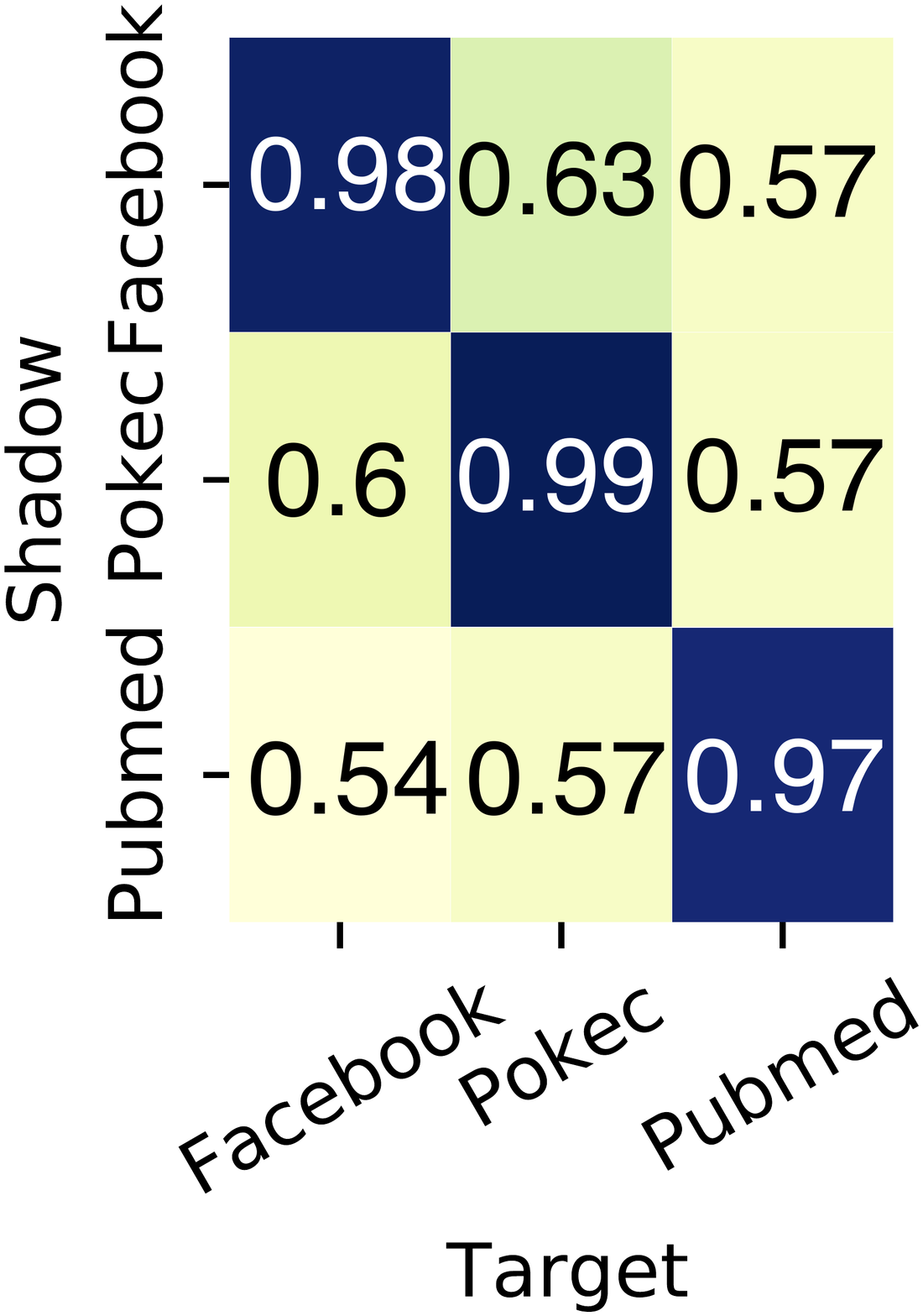}
     \vspace{-0.2in}
    \caption{Node property, $A_3^1$}
    \end{subfigure}
    &
    \begin{subfigure}[b]{.15\textwidth}
    \centering
    \includegraphics[width=\textwidth]{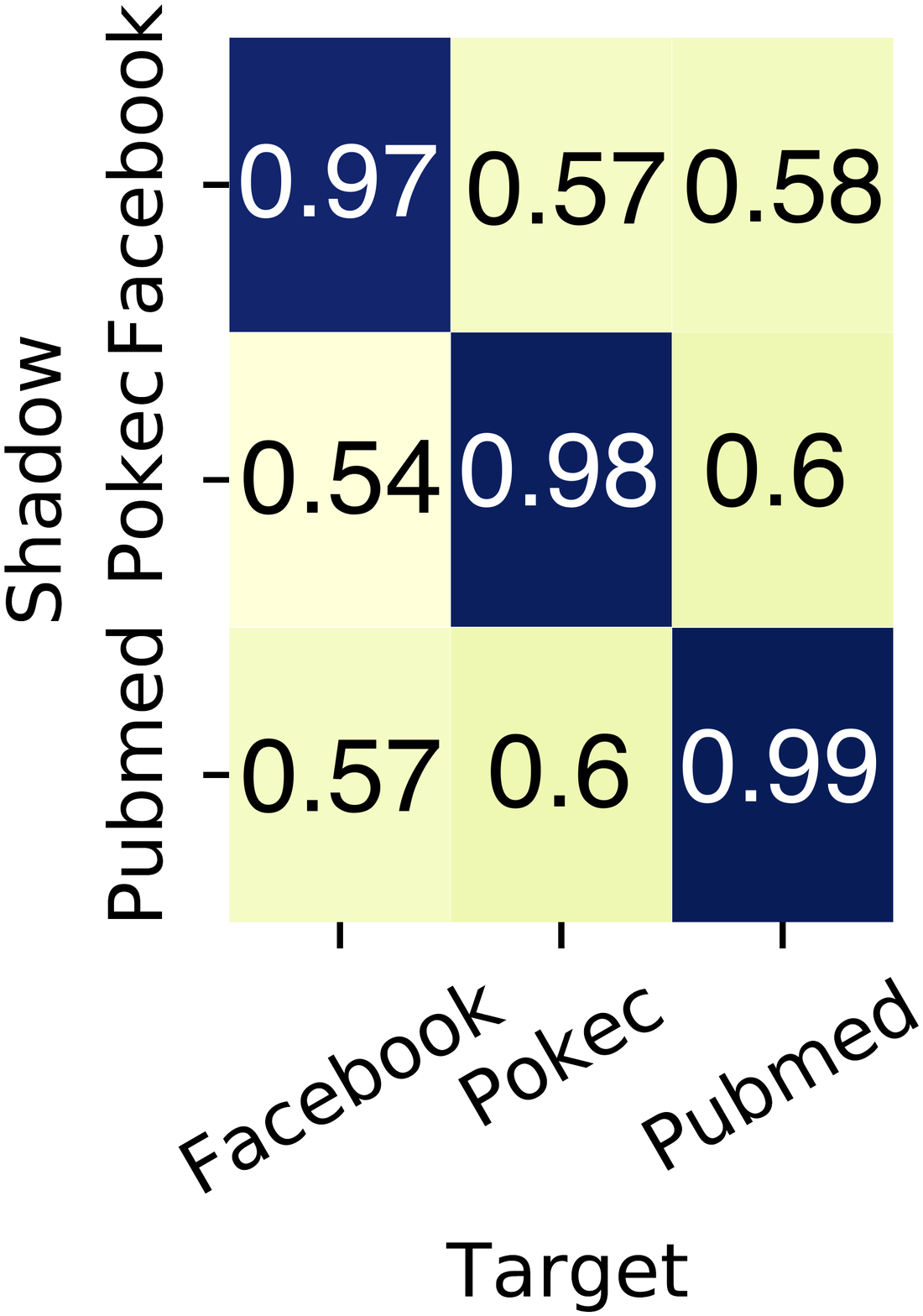}
    \vspace{-0.2in}
    \caption{Node property, $A_3^2$}
    \end{subfigure}
    &
    \begin{subfigure}[b]{.15\textwidth} 
    \centering
    \includegraphics[width=\textwidth]{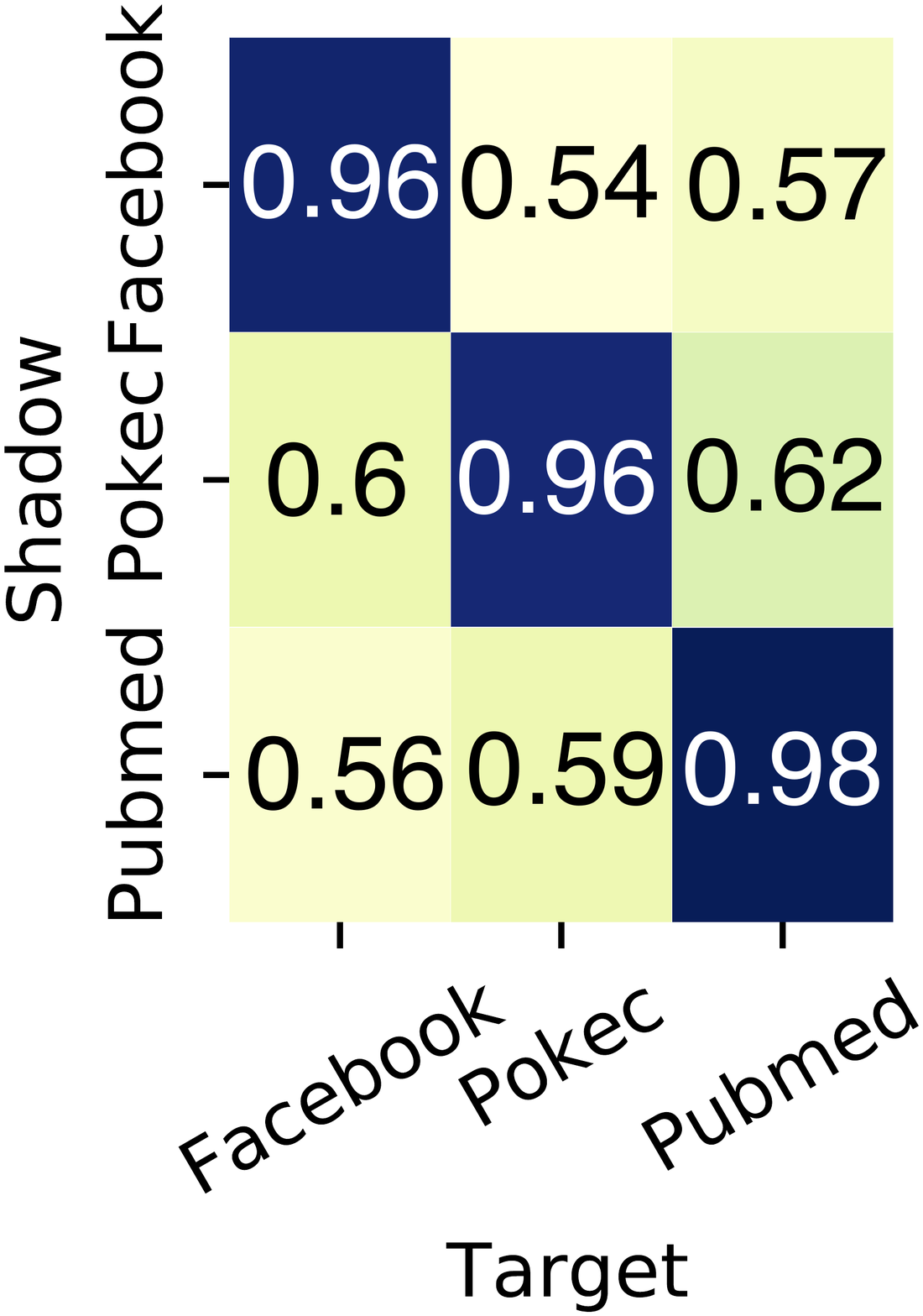}
    \vspace{-0.2in}
    \caption{Node property, $A_4$}
    \end{subfigure}
    &
        \begin{subfigure}[b]{.15\textwidth}
      \centering
    \includegraphics[width=\textwidth]{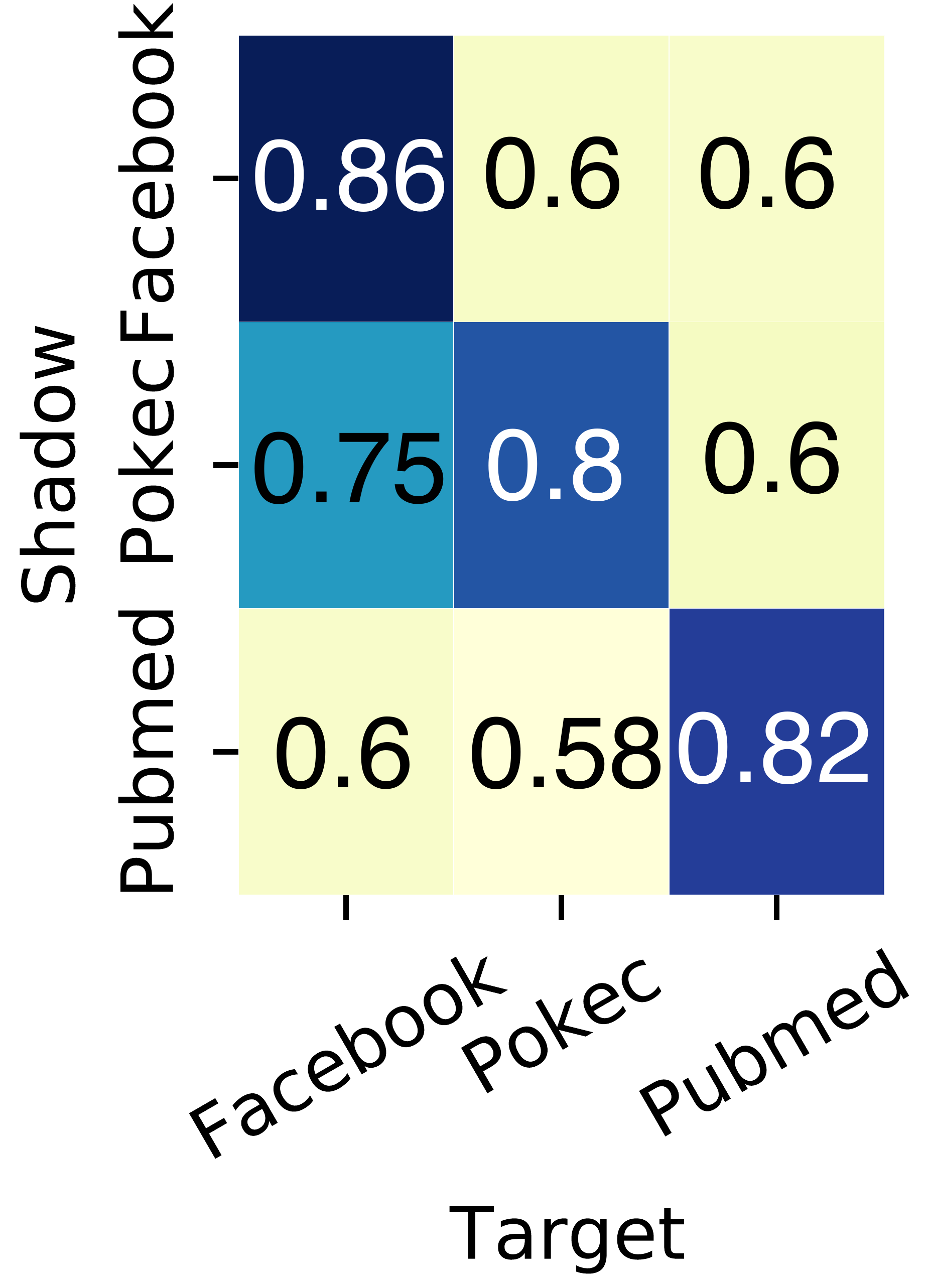}
     \vspace{-0.2in}
    \caption{Link property, $A_3^1$}
    \end{subfigure}
    &
    \begin{subfigure}[b]{.15\textwidth}
    \centering
    \includegraphics[width=\textwidth]{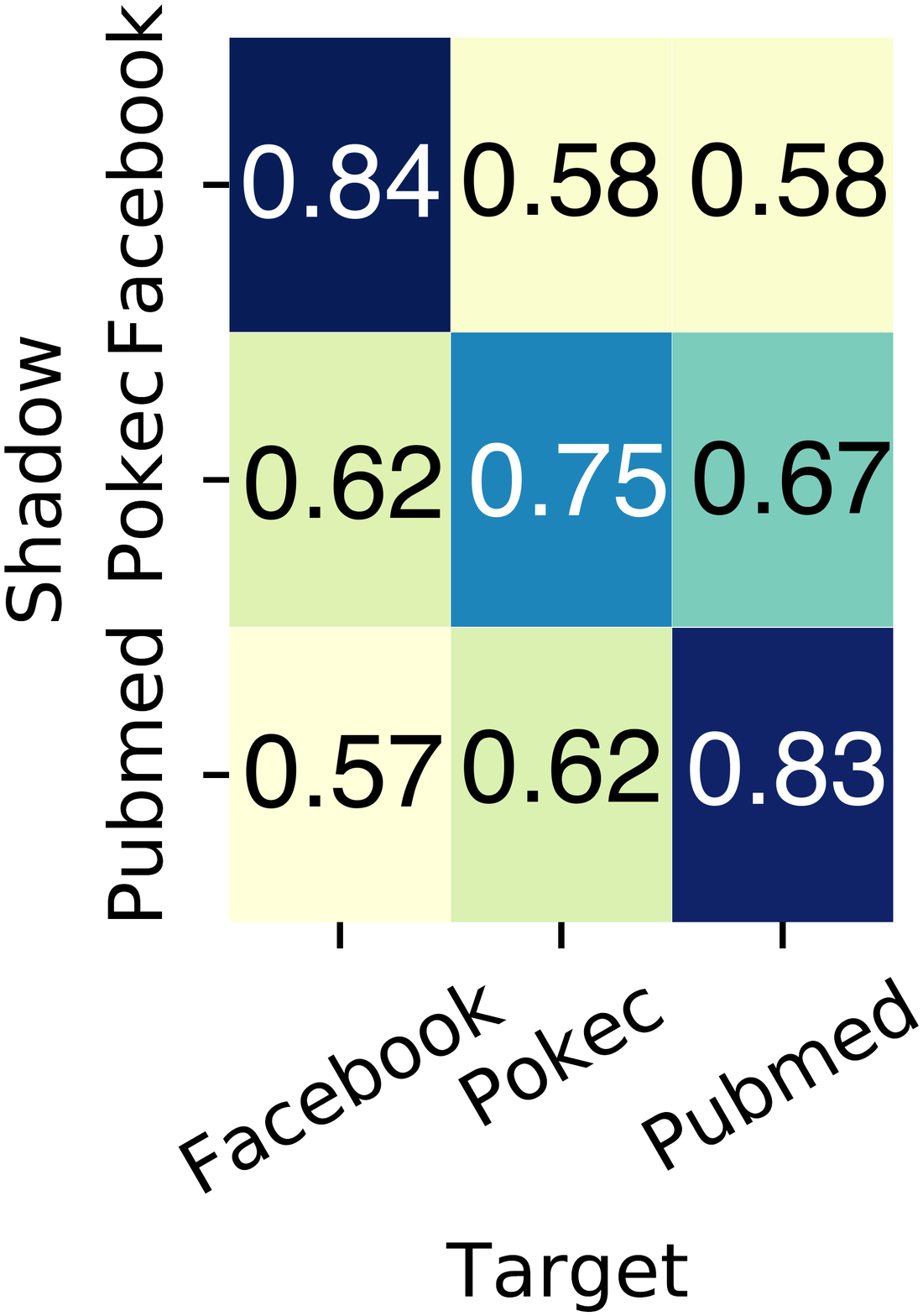}
    \vspace{-0.2in}
    \caption{Link property, $A_3^2$}
    \end{subfigure}
    &
    \begin{subfigure}[b]{.15\textwidth} 
    \centering
    \includegraphics[width=\textwidth]{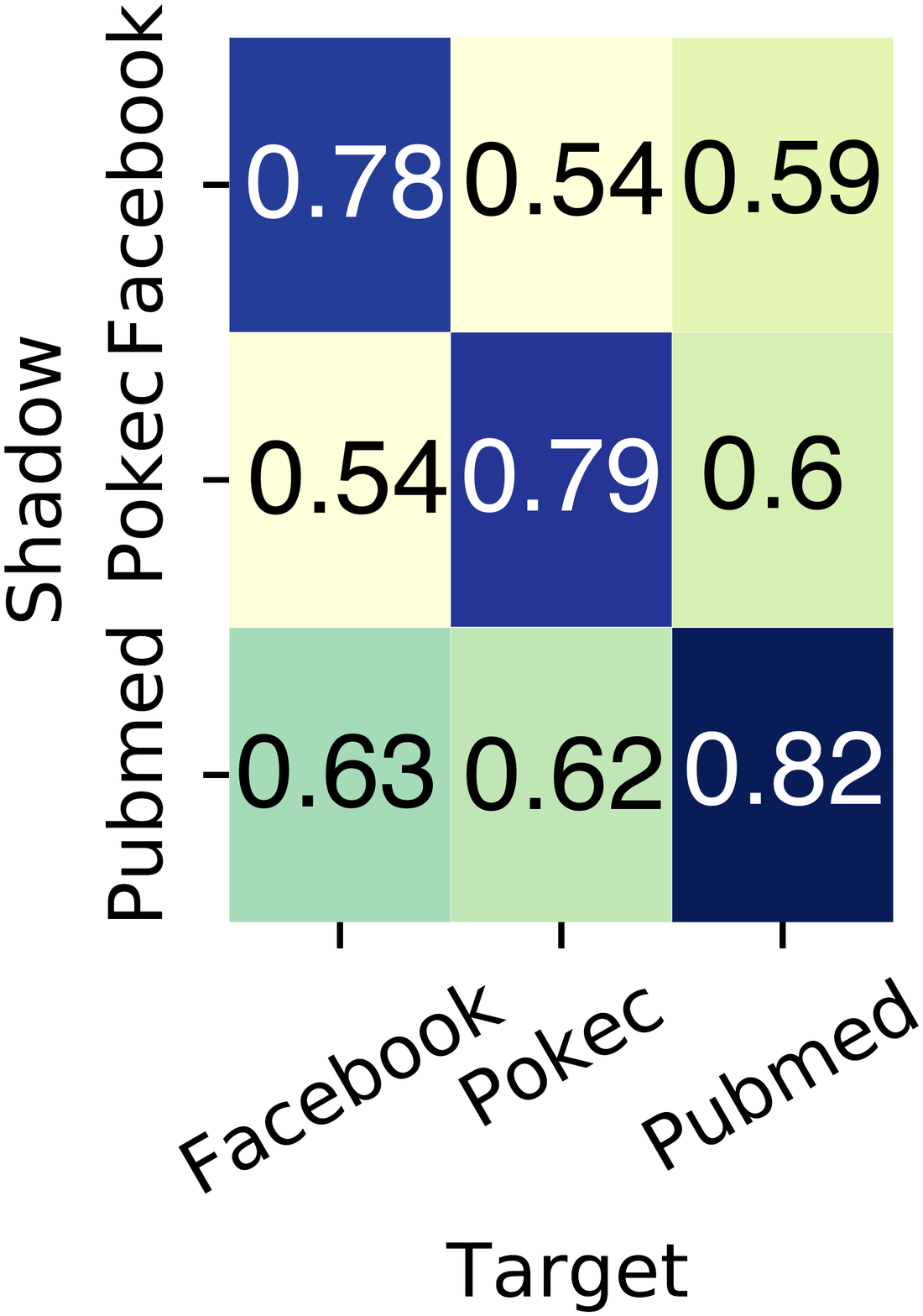}
    \vspace{-0.2in}
    \caption{Link property, $A_4$}
    \end{subfigure}
\end{tabular} 
    \vspace{0.2in}
{\bf GraphSAGE as target model}
    \\ 
    \begin{tabular}{cccccc}
   
    \begin{subfigure}[b]{.15\textwidth}
      \centering
    \includegraphics[width=\textwidth]{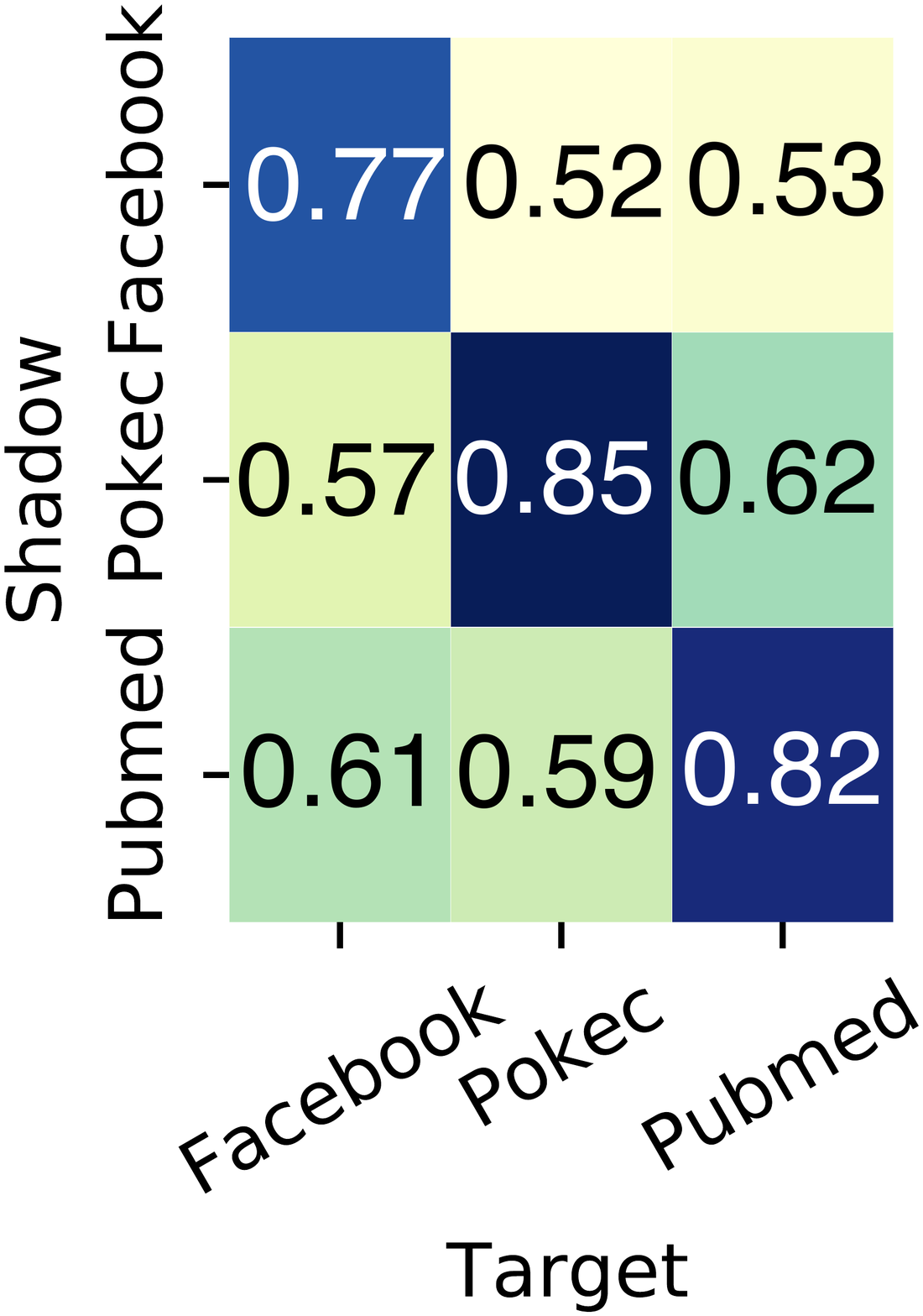}
     \vspace{-0.2in}
    \caption{Node property, $A_3^1$}
    \end{subfigure}
    &
    \begin{subfigure}[b]{.15\textwidth}
    \centering
    \includegraphics[width=\textwidth]{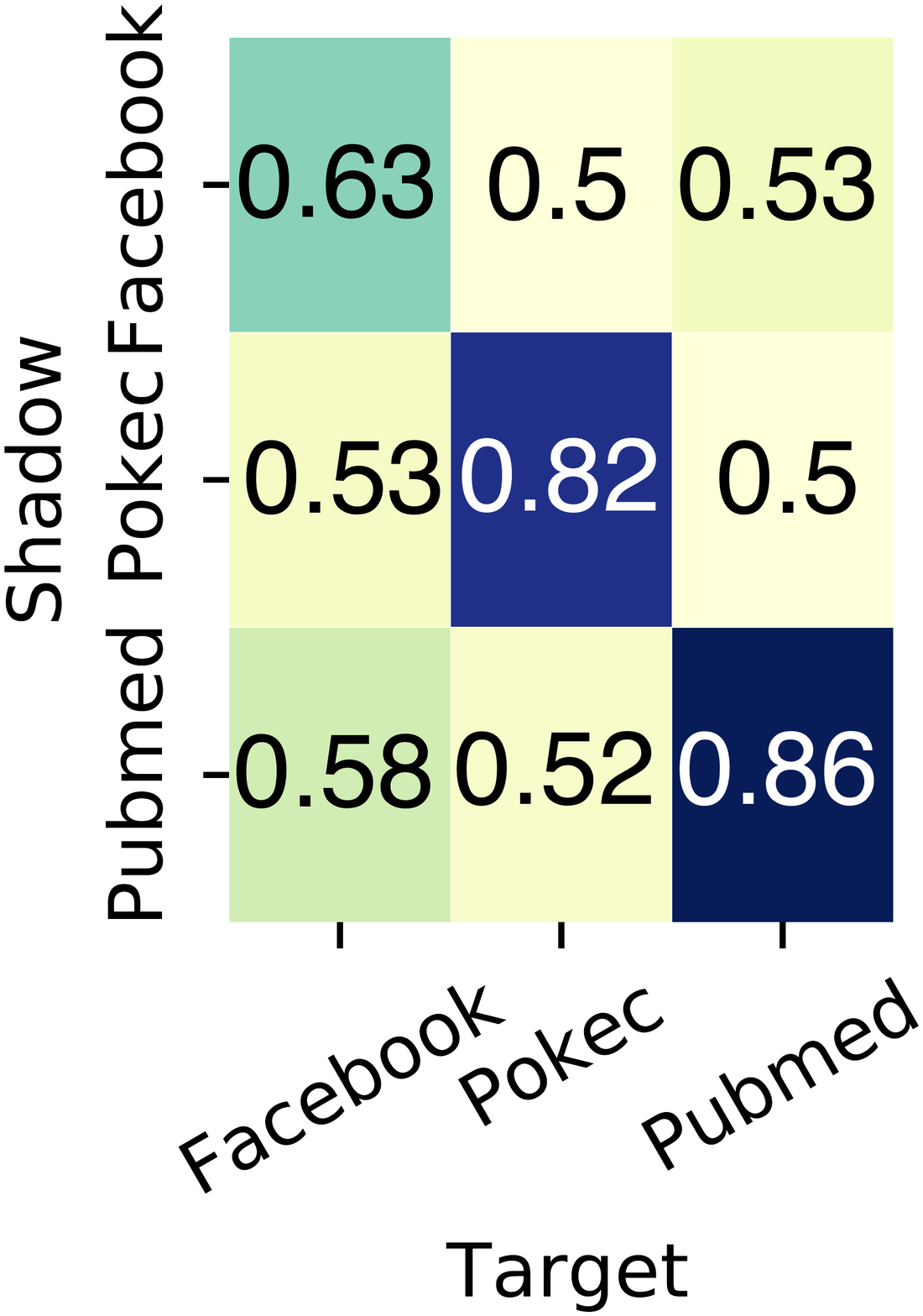}
    \vspace{-0.2in}
    \caption{Node property, $A_3^2$}
    \end{subfigure}
    &
    \begin{subfigure}[b]{.15\textwidth} 
    \centering
    \includegraphics[width=\textwidth]{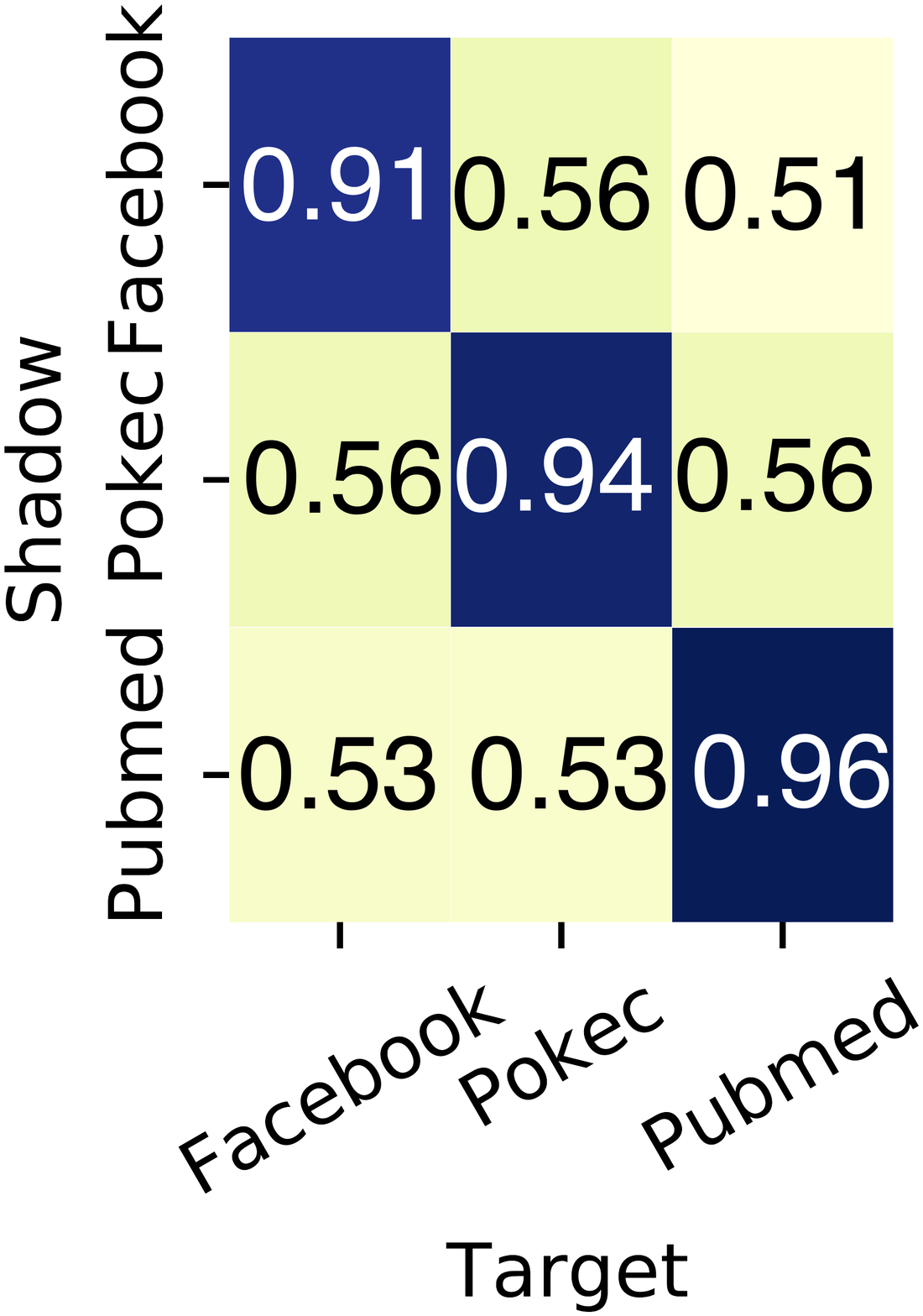}
    \vspace{-0.2in}
    \caption{Node property, $A_4$}
    \end{subfigure}
    &
        \begin{subfigure}[b]{.15\textwidth}
      \centering
    \includegraphics[width=\textwidth]{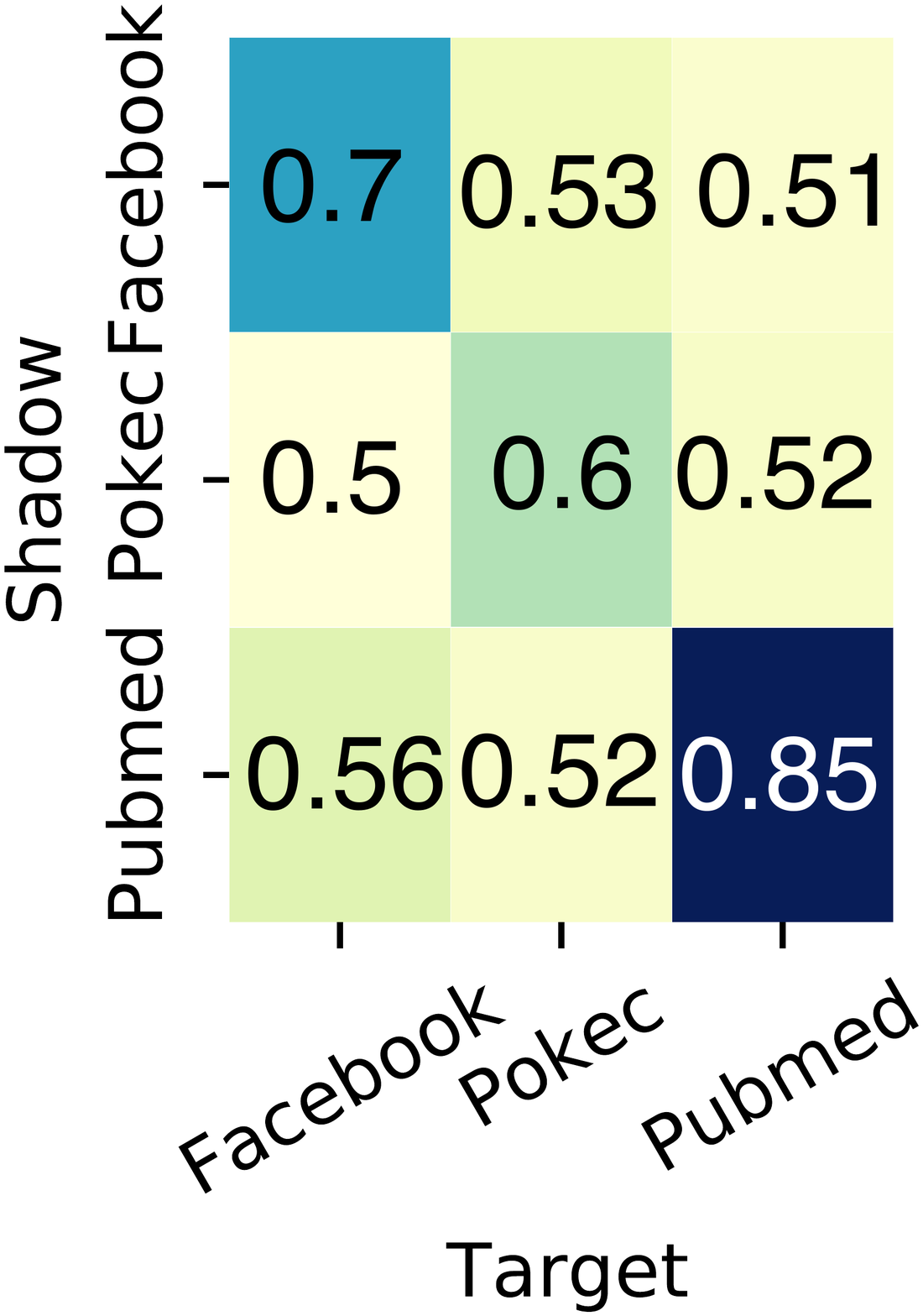}
     \vspace{-0.2in}
    \caption{Link property, $A_3^1$}
    \end{subfigure}
    &
    \begin{subfigure}[b]{.15\textwidth}
    \centering
    \includegraphics[width=\textwidth]{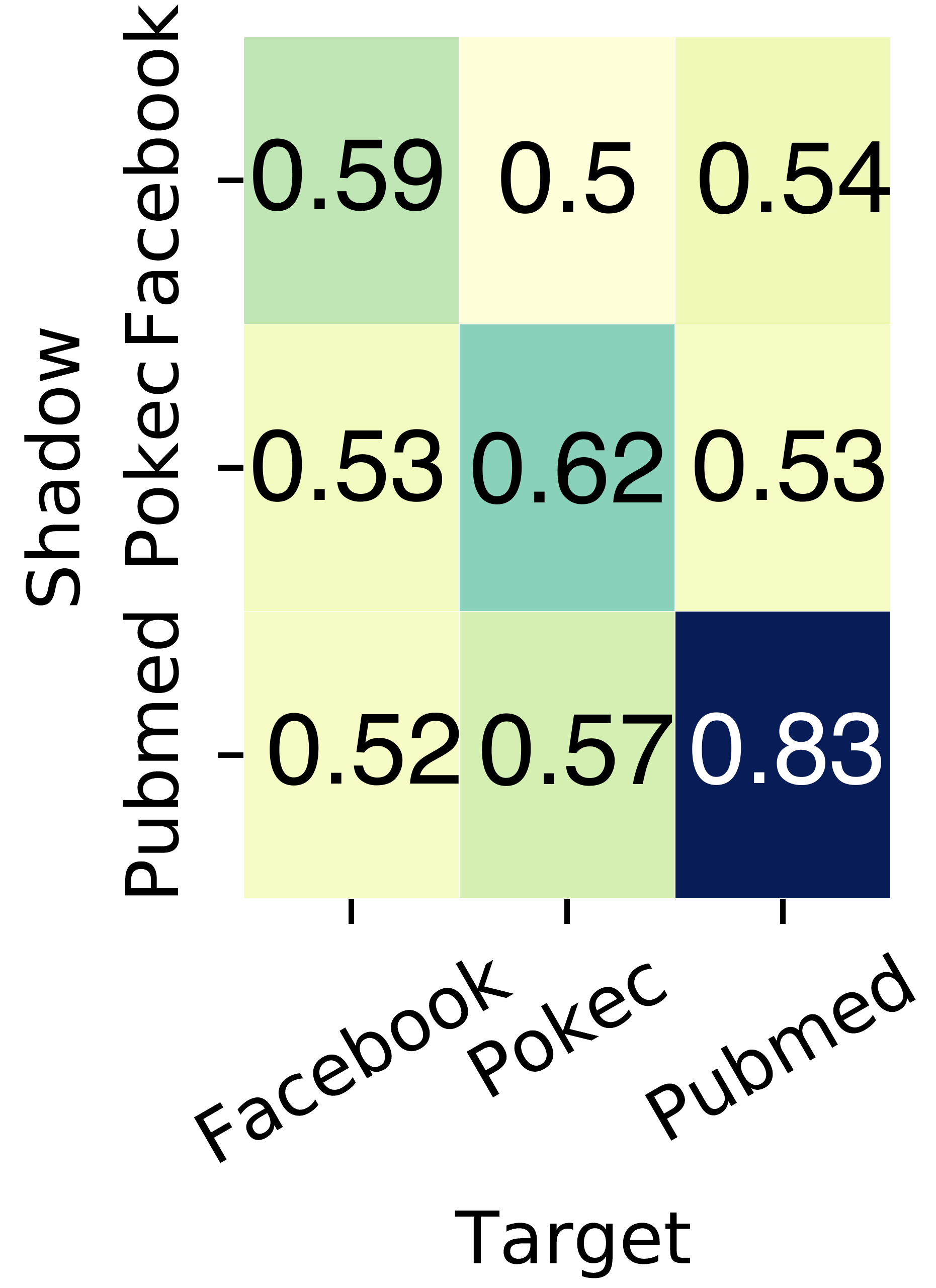}
    \vspace{-0.2in}
    \caption{Link property, $A_3^2$}
    \end{subfigure}
    &
    \begin{subfigure}[b]{.15\textwidth} 
    \centering
    \includegraphics[width=\textwidth]{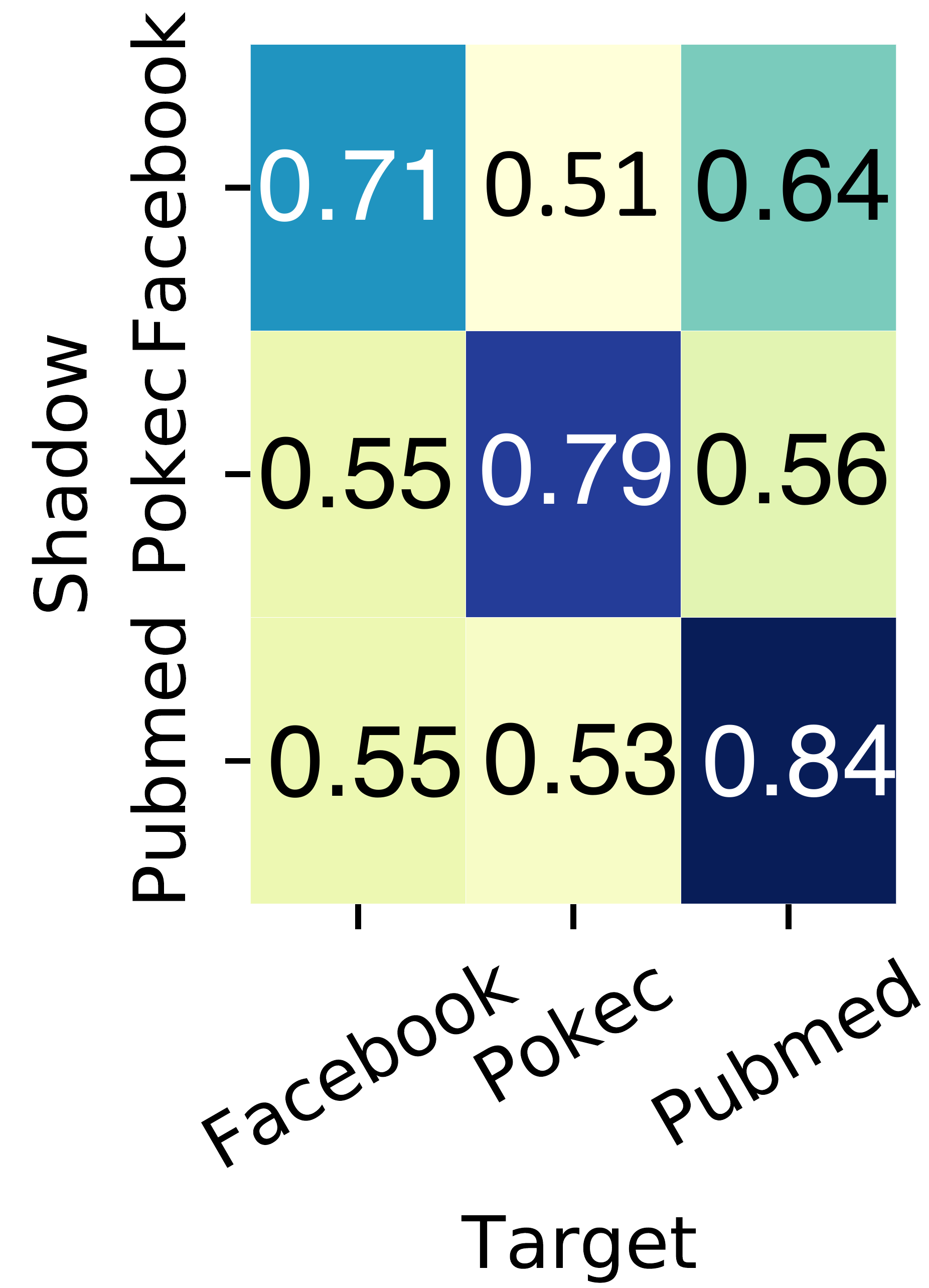}
    \vspace{-0.2in}
    \caption{Link property, $A_4$}
    \end{subfigure}
\end{tabular}  
    {\bf GAT as target model}
    \vspace{-0.05in}
\caption{\label{fig:attack34-acc-gs-gat} Attack accuracy of $A_3$ and $A_4$ when the GraphSAGE and GAT  are the target models. $A_3^1$ and $A_3^2$ indicate the $A_3$ attack that uses the model parameters at Layer 1 and Layer 2 of the target model respectively. } 
\end{figure*}
\begin{figure*}[t!]
    \centering
{\bf Target dataset: Pokec, Shadow dataset: Pubmed}
\\
\begin{subfigure}[b]{.6\textwidth}
      \centering
    \includegraphics[width=\textwidth]{text/figure/att56-pokec-legend11.pdf}\\
    \end{subfigure}
\begin{tabular}{ccc}
    \begin{subfigure}[b]{.3\textwidth}
      \centering
    \includegraphics[width=\textwidth]{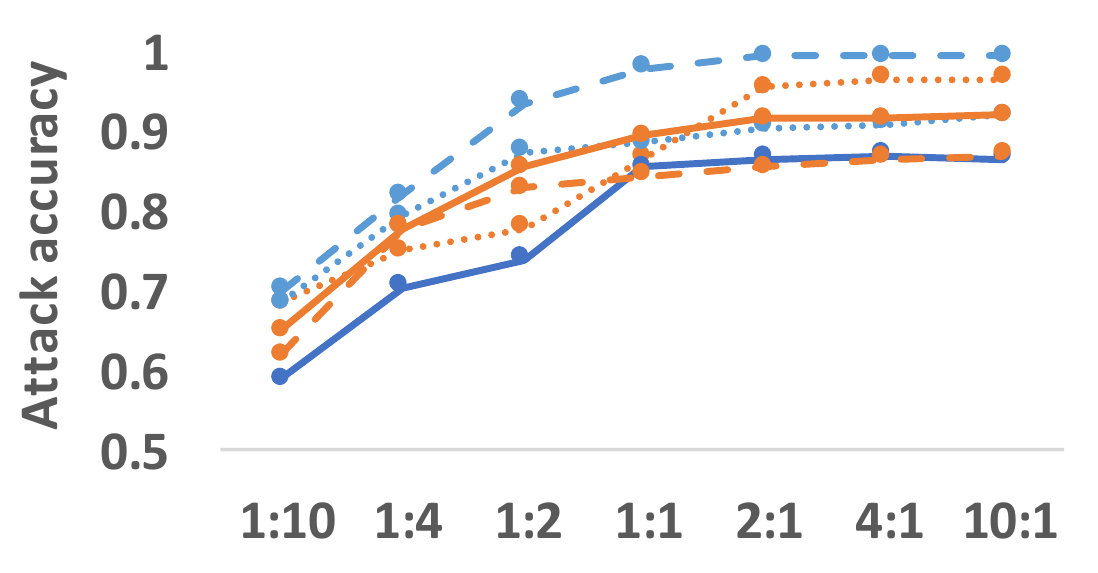}
     \vspace{-0.2in}
    \caption{GCN}
    \end{subfigure}
    &
    \begin{subfigure}[b]{.3\textwidth}
    \centering
    \includegraphics[width=\textwidth]{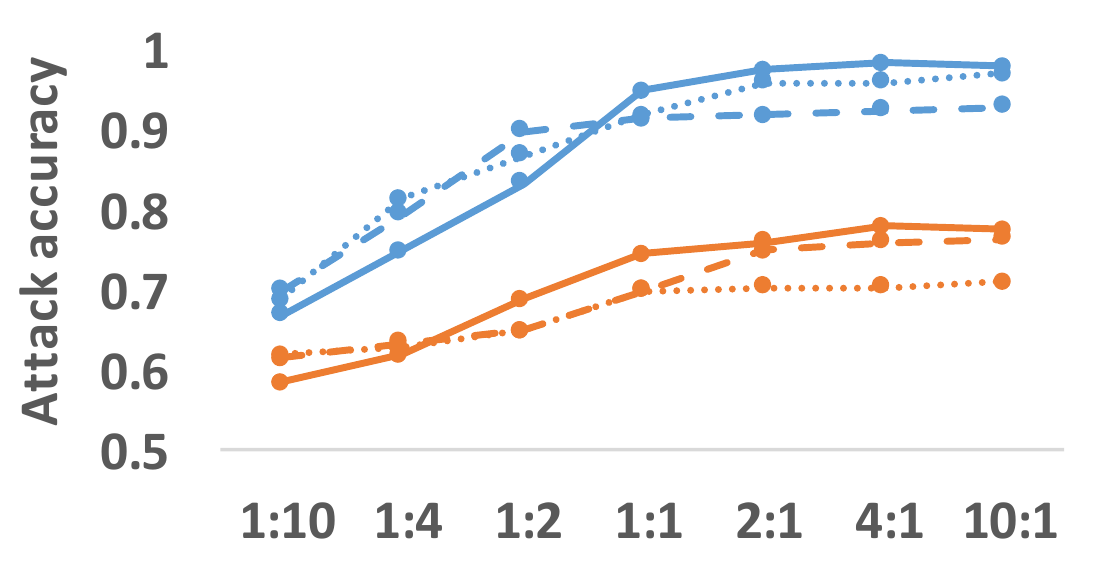}
    \vspace{-0.2in}
    \caption{GraphSAGE}
    \end{subfigure}
    &
    \begin{subfigure}[b]{.3\textwidth} 
    \centering
    \includegraphics[width=\textwidth]{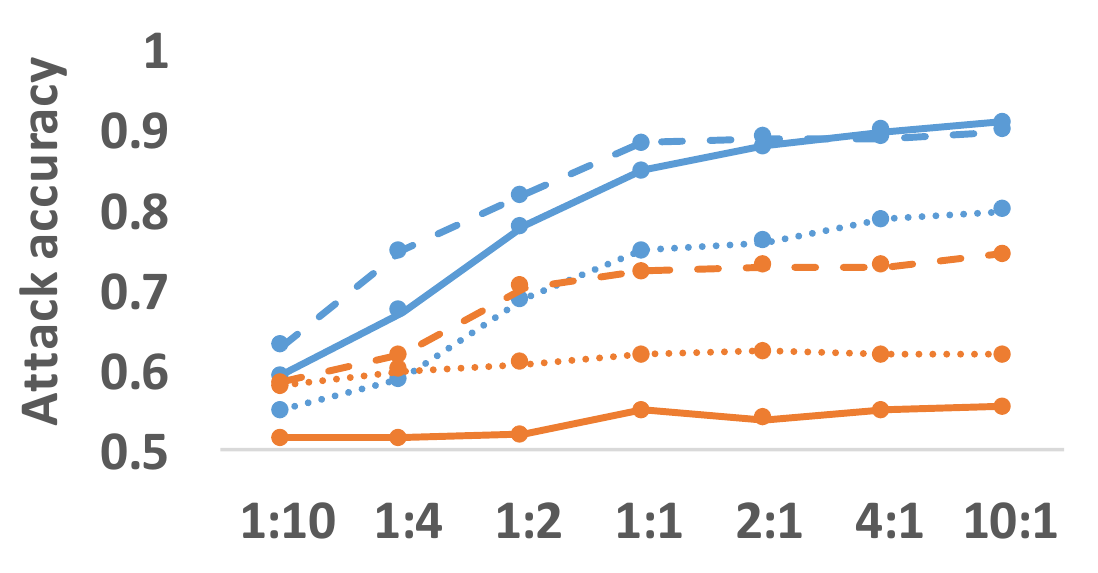}
    \vspace{-0.2in}
    \caption{GAT}
    \end{subfigure}
    \end{tabular} 
    \\
    \vspace{.1in}
    {\bf Target dataset: Facebook, Shadow dataset: Pubmed}
    \\
    \begin{subfigure}[b]{.6\textwidth}
      \centering
    \includegraphics[width=\textwidth]{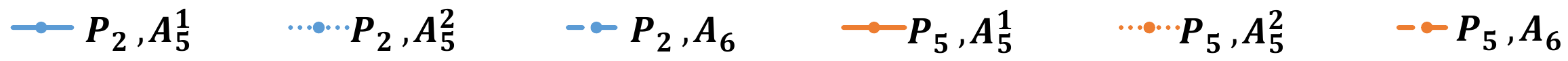}\\
    \end{subfigure}
    \begin{tabular}{ccc}
    \begin{subfigure}[b]{.3\textwidth}
      \centering
    \includegraphics[width=\textwidth]{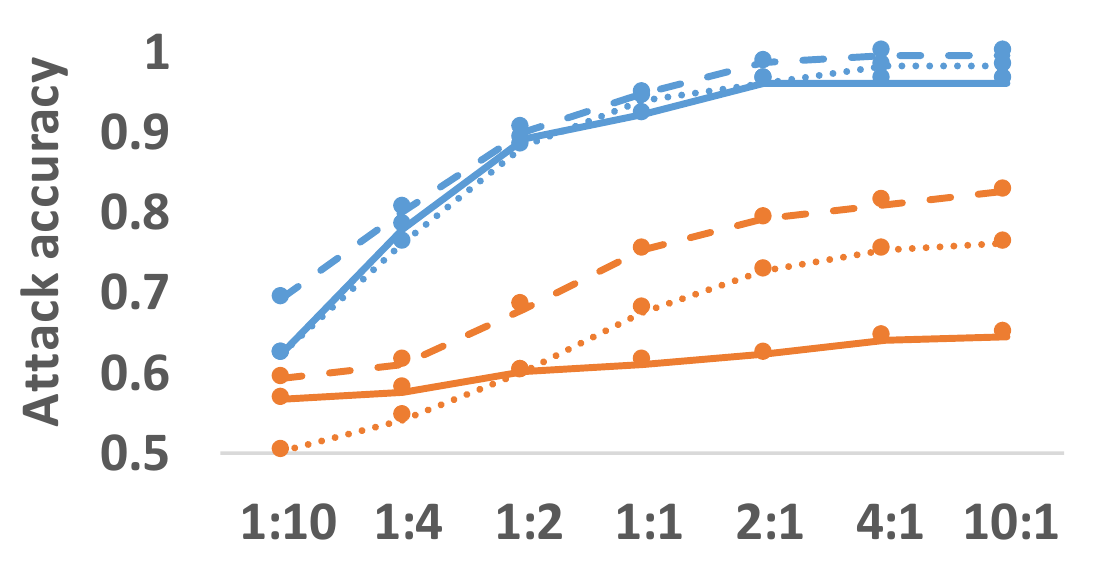}
     \vspace{-0.2in}
    \caption{GCN}
    \end{subfigure}
    &
    \begin{subfigure}[b]{.3\textwidth}
    \centering
    \includegraphics[width=\textwidth]{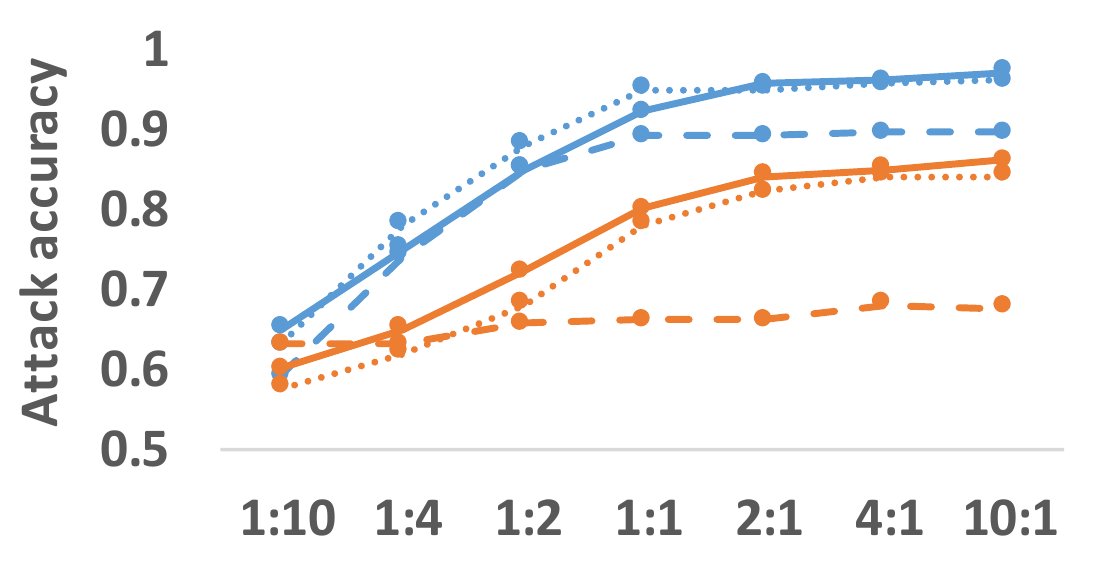}
    \vspace{-0.2in}
    \caption{GraphSAGE}
    \end{subfigure}
    &
    \begin{subfigure}[b]{.3\textwidth} 
    \centering
    \includegraphics[width=\textwidth]{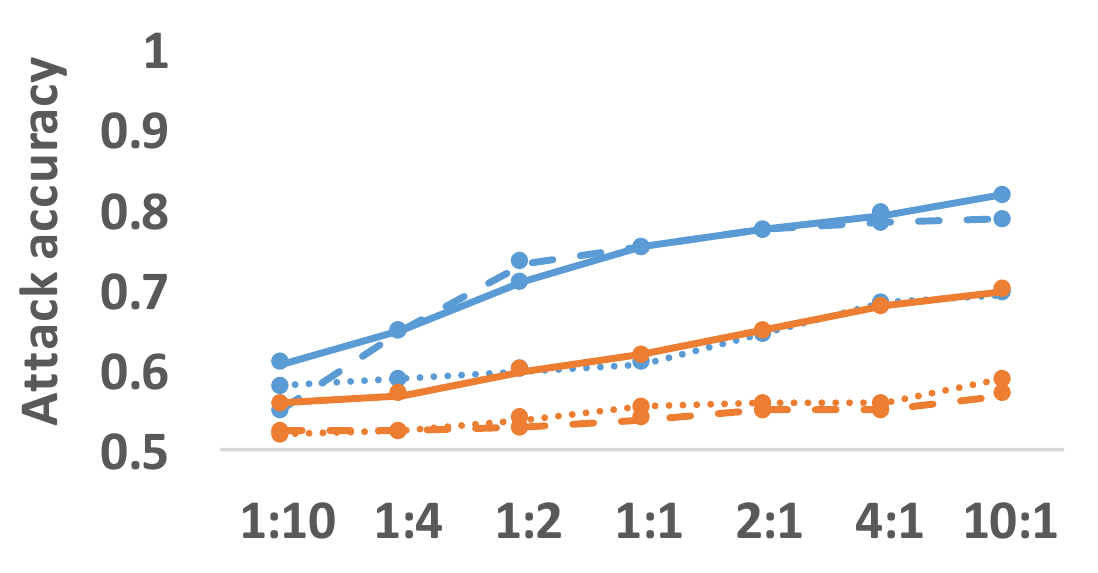}
    \vspace{-0.2in}
    \caption{GAT}
    \end{subfigure}
    
\end{tabular} 
    \\
    \vspace{.1in}
    {\bf Target dataset: Pubmed, Shadow dataset: Pokec}
    \\
    \begin{subfigure}[b]{.6\textwidth}
      \centering
    \includegraphics[width=\textwidth]{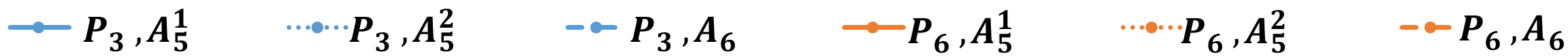}\\
    \end{subfigure}
    \begin{tabular}{ccc}
    \begin{subfigure}[b]{.3\textwidth}
      \centering
    \includegraphics[width=\textwidth]{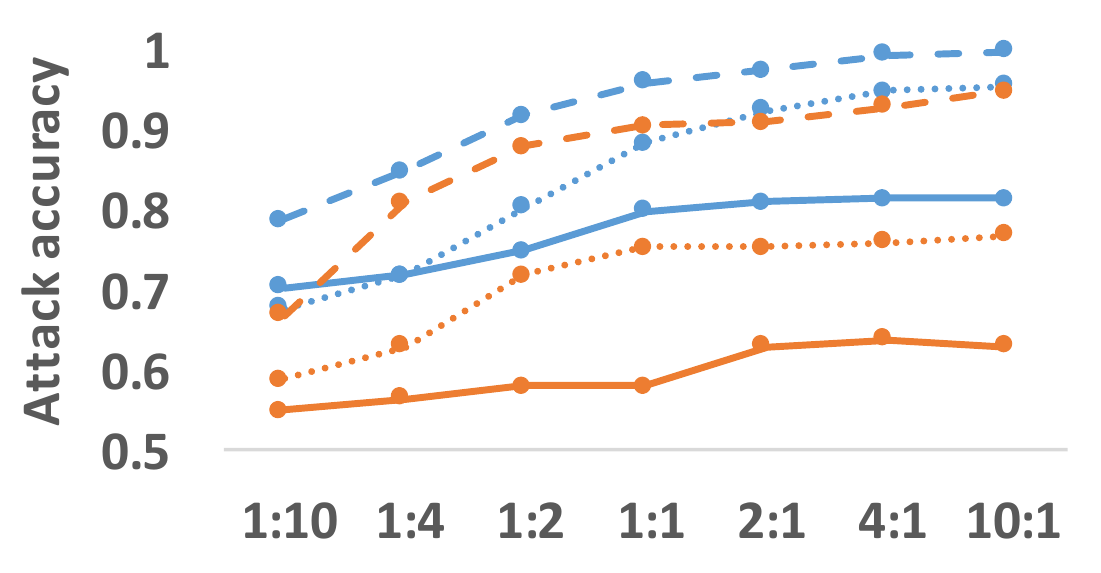}
     \vspace{-0.2in}
    \caption{GCN}
    \end{subfigure}
    &
    \begin{subfigure}[b]{.3\textwidth}
    \centering
    \includegraphics[width=\textwidth]{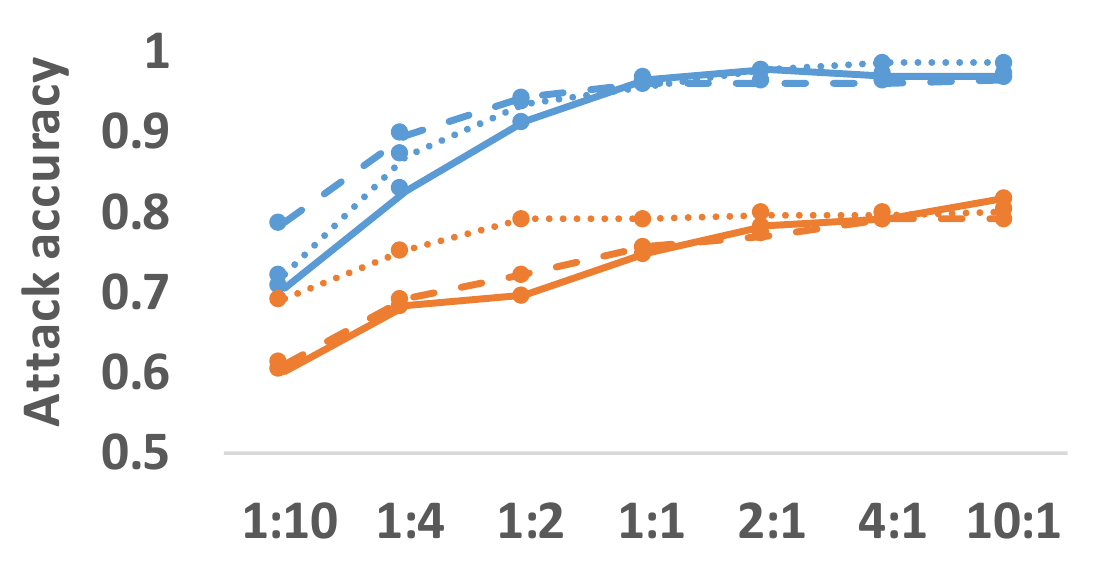}
    \vspace{-0.2in}
    \caption{GraphSAGE}
    \end{subfigure}
    &
    \begin{subfigure}[b]{.3\textwidth} 
    \centering
    \includegraphics[width=\textwidth]{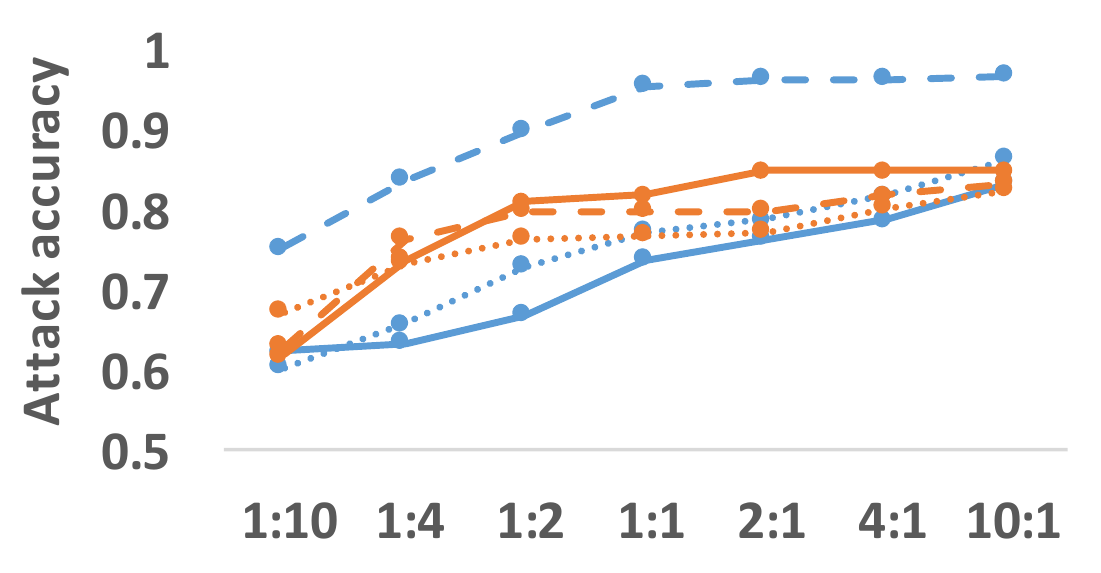}
    \vspace{-0.2in}
    \caption{GAT}
    \end{subfigure}
\end{tabular}    
    \\
    \vspace{.1in}
    {\bf Target dataset: Pubmed, Shadow dataset: Facebook}
    \\
    \begin{subfigure}[b]{.6\textwidth}
      \centering
    \includegraphics[width=\textwidth]{text/figure/att56-pubmed-legend11.pdf}\\
    \end{subfigure}
    \begin{tabular}{ccc}
    \begin{subfigure}[b]{.3\textwidth}
      \centering
    \includegraphics[width=\textwidth]{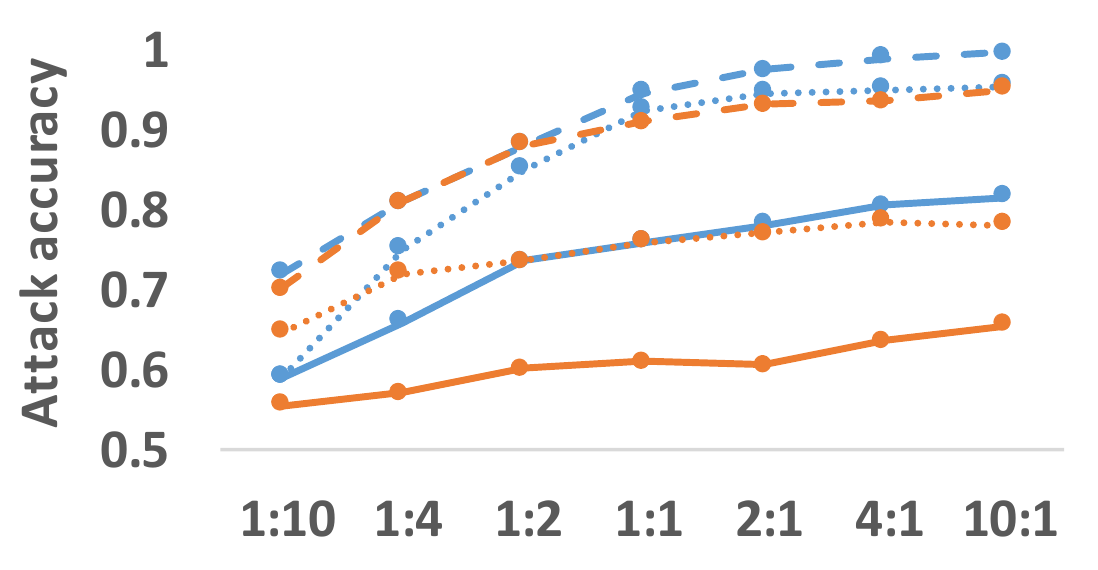}
     \vspace{-0.2in}
    \caption{GCN}
    \end{subfigure}
    &
    \begin{subfigure}[b]{.3\textwidth}
    \centering
    \includegraphics[width=\textwidth]{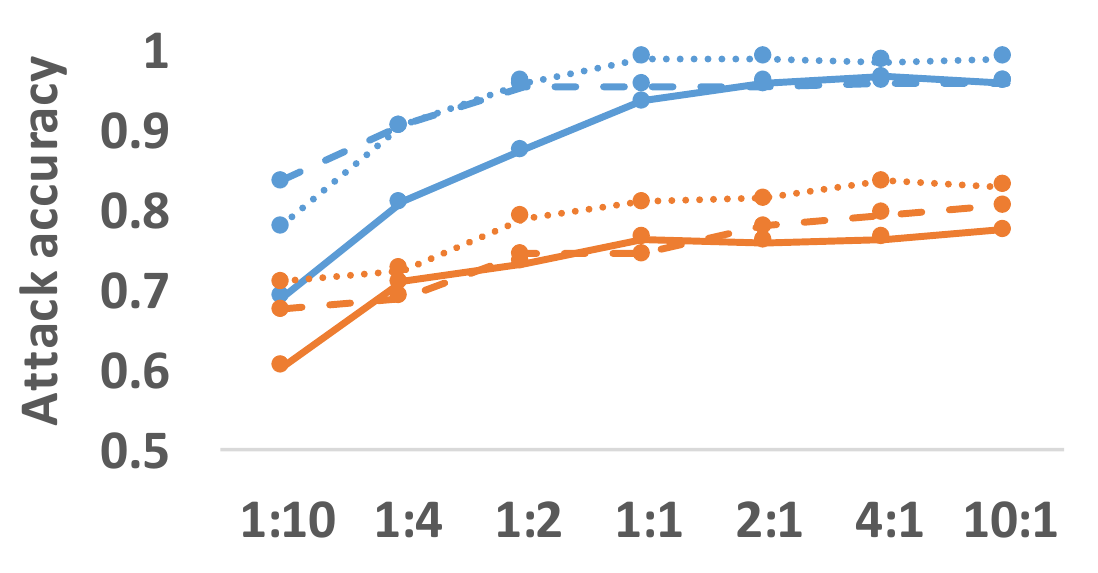}
    \vspace{-0.2in}
    \caption{GraphSAGE}
    \end{subfigure}
    &
    \begin{subfigure}[b]{.3\textwidth} 
    \centering
    \includegraphics[width=\textwidth]{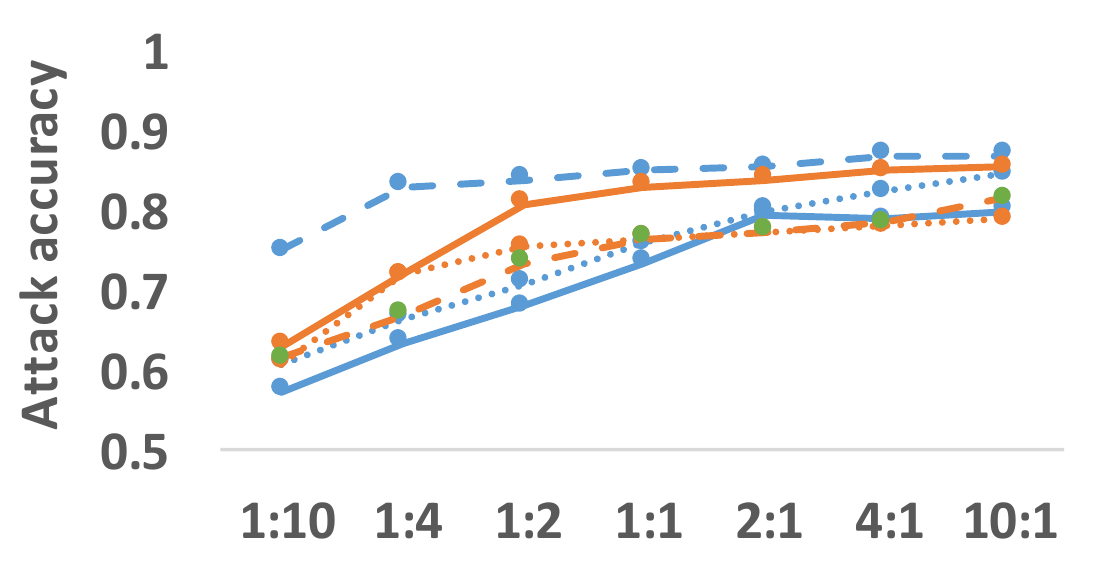}
    \vspace{-0.2in}
    \caption{GAT}
    \end{subfigure}
\end{tabular}    
    \\
    \vspace{.1in}
    {\bf Target dataset: Facebook, Shadow dataset: Pokec}
    \\
    \begin{subfigure}[b]{.6\textwidth}
      \centering
    \includegraphics[width=\textwidth]{text/figure/att56-fb-legend11.pdf}\\
    \end{subfigure}
    \begin{tabular}{ccc}
    \begin{subfigure}[b]{.3\textwidth}
      \centering
    \includegraphics[width=\textwidth]{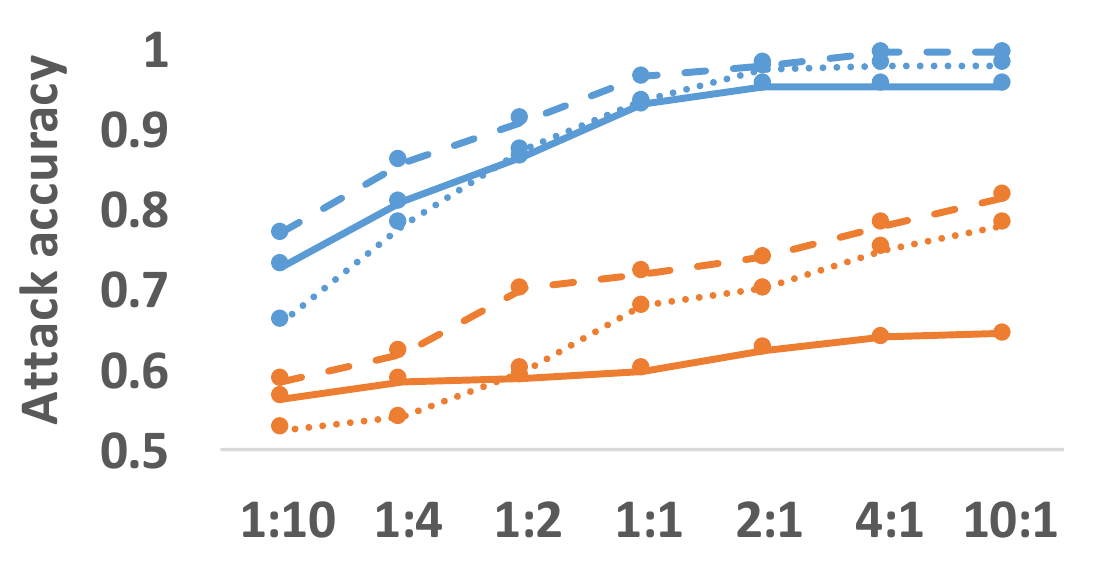}
     \vspace{-0.2in}
    \caption{GCN}
    \end{subfigure}
    &
    \begin{subfigure}[b]{.3\textwidth}
    \centering
    \includegraphics[width=\textwidth]{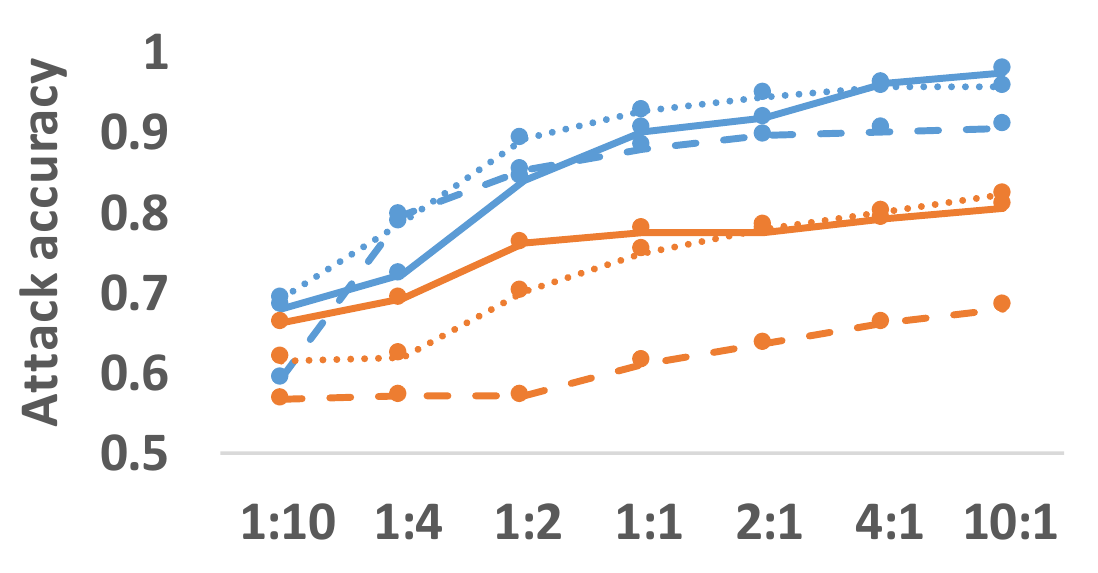}
    \vspace{-0.2in}
    \caption{GraphSAGE}
    \end{subfigure}
    &
    \begin{subfigure}[b]{.3\textwidth} 
    \centering
    \includegraphics[width=\textwidth]{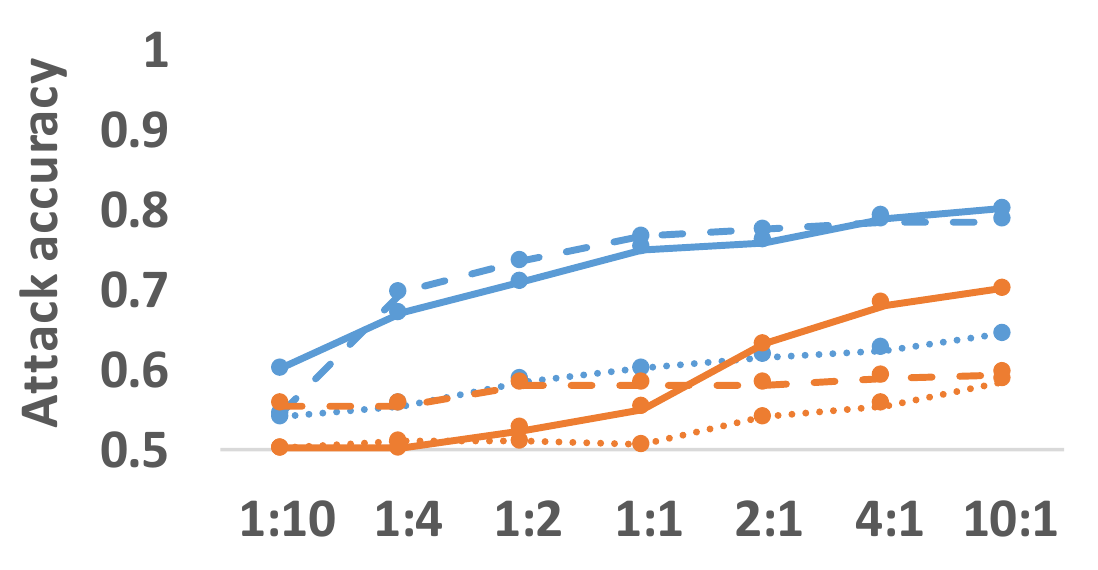}
    \vspace{-0.2in}
    \caption{GAT}
    \end{subfigure}
\end{tabular}    
    \vspace{-0.05in}
\caption{\label{fig:attack56-acc2} Attack accuracy of $A_5$ and $A_6$. X-axis shows the size ratio between partial and shadow graphs. 
Node-level properties and link-level properties are indicated in \textcolor{blue}{blue} and \textcolor{orange}{orange} colors respectively, while $A^1_{5}$, $A^2_{5}$, and $A_{6}$ are indicated in solid, dotted, and dashed lines respectively.} 
\end{figure*}

Figure \ref{fig:attack34-acc-gs-gat} shows attack accuracy of $A_3$ and $A_4$ against GraphSAGE and GAT models. The observation is similar to Figure \ref{fig:attack34-acc} and thus the discussions are omitted.

\subsection{Accuracy of $A_5$ and $A_6$}
\label{appendix:attack56-2}
Figure \ref{fig:attack56-acc2} shows attack accuracy of $A_5$ and $A_6$ when partial graph is sampled from Pokec dataset and shadow graph is from Pubmed dataset. The observation is similar to Figure \ref{fig:attack56-acca-pokec-fb} and thus the discussions are omitted.

\subsection{Additional Baseline Methods}
\label{appendix:more-baseline}

\begin{figure*}[t!]
    \centering
\begin{subfigure}[b]{.6\textwidth}
      \centering
    \includegraphics[width=\textwidth]{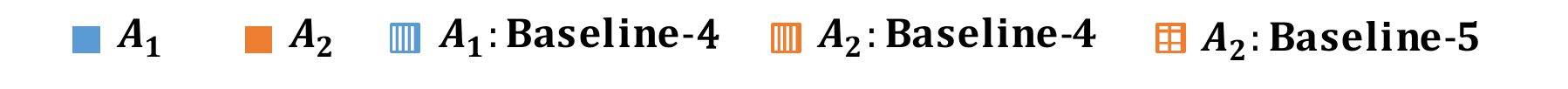}
     \vspace{-0.2in}
    \end{subfigure}\\
\begin{tabular}{ccc}
    \begin{subfigure}[b]{.32\textwidth}
      \centering
    \includegraphics[width=\textwidth]{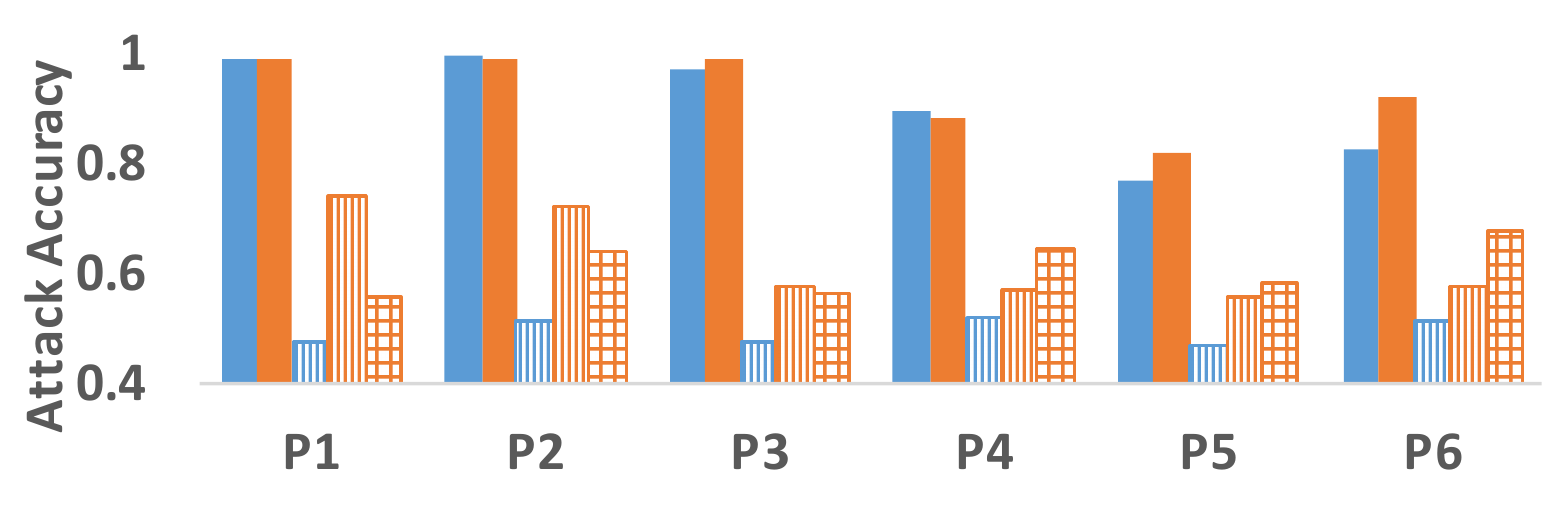}
     \vspace{-0.2in}
    \caption{GCN}
    \end{subfigure}
    &
    \begin{subfigure}[b]{.32\textwidth}
    \centering
    \includegraphics[width=\textwidth]{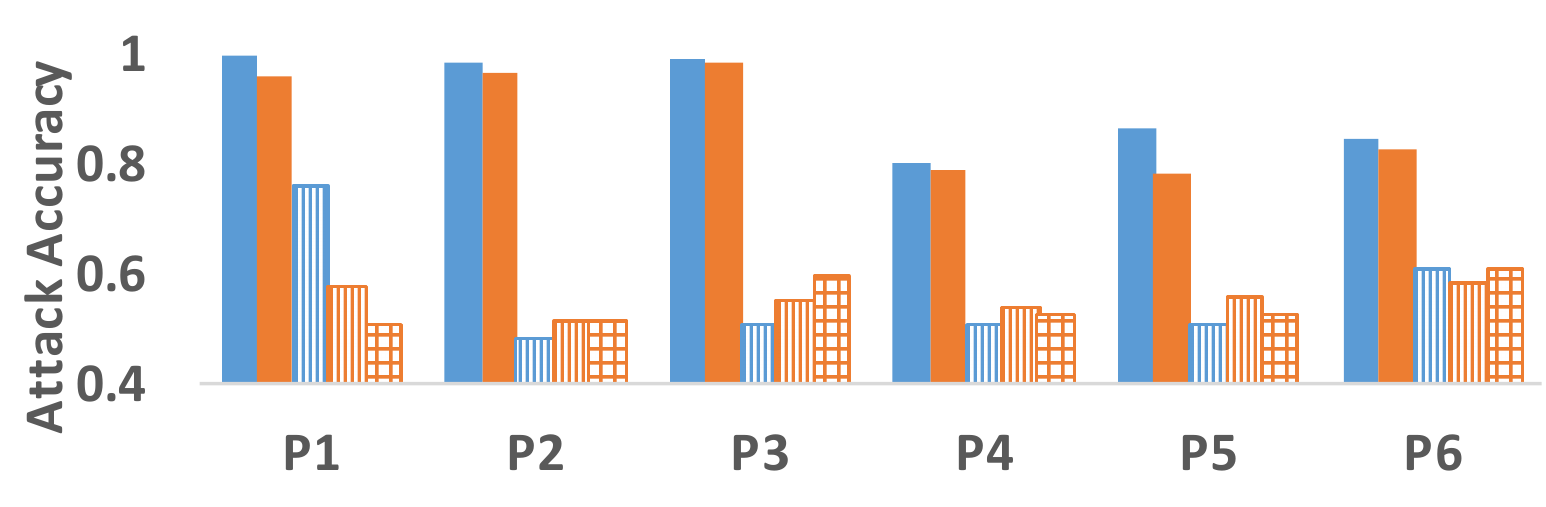}
    \vspace{-0.2in}
    \caption{GraphSAGE}
    \end{subfigure}
    &
    \begin{subfigure}[b]{.32\textwidth} 
    \centering
    \includegraphics[width=\textwidth]{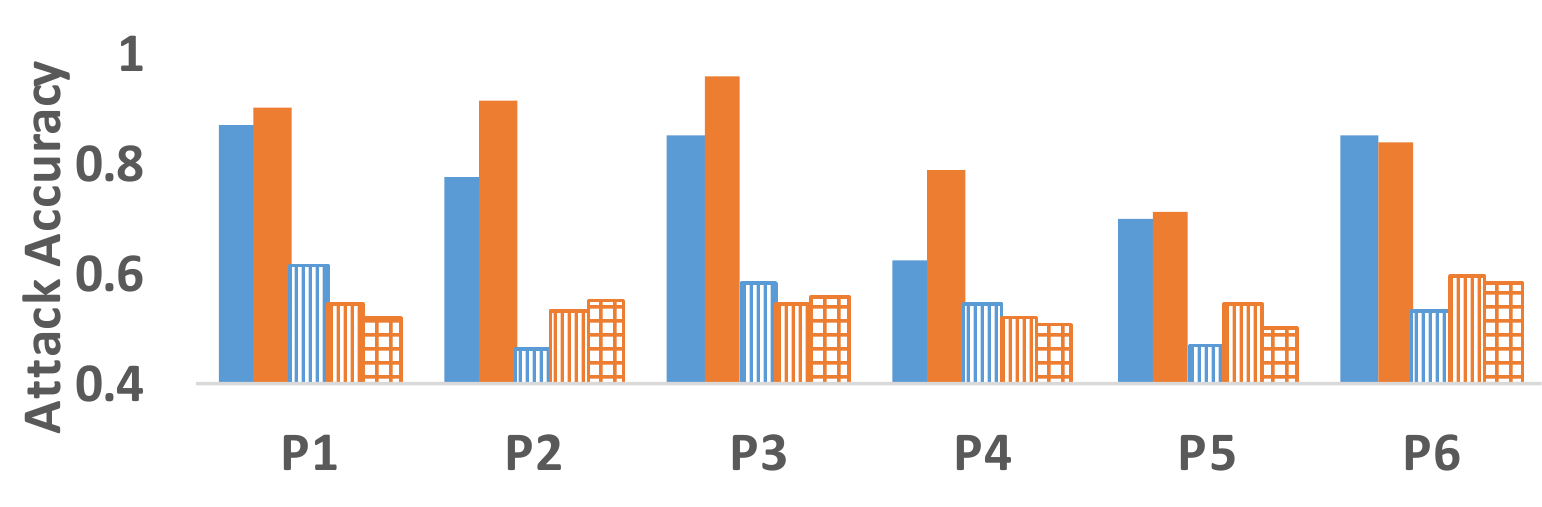}
    \vspace{-0.2in}
    \caption{GAT}
    \end{subfigure}
\end{tabular}    
    \vspace{-0.2in}
\caption{\label{fig:pia_acc_more_baseline} Attack accuracy of $A_1$ and $A_2$ by our attack and two baseline methods (Baseline-4 \& Baseline-5). $A_1$ and $A_2$ are indicated in different colors respectively, while our approaches, Baseline-4 and Baseline-5 are indicated in vertical fill and grid shape fill respectively.}
\end{figure*}

Besides the three baseline methods in Section \ref{sc:exp-pia}, we consider another two threshold-based methods: (1) {\bf Auxiliary summarization (Baseline-4)}: We adapt the Directly Summarizing Auxiliary Dataset (DSAD) method \cite{zhang2021inference} to our setting as the baseline. We use a threshold of the average property values summarized from the adversary knowledge (e.g., partial graph) to predict the property of the target graph instead of training a classifier. We choose the threshold that has the best attack performance; (2) {\bf Loss-gap threshold (Baseline-5)}: First, we calculate the difference in training and testing loss (training-testing gap) for each shadow graph, then pick a threshold that returns the best attack accuracy. We use the picked threshold to predict the property of target graphs. The loss gap that is higher than the threshold will be determined as with the property. The attack accuracy of Baseline-4 and Baseline-5 are shown in Figure \ref{fig:pia_acc_more_baseline}. We observe that the attack accuracy of $A_1$ and $A_2$ of our GPIA is significantly higher than both Baseline-4 and Baseline-5. 


\nop{
\section{TSNE visualization of the distributions of different inputs of PIA model on different tasks (Pokec dataset, GraphSAGE and GAT)}
\label{appendix:dis_pokec2}
\xiuling{Figure \ref{fig:dis_pokec2} is the TSNE visualization of different inputs of PIA model for $P_1$ and $P_4$ on GraphSAGE and GAT. The observation is similar to Figure \ref{fig:dis_pokec} and thus the discussions are omitted.}
\begin{figure*}[t!]
\centering
    \centering
    {\bf GNN model: GraphSAGE}
    \\
    \begin{subfigure}[b]{.15\textwidth}
      \centering
     \includegraphics[width=\textwidth]{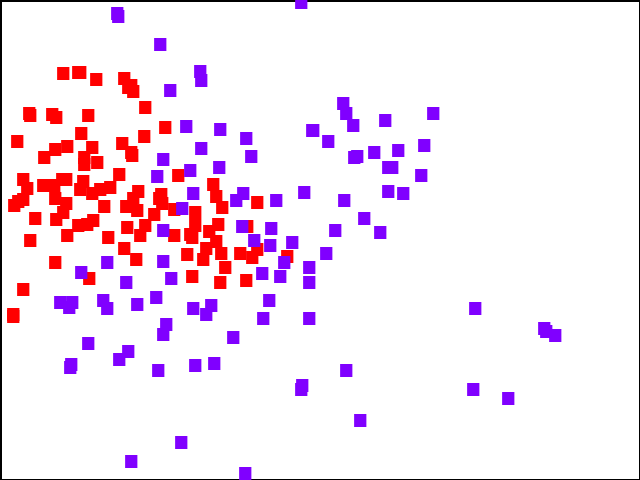}
     \vspace{-0.2in}
    \caption{$T_1,A_1^1$}
    \end{subfigure}
     \begin{subfigure}[b]{.15\textwidth}
      \centering
     \includegraphics[width=\textwidth]{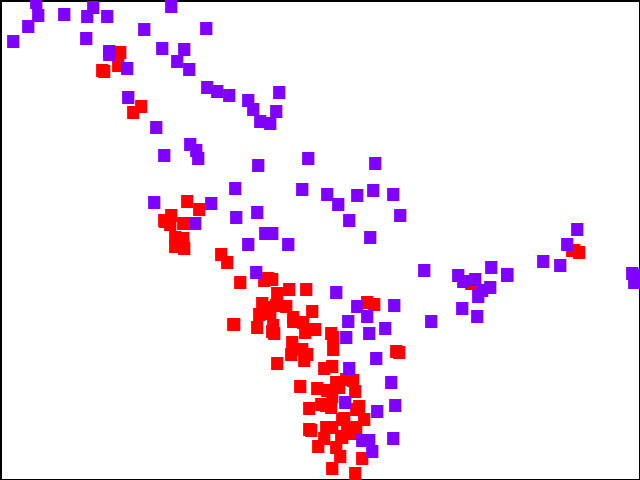}
     \vspace{-0.2in}
    \caption{$T_1,A_1^2$}
    \end{subfigure}
    \begin{subfigure}[b]{.15\textwidth}
      \centering
     \includegraphics[width=\textwidth]{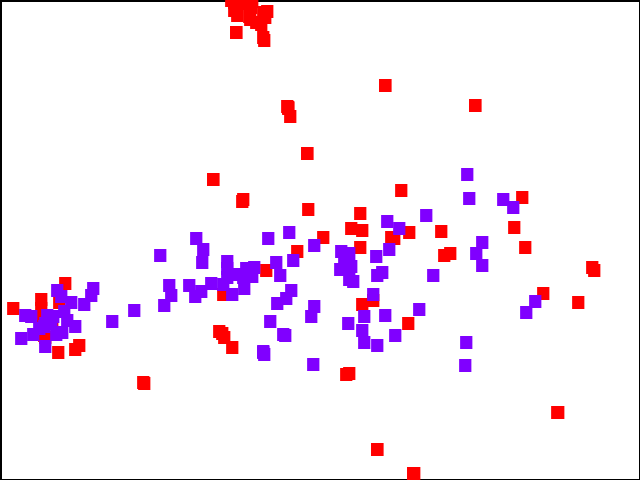}
     \vspace{-0.2in}
    \caption{$T_1,A_2$}
    \end{subfigure}
    \begin{subfigure}[b]{.15\textwidth}
      \centering
     \includegraphics[width=\textwidth]{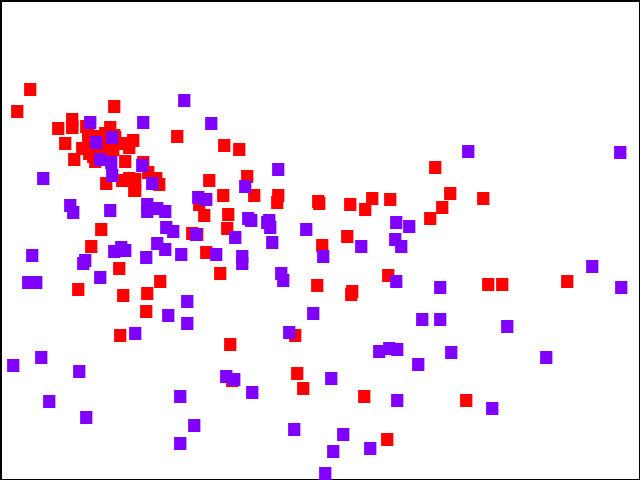}
     \vspace{-0.2in}
    \caption{$T_4,A_1^1$}
    \end{subfigure}
     \begin{subfigure}[b]{.15\textwidth}
      \centering
     \includegraphics[width=\textwidth]{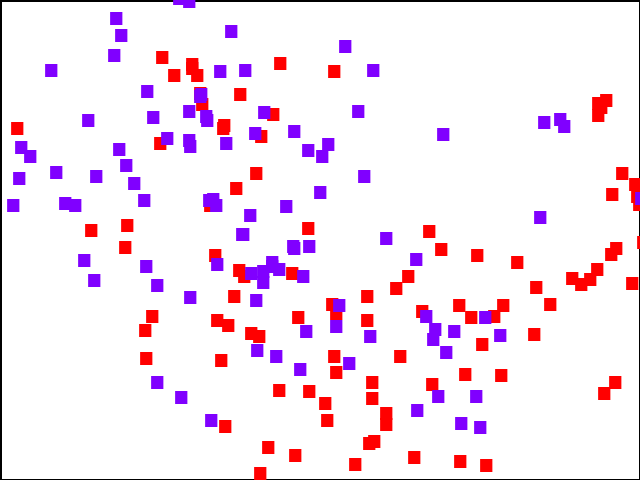}
     \vspace{-0.2in}
    \caption{$T_4,A_1^2$}
    \end{subfigure}
    \begin{subfigure}[b]{.15\textwidth}
      \centering
     \includegraphics[width=\textwidth]{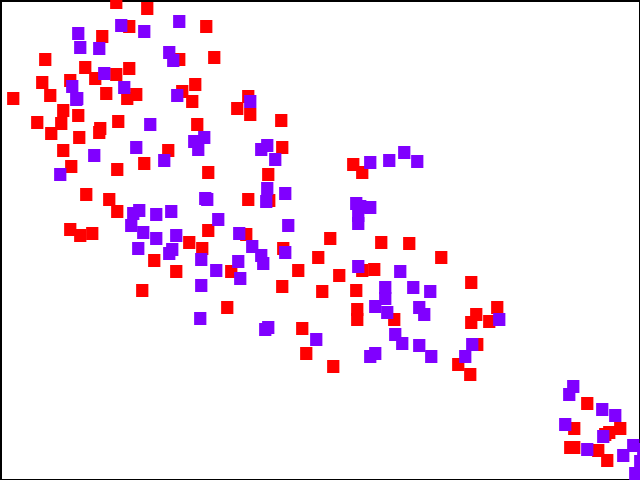}
     \vspace{-0.2in}
    \caption{$T_4,A_2$}
    \end{subfigure}
    \\
    {\bf GNN model: GAT}
    \\
    \begin{subfigure}[b]{.15\textwidth}
      \centering
     \includegraphics[width=\textwidth]{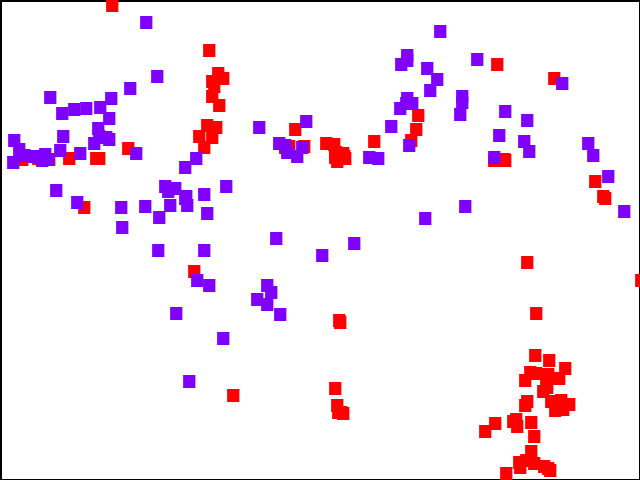}
     \vspace{-0.2in}
    \caption{$T_1,A_1^1$}
    \end{subfigure}
     \begin{subfigure}[b]{.15\textwidth}
      \centering
     \includegraphics[width=\textwidth]{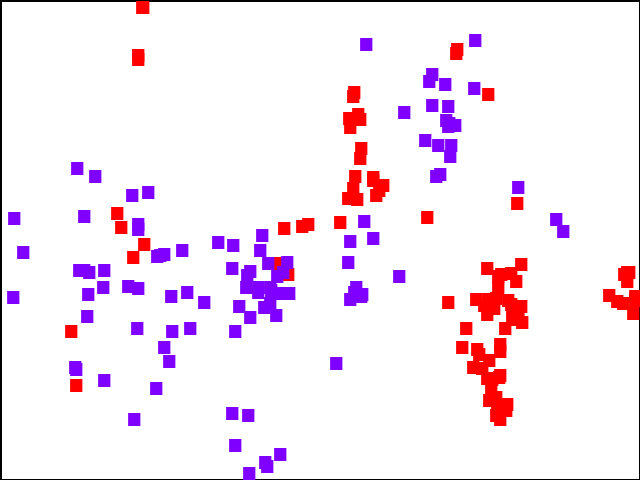}
     \vspace{-0.2in}
    \caption{$T_1,A_1^2$}
    \end{subfigure}
    \begin{subfigure}[b]{.15\textwidth}
      \centering
     \includegraphics[width=\textwidth]{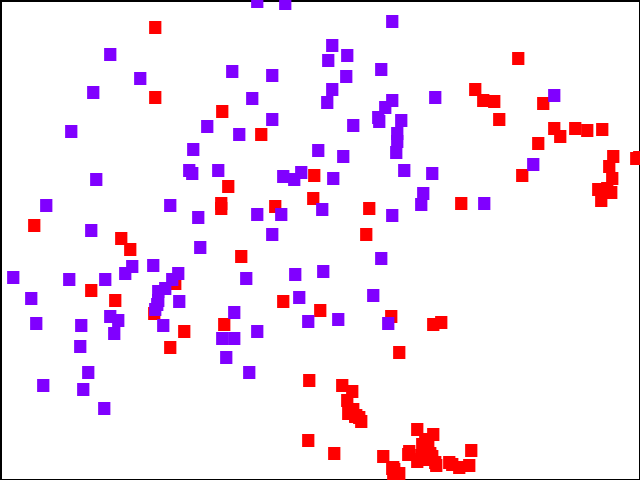}
     \vspace{-0.2in}
    \caption{$T_1,A_2$}
    \end{subfigure}
    \begin{subfigure}[b]{.15\textwidth}
      \centering
     \includegraphics[width=\textwidth]{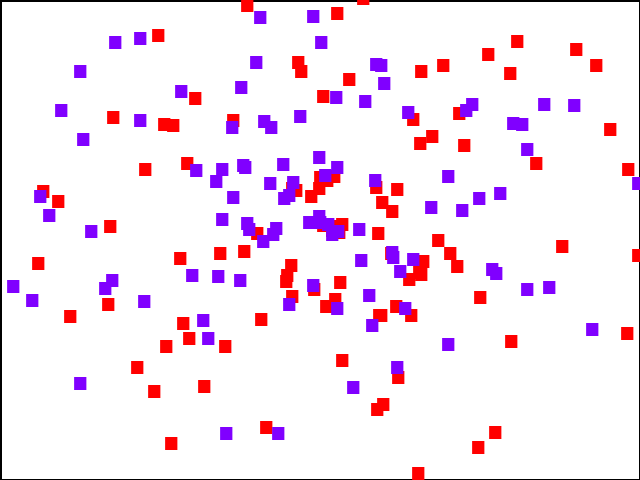}
     \vspace{-0.2in}
    \caption{$T_4,A_1^1$}
    \end{subfigure}
     \begin{subfigure}[b]{.15\textwidth}
      \centering
     \includegraphics[width=\textwidth]{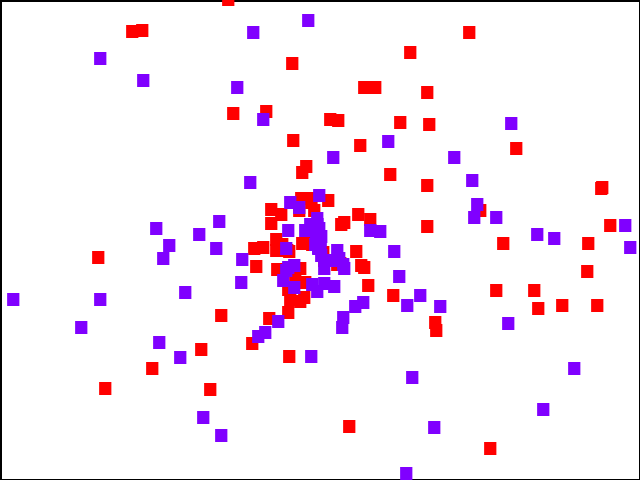}
     \vspace{-0.2in}
    \caption{$T_4,A_1^2$}
    \end{subfigure}
    \begin{subfigure}[b]{.15\textwidth}
      \centering
     \includegraphics[width=\textwidth]{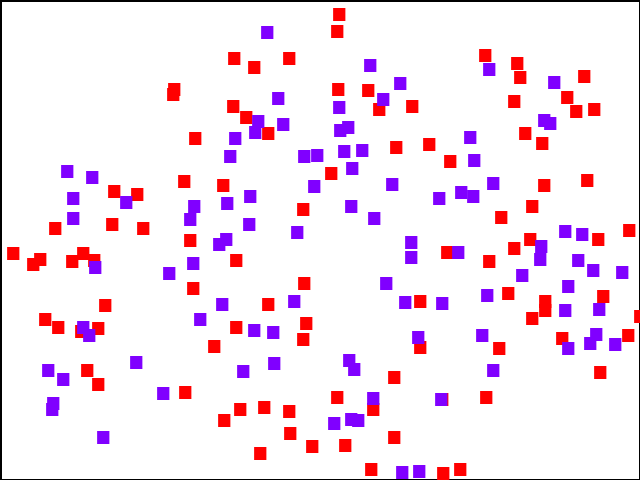}
     \vspace{-0.2in}
    \caption{$T_4,A_2$}
    \end{subfigure}
\caption{\label{fig:dis_pokec2} TSNE visualization of the distribution of node embeddings and target model outputs by GraphSAGE/GAT model on Pokec dataset. Blue and red dots denote the node embeddings or posteriors generated from the graph with or without the target property respectively. }
\end{figure*}
}

\section{Relationship between Train-test Loss Gap over Positive and Negative Graphs and GPIA Accuracy}
\label{appendix:overfit-pia}
\begin{table*}[t!]
    \centering
    {
    \begin{tabular}{c|c|c|c|c|c|c|c|c|c|c|c|c}
    \hline
     \multirow{2}{*}{Range of loss gap}&\multicolumn{2}{c|}{\bf $P_1$} & \multicolumn{2}{c|}{\bf $P_2$}& \multicolumn{2}{c|}{\bf $P_3$}& \multicolumn{2}{c|}{\bf $P_4$}& \multicolumn{2}{c|}{\bf $P_5$}& \multicolumn{2}{c}{\bf $P_6$}  \\ \cline{2-13}
     &Gap&Acc &Gap&Acc &Gap&Acc &Gap&Acc &Gap&Acc &Gap&Acc\\ \cline{1-13}
     Top-25\% &$8.9E^{-4}$&1&0.0012&0.95&$9.2E^{-4}$&0.97&0.0013&0.83&0.0042&0.92&$3.7E^{-4}$&0.84\\ \cline{1-13}
     [Top-25\%, top-50\%) &0.0019&0.97&0.0022&0.95&0.0055&0.97&0.0025&0.77&0.0059&0.97&0.0013&0.97\\ \cline{1-13}
     [Top-50\%, top-75\%) &0.006&1&0.0032&0.99&0.015&0.97&0.0037&0.86&0.0067&0.95&0.0049&0.86\\ \cline{1-13}
     Last 25\% &0.0067&1&0.004&0.98&0.018&0.97&0.0069&0.89&0.013&0.92&0.012&0.89\\\hline
    \end{tabular}
    }
    \caption{Relationship between attack accuracy and train-test loss gap between positive and negative graphs (Gap: average train-test loss gap; Acc: average attack accuracy)}
    \label{tab:overfit-pia-acc}
    \vspace{-0.2in}
\end{table*}

Recent studies \cite{shokri2017membership,li2021membership} have identified the {\em model train-test gap} (i.e., difference between training and testing accuracy) as an essential factor that contributes to membership inference attacks (MIA) \cite{shokri2017membership}. Intuitively, the attacker can infer the membership of some samples because the model behaves differently on the dataset with and without these samples. This raises the following question: {\em Does GPIA work because the GNN models behave differently on the graphs with and without the target properties?} To answer this question, we consider the training data that consists of only positive graphs (i.e., with the target property) and testing data that only include negative graphs (i.e., without the target property). We setup multiple settings of training/testing data that includes different samples of positive and negative graphs, and measure the model train-test gap as the difference between target model loss on training and testing data, as well as the  attack accuracy for these settings. We sort the train-test gaps in the ascending order, and generate four ranges of gaps: top-25\%, [top-25\%, top-50\%), [top-50\%, top-75\%), and the remaining 25\%. We measure both average loss gap and average attack accuracy for each gap range, and show the results in Table \ref{tab:overfit-pia-acc}. We observe that there is no linear relationship between the train-test loss gap of the target model and GPIA accuracy, as GPIA accuracy can be either increasing or decreasing when the loss gap grows.

\section{Distribution of Embedding/Posteriors for Positive and Negative Graphs}
\label{appendix:distribution}

Figure \ref{fig:dis_pokec} visualizes the distribution of GPIA attack features that are aggregated from node embeddings and posteriors output by GNN models on positive and negative graphs. We observe that, for all the three datasets, the (aggregated) embeddings and posteriors from positive graphs are distinguishable from that of the negative graphs. For example, as shown in Figure \ref{fig:dis_pokec} (a), the (aggregated) node embeddings generated from positive graphs (blue dots) are well separated from  those from negative graphs (red dots). This explains why GPIA can infer the existence of properties with high accuracy. 

We also observe that the similarity of the attack features can be transferred from the shadow graph to the target graph which can be of different structure. For example, Figure \ref{fig:dis_pokec}  (a) - (c) and (m) - (o) show that the distribution of the attack features generated from Pokec dataset is more similar to that on Pubmed dataset than Facebook dataset. This explains why GPIA accuracy of the transfer attacks $A_3/A_4$ can be as high as 0.72 when Pubmed and Pokec datasets are the target and shadow datasets respectively (Figure \ref{fig:attack34-acc} (c)).

\nop{
\subsection{Train-test Gap on Graphs with vs. w/o properties}
\label{appendix:overfit-property}

\begin{figure*}[t!]
    \begin{subfigure}[b]{.23\textwidth}
    \includegraphics[width=\textwidth]{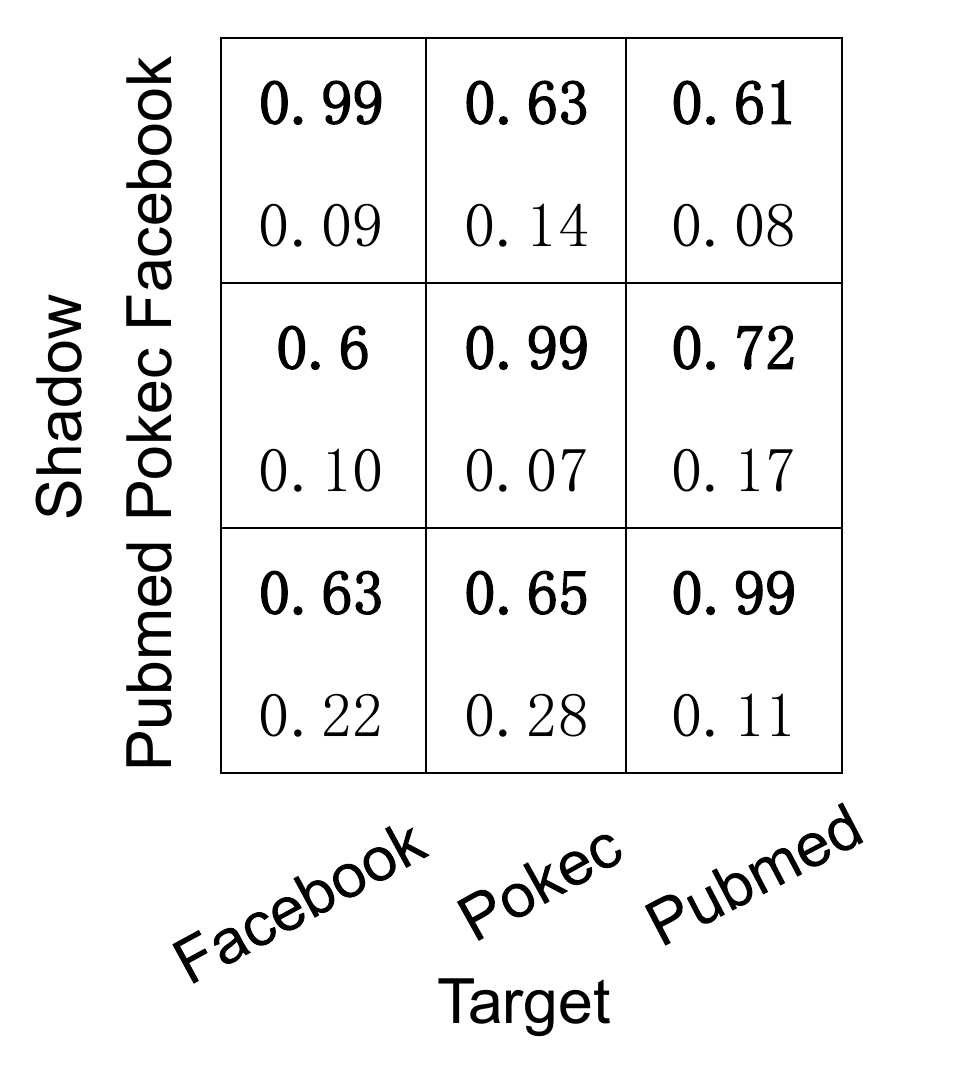}
    \vspace{-0.2in}
    \caption{Accuracy gap - $P_1$}
    \end{subfigure}
    \begin{subfigure}[b]{.23\textwidth} 
    \includegraphics[width=\textwidth]{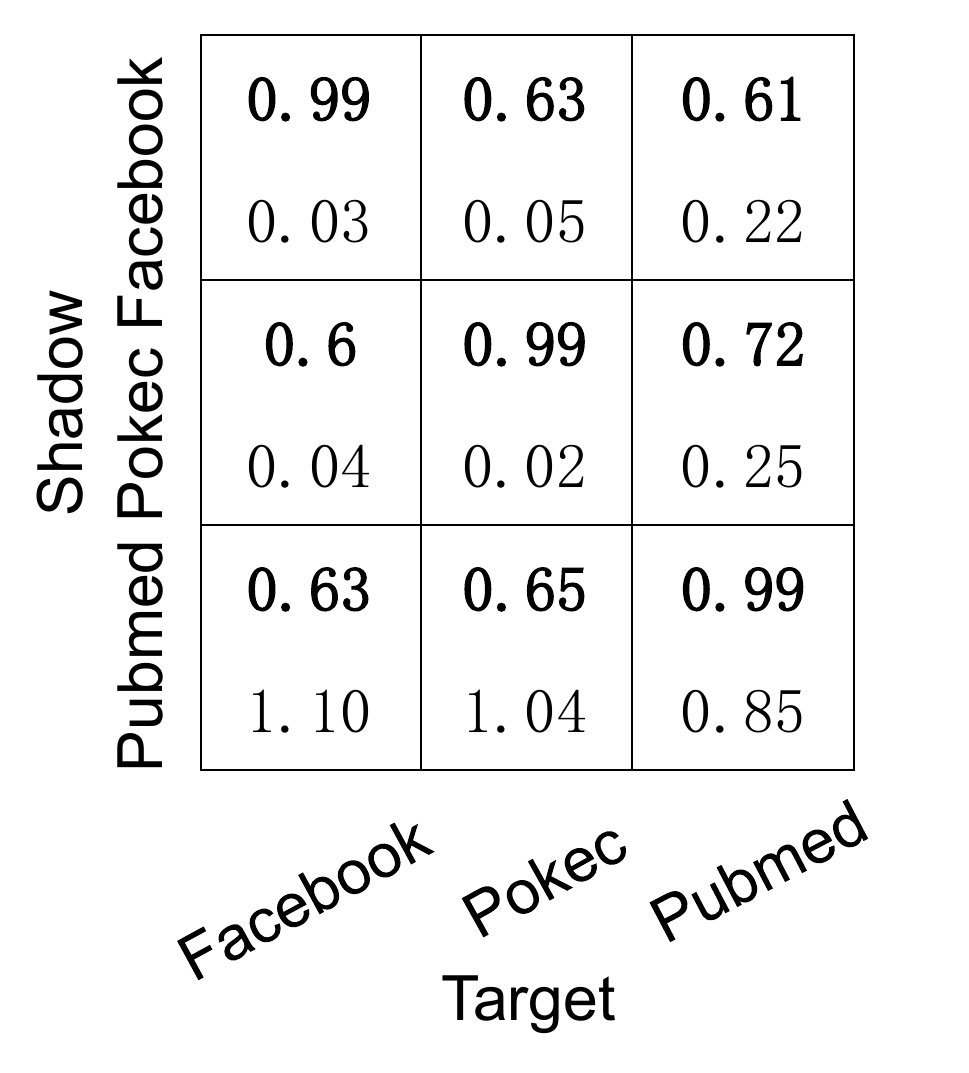}
    \vspace{-0.2in}
    \caption{Loss gap - $P_1$}
    \end{subfigure}
    \begin{subfigure}[b]{.23\textwidth} 
    \includegraphics[width=\textwidth]{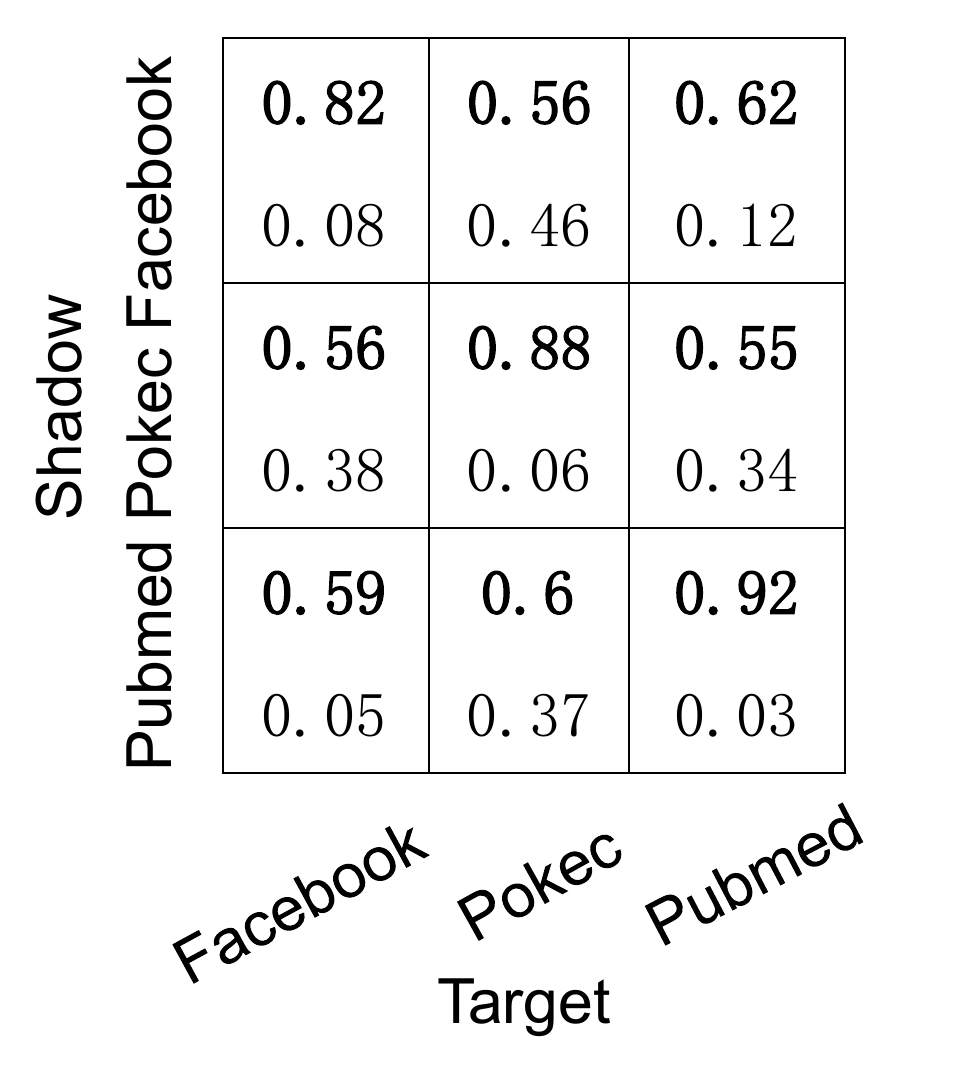}
    \vspace{-0.2in}
    \caption{Accuracy gap - $P_4$}
    \end{subfigure}
    \begin{subfigure}[b]{.23\textwidth} 
    \includegraphics[width=\textwidth]{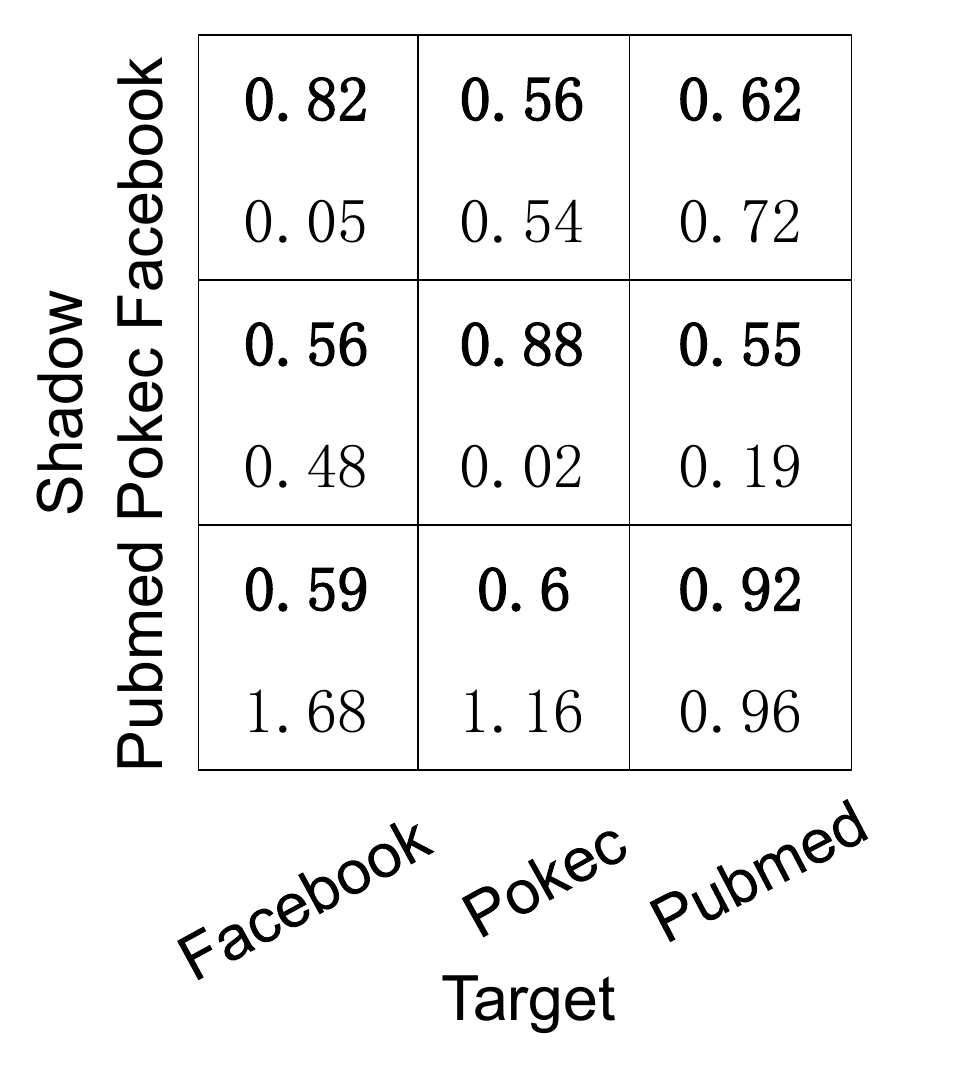}
    \vspace{-0.2in}
    \caption{Loss gap - $P_4$}
    \end{subfigure}
\caption{ \xiuling{Model accuracy gap between positive and negative graphs ($A_2$ Attack). In each cell, the values in first row are the attack accuracy, in second row are the values measured by accuracy gap or loss gap.} \Wendy{Also measure the train-test gap, where training is the shadow dataset, and testing is the target dataset.}}
\label{fig:A3A4-gaps}
\end{figure*}

\nop{
\begin{figure*}[t!]
    \begin{subfigure}[b]{.23\textwidth}
    \includegraphics[width=\textwidth]{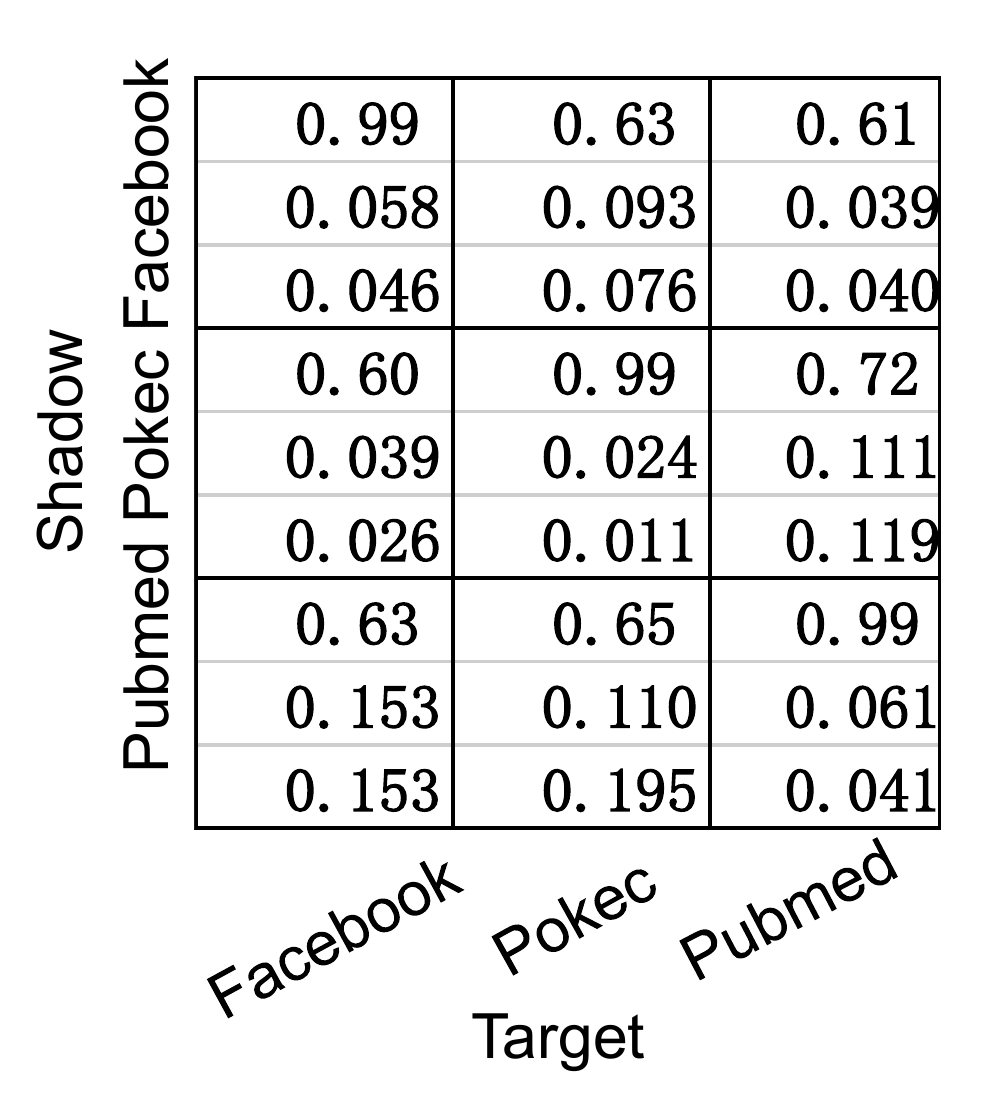}
    \vspace{-0.2in}
    \caption{Gap-1}
    \end{subfigure}
    \begin{subfigure}[b]{.23\textwidth} 
    \includegraphics[width=\textwidth]{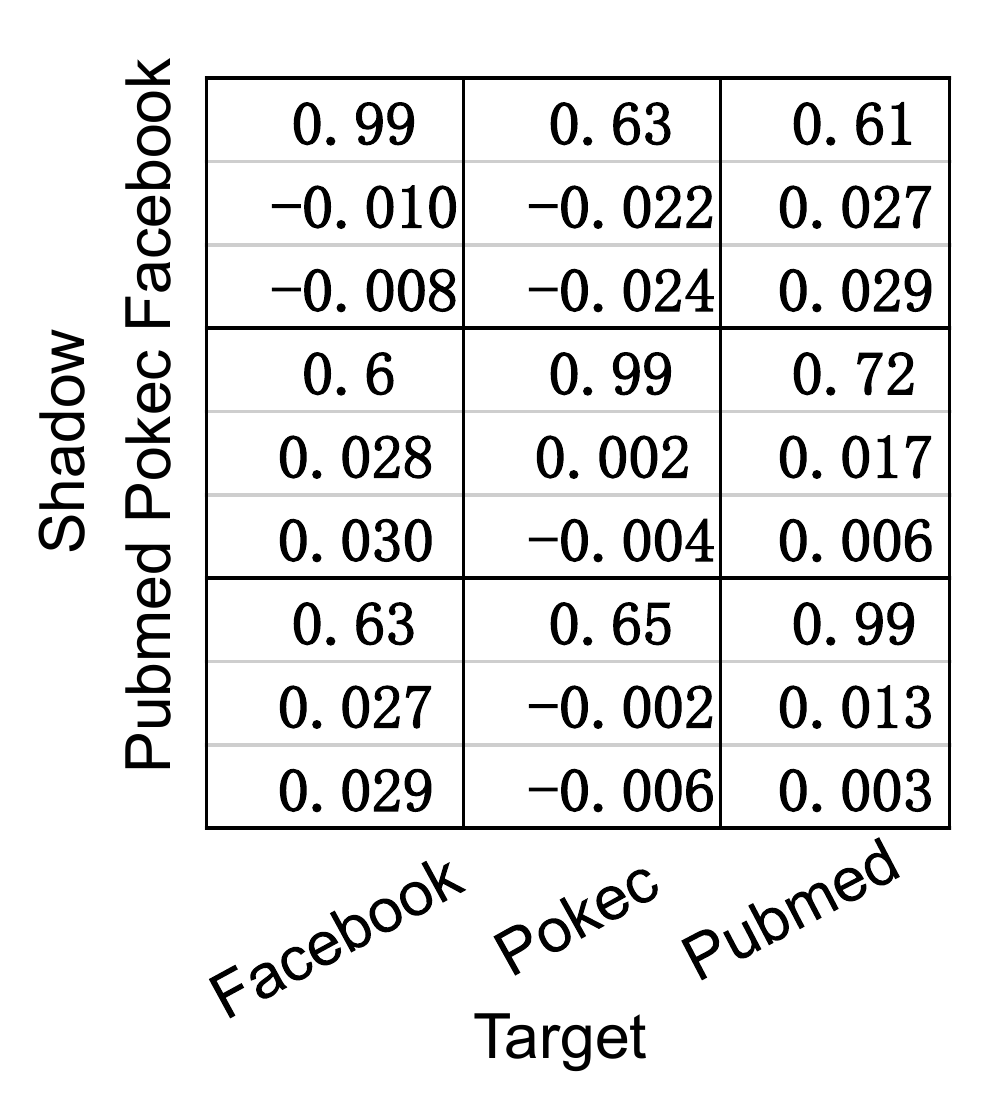}
    \vspace{-0.2in}
    \caption{Gap-2}
    \end{subfigure}
    \vspace{-0.2in}
\caption{ \xiuling{Model accuracy gap between positive and negative graphs ($A_2$ Attack). In each cell, the values in first row are the attack accuracy, in second row are the values measured by Gap-1 or Gap-2 for positive graphs, in third row are the value for negative graphs. For Gap-1, the values in second and third rows in each cell are the GNN model training-testing difference. For Gap-2, the values in second and third rows in each cell are the difference of training-testing gap between classes of property feature.} \xiuling{I need to re-measure the results as $p_{train}$-$(-p)_{test}$}}
\label{fig:A3A4-gaps}
\end{figure*}
}

\xiuling{Table \ref{tab:overfit-property} shows the results of training-testing gap for graphs with or w/o property. \Wendy{Are the results reported as the average over multiple graph samples with or w/o property?} We observe that there is no relation between train-test gap and the whether the graphs have properties. Figure \ref{fig:A3A4-gaps}}\Wendy{unfinished sentence?}
Figure \ref{fig:gap-overfitting}
}

\section{Additional Results for Impact Factors of GPIA}

\begin{figure*}[t!]
    \centering
  \vspace{0.1in}
\begin{subfigure}[b]{.7\textwidth}
      \centering
     \includegraphics[width=\textwidth]{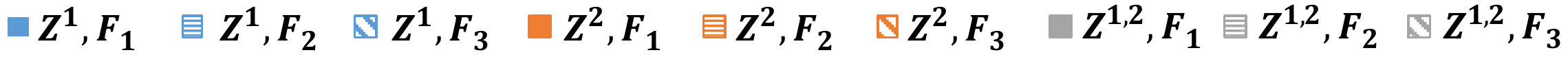}
     \vspace{-0.2in}
    \end{subfigure}\\
    \begin{subfigure}[b]{.33\textwidth}
      \centering
      \includegraphics[width=\textwidth]{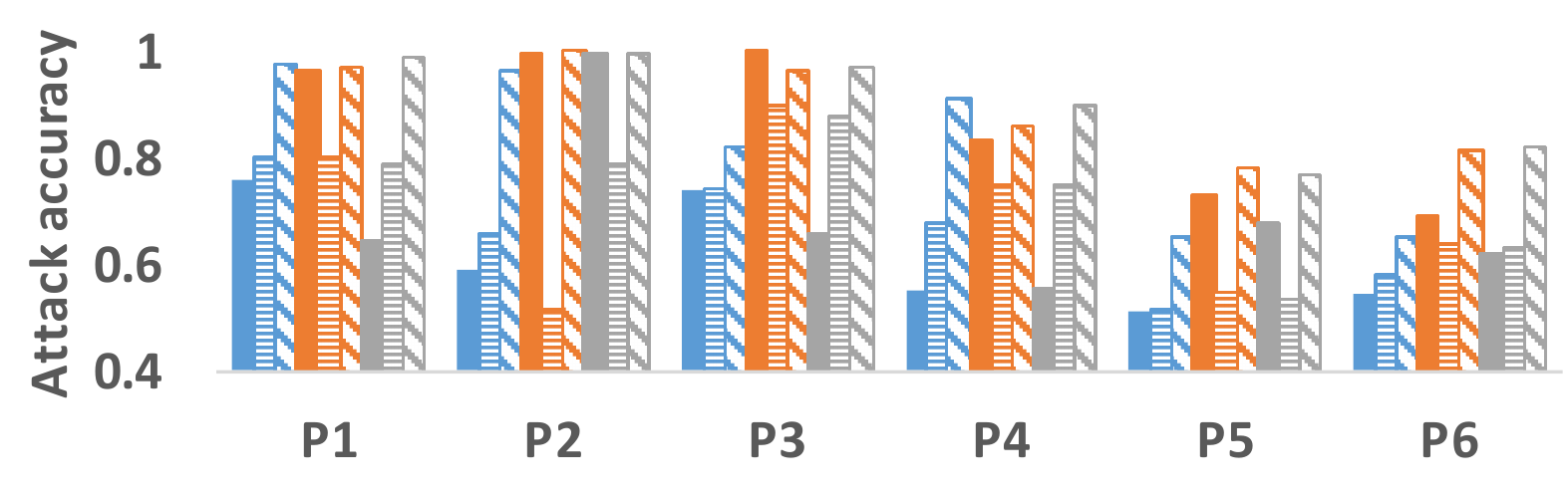}
        \vspace{-0.2in}
    \caption{GCN}
     \end{subfigure}
    \begin{subfigure}[b]{.33\textwidth}
      \centering
     \includegraphics[width=\textwidth]{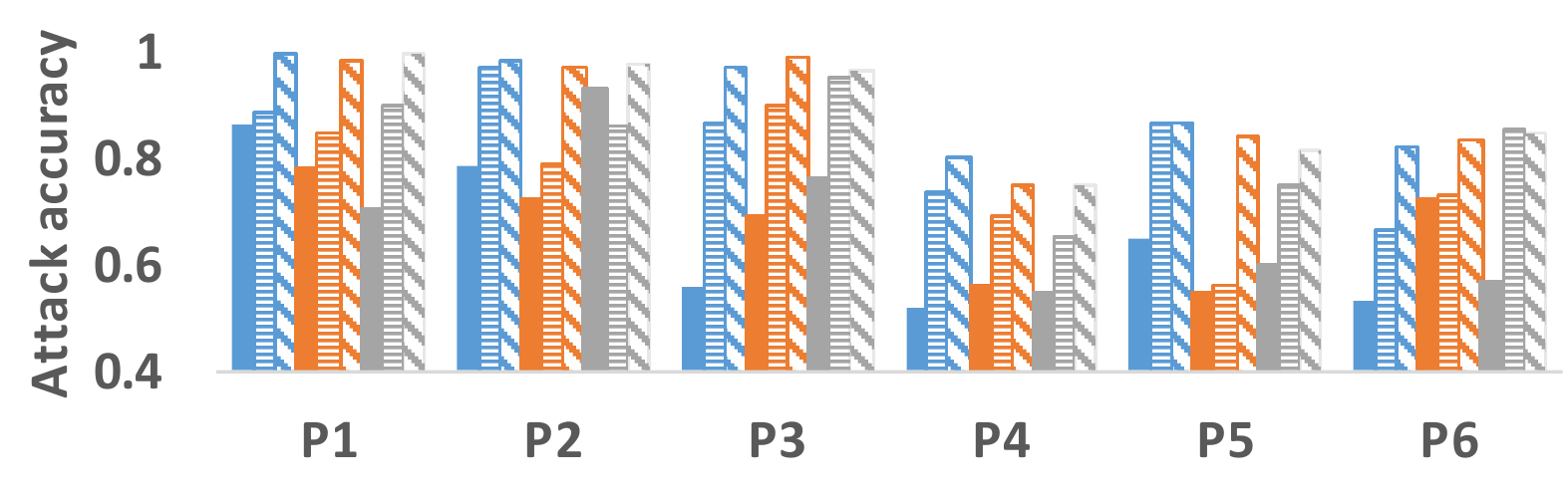}
     \vspace{-0.2in}
    \caption{GraphSAGE}
    \end{subfigure}
    \begin{subfigure}[b]{.33\textwidth}
      \centering
     \includegraphics[width=\textwidth]{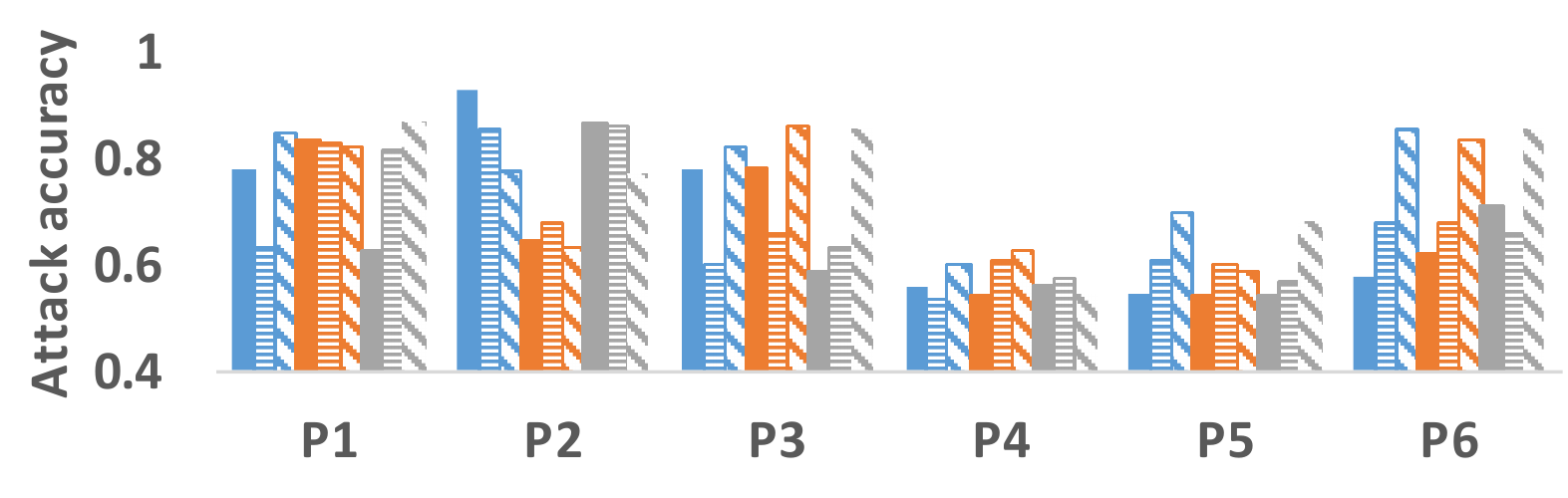}
     \vspace{-0.2in}
    \caption{GAT}
    \end{subfigure}
    \vspace{-0.05in}
\caption{\label{fig:dif-ft-gs-gat-white} Impacts of embedding aggregation methods (concatenation, mean-pooling, max-pooling) on PIA performance. $Z^1$, $Z^2$, and $Z^{1,2}$ are indicated in \textcolor{blue}{blue}, \textcolor{orange}{orange}, and \textcolor{gray}{gray} colors respectively, while $F_1$(concatenation), $F_2$ (mean-pooling), and $F_3$ (max-pooling) are indicated in solid, horizontal stripe, and diagonal stripe fill respectively. }
\end{figure*}

\begin{figure*}[t!]
    \centering
  \vspace{0.1in}
\begin{subfigure}[b]{.4\textwidth}
      \centering
     \includegraphics[width=\textwidth]{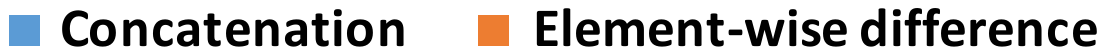}
     \vspace{-0.2in}
    \end{subfigure}\\
    \begin{subfigure}[b]{.3\textwidth}
      \centering
     \includegraphics[width=\textwidth]{text/figure/gcn-post-agg.pdf}
     \vspace{-0.2in}
    \caption{GCN}
    \end{subfigure}
    \begin{subfigure}[b]{.3\textwidth}
      \centering
     \includegraphics[width=\textwidth]{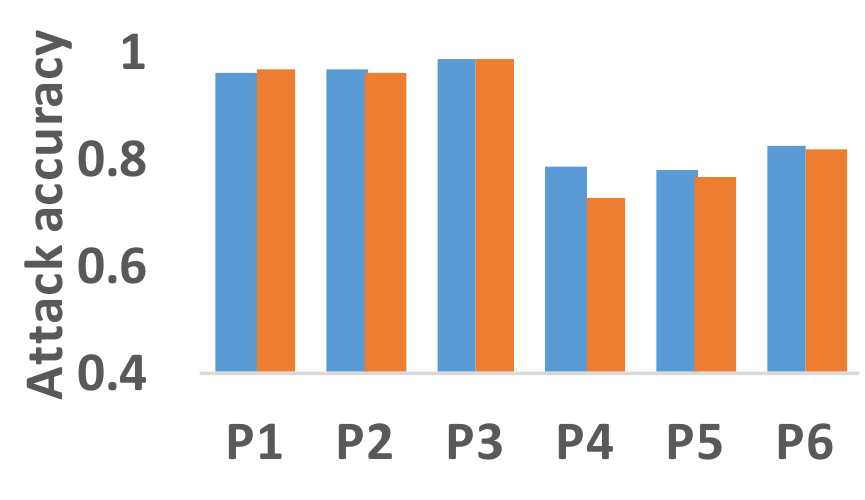}
     \vspace{-0.2in}
    \caption{GraphSAGE}
    \end{subfigure}
    \begin{subfigure}[b]{.3\textwidth}
      \centering
     \includegraphics[width=\textwidth]{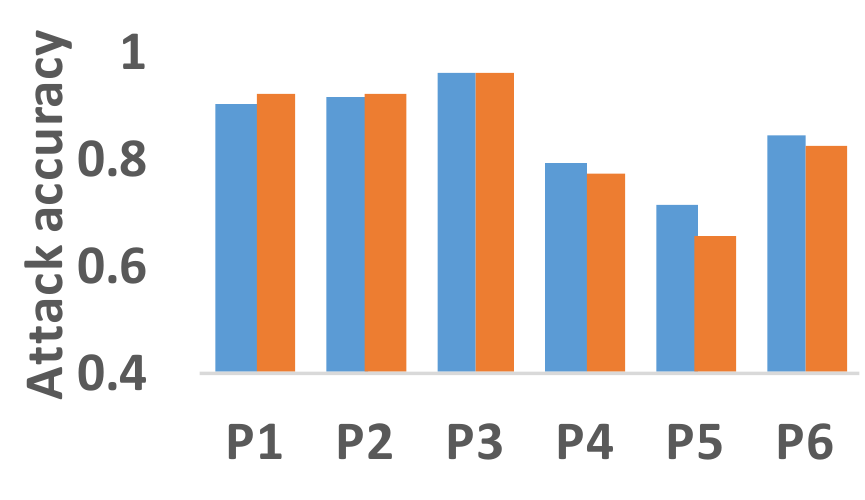}
     \vspace{-0.2in}
    \caption{GAT}
    \end{subfigure}
    \vspace{-0.05in}
\caption{\label{fig:dif-ft-gs-gat-black} Impacts of posterior aggregation methods (concatenation and element-wise difference) on PIA performance (GraphSAGE/GAT as target model). } 
\end{figure*}
\subsection{Amounts of Node Embedding}
\label{appendix:multi-ft2}
\begin{table*}[t!]
    \centering
    \begin{tabular}{c|c|c|c|c|c|c|c|c|c|c|c|c|c|c|c|c|c|c}
    \hline
     \multirow{2}{*}{\bf Embedding}&\multicolumn{6}{c|}{\bf GCN} & \multicolumn{6}{c|}{\bf GraphSAGE}&\multicolumn{6}{c}{\bf GAT}  \\ \cline{2-19}
     &$P_1$&$P_2$&$P_3$&$P_4$&$P_5$&$P_6$&$P_1$&$P_2$&$P_3$&$P_4$&$P_5$&$P_6$&$P_1$&$P_2$&$P_3$&$P_4$&$P_5$&$P_6$\\ \cline{1-19}
     $Z^1$&0.98&0.87&0.69&0.88&0.69&0.78&0.99&0.99&0.91&0.76&0.84&0.81&0.85&0.65&0.81&0.6&0.61&0.84\\ \cline{1-19}
     $Z^2$&0.99&0.92&0.66&0.80&0.66&0.8&0.99&0.97&0.95&0.79&0.79&0.82&0.84&0.63&0.82&0.59&0.63&0.83\\ \cline{1-19}
     $Z^3$&0.97&0.99&0.69&0.76&0.69&0.8&0.97&0.96&0.98&0.8&0.79&0.81&0.82&0.64&0.82&0.58&0.62&0.84\\ \cline{1-19}
     $Z^1$, $Z^2$ &0.99&0.93&0.69&0.88&0.69&0.78&0.99&0.96&0.98&0.76&0.81&0.82&0.85&0.64&0.81&0.58&0.61&0.84\\ \cline{1-19}
     $Z^1$, $Z^3$ &0.98&1&0.68&0.89&0.68&0.79&0.99&0.97&0.94&0.78&0.81&0.81&0.83&0.64&0.81&0.61&0.61&0.86\\\cline{1-19}
     $Z^2$, $Z^3$ &0.99&0.97&0.68&0.80&0.68&0.8&0.99&0.98&0.98&0.79&0.78&0.81&0.85&0.64&0.81&0.58&0.62&0.85\\ \cline{1-19}
     $Z^1$, $Z^2$, $Z^3$ &0.99&0.98&0.68&0.87&0.68&0.8&0.99&0.98&0.97&0.78&0.81&0.81&0.86&0.64&0.82&0.57&0.6&0.86\\ \hline
    \end{tabular}
    \caption{Impact of amounts of node embeddings on PIA performance (Facebook and Pubmed dataset). All the GNNs have three hidden layers. Max-pooling is used as the embedding aggregation method.}
    \label{tab:more-paras-fb-pubmed}
    \vspace{-0.2in}
\end{table*}

\nop{
\begin{figure*}[t!]
    \centering
    \begin{subfigure}[b]{.6\textwidth}
      \centering
    \includegraphics[width=\textwidth]{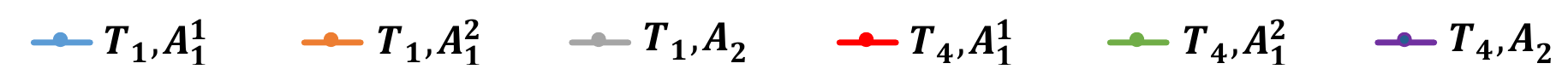}\\
    \end{subfigure}
    \begin{subfigure}[b]{.4\textwidth}
      \centering
    \includegraphics[width=\textwidth]{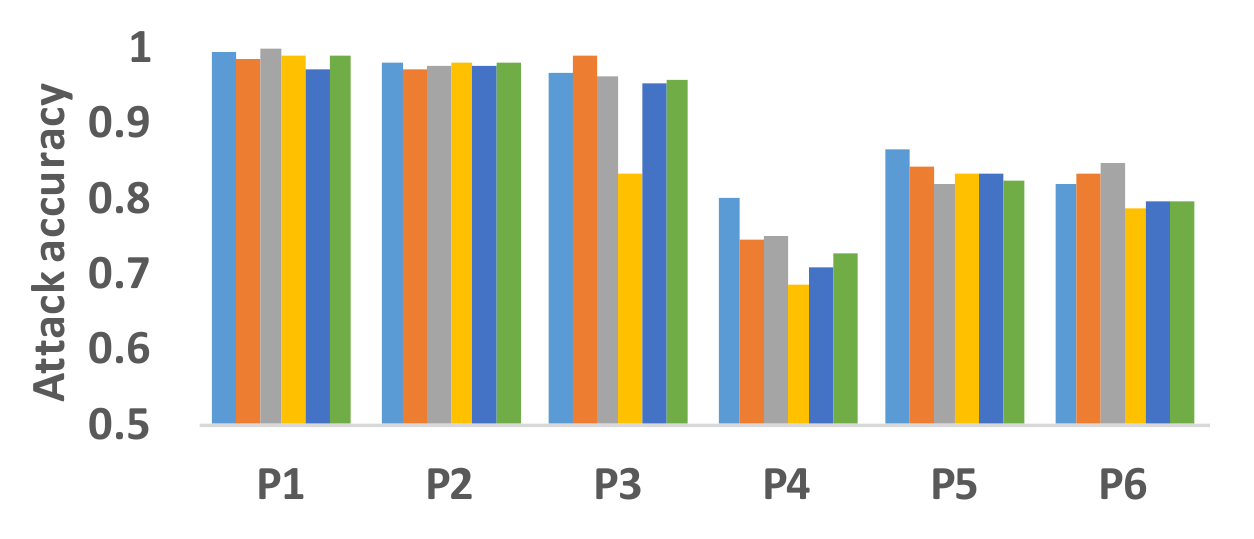}
     \vspace{-0.2in}
    \caption{GraphSage}
    \end{subfigure}
    \begin{subfigure}[b]{.4\textwidth}
      \centering
    \includegraphics[width=\textwidth]{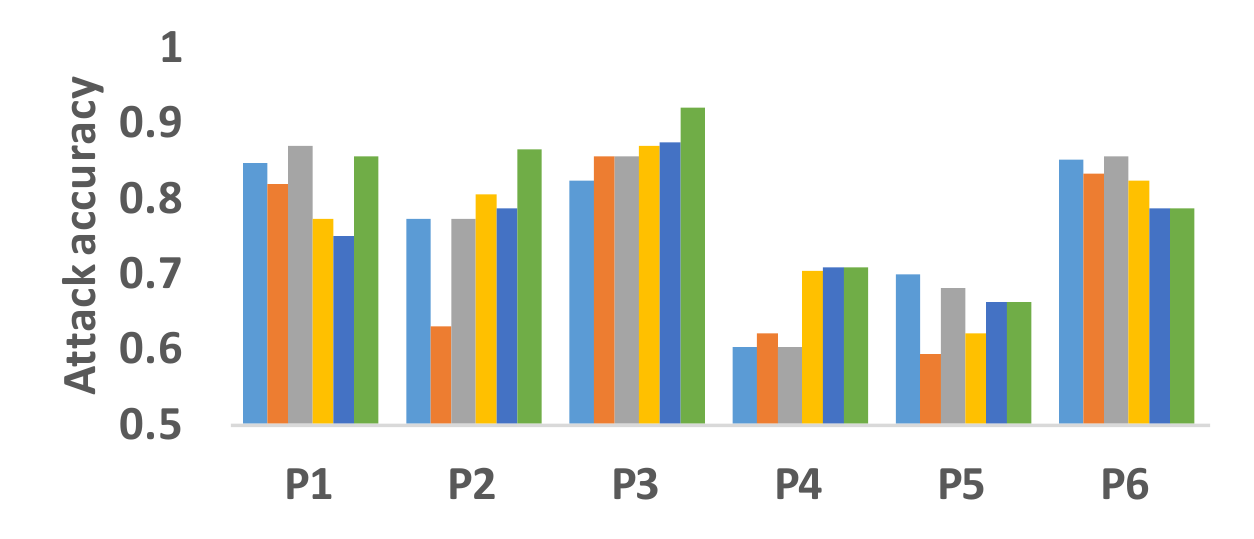}
     \vspace{-0.2in}
    \caption{GAT}
    \end{subfigure}
     \vspace{-0.1in}
\caption{\label{fig:multi-ft2} Attack performance of different input of PIA model on GraphSage and GAT ($A_1^{1}$: embedding from $1^{st}$ layer, $A_1^{2}$: embedding from $2^{nd}$ layer, $A_1^{1,2}$: embedding from $1^{st}$ \& $2^{nd}$ layer, $A_1^{1,O}$: embedding from $1^{st}$ layer \& posterior, $A_1^{2,O}$: embedding from $2^{nd}$ layer \& posterior, $A_1^{1,2,O}$: embedding from $1^{st}$ layer \& $2^{nd}$ layer \& posterior)} 
\end{figure*}
}

In this part of the experiments, we consider GraphSAGE and GAT, and use  GNN models that consist of three hidden layers, and consider the seven possible settings of choosing embeddings from any subset of the three layers when launching $A_1$. 
We use max-pooling as the embedding aggregation method, given its best performance among all aggregation methods. 
Table \ref{tab:more-paras-fb-pubmed} shows the attack performance with different amounts of parameters collected from these layers of the target model.
We observe that, for a given target model and the property to be attacked, GPIA performance is similar across all seven embedding settings. 
The only exception is when GCN as the target model and $P_4$ as the target property, where the GPIA performance changes  significantly from 0.76 ($Z^3$) to 0.89 ($Z^1, Z^3$). 
Second, interestingly, GPIAs that utilize the embeddings from more layers do not necessarily outperform those that use the embeddings from fewer layers. For example, consider GAT as the target model and $P_4$ as the target property, GPIA accuracy is only 0.57 when $Z^1, Z^2, Z^3$ are utilized, but it is 0.6 when only $Z^1$ is used. Note that the GPIA features are generated from the max-pooling aggregation of these embeddings. Therefore, the embeddings aggregated from more layers do not necessarily contain more encoded information of the graph than those from fewer layers. This explains why GPIA attack accuracy does not improve when more embeddings are utilized as the adversary knowledge.

\subsection{Type of Attack Classifiers}
\label{appendix:pia-classifier-graphsage-gat}

\begin{figure*}[t!]
    \centering
      \centering
    \includegraphics[width=0.5\textwidth]{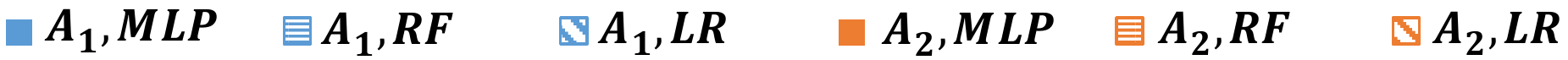}\\
    \begin{subfigure}[b]{.33\textwidth}
      \centering
     \includegraphics[width=\textwidth]{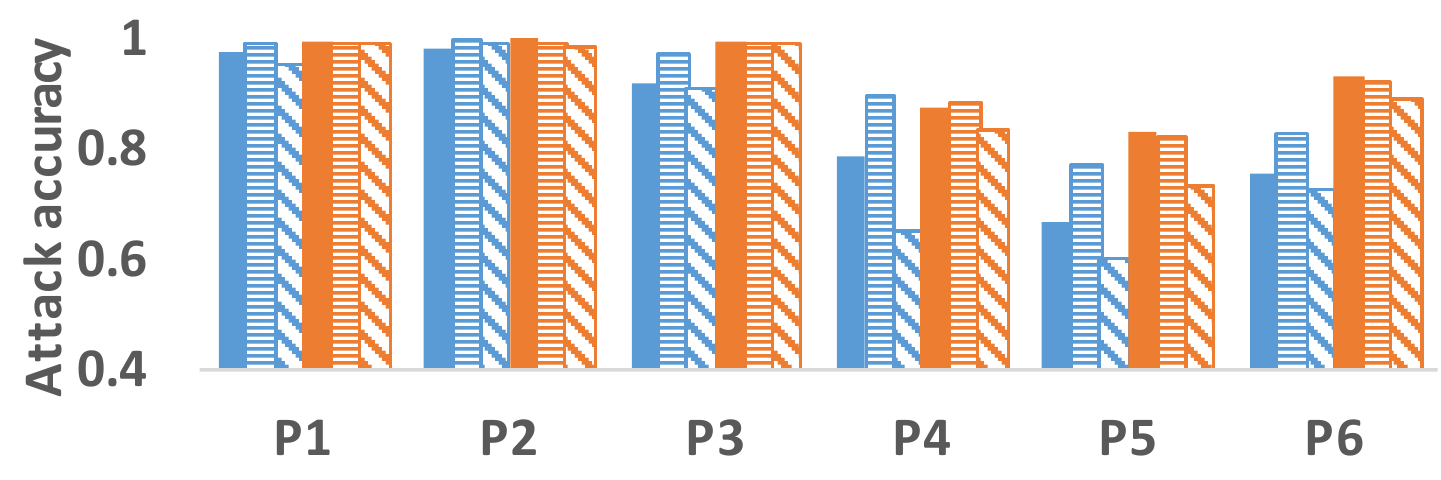}
    \caption{GCN}
    \end{subfigure}
 \begin{subfigure}[b]{.33\textwidth}
      \centering
    \includegraphics[width=\textwidth]{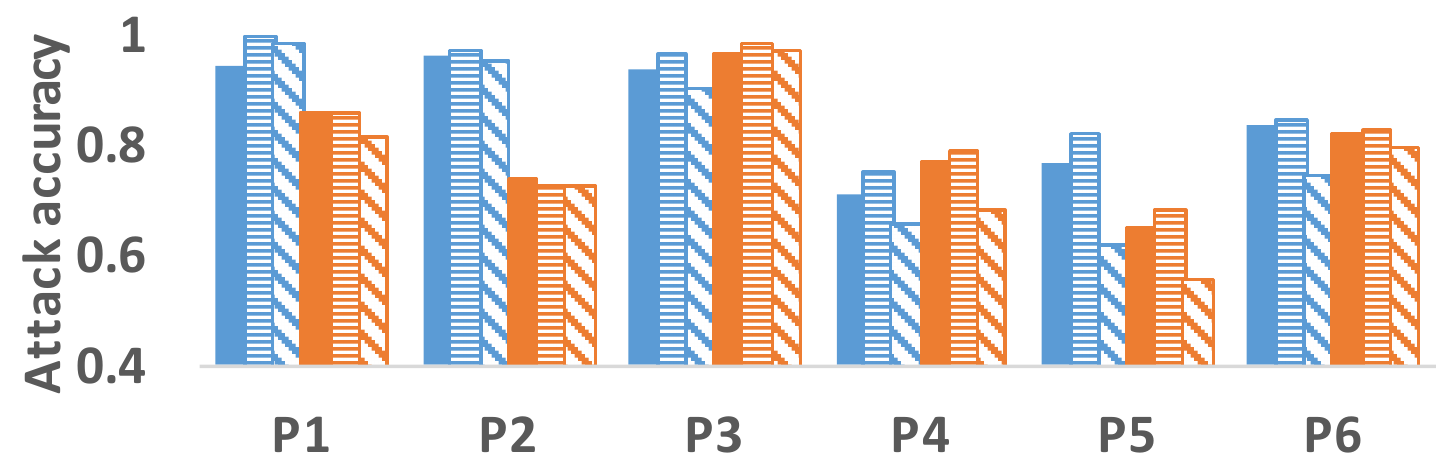}
     \vspace{-0.2in}
    \caption{GraphSAGE}
    \end{subfigure}
    \begin{subfigure}[b]{.33\textwidth}
      \centering
    \includegraphics[width=\textwidth]{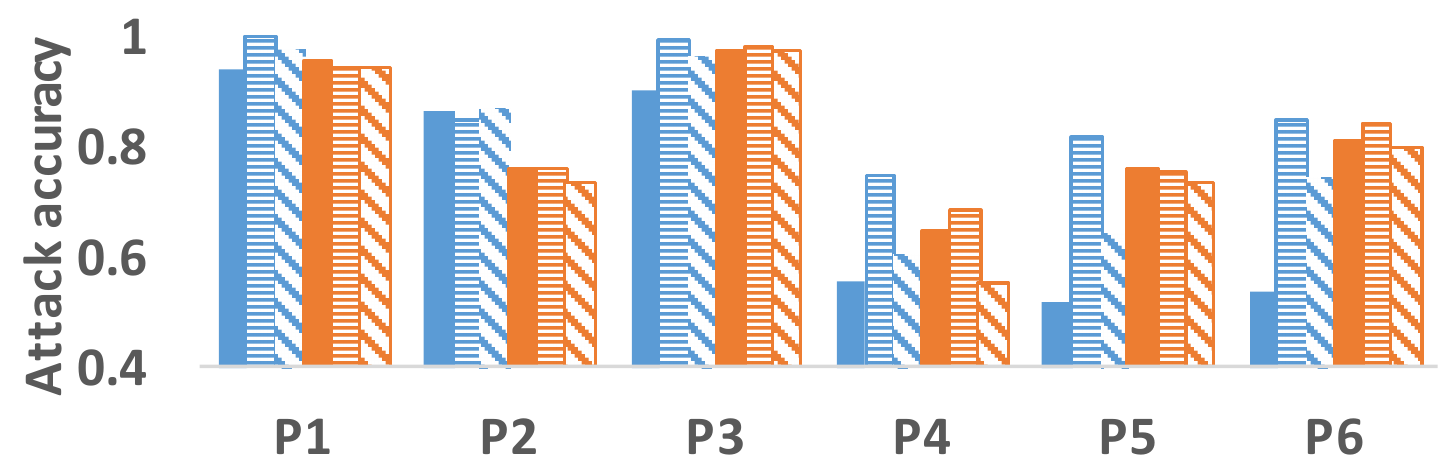}
     \vspace{-0.2in}
    \caption{GAT}
    \end{subfigure}
     \vspace{-0.1in}
\caption{\label{fig:impact-classifier-gs-gat}  Impact of type of attack classifiers on PIA performance. $A_1$ and $A_2$ are indicated in \textcolor{blue}{blue} and \textcolor{orange}{orange} colors respectively, while \textsf{MLP}, \textsf{RF}, and \textsf{LR} are indicated in solid, horizontal stripe, and diagonal stripe fill respectively. }
\end{figure*}

Figure \ref{fig:impact-classifier-gs-gat} shows how different classifier models impact GPIA performance. The main observation is that, while the three attack classifiers deliver similar performance in most of the settings, LR never outperforms MLP and RF. Furthermore, while RF and MLP deliver similar  performance in most of the settings, RF outperforms MLP  in  most of the white-box attacks, and MLP has slightly better performance than RF for the  black-box attack.  Based on these results, we recommend RF and MLP as the white-box and black-box attack classifier respectively. 

\subsection{Embedding Aggregation Methods on PIA
Performance}
\label{appendix:agg-embedding-graphsage-gat}

Figure \ref{fig:dif-ft-gs-gat-white} presents the attack performance of various embedding aggregation methods for the white-box attacks. The main observation is that  max-pooling method outperforms the other two aggregation methods in terms of PIA performance. 

\subsection{Posterior Aggregation Methods}
\label{appendix:agg-posterior-graphsage-gat}

Figure \ref{fig:dif-ft-gs-gat-black} presents the attack performance of various posterior aggregation methods for the black-box attacks against three GNN models. We observe that, the concatenation method can outperform the element-wise difference method significantly (attack accuracy difference 7.1\%) in some settings, while has comparable performance as the element-wise difference method for the remaining cases (attack accuracy difference less than 2\%). 
Therefore, we recommend the concatenation method to aggregate the posterior output and generate GPIA features for the black-box setting.

\subsection{Embedding Dimension Alignment Methods}
\label{sc:pca-encoder-acc}

\begin{table*}[t!]
    \centering
    \scalebox{0.9}{\begin{tabular}{c|c|c|c|c|c|c|c|c|c|c|c|c}\hline
      \multicolumn{13}{c}{{\bf Target model: GCN}}\\\hline
  \multirow{3}{*}{\bf Target dataset}&\multicolumn{12}{c}{\bf Shadow dataset}\\\cline{2-13}
   &\multicolumn{4}{c|}{Pokec}&\multicolumn{4}{c|}{Facebook}&\multicolumn{4}{c}{Pubmed}\\\cline{2-13}      
   &Sampling&AutoEncoder&PCA&TSNE&Sampling&AutoEncoder&PCA&TSNE&Sampling&AutoEncoder&PCA&TSNE\\\hline
   Pokec&\multicolumn{4}{c|}{N/A}&0.55&0.54&0.53&0.54&0.52&0.50&0.56&0.61\\\hline
   Facebook&0.60&0.51&0.55&0.66&\multicolumn{4}{c|}{N/A}&0.54&0.5&0.56&0.54\\\hline  
   Pubmed&0.58&0.51&0.6&0.71&0.57&0.53&0.58&0.59&\multicolumn{4}{c}{N/A}\\\hline
    \multicolumn{13}{c}{{\bf Target model: GraphSAGE}}\\\hline
  \multirow{3}{*}{\bf Target dataset}&\multicolumn{12}{c}{\bf Shadow dataset}\\\cline{2-13}
   &\multicolumn{4}{c|}{Pokec}&\multicolumn{4}{c|}{Facebook}&\multicolumn{4}{c}{Pubmed}\\\cline{2-13}      
   &Sampling&AutoEncoder&PCA&TSNE&Sampling&AutoEncoder&PCA&TSNE&Sampling&AutoEncoder&PCA&TSNE\\\hline
   Pokec&\multicolumn{4}{c|}{N/A}&0.5&0.54&0.56&0.63&0.5&0.51&0.57&0.57\\\hline
   Facebook&0.0.5&0.55&0.57&0.60&\multicolumn{4}{c|}{N/A}&0.5&0.5&0.63&0.54\\\hline  
   Pubmed&0.52&0.54&0.61&0.57&0.5&0.51&0.63&0.57&\multicolumn{4}{c}{N/A}\\\hline
   \multicolumn{13}{c}{{\bf Target model: GAT}}\\\hline
  \multirow{3}{*}{\bf Target dataset}&\multicolumn{12}{c}{\bf Shadow dataset}\\\cline{2-13}
   &\multicolumn{4}{c|}{Pokec}&\multicolumn{4}{c|}{Facebook}&\multicolumn{4}{c}{Pubmed}\\\cline{2-13}      
   &Sampling&AutoEncoder&PCA&TSNE&Sampling&AutoEncoder&PCA&TSNE&Sampling&AutoEncoder&PCA&TSNE\\\hline
   Pokec&\multicolumn{4}{c|}{N/A}&0.52&0.58&0.54&0.52&0.53&0.6&0.58&0.59\\\hline
   Facebook&0.54&0.55&0.56&0.57&\multicolumn{4}{c|}{N/A}&0.5&0.60&0.53&0.61\\\hline  
   Pubmed&0.5&0.55&0.58&0.62&0.51&0.52&0.51&0.53&\multicolumn{4}{c}{N/A}\\\hline
   \end{tabular}}
   \caption{\label{tab:attack34-compare2} Impact of the dimension alignment methods on PIA performance (GraphSAGE/GAT as the GNN model, Attacks 3 as the attacks, and properties $P_1 - P_3$). The three N/A cases do not need dimension alignment as both target and shadow datasets are the same.}
   \end{table*}

Table \ref{tab:attack34-compare2} presents the attack performance of the attacks $A_3$ and $A_4$ with the four dimension alignment methods, namely, sampling, TSNE projection, PCA dimension reduction, and Autoencoder dimension compression, on three models. 
For PCA, we set the amount of variance that needs to be explained as  90\%, 95\%, and 99\% (i.e., the information that the principal components represented).  For the Autoencoder dimension compression method, we use mean squared error (MSE) to measure the loss between original data and the reconstructed data. The main observation is similar to all GNN models - the TSNE dimension alignment method outperforms other three methods  in most of the settings. 

\subsection{Group Size Ratio}
\label{appendix:group-size}
\begin{table}[t!]
  \small
    \centering
    {
    \begin{tabular}{c|c|c|c|c|c|c}
    \hline
     \multicolumn{2}{c|}{\backslashbox{Attack settings}{Group ratio}}&1:1&1:2&1:3&1:4&1:5  \\ \cline{1-7}
     \multirow{2}{*}{GCN}&$A_1$&0.67&1&1&1&1\\ \cline{2-7}
     &$A_2$&0.44&0.95&1&1&1\\ \cline{1-7}
    \multirow{2}{*}{GraphSAGE}&$A_1$&0.56&0.78&0.82&0.89&1\\ \cline{2-7}
     &$A_2$&0.5&0.75&0.9&0.92&1\\ \hline
     \multirow{2}{*}{GAT}&$A_1$&0.5&0.66&0.72&0.77&0.8\\ \cline{2-7}
     &$A_2$&0.5&0.79&0.85&0.92&0.94\\ \cline{1-7}

    \end{tabular}}
  \caption{Impact of group size ratio on GPIA performance (Male:Female for property $P_2$ on Facebook dataset).}
    \label{tab:group-size-acc}
    \vspace{-0.2in}
\end{table}
To measure the impact of group size ratio on GPIA accuracy, we evaluate the attack accuracy of the property $P_2$ on Facebook dataset with various group size ratios, and show the results in Table \ref{tab:group-size-acc}. When the group size ratio is 1:1, the attack accuracy is low (never exceeds 0.6). However, the attack accuracy grows with the increase of the group ratio. It can be as high as 1 when the group size ratio increase to 1:3. This demonstrates that GPIA performance is affected by group prevalence - it may fail if the property has a near 50\% prevalence. 
\subsection{Node Non-Overlap in GPIA Training \& Testing Data}
\label{appendix:non-overlap_pokec}

We generate the non-overlapping GPIA training and testing data by  randomly splitting the nodes in the original graph into two non-overlapping sets, one set $S_1$ for the sampling of subgraphs for training, and the other set $S_2$ for the sampling of subgraphs for testing. Then we randomly sample 700 subgraphs (50/50 split between positive and negative graphs) from the node set $S_1$ as the training dataset, and 300 subgraphs (50/50 split between positive/negative graphs) from $S_2$ as the testing dataset. We have to point out that the node  non-overlapping sampling method cannot meet the requirement of 50/50 split between positive and negative graphs due to the sparsity of the graph. Thus we randomly add some edges to the sampled subgraphs to make them meet the requirement. Table \ref{tab:pia-acc-no-overlap} shows the results of the node non-overlap setting for Pokec dataset. The attack accuracy is very similar to that of the overlapping setting (Figure \ref{fig:pia_acc}). The difference between the attack accuracy for overlapping and non-overlapping settings never exceeds 0.09. This demonstrates that GPIA accuracy is not impacted much when there are a small portion of overlapping nodes in its training and testing data. 

\nop{
 \begin{table}[t!]
    \centering
     \color{blue}{
    \begin{tabular}{c|c|c|c|c|c|c|c|c}
    \hline
     \multirow{2}{*}{\bf Attack} & \multicolumn{2}{c|}{\bf $P_1$}& \multicolumn{2}{c|}{\bf $P_2$}&\multicolumn{2}{c}{\bf $P_4$}& \multicolumn{2}{c}{\bf $P_5$}\\ \cline{2-9}
     &{\bf $A_1$} &{\bf $A_2$}&{\bf $A_1$} &{\bf $A_2$} &{\bf $A_1$} &{\bf $A_2$}&{\bf $A_1$} &{\bf $A_2$} \\ \hline
     GCN&1&0.95&0.95&0.99&0.93&0.96&&\\ \hline
     GraphSAGE&0.99&0.97&0.92&0.96&0.83&0.83&&\\ 
     \hline
     GAT&0.94&0.99&0.77&0.89&0.77&0.82&&\\ 
     \hline
    \end{tabular}
    \caption{   \label{tab:pia-acc-no-overlap}\textcolor{blue}{Comparison of PIA performance for graphs without node overlap (Pokec dataset) \Wendy{Add P4, and other 2 datasets.}} }
    }
    \vspace{-0.2in}
\end{table}
}

\begin{table}[!t]
    \centering
    {
    \begin{tabular}{c|c|c|c|c|c|c}
    \hline
     \multirow{2}{*}{\bf Attack}&\multicolumn{2}{c|}{\bf GCN} &\multicolumn{2}{c|}{\bf GraphSAGE}&\multicolumn{2}{c}{\bf GAT}  \\ \cline{2-7}
     &{\bf $A_1$} &{\bf $A_2$}&{\bf $A_1$} &{\bf $A_2$} &{\bf $A_1$} &{\bf $A_2$}\\ \hline
     $P_1$&1&0.95&0.99&0.97&0.94&0.99\\ \hline
     $P_2$&0.95&0.99&0.92&0.96&0.77&0.89\\ \hline
     $P_3$&0.91&0.99&0.97&0.96&0.87&0.92\\ \hline
     $P_4$&0.93&0.96&0.83&0.83&0.77&0.82\\ \hline
     $P_5$&0.74&0.78&0.88&0.77&0.66&0.69\\ \hline
     $P_6$&0.83&0.94&0.82&0.85&0.78&0.83\\ \hline
    \end{tabular}
    }
    \caption{Attack accuracy of $A_1$ and $A_2$ when there is no node overlap between GPIA training and testing data}
    \label{tab:pia-acc-no-overlap}
    \vspace{-0.2in}
\end{table}

\begin{table*}[!h]
    \centering
    {
    \scalebox{1}{\begin{tabular}{c|c|c|c|c|c|c|c|c|c|c|c|c}
    \hline
      \multirow{2}{*}{Setting}&\multicolumn{2}{c|}{$b$=0}&\multicolumn{2}{c|}{$b$=0.1}&\multicolumn{2}{c|}{$b$=0.5}&\multicolumn{2}{c|}{$b$=1}&\multicolumn{2}{c|}{$b$=5}&\multicolumn{2}{c}{$b$=10}\\ \cline{2-13}
          &$Acc$&$Diff$&$Acc$&$Diff$&$Acc$&$Diff$&$Acc$&$Diff$&$Acc$&$Diff$&$Acc$&$Diff$\\ \hline
       GCN&1&0.109&1&0.109&0.933&0.101&0.687&0.07&0.52&0.067&0.507&-0.022\\ \hline
       GraphSAGE&0.978&0.112&0.957&0.111&0.877&0.110&0.777&0.097&0.547&0.047&0.497&0.01\\ \hline
       GAT&0.63&-0.122&0.593&-0.118&0.577&-0.11&0.57&-0.061&0.533&-0.018&0.502&-0.001\\ \hline
     
    \end{tabular}
    }}
    \caption{Influence of different noise scales on disparity of group loss - Facebook dataset ($Acc$: attack accuracy; $Diff$: loss difference between Male \& Female groups. Larger $b$ values indicate stronger defense.}
    \label{tab:intuition-defense}
    \vspace{-0.2in}
\end{table*}

\section{More Results on Defense}

\begin{figure*}[t!]
\centering
    \centering
  {\bf GraphSAGE}
    \\
    \vspace{0.1in}
    \begin{subfigure}[b]{.7\textwidth}
      \centering
     \includegraphics[width=\textwidth]{text/figure/def12-legend1.pdf}
     \vspace{-0.2in}
    \end{subfigure}
    \begin{tabular}{cc}
    \begin{subfigure}[b]{.3\textwidth}
      \centering
     \includegraphics[width=\textwidth]{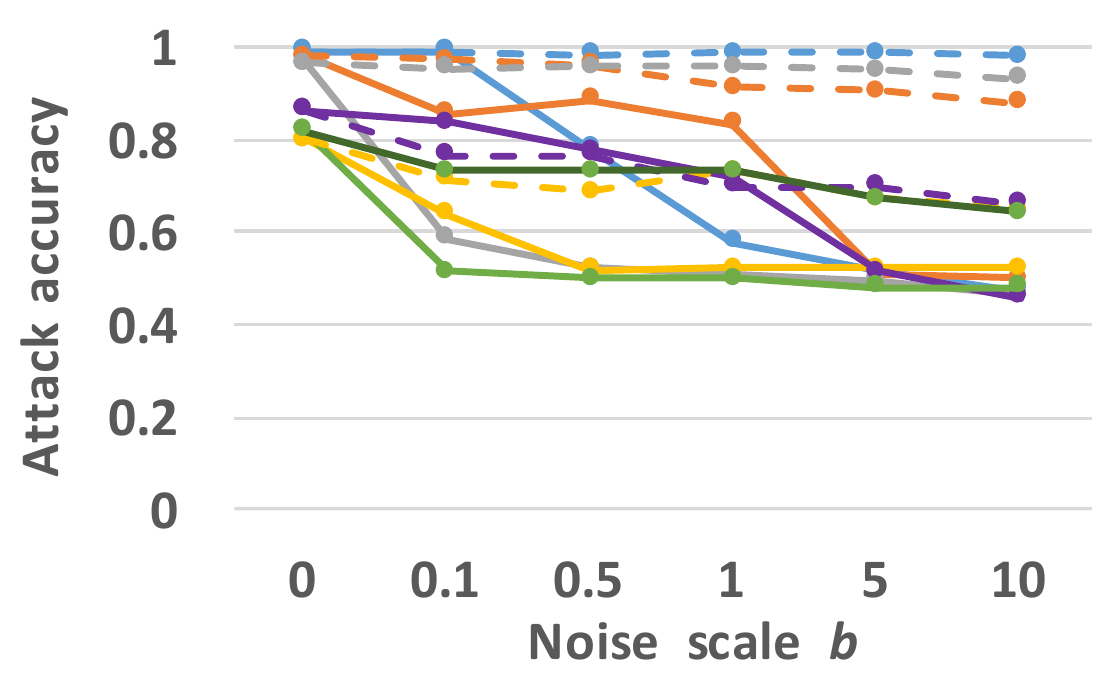}
     \vspace{-0.2in}
    \caption{\label{fig:gs-lap-embed1}Embedding $Z^1$}
    \end{subfigure}
     \begin{subfigure}[b]{.3\textwidth}
      \centering
     \includegraphics[width=\textwidth]{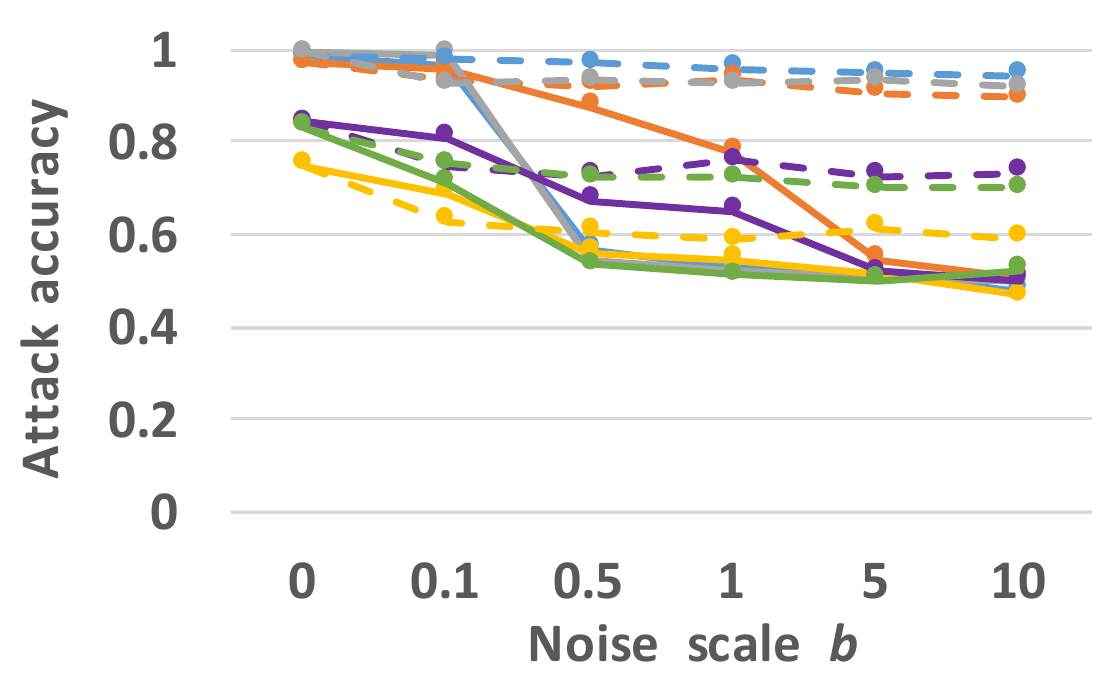}
     \vspace{-0.2in}
    \caption{\label{fig:gs-lap-embed2}Embedding $Z^2$}
    \end{subfigure}
    \begin{subfigure}[b]{.3\textwidth}
      \centering
     \includegraphics[width=\textwidth]{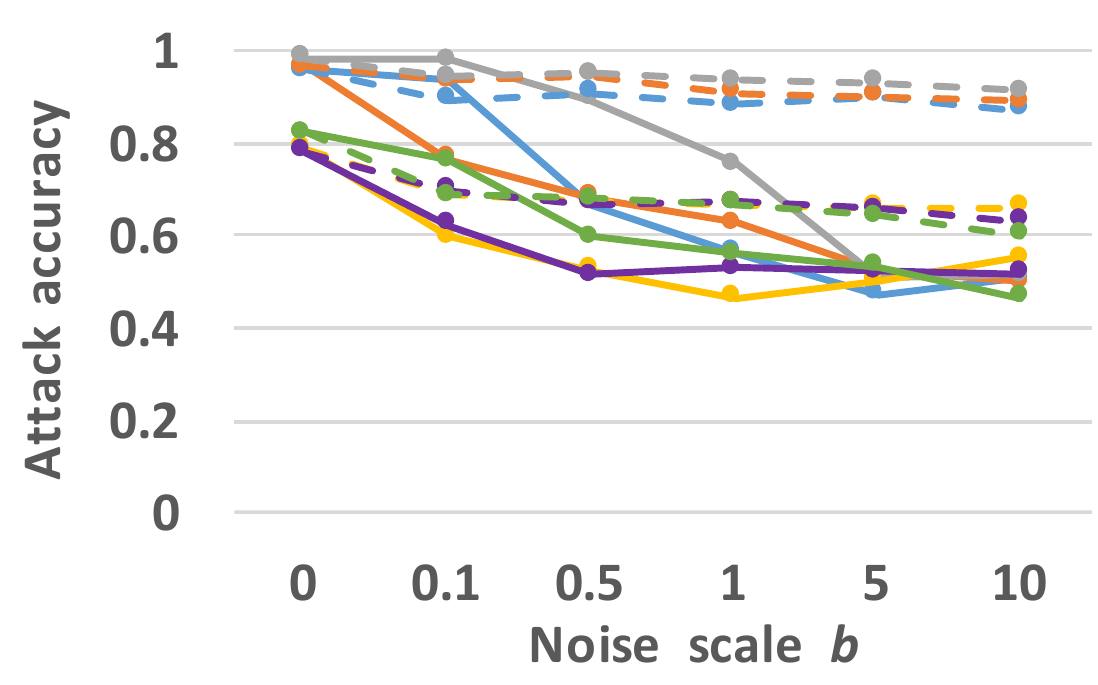}
     \vspace{-0.2in}
    \caption{\label{fig:gs-lap-post}Posteriors}
    \end{subfigure}
    \end{tabular}
    \\
  \vspace{0.05in}
    {\bf GAT}
    \\
    \vspace{0.1in}
        \begin{subfigure}[b]{.6\textwidth}
      \centering
     \includegraphics[width=\textwidth]{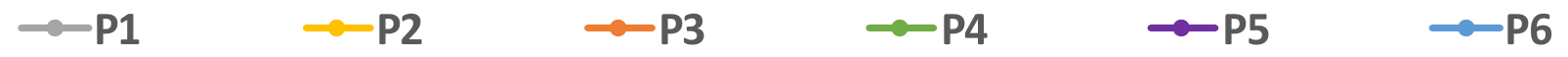}
     \vspace{-0.2in}
    \end{subfigure}
    \begin{tabular}{cc}
    \begin{subfigure}[b]{.3\textwidth}
      \centering
     \includegraphics[width=\textwidth]{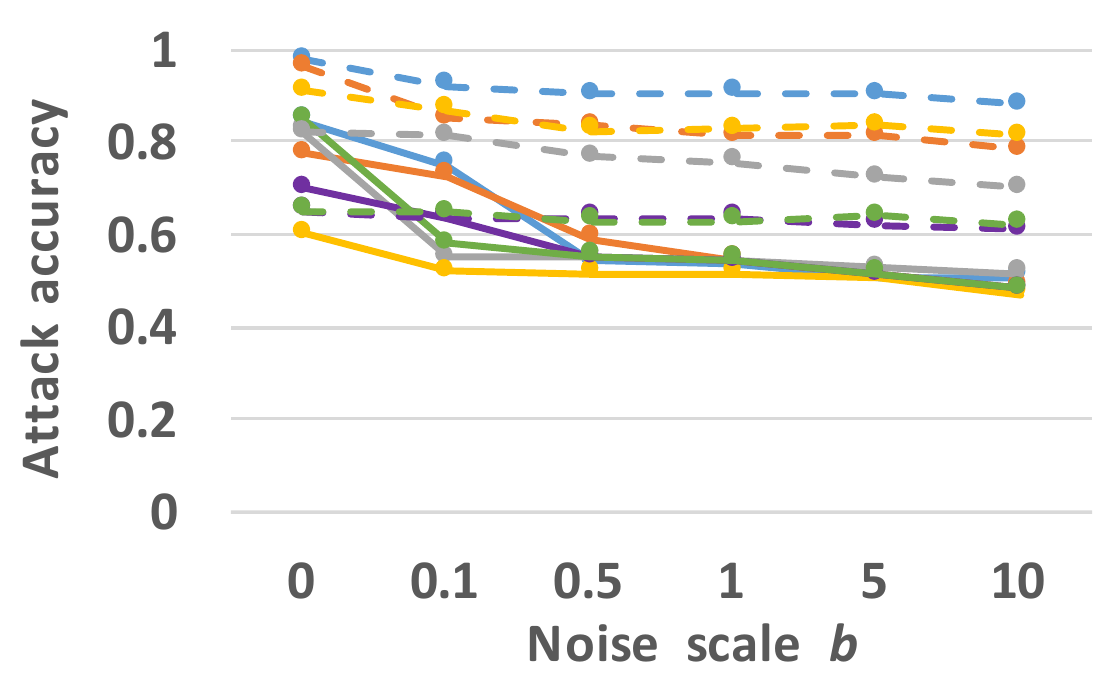}
     \vspace{-0.2in}
    \caption{\label{fig:gat-lap-embed1}Embedding $Z^1$}
    \end{subfigure}
     \begin{subfigure}[b]{.3\textwidth}
      \centering
     \includegraphics[width=\textwidth]{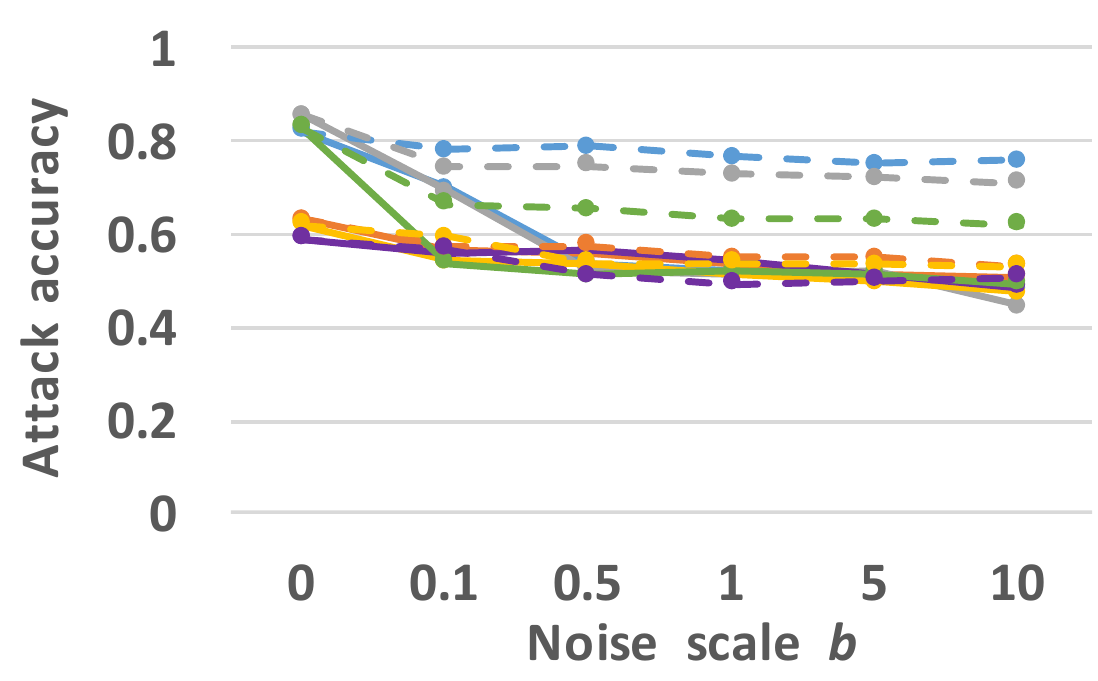}
     \vspace{-0.2in}
    \caption{\label{fig:gat-lap-embed2}Embedding $Z^2$}
    \end{subfigure}
    \begin{subfigure}[b]{.3\textwidth}
      \centering
     \includegraphics[width=\textwidth]{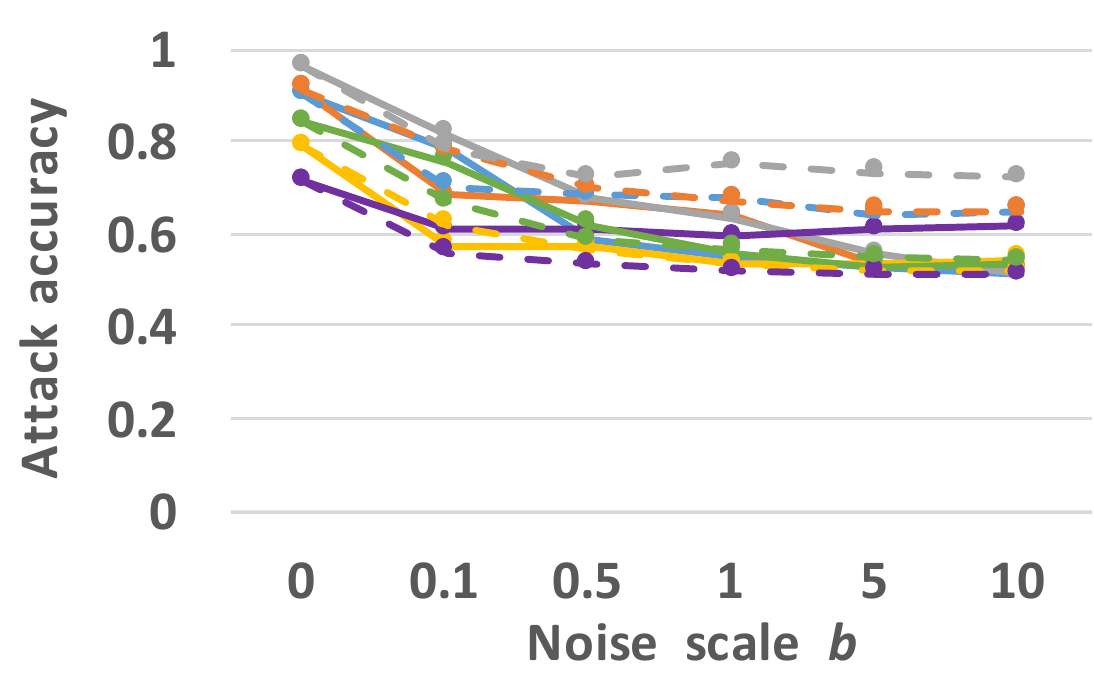}
     \vspace{-0.2in}
    \caption{\label{fig:gat-lap-post}Posteriors}
    \end{subfigure}
    \end{tabular}
    \vspace{-0.1in}
\caption{\label{fig:gs-lap} Defense effectiveness of the noisy posterior/embedding defense method on both GraphSAGE and GAT models. }
\vspace{-0.1in}
\end{figure*}

\nop{
\begin{figure*}[t!]
\centering
    \centering
    \begin{subfigure}[b]{.6\textwidth}
      \centering
     \includegraphics[width=\textwidth]{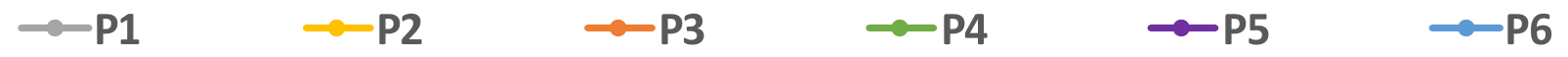}
     \vspace{-0.2in}
    \end{subfigure}
    \begin{tabular}{cc}
    \begin{subfigure}[b]{.3\textwidth}
      \centering
     \includegraphics[width=\textwidth]{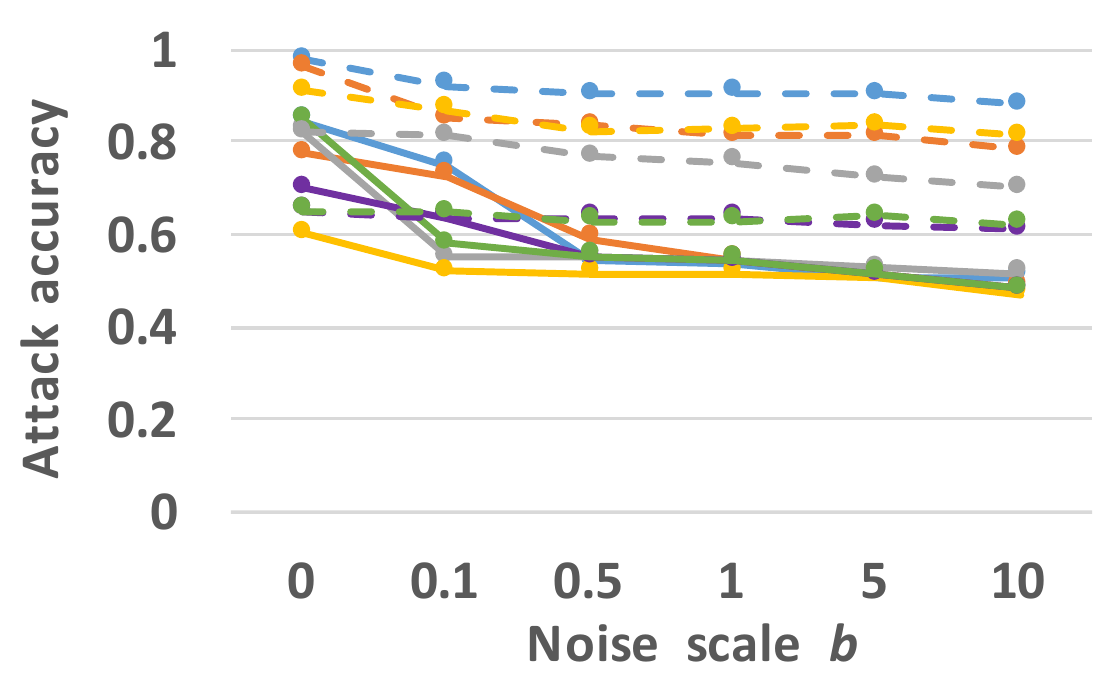}
     \vspace{-0.2in}
    \caption{\label{fig:gat-lap-embed1}Embedding $Z^1$}
    \end{subfigure}
     \begin{subfigure}[b]{.3\textwidth}
      \centering
     \includegraphics[width=\textwidth]{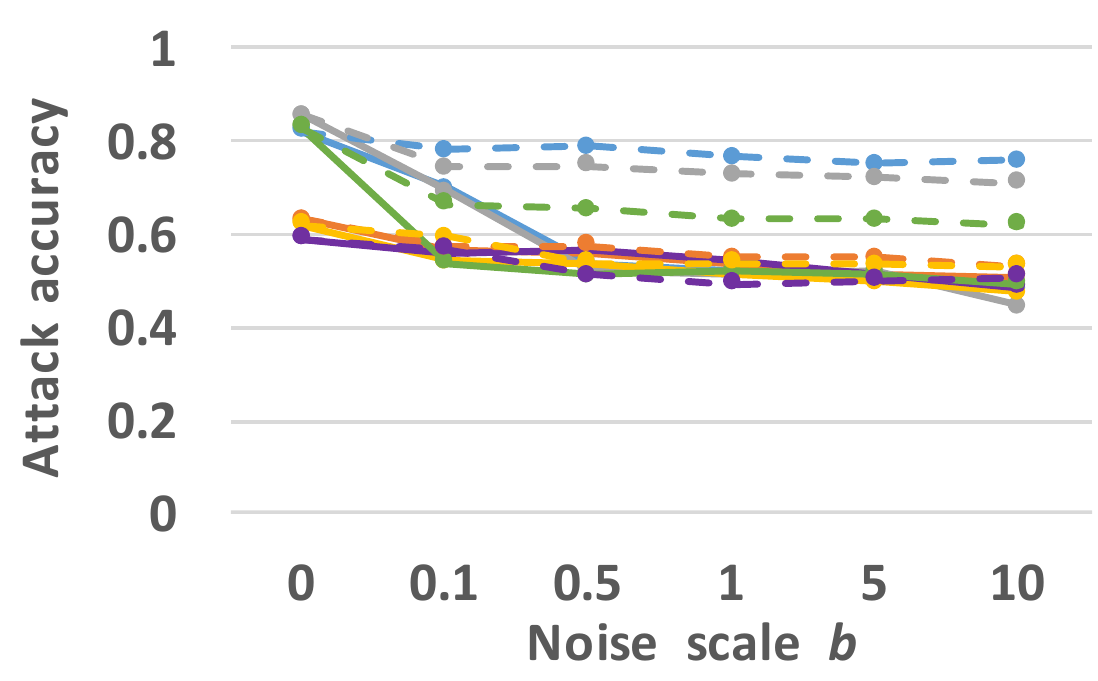}
     \vspace{-0.2in}
    \caption{\label{fig:gat-lap-embed2}Embedding $Z^2$}
    \end{subfigure}
    \begin{subfigure}[b]{.3\textwidth}
      \centering
     \includegraphics[width=\textwidth]{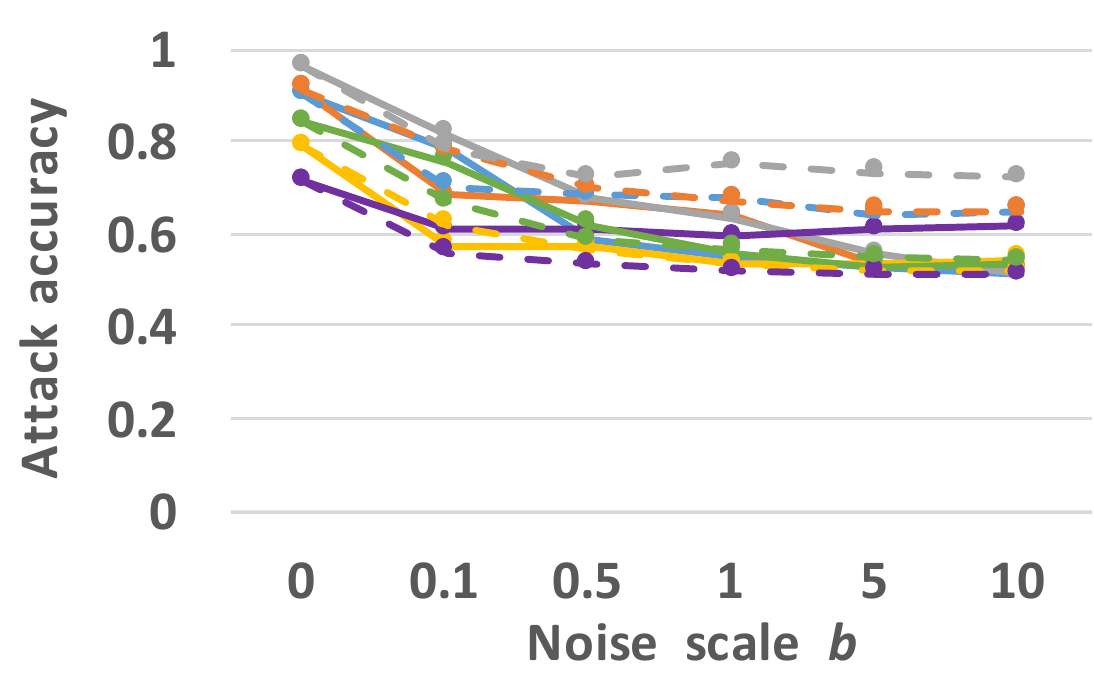}
     \vspace{-0.2in}
    \caption{\label{fig:gat-lap-post}Posteriors}
    \end{subfigure}
    \end{tabular}
\caption{\label{fig:gat-lap}  Defense effectiveness of the noisy posterior/embedding defense method (GAT as the target model). The noisy embedding defense is used against $A^1_1$ and $A^2_1$, while the noisy posterior defense is used against $A_2$.  }
\end{figure*}
}

\begin{figure*}[t!]
\centering
   \vspace{0.1in}
    \centering
    {\bf GraphSAGE}
    \\
       \begin{subfigure}[b]{.8\textwidth}
      \centering
     \includegraphics[width=\textwidth]{text/figure/def12-legend11.pdf}
    \end{subfigure}
    \begin{tabular}{cc}
    \begin{subfigure}[b]{.3\textwidth}
      \centering
     \includegraphics[width=\textwidth]{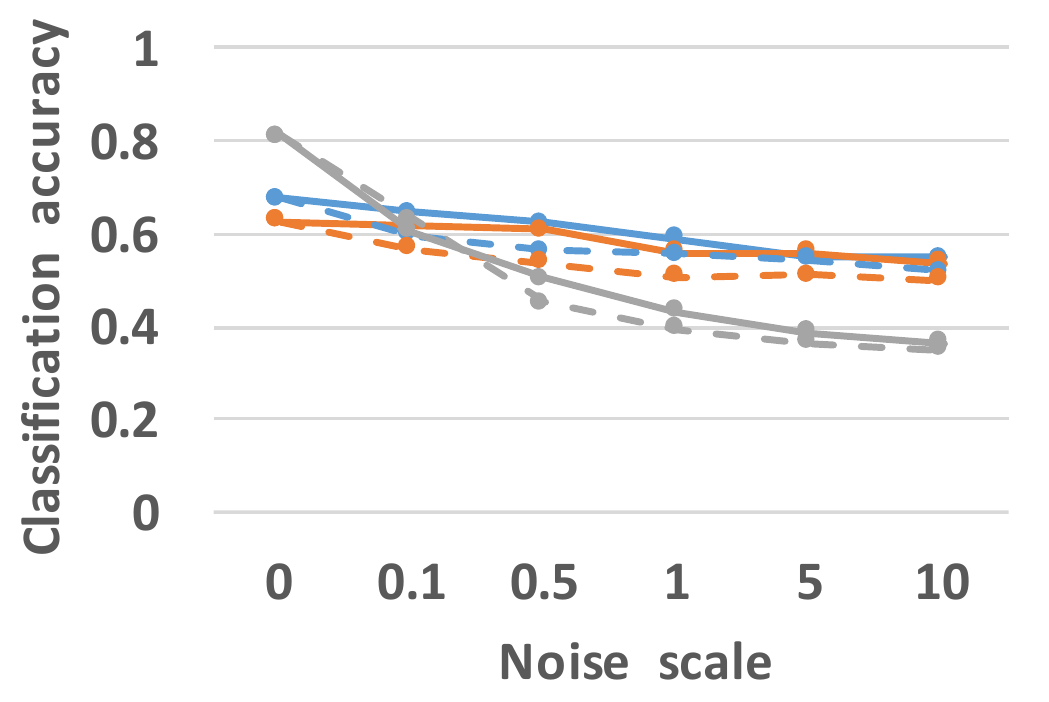}
     \vspace{-0.2in}
    \caption{\label{fig:gs-lap-embed1-acc}Embedding $Z^1$}
    \end{subfigure}
     \begin{subfigure}[b]{.3\textwidth}
      \centering
     \includegraphics[width=\textwidth]{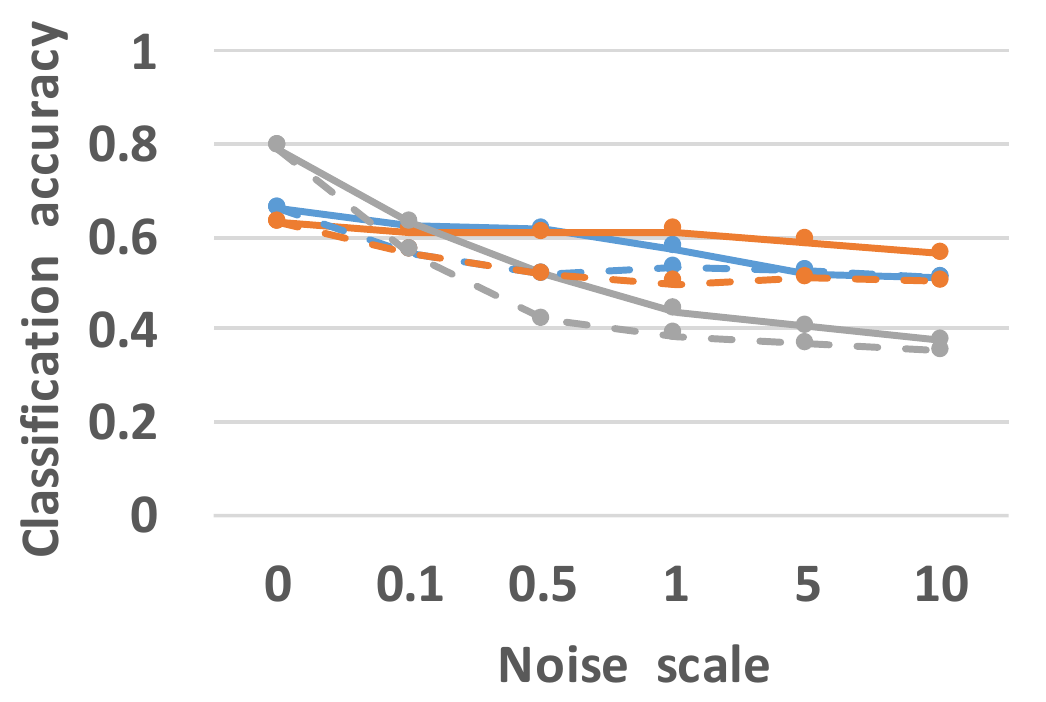}
     \vspace{-0.2in}
    \caption{\label{fig:gs-lap-embed2-acc}Embedding $Z^2$}
    \end{subfigure}
    \begin{subfigure}[b]{.3\textwidth}
      \centering
     \includegraphics[width=\textwidth]{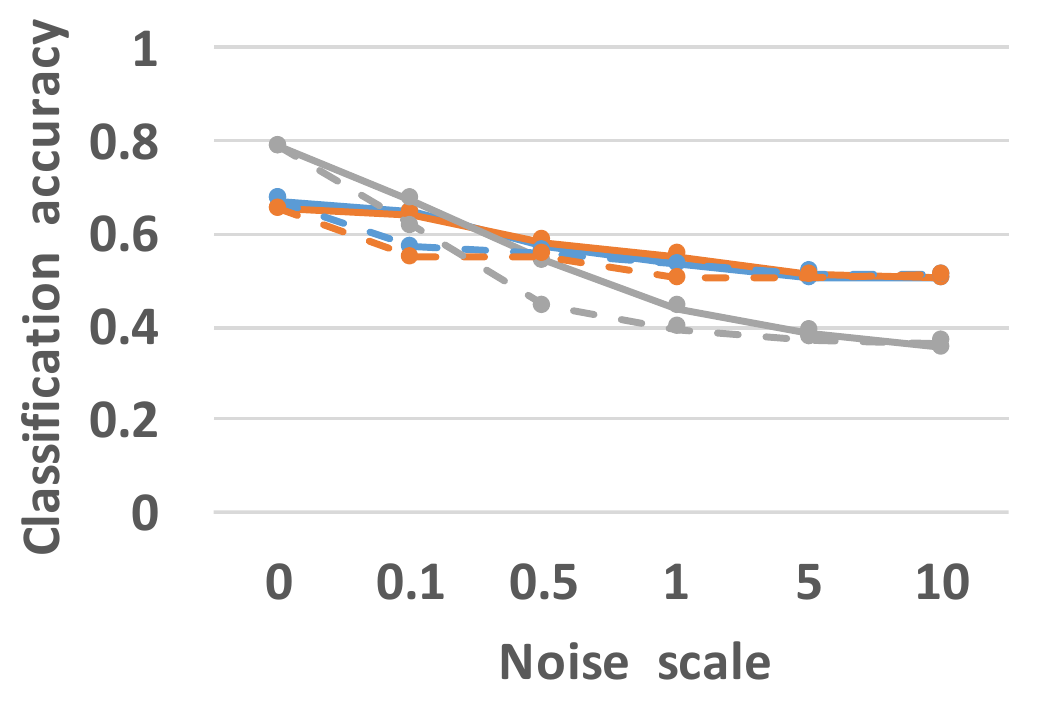}
     \vspace{-0.2in}
    \caption{Posteriors\label{fig:gs-lap-post-acc}}
    \end{subfigure}
    \end{tabular}
    \\
    \vspace{0.05in}
    {\bf GAT}
    \\
        \begin{subfigure}[b]{.8\textwidth}
      \centering
     \includegraphics[width=\textwidth]{text/figure/def12-legend11.pdf}
     \vspace{-0.2in}
    \end{subfigure}
    \begin{tabular}{cc}
    \begin{subfigure}[b]{.3\textwidth}
      \centering
     \includegraphics[width=\textwidth]{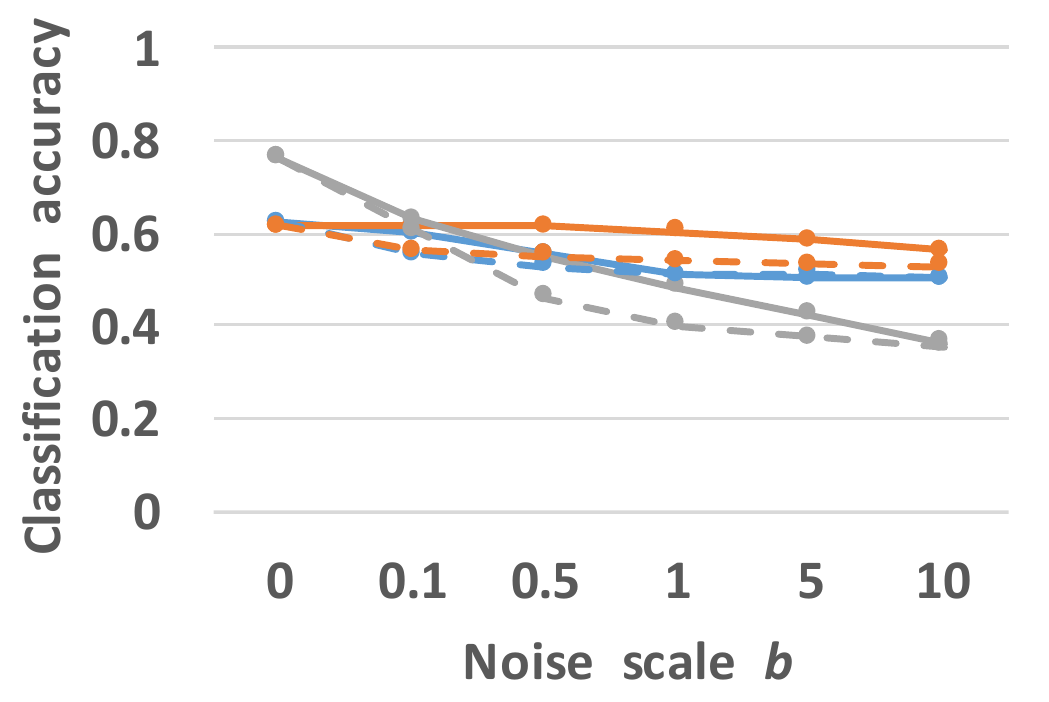}
     \vspace{-0.2in}
    \caption{\label{fig:gat-lap-embed1-acc}Embedding $Z^1$}
    \end{subfigure}
     \begin{subfigure}[b]{.3\textwidth}
      \centering
     \includegraphics[width=\textwidth]{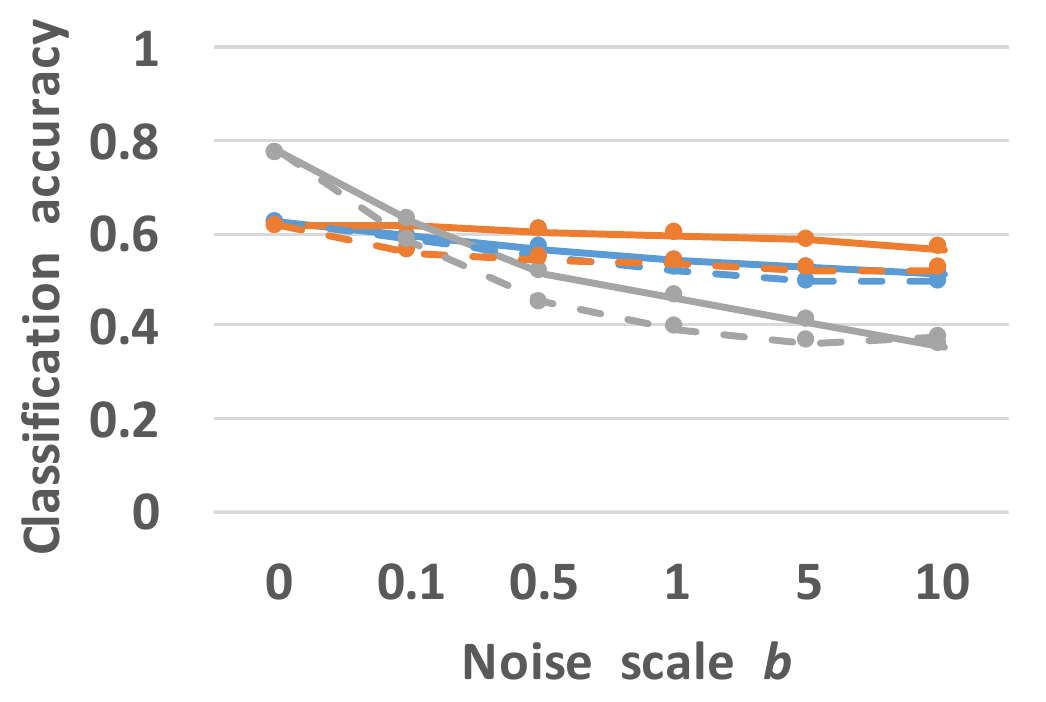}
     \vspace{-0.2in}
    \caption{\label{fig:gat-lap-embed2-acc}Embedding $Z^2$}
    \end{subfigure}
    \begin{subfigure}[b]{.3\textwidth}
      \centering
     \includegraphics[width=\textwidth]{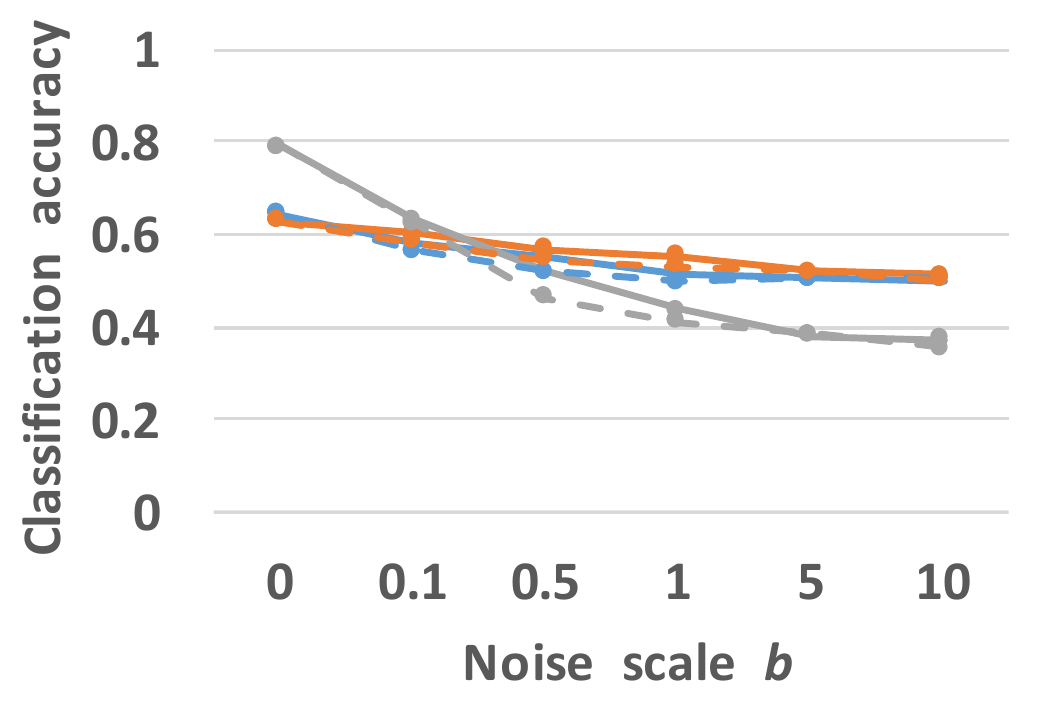}
     \vspace{-0.2in}
    \caption{Posteriors\label{fig:gat-lap-post-acc}}
    \end{subfigure}
    \end{tabular}
    \vspace{-0.1in}
\caption{\label{fig:gat-lap-acc} Target model accuracy under the noisy posterior/embedding defense for both GraphSAGE and GAT models.}
\end{figure*}

\subsection{Noisy embedding/posterior defense performance on GraphSAGE and GAT}
\label{appendix:gs-gat-defense1}
Figure \ref{fig:gs-lap} shows the attack performance after adding Laplace noise against GraphSAGE and GAT models. The observation is similar to Figure \ref{fig:gcn-lap} that the attack accuracy can be reduced to close to 0.5 for both black-box and white-box attacks when noise scale $b \leq 5$ for GraphSAGE and $b \leq 0.5$ for GAT, while DP is weaker than our defense in most of the cases). 

Figure \ref{fig:gat-lap-acc} shows the node classification accuracy after adding Laplace noise against GraphSAGE and GAT. The observation is similar to Figure \ref{fig:gcn-lap-acc} that the target model accuracy downgrades when more noise is added to the embeddings/posteriors. And the target model accuracy of our methods always outperforms that of DP. 


\subsection{Performance of Embedding Truncation Defense on GraphSAGE and GAT}
\label{appendix:gs-gat-defense2}
\begin{figure*}[t!]
\centering
   \vspace{0.1in}
    {\bf Attack accuracy}\\
    \begin{subfigure}[b]{.45\textwidth}
      \centering
     \includegraphics[width=\textwidth]{text/figure/trunc-legend2.pdf}
     \vspace{-0.2in}
    \end{subfigure}
    \\
    \begin{tabular}{cccc}
    \begin{subfigure}[b]{.23\textwidth}
      \centering
     \includegraphics[width=\textwidth]{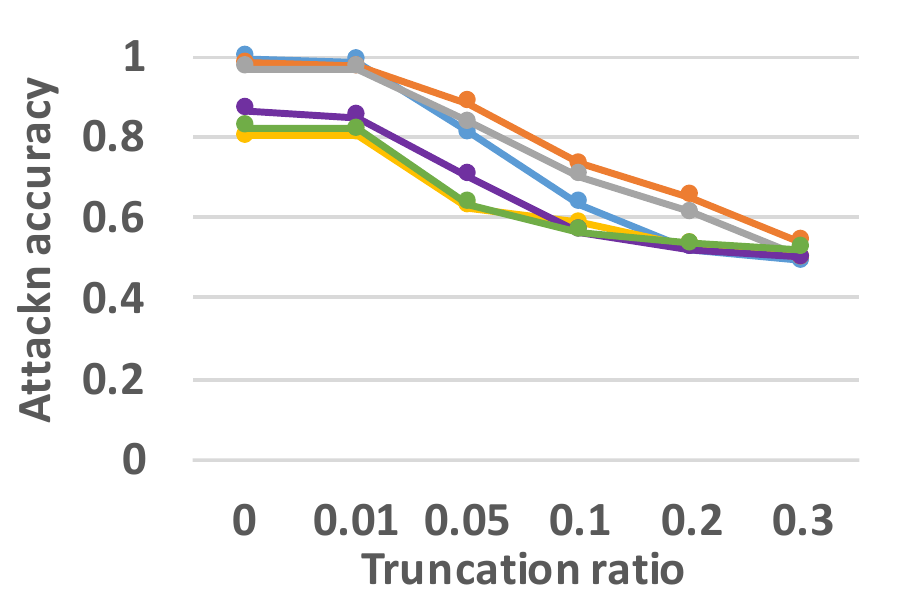}
     \vspace{-0.2in}
    \caption{\label{fig:gs-trunc-embed1} Attack $A^1_1$, GraphSAGE}
    \end{subfigure}
     \begin{subfigure}[b]{.23\textwidth}
      \centering
     \includegraphics[width=\textwidth]{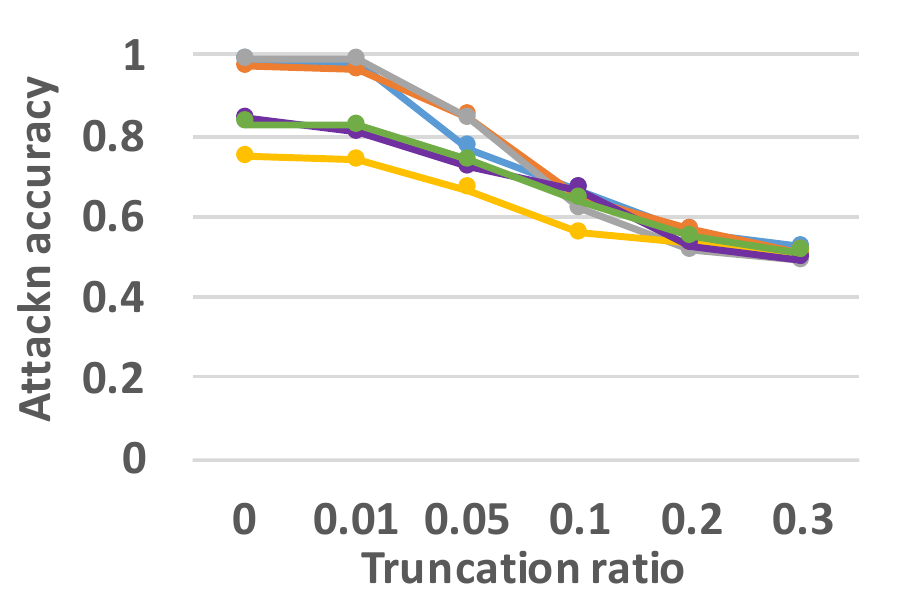}
     \vspace{-0.2in}
    \caption{\label{fig:gs-trunc-embed2} Attack $A^2_1$, GraphSAGE}
    \end{subfigure}
    \begin{subfigure}[b]{.23\textwidth}
      \centering
    \includegraphics[width=\textwidth]{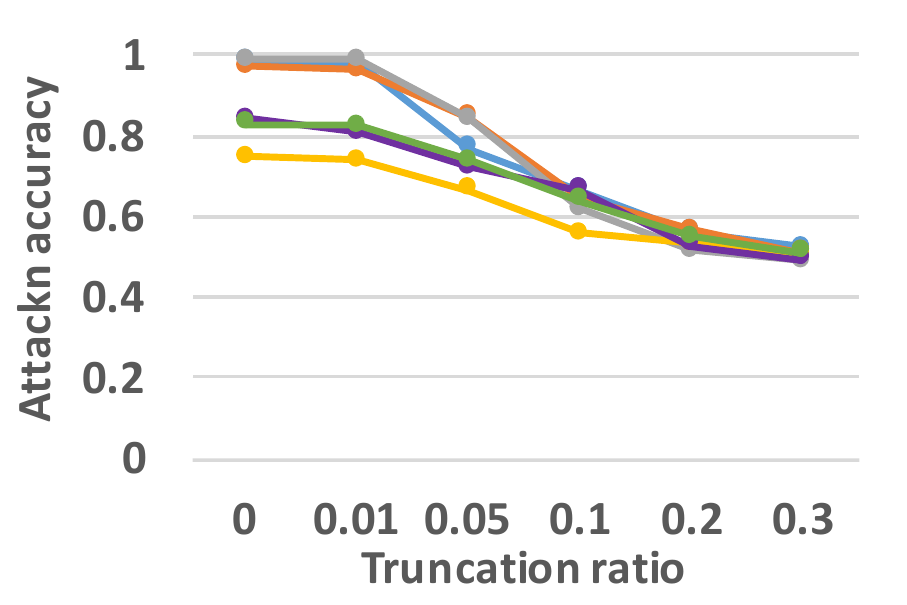}
     \vspace{-0.2in}
    \caption{\label{fig:gat-trunc-embed1} Attack $A^1_1$, GAT}
    \end{subfigure}
     \begin{subfigure}[b]{.23\textwidth}
      \centering
     \includegraphics[width=\textwidth]{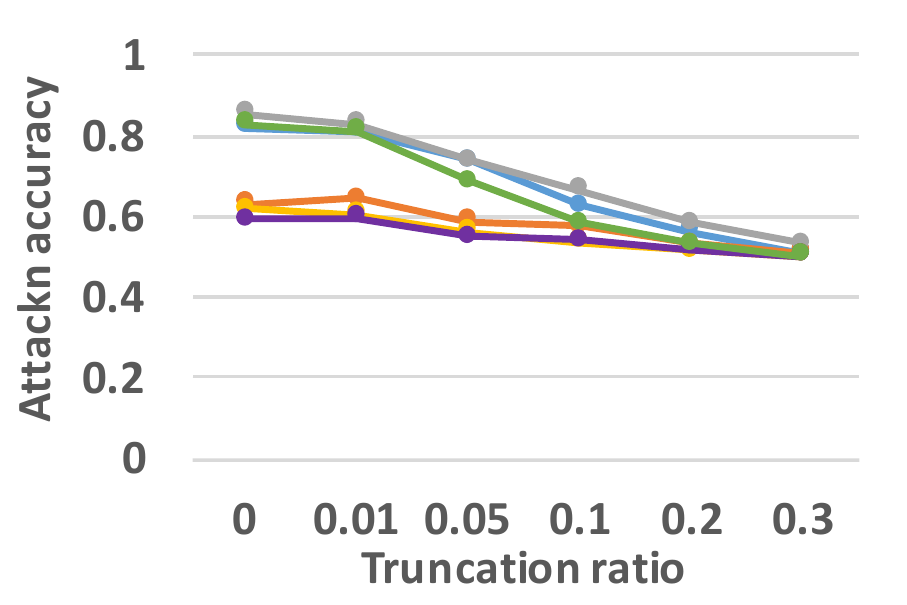}
     \vspace{-0.2in}
    \caption{\label{fig:gat-trunc-embed2} Attack $A^2_1$, GAT}
    \end{subfigure}
    \\
    \vspace{0.1in}
    {\bf Target model accuracy}\\
       \begin{subfigure}[b]{.3\textwidth}
      \centering
     \includegraphics[width=\textwidth]{text/figure/trunc-target-acc-legend1.pdf}
     \vspace{-0.2in}
    \end{subfigure}
    \\
    \begin{subfigure}[b]{.23\textwidth}
      \centering
     \includegraphics[width=\textwidth]{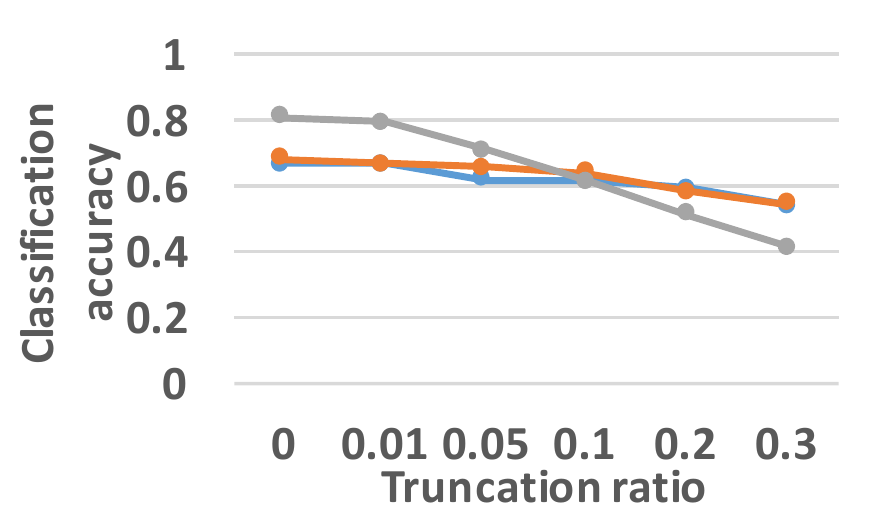}
     \vspace{-0.2in}
    \caption{\label{fig:gs-trunc-quality1}Embedding $Z^1$, GraphSAGE}
    \end{subfigure}
    \begin{subfigure}[b]{.23\textwidth}
      \centering
     \includegraphics[width=\textwidth]{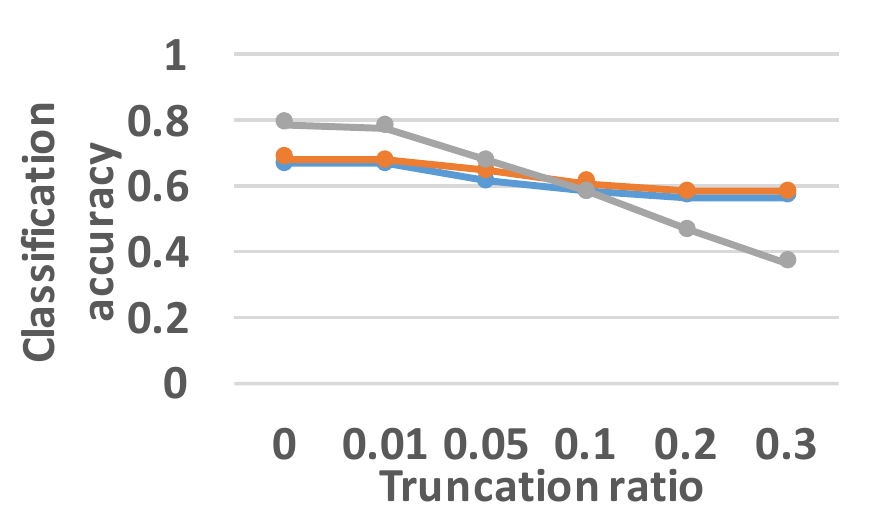}
     \vspace{-0.2in}
    \caption{\label{fig:gs-trunc-quality2}Embedding $Z^2$, GraphSAGE}
    \end{subfigure}
        \begin{subfigure}[b]{.23\textwidth}
      \centering
     \includegraphics[width=\textwidth]{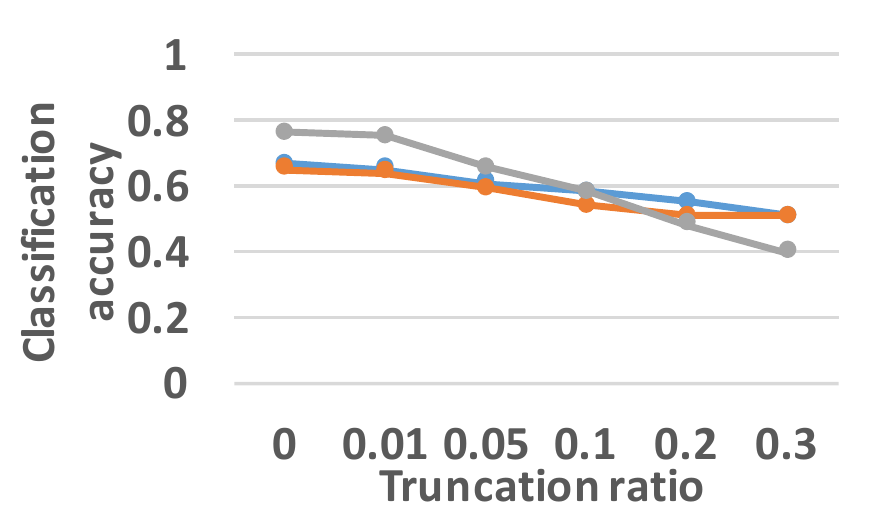}
     \vspace{-0.2in}
    \caption{\label{fig:gat-trunc-quality1}Embedding $Z^1$, GAT}
    \end{subfigure}
    \begin{subfigure}[b]{.23\textwidth}
      \centering
     \includegraphics[width=\textwidth]{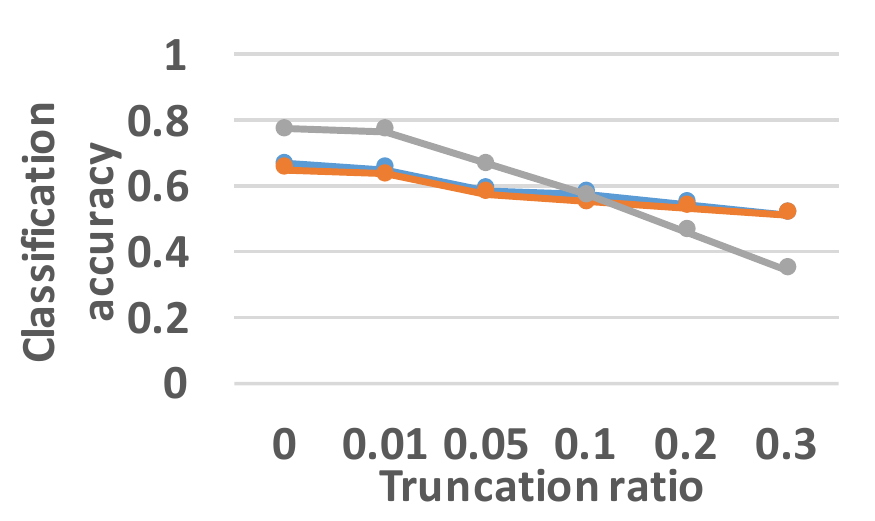}
     \vspace{-0.2in}
    \caption{\label{fig:gat-trunc-quality2}Embedding $Z^2$, GAT}
    \end{subfigure}
    \end{tabular}
\caption{\label{fig:gs-gat-trunc} Performance of the embeddings truncation defense for both GraphSAGE and GAT models.}
 \vspace{-0.2in}
\end{figure*}

Figure \ref{fig:gs-gat-trunc} shows the attack performance of Embedding truncation defense against GraphSAGE and GAT. The observation is similar to Figure \ref{fig:gcn-trunc} and thus the discussions are omitted.

\subsection{Mitigation of Disparity of Group Influence by Defense}
\label{appendix:defense-mitigation}

To explain how adding perturbations on individual embeddings and posteriors can defend against property inference at group level, we measure the impact of perturbations on each group. We consider Facebook dataset and property $P_2$, and measure the average loss  of male and female groups respectively as their group loss. We note that it is difficult to calculate the gradient-based influence score  (Eqn. \ref{eqn:if}) of groups as adding noise on embeddings/posteriors will not change the gradients (i.e., group influence score remains unchanged).
Table \ref{tab:intuition-defense} shows the results of the gap between group loss of Male and Female groups in Facebook dataset before and after adding Laplace noise to node embeddings. The gap is measured as $\text{LG} = L_{\text{Male}} - L_{\text{Female}}$, where $L_{\text{Male}}$ and $L_{\text{Female}}$ are the loss of male and female groups respectively. We observed that  more noise leads to smaller disparity in group loss. Since one of root causes of GPIA is the disparate influence and loss across different groups (Section \ref{sc:exp-pia}), adding noise can defend against GPIA by mitigating such  loss gap.

\end{document}